%% file: MAIN.tex
\documentclass[lettersize,twoside,journal]{IEEEtran}
\usepackage{amsmath,amsfonts}
\usepackage{array}
\usepackage{stfloats}
\usepackage{url}
\usepackage{verbatim}
\usepackage{graphicx}
\usepackage{cite}
\usepackage{amssymb}
\usepackage{booktabs}
\usepackage{epsfig}
\usepackage{grffile}
\usepackage{multirow}
\usepackage{comment}
\usepackage[export]{adjustbox}
\usepackage[dvipsnames]{xcolor} %
\usepackage{tabularray-2021}
\usepackage{makecell}
\usepackage[utf8]{inputenc}
\usepackage[T1]{fontenc}    %
\usepackage{mathrsfs}
\usepackage{wrapfig}

\usepackage{url}

\usepackage{dsfont}
\usepackage{caption}

\definecolor{ultramarine}{RGB}{0,32,96}
\definecolor{tred}{RGB}{191,10,10}
\definecolor{tpink}{RGB}{232,114,114}
\newcommand{\tred}[1]{\textbf{\textcolor{tred}{#1}}}
\newcommand{\tpink}[1]{\textbf{\textcolor{tpink}{#1}}}

\usepackage[ruled,vlined,linesnumbered]{algorithm2e}

\makeatletter
\newcommand{\removelatexerror}{\let\@latex@error\@gobble}
\makeatother

\makeatletter
\@namedef{ver@everyshi.sty}{}
\makeatother
\usepackage{tikz}
\usetikzlibrary{shapes, arrows, calc, positioning}

\definecolor{ForestGreeny}{RGB}{37,141,35}
\usepackage[pagebackref,breaklinks,colorlinks,urlcolor=black,citecolor=ForestGreeny]{hyperref}

\usepackage{array}
\usepackage[export]{adjustbox}
\usepackage{amsthm}

\usepackage{thmtools,thm-restate}
\usepackage[english]{babel}

\usepackage{tcolorbox}
\tcbuselibrary{theorems}
\tcbuselibrary{magazine}%

\usepackage[e]{esvect}
\newcommand{\myvec}[2][1mu]{\vec{#2\mkern-#1}\mkern#1}

\definecolor{cadmiumgreen}{rgb}{0.0, 0.42, 0.24}
\definecolor{darkpink}{rgb}{0.91, 0.33, 0.5}
\definecolor{mygreen}{RGB}{20, 99, 4}
\definecolor{ForestGreen}{RGB}{27,132,27}
\definecolor{ultramarine}{RGB}{0,32,96}

\newcommand{\lrtm}{6pt}
\newcommand{\tbtm}{3pt}

\newcommand{\tempp}[1]{#1} %

\newtcbtheorem[number within=section]{myprop}{Proposition}%
{colback=blue!5,colframe=blue!50!black,fonttitle=\bfseries, label type=myprop, theorem hanging indent=0pt,left=\lrtm,right=\lrtm,top=\tbtm,bottom=\tbtm}{mp}

\newtcbtheorem[number within=section]{mythm}{Theorem}%
{colback=red!5,colframe=red!35!black,fonttitle=\bfseries, theorem hanging indent=0pt,left=\lrtm,right=\lrtm,top=\tbtm,bottom=\tbtm}{th}

\newtcbtheorem[number within=section]{mydef}{Definition}%
{colback=TealBlue!5,colframe=TealBlue!35!black,fonttitle=\bfseries, theorem hanging indent=0pt,left=\lrtm,right=\lrtm,top=\tbtm,bottom=\tbtm}{df}

\newtcbtheorem[number within=section]{mycdef}{Constraint Definition}%
{colback=cadmiumgreen!5,colframe=cadmiumgreen,fonttitle=\bfseries, theorem hanging indent=0pt,left=\lrtm,right=\lrtm,top=\tbtm,bottom=\tbtm}{cdf}

\newtcbtheorem[number within=section]{mycor}{Corollary}%
{colback=Peach!10,colframe=Orange!80!black,fonttitle=\bfseries, theorem hanging indent=0pt,left=\lrtm,right=\lrtm,top=\tbtm,bottom=\tbtm}{mc}

\newtcbtheorem[number within=section]{mylem}{Lemma}%
{colback=darkpink!10,colframe=darkpink,fonttitle=\bfseries, theorem hanging indent=0pt,left=\lrtm,right=\lrtm,top=\tbtm,bottom=\tbtm}{lma}

\newcommand{\myrecall}[2][1]{\par\noindent\useboxarray[#2]{#1}}
\usepackage{environ}
\NewEnviron{restatlem}[2]{%
	\newboxarray{#2}%
	\begin{mylem}[reset box array=#2, store to box array=#2]{#1}{#2}
		\BODY%
	\end{mylem}%
	\myrecall{#2}%
}
\NewEnviron{restatthm}[2]{%
	\newboxarray{#2}%
	\begin{mythm}[reset box array=#2, store to box array=#2]{#1}{#2}
		\BODY%
	\end{mythm}%
	\myrecall{#2}%
}
\NewEnviron{restatcor}[2]{%
	\newboxarray{#2}%
	\begin{mycor}[reset box array=#2, store to box array=#2]{#1}{#2}
		\BODY%
	\end{mycor}%
	\myrecall{#2}%
}

\newcommand{\acrshort}[1]{\uppercase{#1}}
\newcommand{\abs}{Analysis-by-Synthesis}

\newcommand{\xirepname}{{indicator}}
\newcommand{\Xirepname}{{Indicator}}

\newcommand{\closure}{\mathscr{C}}

\newcommand{\visrays}{\mathcal{R}_V}
\newcommand{\nonvisrays}{\mathcal{R}_N}

\newcommand{\lvz}{Q_{d,\xi}}

\newcommand{\dvdz}{\mathcal{D}_{d,\xi}}

\renewcommand{\hat}[1]{\widehat{#1}}

\newcommand{\real}{\mathbb{R}}

\newcommand{\indic}[1]{\mathds{1}\left[ #1 \right]}

\newcommand{\rulesep}{\unskip\ \vrule\ }
\newcommand{\joinR}{\hspace{-.1em}}
\newcommand{\RomanI}{I}
\newcommand{\RomanII}{\mbox{\RomanI\joinR\RomanI}}
\DeclareMathOperator*{\argmin}{\arg\!\min}
\DeclareMathOperator*{\argmax}{\arg\!\max}

\begin{document}

\newcommand{\titlename}{Probabilistic Directed Distance Fields for Ray-Based Shape Representations}

\title{\titlename}

\author{
	Tristan Aumentado-Armstrong,
	Stavros Tsogkas,
	Sven Dickinson,
	Allan Jepson
	\thanks{
		All authors were with the Samsung AI Centre, Toronto, at the time of this work.
		Tristan Aumentado-Armstrong, %
		Sven Dickinson, and
		Allan Jepson are also with the Department of Computer Science,  University of Toronto. %
	}
}

\newif\ifShowSupp
\ShowSupptrue %

\newif\ifRefAtEnd
\RefAtEndfalse %

\newcommand{\ts}[2]{%
	\ifShowSupp%
	#1%
	\else%
	#2%
	\fi%
}

\newcommand{\tr}[2]{%
	\ifRefAtEnd%
	#1%
	\else%
	#2%
	\fi%
}

\markboth{PDDFs for Ray-Based Shape Representations.}{Aumentado-Armstrong \MakeLowercase{\textit{et al.}}}

\maketitle
\thispagestyle{empty}

\begin{abstract}
In modern computer vision, the optimal representation of 3D shape remains task-dependent. One fundamental operation applied to such representations is differentiable rendering, which enables learning-based inverse graphics approaches. Standard explicit representations are often easily rendered, but can suffer from limited geometric fidelity, among other issues. On the other hand, implicit representations generally preserve greater fidelity, but suffer from difficulties with rendering, limiting scalability. In this work, we devise \textit{Directed Distance Fields} (DDFs), which map a ray or oriented point (position and direction) to surface visibility and depth. This enables efficient differentiable rendering, obtaining depth with a single forward pass per pixel, as well as higher-order geometry with only additional backward passes. 
Using probabilistic DDFs (PDDFs), we can model the inherent discontinuities in the underlying field. 
We then apply DDFs to single-shape fitting, generative modelling, and 3D reconstruction, showcasing strong performance with simple architectural components via the versatility of our representation. Finally, since the dimensionality of DDFs permits view-dependent geometric artifacts, we conduct a theoretical investigation of the constraints necessary for view consistency. We find a small set of field properties that are sufficient to guarantee a DDF is consistent, \textit{without knowing} which shape the field is expressing.
\end{abstract}


\input{intro} %

\input{ddfs}

\input{applications}

\input{theory}

\input{discussion}

\section*{Acknowledgments} %
We are grateful for support from
NSERC (CGSD3-534955-2019) and Samsung Research.

\tr{}{\input{bib_and_bio}}

\ts{}{ 
	\tr{}{
		\clearpage
		\newpage
	} 
}

\ts{ %
	\tr{
		\clearpage
		\newpage
		\setcounter{page}{1}
	}{} 
}{}

\ts{

\begin{center}
	{
	\Large 
	\titlename:
	}\\ {\Large Supplemental Material}
\end{center}

\appendices

\input{supp-geomprops}

\input{supp-nonptapps}

\input{supp-pathtrace}

\input{supp-theory}

}{}

\tr{\input{bib_and_bio}}{}

\end{document}

%% file: intro.tex
\section{Introduction}

\newcommand{\fwone}{0.16\textwidth}

\IEEEPARstart{T}{he} representation of 3D geometry remains an open problem, 
	with significant implications
	across computer vision, graphics, and artificial intelligence.
Generally, the appropriate representation is application-dependent,
	determined by the ease of relevant operations,
	such as deformation, 
	composition,
	learnability,
	and rendering.
The latter operation is particularly important:
	since the earliest days of computer vision
	(e.g., \cite{roberts1963machine,baumgart1974geometric}),
	extraction of 3D structure by matching 2D observations
	has been an essential task. 
In the modern context,
	the analysis-by-synthesis (AbS) approach to representation learning \cite{yuille2006vision}, 
	which treats vision as ``inverse graphics'', 
	relies on rendering to form the bridge between 3D and 2D.
For example, at the single-scene scale, 
differentiable volume rendering in neural radiance fields (NeRFs)
	\cite{mildenhall2020nerf}
	enables AbS extraction of a 3D representation
	that is highly effective for novel view synthesis (NVS).
Similarly, implicit geometric fields, 
such as occupancy fields \cite{mescheder2019occupancy}
and
signed distance fields (SDFs) \cite{park2019deepsdf},
have been used in conjunction with differentiable rendering as well
(e.g., \cite{liu2020dist,niemeyer2020differentiable,jiang2020sdfdiff,vicini2022differentiable,tewari2022advances}).
Finally, more abstract
representations, learned through generative models,
have also provided impressive results, 
connecting 2D and 3D via rendering
\cite{liu2023zero,bhattad2024stylegan,tewari2024diffusion,muller2023diffrf,wang2023score,poole2022dreamfusion}.

\begin{figure}[t]
	\centering
	\includegraphics[width=0.99\linewidth]{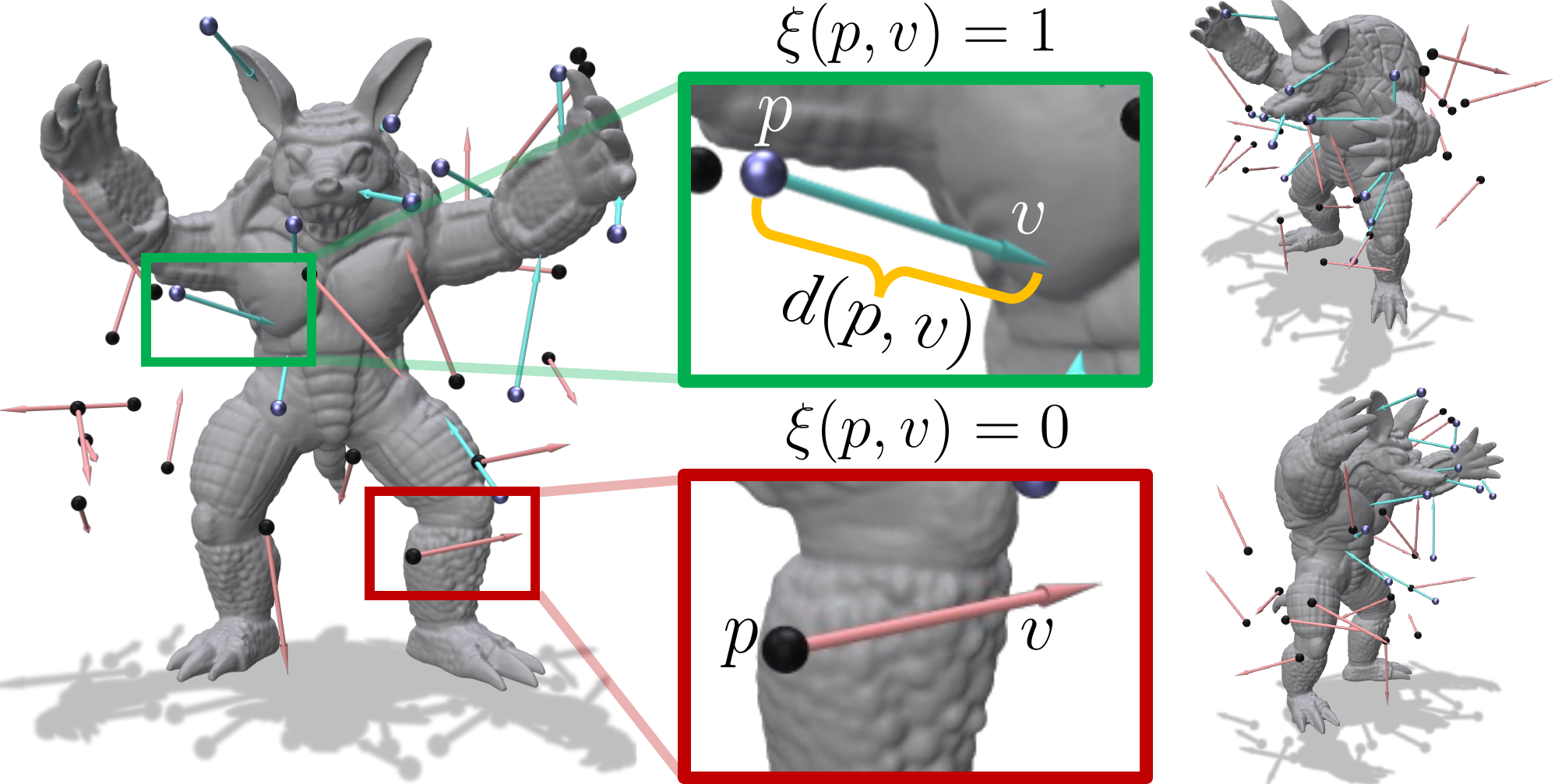}
	\caption{
		Basic depiction of a Directed Distance Field (DDF). 
		Oriented points (or rays) are shown as a position $p$ and direction $v$. 
		Each ray is assigned a \textit{visibility} value:
		$\xi(p,v)=1$ means it hits the shape, 
		while $\xi(p,v)=0$ means it missed.
		For rays with $\xi(p,v)=1$, 
		the \textit{distance field}, $d(p,v)$, returns the distance between $p$ and that intersection point (green box). 
	}
	\label{fig:armadillo}
\end{figure}

\begin{figure*}[t]
	\centering
	\begin{minipage}[c]{0.30\textwidth}%
		\includegraphics[width=0.99\textwidth]{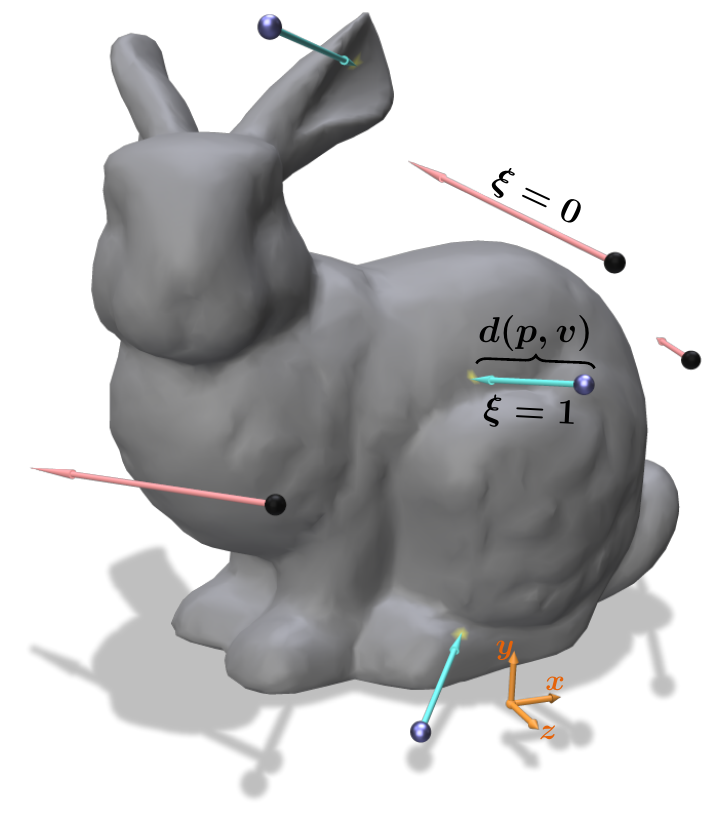}
	\end{minipage}\hfill%
		\begin{minipage}[l]{0.65\textwidth}%
			\includegraphics[height=\fwone]{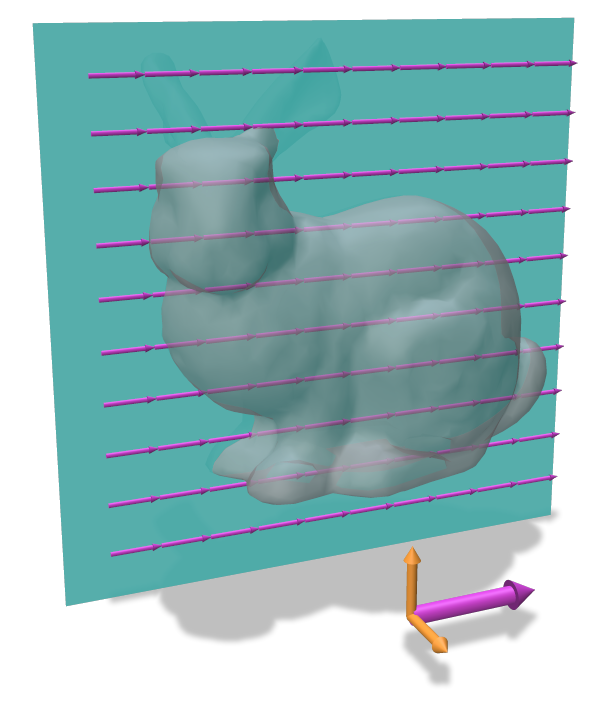}\hfill
			\includegraphics[height=\fwone]{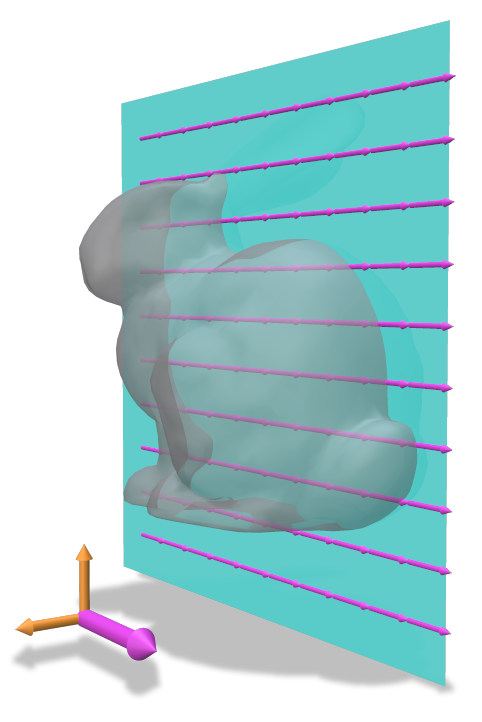}\hfill
			\includegraphics[height=\fwone]{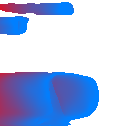}
			\includegraphics[height=\fwone]{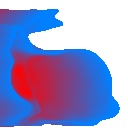}
			\includegraphics[height=\fwone]{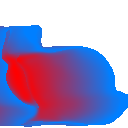}
			\includegraphics[height=\fwone]{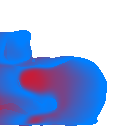}
			\\
			\includegraphics[height=\fwone]{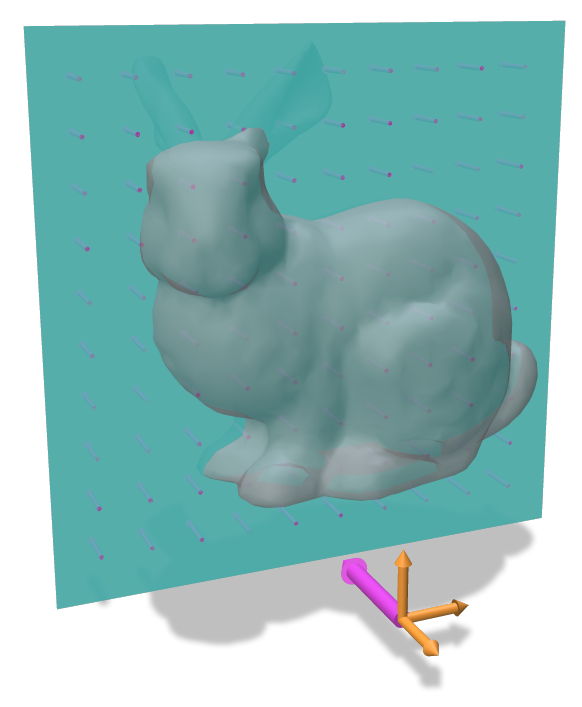}\hfill
			\includegraphics[height=\fwone]{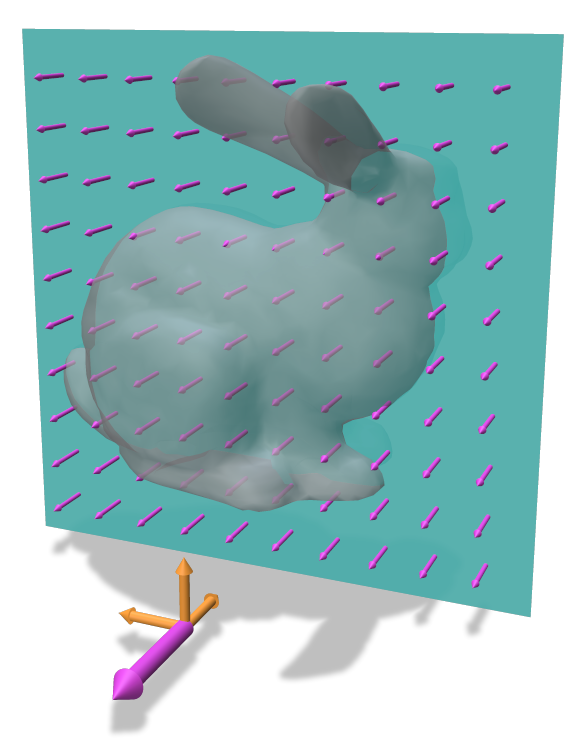}\hfill
			\includegraphics[height=\fwone]{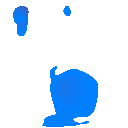}       
			\includegraphics[height=\fwone]{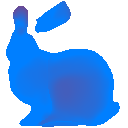}
			\includegraphics[height=\fwone]{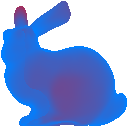}
			\includegraphics[height=\fwone]{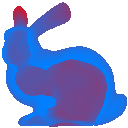}
			\\
			\includegraphics[height=\fwone]{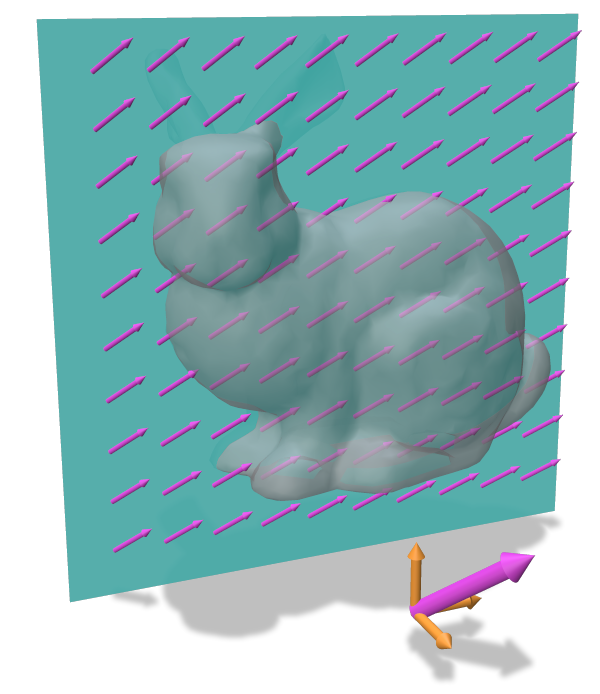}\hfill
			\includegraphics[height=\fwone]{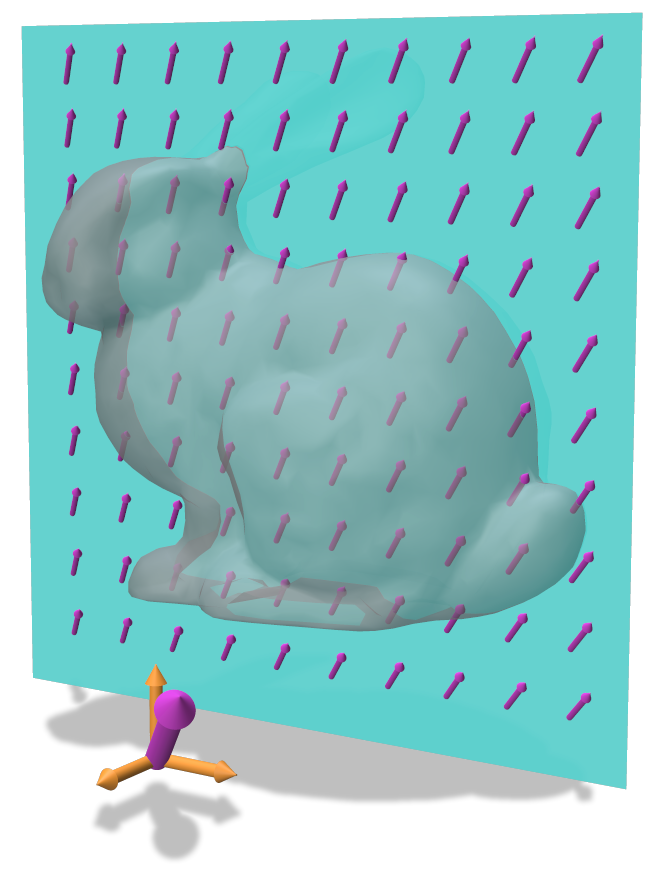}\hfill
			\includegraphics[height=\fwone]{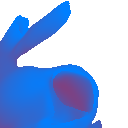}
			\includegraphics[height=\fwone]{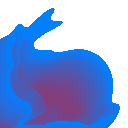}
			\includegraphics[height=\fwone]{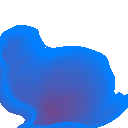}
			\includegraphics[height=\fwone]{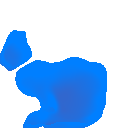}
		\end{minipage}
		\makebox[0pt][r]{%
			\begin{tikzpicture}[overlay]
				\node[xshift=0cm,yshift=-0.6mm] (hidden) at (0,0) {};
				\node[xshift=1cm,yshift=-0.6mm] (s) at (-9.18,-2.95) {};
				\node[xshift=1cm,yshift=-0.6mm] (e) at (-1.00,-2.95) {};
				\node[xshift=1cm,yshift=-0.6mm] (t) at (-5.09,-2.95) {\footnotesize Increasing $z$};
				\draw [-latex,thick] (s.center) --  node[right,above] {}(t.west) ;
				\draw [-latex,thick] (t.east) --  node[right,above] {}(e.center) ;
			\end{tikzpicture}
		}
		\caption{
			Two-dimensional spatial slices of a directed distance field (DDF).
			\textit{Left}: depiction of visible oriented points (blue points, turquoise directions) that intersect the shape and those that miss the shape (black points, red directions) with $\xi = 0$. 
			\textit{Middle}: per row, illustrations of one slice plane (from two different views) and the fixed $v$ vector per slice plane (pink arrows), corresponding to the insets on the right (i.e., $v$ is the same across all $p$ for each row). 
			\textit{Right}: resulting depth field evaluated across positions $p$ at fixed orientations $v$ (rows: top, middle, and bottom show different $v$ values, parallel to $(1,0,0)$, $(0,0,-1)$, and $(1,1,1)$, respectively; columns: different slices in 3D with each having fixed $z$, effectively sliding the turquoise plane from the middle inset in $z$). 
			Each pixel value is coloured with the distance value $d(p,v)$ obtained for that position $p$ and direction $v$ (red to blue meaning further to closer).
			Non-visible oriented points ($\xi=0$) are shown as white. 
		} %
	\label{fig:teaser}
\end{figure*}
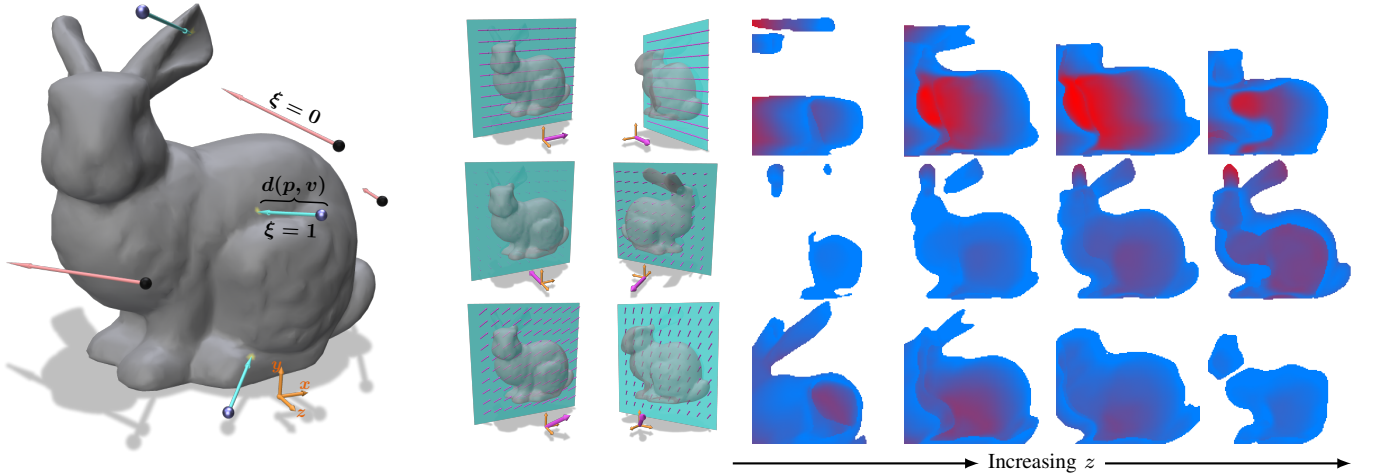

Nevertheless, 
it is still not always clear which representation is best for a given task.
Voxels and point clouds tend to have reduced geometric fidelity, while meshes suffer from the difficulties inherent in discrete structure generation 
(e.g., \cite{pan2019deep,tang2019skeleton}), often leading to topological and textural fidelity constraints, 
Meshes can also struggle with the dependence of rendering efficiency on shape complexity, and the ad hoc ``softening'' procedures need to enable differentiability (e.g., \cite{kato2018neural,liu2019soft}).
While implicit shapes can have superior fidelity, they struggle with complex or inefficient rendering procedures, requiring multiple network forward passes and/or complex calculations per pixel \cite{mildenhall2020nerf,sitzmann2019scene,liu2020dist}, and may be difficult to use for certain tasks (e.g., deformation, segmentation, or correspondence).
Thus, a natural question is how to design a method capable of fast differentiable rendering, yet still retaining high-fidelity geometric information 
that is useful for a variety of downstream applications. %
In general, rendering relies on ``directed geometric queries'' (DGQs), 
which ask whether and where a surface lies on a given ray;
our strategy is to make such queries the basic operation of the representation.

More specifically, in this work, 
we explore \textit{directed distance fields} (DDFs), 
a neural field representation of shape that 
	uses only a single forward pass per DGQ.
The definition is simple (see Fig.~\ref{fig:armadillo}): 
	for a given shape, we learn a 5D field that maps any 
	position and orientation (i.e., oriented point or ray) to
	\textit{visibility} 
	(i.e., whether the shape exists from that position along that direction) 
	and 
	\textit{distance}  
	(i.e., how far the shape is along that ray, if it is visible).
This enables efficient differentiable geometric rendering, 
		compared to other common implicit approaches, but
	can still capture detailed geometry, 
	including higher-order differential quantities and internal structure.

Fig.~\ref{fig:teaser} illustrates how DDFs can be viewed as implicitly storing all possible depth images of a given shape (i.e., from all possible cameras), 
reminiscent of a light field, but with geometric distance instead of radiance.
Such a field is inherently discontinuous (see Fig.\ \ref{fig:discontillus}), 
presenting issues for differentiable neural networks, 
but is advantageous in rendering, 
since a depth image can be computed with a single forward pass per pixel.
More significantly, however, 
the increased dimensionality permits \textit{view inconsistencies}
(meaning the DDF can represent an aberrant 5D entity, where \textit{geometry is view-dependent}, rather than a consistent 3D shape).
We mitigate these in our applications 
by regularizing properties of the field,
including analogues of the Eikonal constraint for classical distance fields,
though these alone cannot \textit{guarantee} view consistency (VC) in a mathematical sense.

Yet, knowing the theoretical conditions (i.e., field properties) 
	necessary to avoid such inconsistencies
	can be practically valuable
	(e.g., for formally defining geometric consistency, or for regularizer design).
Hence, we examine sufficient conditions to ensure VC,
	by introducing a notion of ``inducement'',
	whereby a shape can be used to construct a perfect DDF.
Then, starting from an \textit{un}constrained field (not necessarily representing a shape), we devise local conditions (independent of a fixed pre-defined shape)
	that ensure the existence of a shape that induces the field.
Specifically, the DDF subfields (i.e., visibility $\xi$ and depth $d$) 
	each have three corresponding constraints, 
	which independently ensure each field is ``internally'' consistent, 
	as well as two \textit{compatibility} conditions that mutually 
	constrain $\xi$ and $d$.
To our knowledge, this analysis is the first 
	theoretical investigation of 
	when the higher-dimensional, ray-based DDFs	
	formally represent 3D shapes.
Thus, our contributions can be summarized as follows:
\begin{enumerate}
	\item We define the DDF, a 5D mapping from any position and viewpoint to depth and visibility (\S\ref{sec:ddf}). 
	\item By construction, our representation allows differentiable rendering via a single forward pass per pixel (\S\ref{sec:app:rendering}), 
	without restrictions on the shape (topology, water-tightness) or field queries (internal structure).
	\item
	We construct probabilistic DDFs (PDDFs), 
	which differentiably model
	discontinuities 
	(\S\ref{sec:app:pddfs}).    %
	\item We examine several geometric properties of DDFs %
	(\S\ref{sec:ddf:properties}--\ref{sec:app:rendering})
	and use them in our method (\S\ref{sec:app:learning}). 
	\item We apply DDFs to fitting shapes (\S\ref{sec:results:singlefieldfitting}), 
	single-image reconstruction (\S\ref{sec:app:si3dr}), %
	and generative modelling (\S\ref{sec:genmodel}).
	\item 
	We show that DDFs can be used 
		for path tracing,
		using internal structure modelling capabilities 
		(\S\ref{sec:pathtracing}).
	\item 
	We present a theoretical analysis 
		of the conditions under which DDFs
		can guarantee multiview consistency 
		(i.e., act as an shape representation)
				in \S\ref{sec:theory}.
	These results hold for any ray-based geometry field,
		including concurrent work independent from DDFs
		(see also \S\ref{sec:relwork}).
\end{enumerate}

We remark that this work is an extension of a previous conference paper 
	\cite{aumentado2022representing}.
As noted just above in (6), the largest novel contribution of this work
	is a theoretical analysis of the conditions required for view consistency
	(\S\ref{sec:theory}).
Since the DDF has a direction-dependence (i.e., is 5D),
	it can produce fields that do not properly correspond with a 3D shape,
	due to inconsistencies 
	(i.e., geometry existing from one viewpoint but not others).
Hence, our interest is in what constraints on the field can be imposed,
	to ensure consistency and thus a representation of shape.
In addition,
	we demonstrate rendering efficiency in
	\S\ref{sec:main:renderingefficiency},
	and 
	provide a new interpretation of
	recursive DDF calls as  
	inter-reflections between surfaces,
	and apply this to path tracing 
	(\S\ref{sec:pathtracing}).

\section{Related Work}
\label{sec:relwork}

Our work falls under distance field representations of shape, 
which have a long history in computer vision \cite{rosenfeld1968distance}, 
recently culminating in signed/unsigned distance fields (S/UDFs) \cite{park2019deepsdf,chibane2020neural,venkatesh2020dude} and related methods 
\cite{cai2020learning,venkatesh2021deep}. 
Compared to explicit ones, 
implicit shapes can capture arbitrary topologies with high fidelity
\cite{park2019deepsdf,mescheder2019occupancy,liu2019learning}.
Several works examine differentiable rendering of implicit fields \cite{liu2020dist,jiang2020sdfdiff,niemeyer2020differentiable,sitzmann2019scene,liu2019learning,takikawa2021neural,zhang2021learning} (or combine it with neural volume rendering \cite{Kellnhofer:2021:nlr,oechsle2021unisurf,yariv2021volume,wang2021neus}), 
generally based on approaches that require many field queries per pixel, such as sphere tracing.
In contrast, by conditioning on both viewpoint and position, DDFs can flexibly render depth, with a \textit{single} field query per pixel.
Further, a UDF can actually be extracted from a DDF
(see \S\ref{sec:ddf:properties} and \S\ref{sec:app:udfextract}).

Separately, neural radiance fields (NeRFs) \cite{mildenhall2020nerf} 
provide a powerful 3D representation of both geometry and texture,
with increasingly impressive results in NVS (e.g., \cite{barron2023zip}).
However, the standard differentiable volume rendering used by NeRFs is computationally expensive, requiring many forward passes per pixel, though recent work has improved on this 
(e.g., \cite{garbin2021fastnerf,hedman2021baking,reiser2021kilonerf,yu2021plenoctrees,aumentado2023reconstructive,wan2023learning,kerbl20233d}).
Furthermore, the distributed nature of the density makes extracting explicit geometric details 
(including higher-order surface information, such as normals and curvatures) more difficult 
(e.g., \cite{yariv2021volume,oechsle2021unisurf}).
In contrast, though focused purely on geometry, DDFs are rendered with a single forward pass and enable non-local computation of higher-order surface properties.
Further, by noting that light paths between surfaces correspond to recursive DDF calls,
we show that DDFs are amenable to path-tracing (see \S\ref{sec:pathtracing}),
whereas NeRFs require substantial modification 
(e.g., \cite{zhang2021nerfactor}).

More similar to DDFs are Light Field Networks (LFNs) \cite{sitzmann2021light},
which render with a single forward pass per pixel, 
and permit sparse depth map extraction (assuming a Lambertian scene).
Unlike LFNs, DDFs model geometry rather than radiance as the primary quantity, 
computing depth with a single forward pass, and surface normals with a single backward pass,
while LFNs predict RGB and sparse depth from such a forward-backward operation.
Finally, %
LFNs cannot render
from viewpoints between occluded objects.
The neural 4D light field (NeuLF) \cite{li2021neulf} focuses on NVS and
leverages per-ray depth estimates, but is also only 4D, like LFNs.
More recently, Attal et al.~\cite{attal2022learning} improve on efficient NVS via a ray-space embedding within the light field.
In this work, we 
instead focus on 
(i) the representation of geometry and 
(ii) theoretical analysis of the conditions under which 
	higher-dimensional shape fields can ensure multiview consistency.

Several ray-based shape representations have appeared concurrently or subsequently 
	to our prior work \cite{aumentado2022representing}.
The Signed Directional Distance Field (SDDF) \cite{zobeidi2021deep} 
also maps position and direction to depth, but
introduces a fundamental difference 
in structure modelling, due to its signed nature.
Starting from a point $p$, consider a ray that intersects with a wall; 
evaluating a DDF at a point after the intersection provides 
the distance to the \textit{next} object, 
	while the SDDF continues to measure the signed distance to the wall.
This reduces complexity and dimensionality, 
but may limit representational utility for some tasks and/or shapes, 
including a lack of internal structure, in a manner similar to LFNs.
Houchens et al.~\cite{houchens2022neuralodf} %
showcase the utility of DDFs (called neural omnidirectional distance fields)
in inter-converting between a variety of shape representations.
The Primary Ray-based Implicit Function (PRIF) \cite{feng2022prif} representation also suggested the use of DDFs, using a novel ray parameterization, on several tasks; 
however, it did not investigate the theoretical requirements of exact representation, as we consider in this work, nor certain other issues (e.g., discontinuity modelling, relations to light transport). 

Several other new representations and models are closely related to DDFs.
Neural vector fields (NVFs) \cite{yang2023neural} compute the displacement from any given query to the \textit{closest} surface point. 
Like UDFs, this representation is closely connected to DDFs, in that
it corresponds to fixing a special direction field, 
which we call the Minimal Direction Field (MDF),
$v^* : \real^3\rightarrow\mathbb{S}^2$, 
which points to the closest surface point from $p$.
The NVF can then be computed via 
$\mathrm{NVF}(p) = d(p,v^*(p))v^*(p)$, for a given DDF $d$.
Note also that $\mathrm{UDF}(p) = d(p,v^*(p))$; see \S\ref{sec:app:udfextract} for details.
In terms of other tasks,
	Directed Ray Distance Functions \cite{kulkarni2022directed} %
	model surfaces via per-ray depth functions and their zeroes,
	to obtain scene representations from posed RGBD images.
RayMVSNet \cite{xi2022raymvsnet} %
	learns to generate per-ray depths, 
	but in a multiview stereo setting, based on learned features from images.
In a robotics context,
	Zhang et al.~\cite{zhang2024ddf} apply DDFs to modelling hand-object interaction.
Finally, the FIRe \cite{yenamandra2024fire} model combines an SDF with a DDF,
	in order to obtain efficiency while maintaining multiview consistency.

In contrast to these works, the primary novel contribution of this paper is the
	comprehensive characterization of theoretical requirements for view consistency 
	of DDFs (\S\ref{sec:theory}).
Since the aforementioned works %
	largely describe the same field mathematically,
	our results are equally applicable to them as well,
	and may suggest regularizations and architectural designs with 
	better theoretical properties.

%% file: ddfs.tex
\section{Directed Distance Fields}

\label{sec:ddf}

Let $S \subset \mathcal{B}$ be a 3D shape, 
where $\mathcal{B} \subset \mathbb{R}^3$ is a bounding volume 
acting as the field domain.
Consider a position $p\in\mathcal{B}$ and view direction $v\in\mathbb{S}^2$.
We define $S$ to be \textit{visible} from $(p,v)$ 
if the line $\ell_{p,v}(t) = p + tv$ 
intersects $S$ for some $t \geq 0$.

\noindent$\bullet\,$\textbf{Visibility.}
We write the binary visibility field (VF) for $S$ as
$\xi(p,v) = \mathds{1}[S\text{ is visible from }(p,v)]$. 
For convenience, we refer to an oriented point $(p,v)$ as visible if $\xi(p,v)=1$.

\noindent$\bullet\,$\textbf{Depth.}
We then define a distance or depth field as a non-negative scalar field $d : \mathcal{B} \times \mathbb{S}^2 \rightarrow \mathbb{R}_+$,
which maps from any visible 
oriented point
to the minimum distance from $p$ to $S$ along $v$ (i.e., the first intersection of $\ell_{p,v}(t)$ with $S$).
In other words,
$q(p,v) = d(p,v) v + p$ 
is a map to the shape, and thus
satisfies $q(p,v) \in S$ for visible $(p,v)$ (i.e., $\xi(p,v) = 1$).

\noindent$\bullet\,$\textbf{DDF.}
A \textit{Directed Distance Field} (DDF) is simply a 
tuple of fields $(\xi,d)$.
See Figs.~\ref{fig:armadillo} and \ref{fig:teaser} for illustrations.

\begin{figure}
	\centering
	\includegraphics[width=0.19\textwidth]{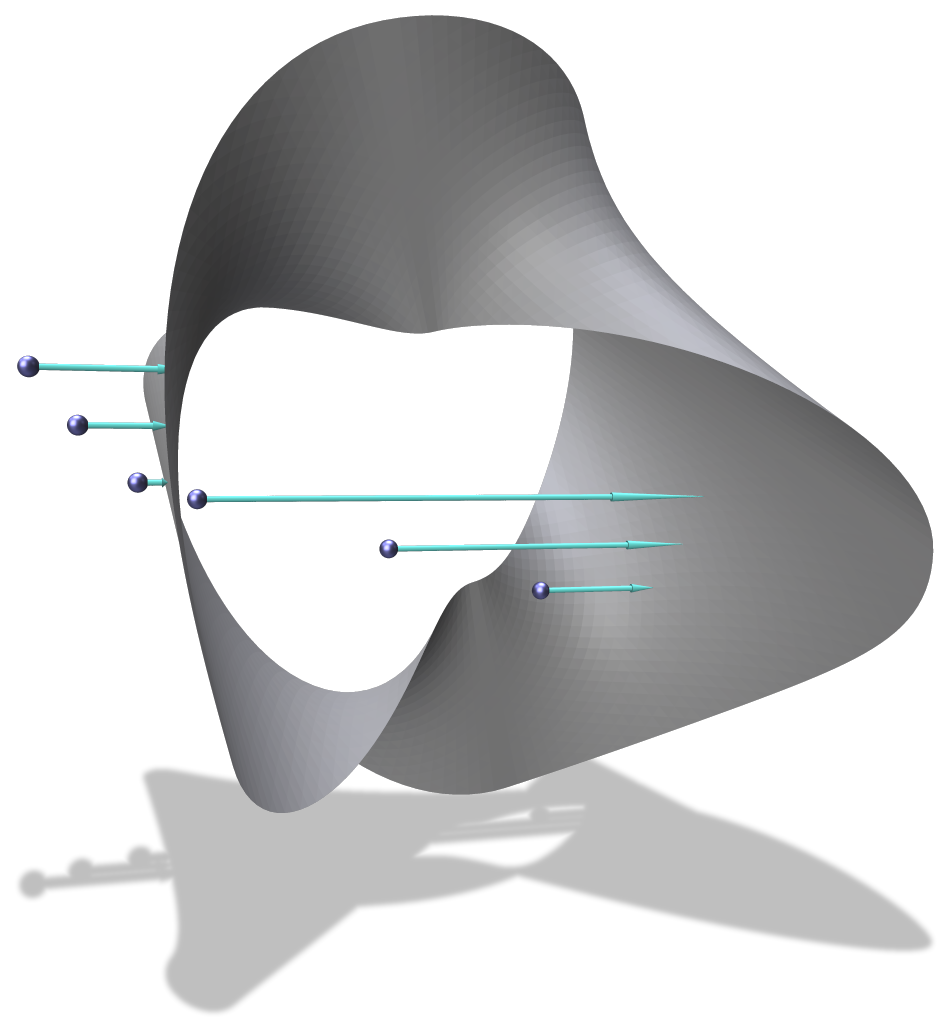}\rulesep
	\includegraphics[width=0.195\textwidth]{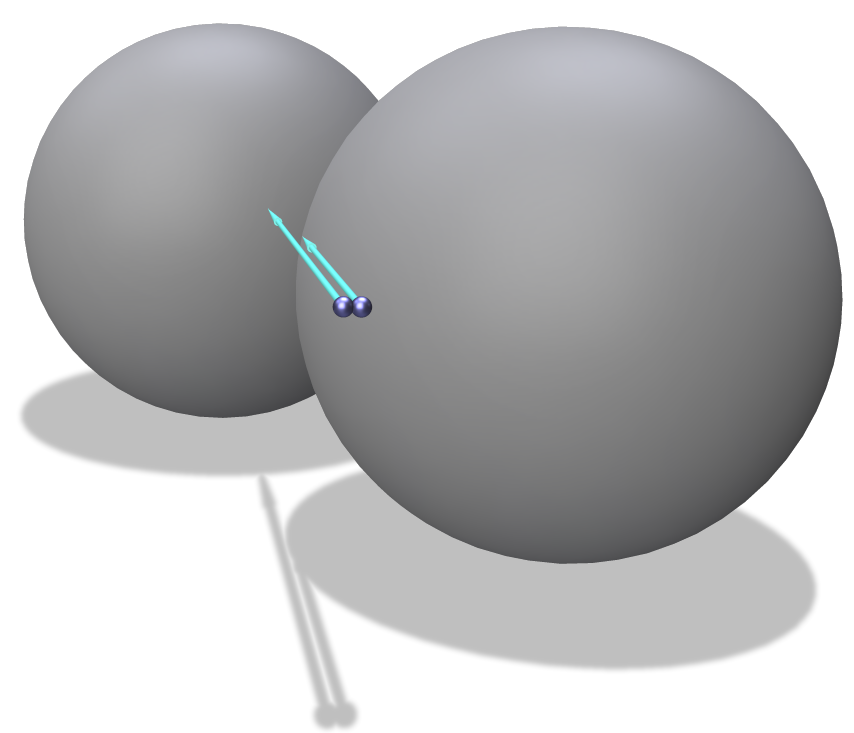}
	{%
		\setlength{\fboxsep}{0pt}%
		\setlength{\fboxrule}{0.5pt}%
		\raisebox{0.25\height}{%
			\fbox{\includegraphics[width=0.04\textwidth]{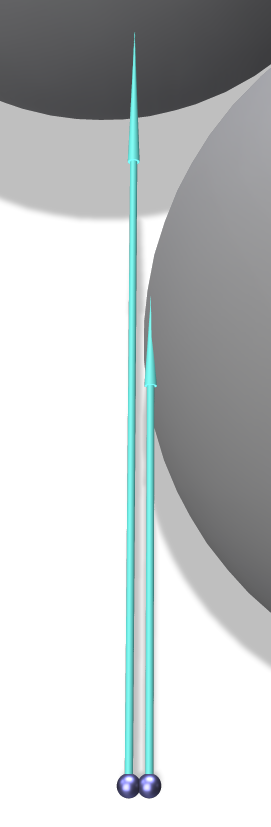}}%
		}%
	}%
	\caption{
		Inherent DDF discontinuities.
		Left: \textit{surface discontinuities}, where $p$ passes through $S$. Right: \textit{occlusion discontinuities}, where $v$ or $p$ moves over an occlusion boundary. 
	}
	\label{fig:discontillus}
\end{figure}

\subsection{Geometric Properties}
\label{sec:ddf:properties}

DDFs satisfy several useful properties
(see \S\ts{\ref{appendix:proofs}}{A} for proofs).

\phantomsection
\label{property1}
\noindent$\bullet\,$\textbf{Property I: Directed Eikonal Equation.}
Similar to SDFs,
which satisfy the eikonal equation 
$||\nabla_p \text{SDF}(p)||_2 = 1$, 
a DDF enforces a directed version of this property.
In particular, 
for any visible $(p,v)$,
we have 
$ \nabla_p d(p,v) v = -1 $, with $\nabla_p d(p,v)\in\mathbb{R}^{1\times 3}$.
This also implies $||\nabla_p d(p,v)||_2 \geq 1$.
The equivalent property for $\xi$ asks that 
locally moving along the viewing line cannot change visibility\footnote{Except when moving \textit{through} certain surfaces (see \S\ref{sec:theory}).}:
$ \nabla_p \xi(p,v) v = 0 $.

\phantomsection
\label{property2}
\noindent$\bullet\,$\textbf{Property II: Surface Normals.}
The derivatives of implicit fields are closely related to the normals $n \in\mathbb{S}^2$ 
of $S$;
e.g., $\nabla_q \text{SDF}(q)^T = n(q)$ $\forall$ $q\in S$.
For DDFs, a similar relation holds (\textit{without} requiring $p\in S$):
$ %
\nabla_p d(p,v) = {-n(p,v)^T} / ({n(p,v)^T v}),
$ %
for any visible $ (p,v) $ such that
$n(p,v) := n(q(p,v))$
are the normals at
$q(p,v)\in S$
and $n(p,v) \not\perp v$ 
(i.e., the change in $d$ moving \textit{off} the surface is undefined).
From any $(p,v)$ that ``looks at'' $q\in S$, 
the normals at $q$
can be computed via
$ n(p,v) = 
\varsigma \nabla_p d(p,v)^T / ||\nabla_p d(p,v)||_2 $,
where we choose $\varsigma \in \{-1,1\}$
such that $n^Tv < 0$ 
(so that $n$ always points back to the query)\footnote{This defines the normal via $v$, even for non-orientable surfaces.}.

\phantomsection
\label{property3}
\noindent$\bullet\,$\textbf{Property III: Gradient Consistency.}
Intuitively, given a visible $(p,v)$, infinitesimally perturbing the \textit{viewpoint} $v$ by $\delta_v$ should be similar to pushing the \textit{position} $p$ along $\delta_v$.
In fact, the following differential constraint 
holds
(see Supp.~\S\ts{\ref{appendix:proofs:prop3}}{A-C}%
):
	$\nabla_v d(p,v) = d(p,v) \nabla_p d(p,v) \mathcal{P}_v,$
where $\mathcal{P}_v = I - vv^T$  is an orthogonal projection, 
	since $v$ cannot be perturbed along itself. %

\phantomsection
\label{property4}
\noindent$\bullet\,$\textbf{Property IV: Deriving Unsigned Distance Fields.}
An unsigned distance field (UDF) can be extracted from a DDF via the following optimization:
$ \text{UDF}(p) = \min_{v \in \mathbb{S}^2} d(p,v) $, constrained such that $\xi(p,v) = 1$, allowing them to be procured if needed (see \S\ref{sec:app:udfextract}).
UDFs remove the discontinuities from DDFs (see \S\ref{sec:app:pddfs} and Fig.\ \ref{fig:discontillus}), but cannot be rendered as easily nor be queried for distances in arbitrary directions.

\phantomsection
\label{property5}
\noindent$\bullet\,$\textbf{Property V: Local Differential Geometry.}
For any visible $(p,v)$, the geometry of a 2D manifold $S$ near $q(p,v)$ is completely characterized by $d(p,v)$ and its derivatives. 
In particular,
we can estimate the first and second fundamental forms 
using the gradient and Hessian of $d(p,v)$ 
(see Supp.~\S\ts{\ref{appendix:proofs:prop5}}{A-D}).
This allows computing surface properties, 
such as curvatures,
from any visible oriented position, simply by querying the network; 
see Fig.\ \ref{fig:singobjfits}
for an example.

\noindent$\bullet\,$\textbf{Neural Geometry Rendering.} 
Differentiable generation of geometry, 
such as depth and normals 
(e.g., \cite{yan2016perspective,wu2017marrnet,tulsiani2017multi,nguyen2018rendernet}),
can sometimes be written as parallelized DDFs.
In such cases, regardless of architecture,  
the properties of DDFs discussed in this paper still hold
(see also Supp.~\S\ts{\ref{appendix:neurren}}{A-E}).

\subsection{The View Consistency Inequality}
\label{pddf:appendix:viewconsis}
{
	Ideally, DDFs should 
	maintain
	\textit{view} \textit{consistency} (VC).
	One form of VC can be expressed by a simple inequality, which
	demands that an \textit{opaque} position viewed from one direction 
	(e.g., the point $q_1$, from $v_1$, in Fig.~\ref{pddf:vcv1})
	must be opaque from all directions.
	That is, when a point $q_1$ is opaque from one direction,
	but is also on another directed ray from, say, $\tau_2 = (p_2,v_2)$, then
	its depth (from $\tau_2$) is lower-bounded by the distance to that known surface position.
	
	More specifically, 
	consider two oriented points $\tau_1 = (p_1,v_1)$ and $\tau_2 = (p_2,v_2)$.
	Assume that
	$\tau_1$ is visible, 
	so $q_1 = p_1 + d(\tau_1)v_1 \in S$,
	and 
	$\exists\;t>0$ 
	such that $\ell_{\tau_2}(t) = p_2 + t v_2 = q_1$.
	\setlength{\intextsep}{3.1pt}%
\setlength{\columnsep}{14.0pt}%
\begin{wrapfigure}{l}{0.38\linewidth}
	\begin{center}
		\includegraphics[width=0.89\linewidth]{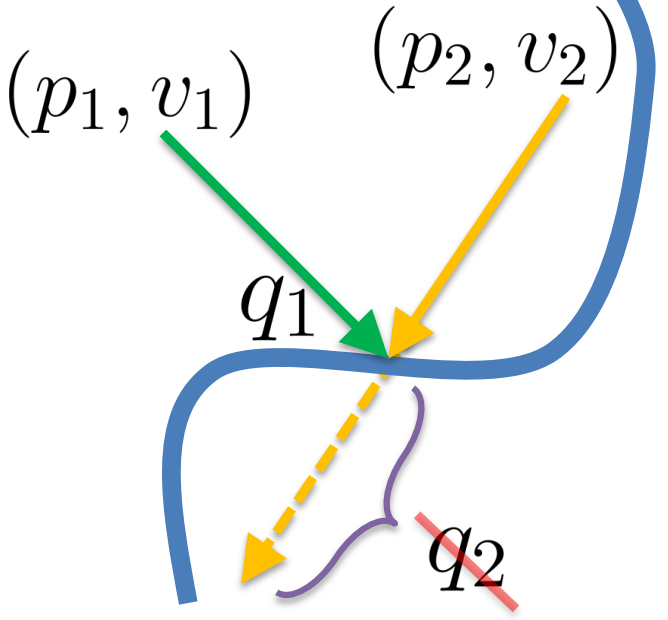}
	\end{center}
	\caption[Depiction of View Consistency Inequality]{
		The view consistency inequality,
		$ d(p_2,v_2) \leq ||p_2 - q_1||_2 $. 
	}\label{pddf:vcv1}
\end{wrapfigure}%
	I.e.,
	the lines of sight from $\tau_1$ and $\tau_2$ 
	intersect at the surface point $q_1$. 
	Then,  
	(i) %
	$\xi(\tau_2) = 1$,
	and
	(ii) $d(\tau_2) \leq ||p_2 - q_1||_2 = t$.
	These can be seen by the definition of $\xi$ and $d$.
	For (i), since there exists a surface (at $q_1$) along $\ell_{\tau_2}$, the oriented point $(\tau_2)$ must be visible. 
	For (ii), $d(\tau_2)$ can be no further than $t$, since DDFs return the minimum distance  to a point on $S$ along the line $\ell_{\tau_2}(t)$, and hence its output can be no greater than the assumed distance $t$. %
	In \S\ref{sec:theory}, we examine the question of view consistency in greater detail.
	
}

\subsection{Rendering} 
\label{sec:app:rendering}

A primary application of DDFs is rapid differentiable rendering.
In contrast to some mesh rasterizers (e.g., \cite{liu2019soft}), there is no dependence on the complexity of the underlying shape, after training.
Unlike most implicit shape fields \cite{liu2020dist,sitzmann2019scene,mildenhall2020nerf}, 
DDFs only require a single forward pass per pixel.

DDF rendering is simply ray-casting.
Given a camera $\Pi$ with position $p_0\in\mathcal{B}$, for a pixel with 3D position $\rho$, we cast a ray $r(t) = p_0 + t v_\rho $, where $v_\rho = (\rho - p_0)/||\rho - p_0||_2$, into the scene via a single query $d(p_0, v_\rho)$, which provides the depth pixel value.
For $p\notin \mathcal{B}$,
we first compute the intersection $p_r\in\partial\mathcal{B}$ between the ray $r$ and the boundary $\partial \mathcal{B}$. 
We then use $d(p_r,v) + ||p - p_r||_2$ as the output depth (or set $\xi(p_r,v)=0$ if no intersection exists).
This allows querying $(\xi,d)$ from arbitrary oriented points, including those unseen during fitting.

\newcommand{\ctwa}{0.19\textwidth}
\newcommand{\ctwb}{0.159\textwidth}
\begin{figure}
	\centering
	\includegraphics[width=0.49\textwidth]{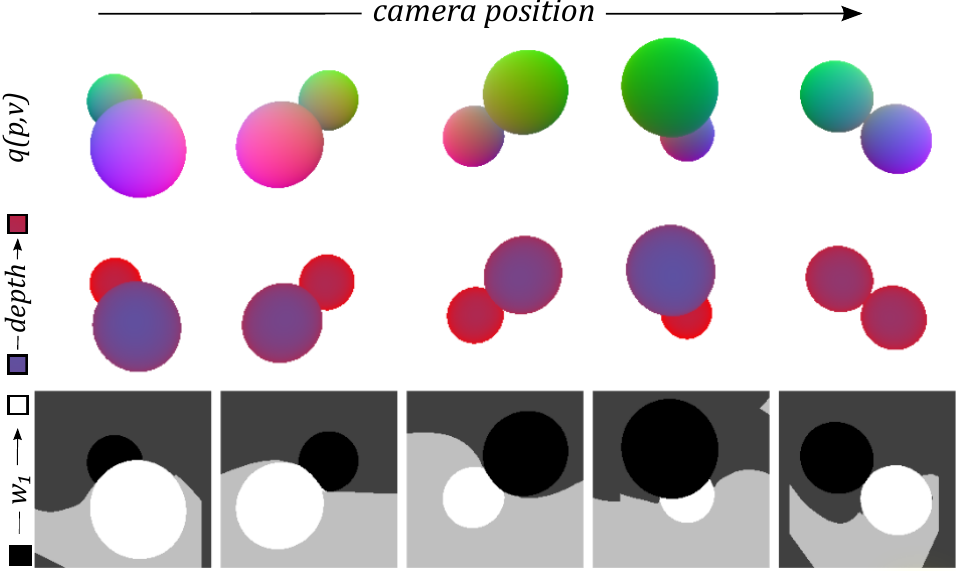} \\
	\includegraphics[width=0.49\textwidth]{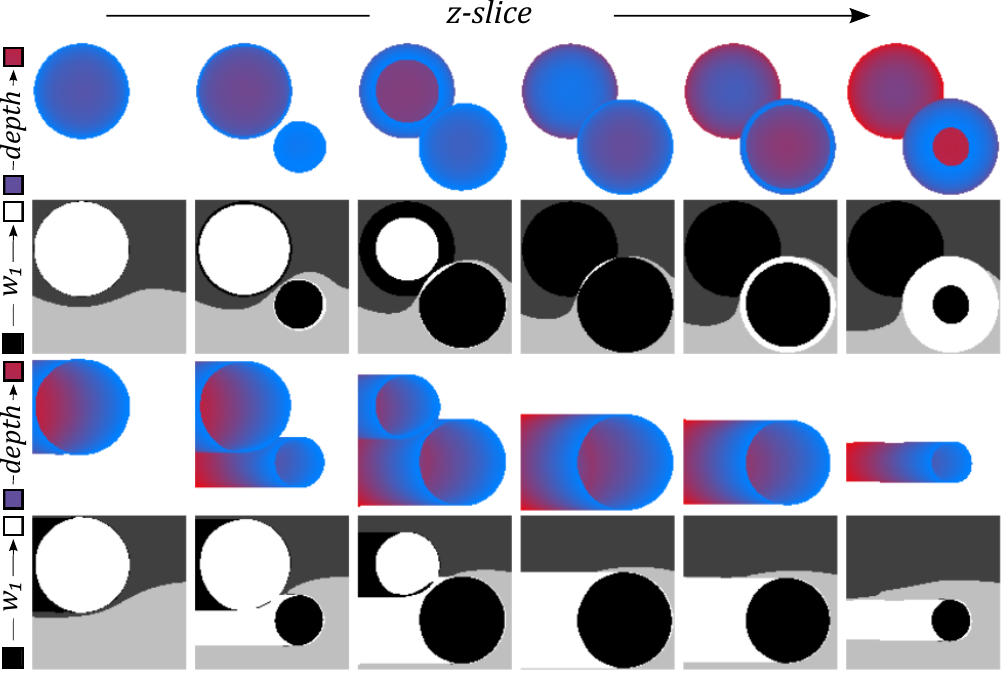}
	\caption{
		Discontinuous depth with PDDFs, 
			via the weight field (WF).
		Here, $K = 2$, so $w_1 = 1 - w_2$ (see \S\ref{sec:app:pddfs}).
		The \textit{upper inset} (rows 1--3) shows
			WF transitions in renders. 
		In the third row, white vs black mark high vs low $w_1$,
		and thus which surface ($d_1$ vs $d_2$) is active,
		for high $\xi$.
		Light and dark grey demarcate the non-visible (low $\xi$) counterparts of white and black. 
		The change in dominant weight ($w_1$ vs $w_2$) at occlusion edges permits discontinuities.
		The \textit{lower inset} 
		(rows 4--7) shows 
		WF transitions using slices in $z$. 
		Rows four and six depict distances, with fixed $v$ 
		($(0,0,-1)$ and $(1,0,0)$, respectively) and varying $p$ across the image. 
		Rows five and seven show WF values, 
		as in row three.
		Notice the WF switching upon $p$ transitioning through a surface.
	}
	\label{fig:prob_illus}
\end{figure}

\subsection{Discontinuity Handling: Probabilistic DDFs}
\label{sec:app:pddfs}

\newcommand{\rmtl}{0.071}
\newcommand{\rmtll}{0.066}
\begin{figure}
	\centering
	\includegraphics[width=\linewidth]{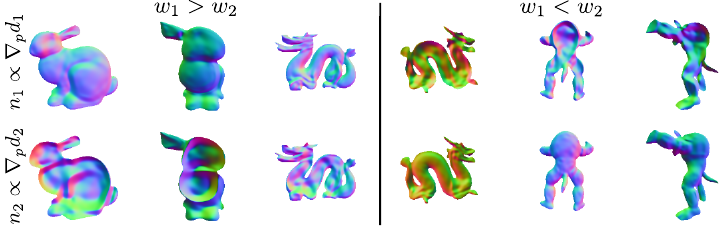}
	\caption{
		PDDF renders of $n_1$ and $n_2$.
		Though not explicitly enforced, a ``see-through effect'' occurs when the lower-weight field models the surface \textit{behind} the currently visible one
		(i.e., the PDDF components model separate surfaces).
	}
	\label{fig:seethrough}
\end{figure}

DDFs are inherently discontinuous functions of $p$ and $v$. 
Whenever 
(i) $p$ passes through the surface $S$ or 
(ii) $p$ or $v$ moves across an occlusion boundary, 
a discontinuity in $d(p,v)$ occurs
(see Fig.\ \ref{fig:discontillus}).
We therefore modify the DDF, to allow a $C^1$ 
network to represent the discontinuous field.
In particular, we alter $d$ to output probability distributions over depths, rather than a single value.
Let $\mathbb{P}_\ell$ be the set of probability distributions with support on some ray $\ell_{p,v}(t) = p+tv,\, t\geq 0$.
Then $d : \mathcal{B} \times \mathbb{S}^2 \rightarrow \mathbb{P}_\ell$ is a \textit{probabilistic DDF} (PDDF). 
The visibility field, $\xi(p,v)$, is unchanged in the PDDF.

For simplicity, herein we restrict $\mathbb{P}_\ell$ to be the set of mixtures of Dirac delta functions with $K$ components.
Thus, the network output is %
$
P_{p,v}(d) = %
\sum_i w_i \delta(d - d_i)
$ 
over depths,
where $w_i$'s are the mixture weights, with $\sum_i w_i = 1$, and $d_i$'s are the delta locations.
Our output depth is then $d_{i^*}$, where
$i^* = \argmax_i w_i$; i.e., the highest weight delta function marks the final output location.
As $w_i$ changes continuously, $w_{i^*}$ will switch from one $d_i$ to another $d_j$, which may be arbitrarily far apart, resulting in a discontinuous jump.
Thus, by having the weight field $w(p,v)$ smoothly \textit{transition} from one index ${i^*}$ to another, at the site of a surface or occlusion discontinuity, we can obtain a discontinuity in $d$ as desired.
In this work, we use $K=2$, to represent discontinuities without sacrificing efficiency.
Fig.\ \ref{fig:prob_illus} showcases example transitions, with respect to 
occlusion (a) and surface collision (b) discontinuities; 
Fig.\ \ref{fig:seethrough} visualizes the normals field for each depth component.
Notationally, a PDDF is simply a DDF: $d(p,v) := d_{i^*}(p,v)$.

\newcommand{\trw}{0.14\textwidth}
\newcommand{\trr}{0.88in}

\begin{figure*}[ht]
	\centering
	\includegraphics[clip,trim={0.0cm 0.0cm 0.0cm 0.0cm},width=0.11\textwidth]{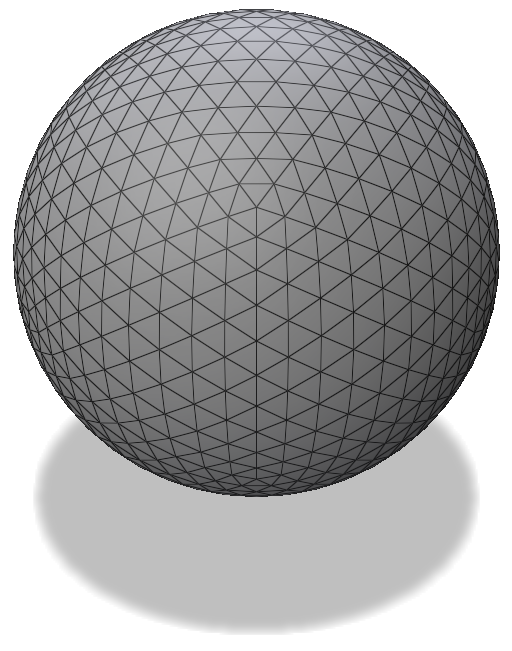} \hfill
	\includegraphics[clip,trim={12.2cm 1cm 12.2cm 1cm},width=\trw]{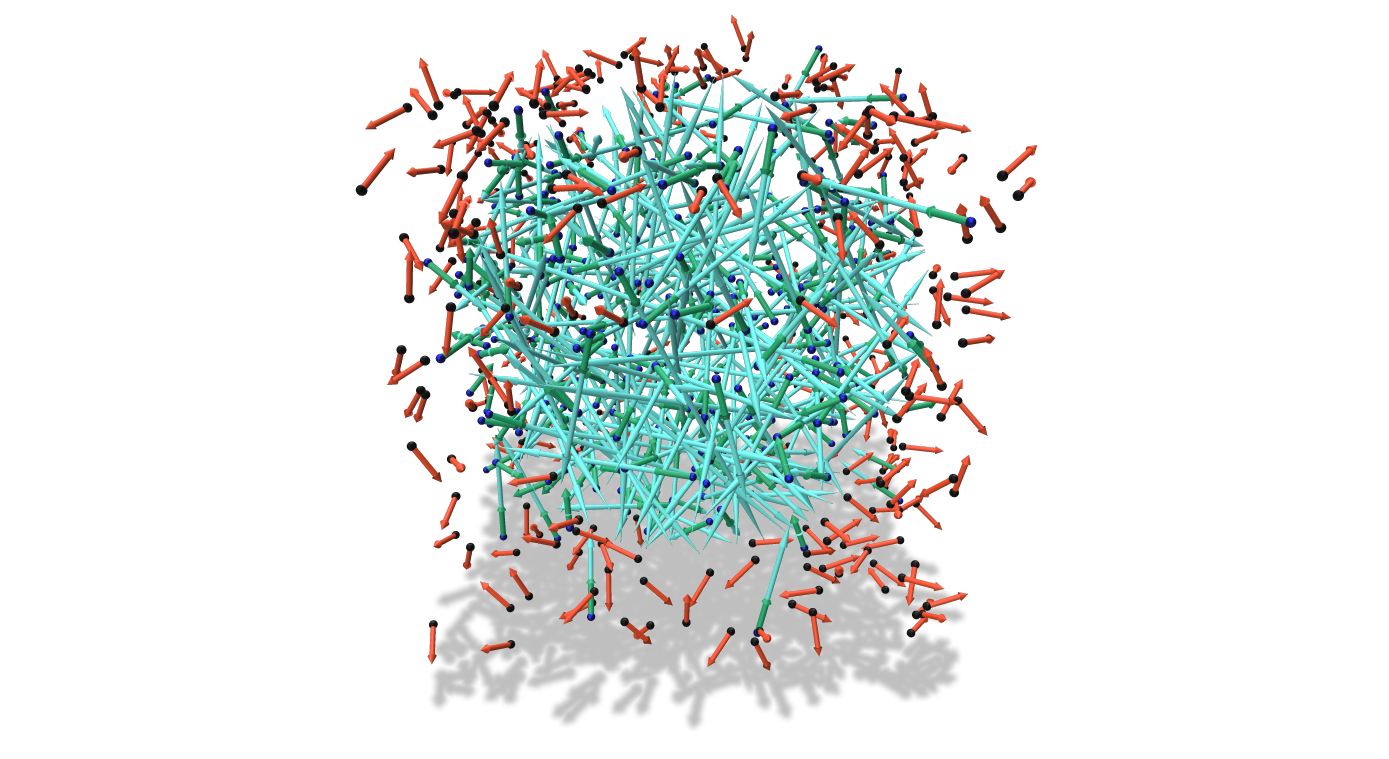} \hfill
	\includegraphics[clip,trim={12.2cm 1cm 12.2cm 1cm},width=\trw]{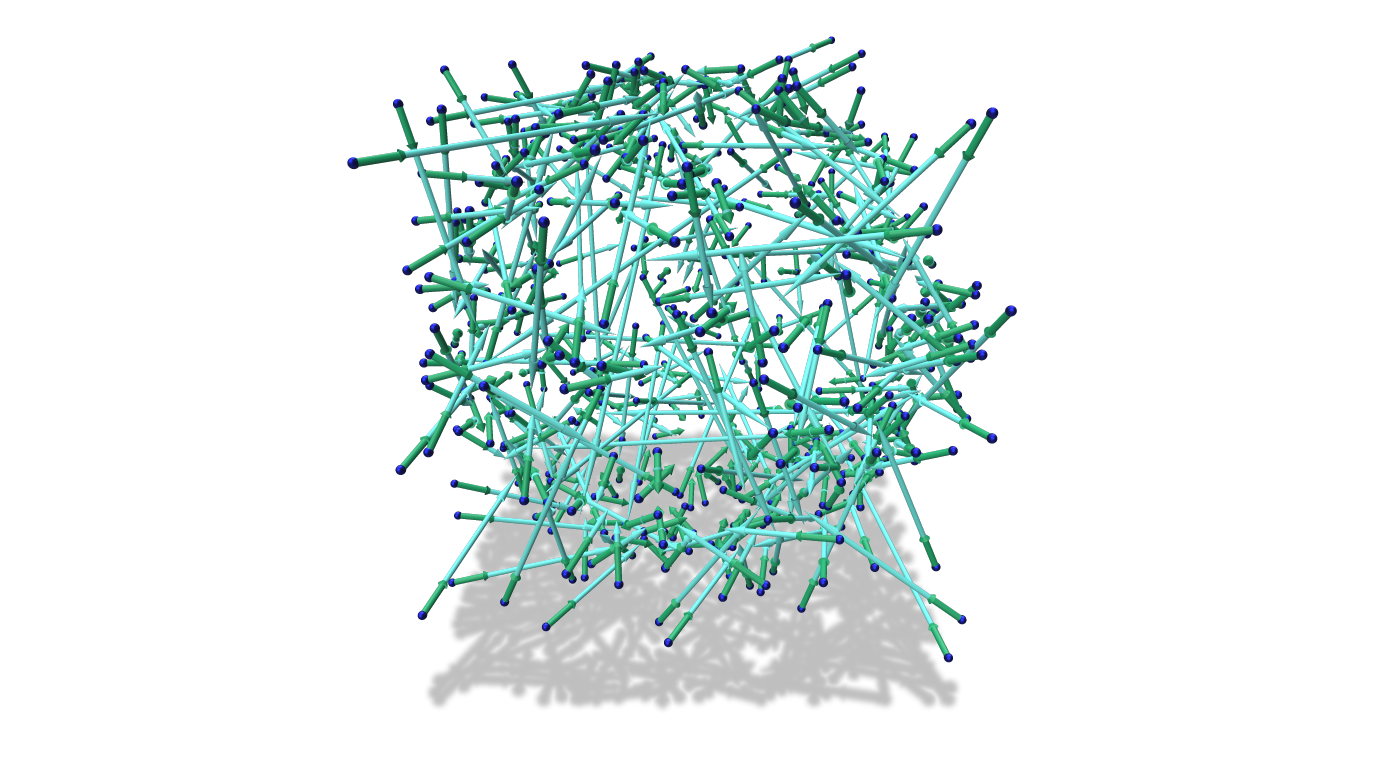} \hfill
	\includegraphics[clip,trim={12.2cm 1cm 12.2cm 1cm},width=\trw]{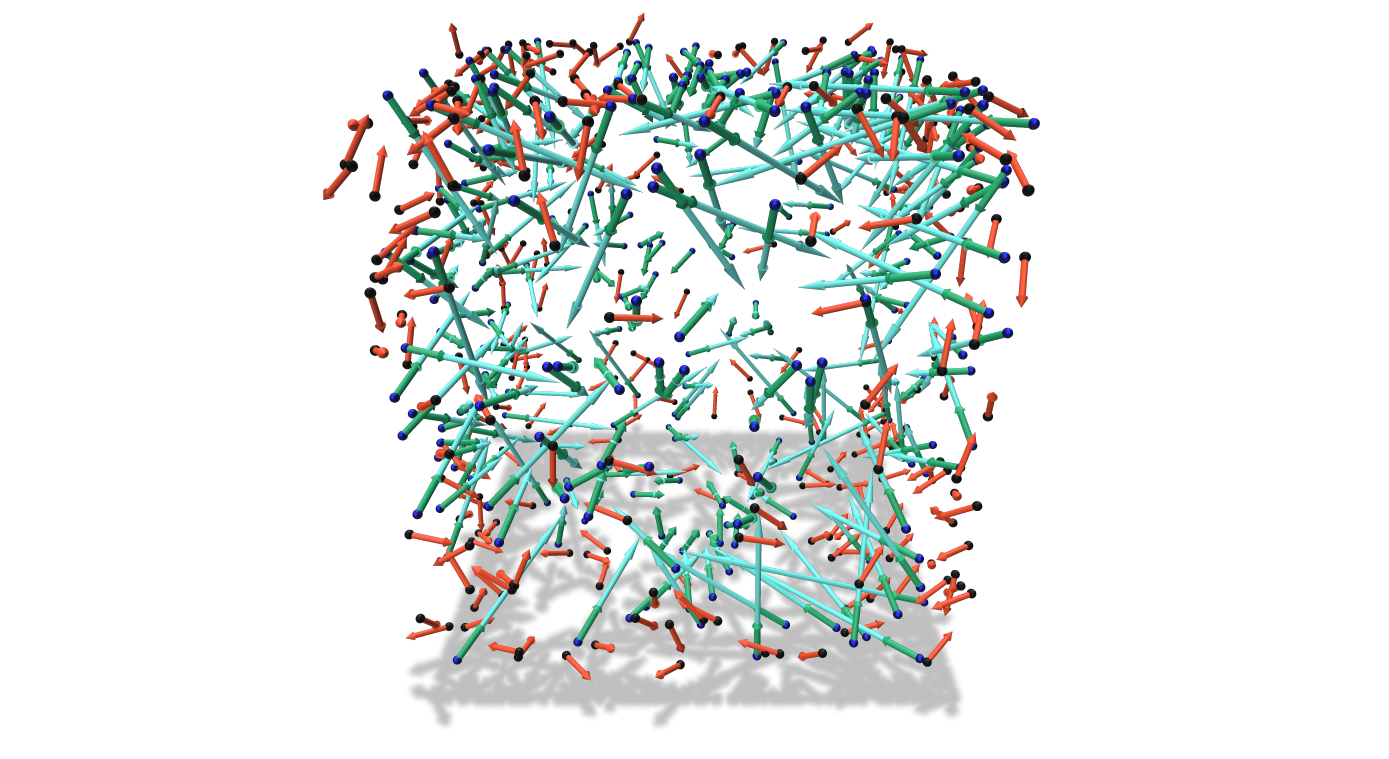} \hfill
	\includegraphics[clip,trim={12.2cm 1cm 12.2cm 1cm},width=\trw]{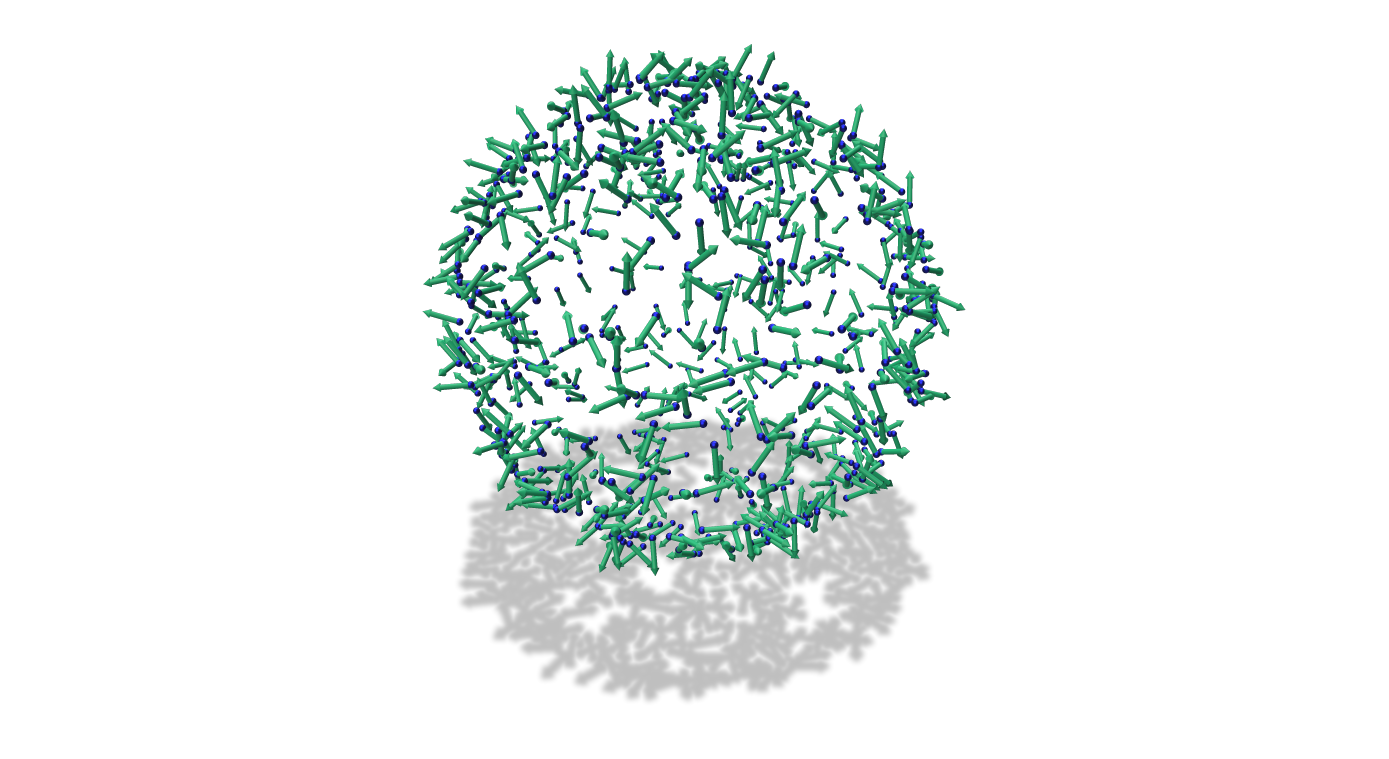} \hfill
	\includegraphics[clip,trim={12.2cm 1cm 12.2cm 1cm},width=\trw]{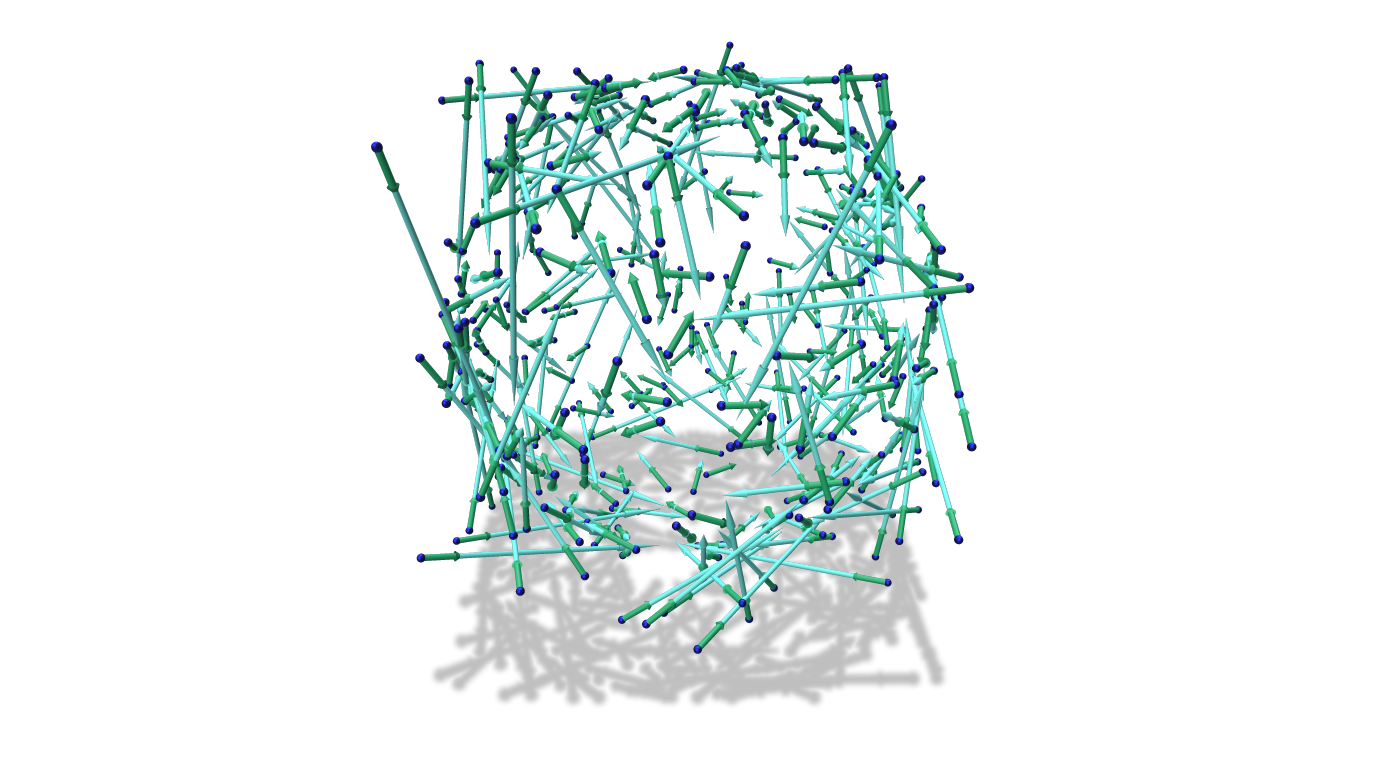} \hfill
	\includegraphics[clip,trim={12.2cm 1cm 12.2cm 1cm},width=\trw]{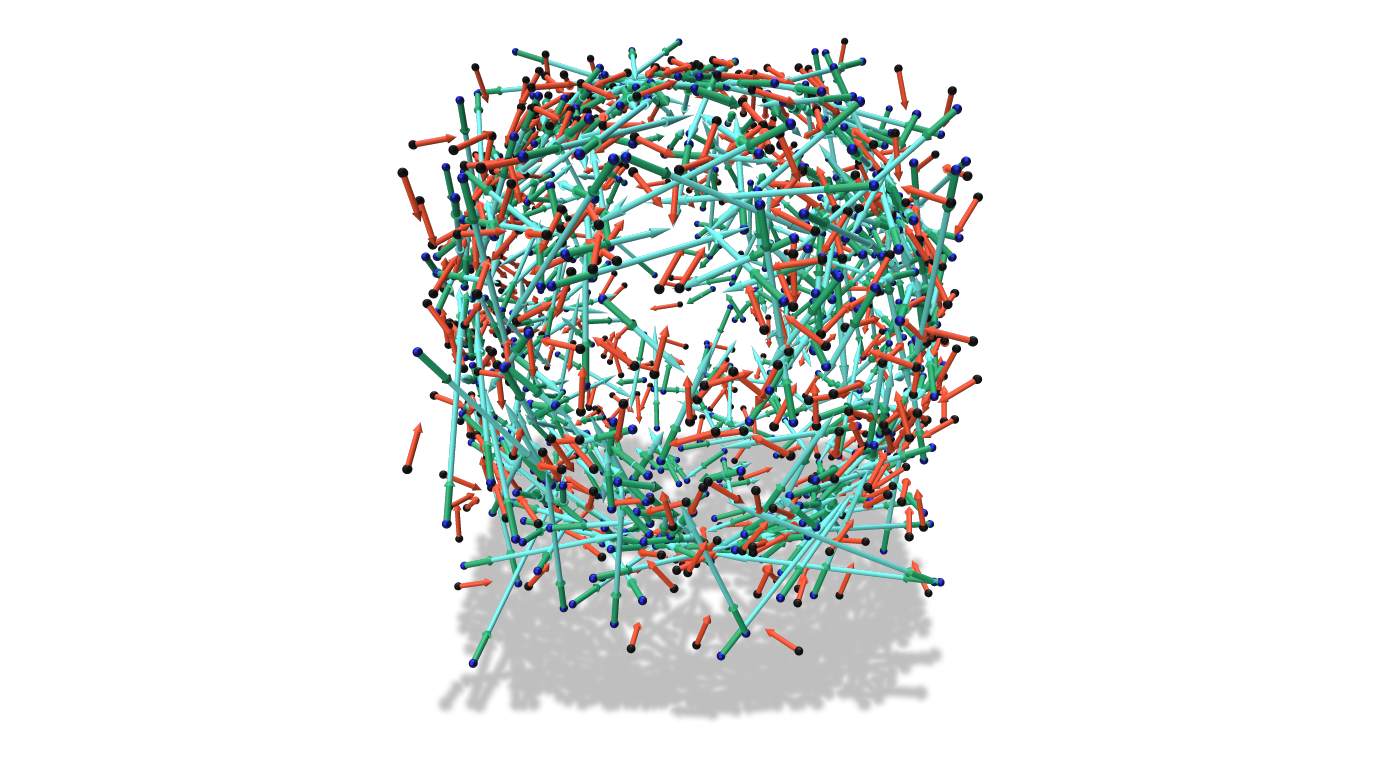}
	\makebox[0pt][r]{%
		\begin{tikzpicture}[overlay]%
			\node[xshift=0cm] (hidden) at (0,0) {};
			\node[xshift=0cm,yshift=0cm] (M) at (-16.50,0.05) {\footnotesize Mesh};
			\node[xshift=0cm,yshift=0cm] (U) [right=0.8in of M] {\footnotesize U};
			\node[xshift=0cm,yshift=0cm] (A) [right=\trr of U] {\footnotesize A};
			\node[xshift=0cm,yshift=0cm] (B) [right=\trr of A] {\footnotesize B};
			\node[xshift=0cm,yshift=0cm] (S) [right=\trr of B] {\footnotesize S};
			\node[xshift=0cm,yshift=0cm] (T) [right=\trr of S] {\footnotesize T};
			\node[xshift=0cm,yshift=0cm] (O) [right=\trr of T] {\footnotesize O};
		\end{tikzpicture}%
	}%
	\caption{Illustration of mesh-derived data types.
		Left to right: input sphere mesh, U, A, B, S, T, and O data. Visible points depict $p$ in blue, $v$ in green, and a line from $p$ to $q$ in turquoise; non-visible points depict $p$ in black and $v$ in red. 
	}
	\label{fig:datatypes}
\end{figure*}
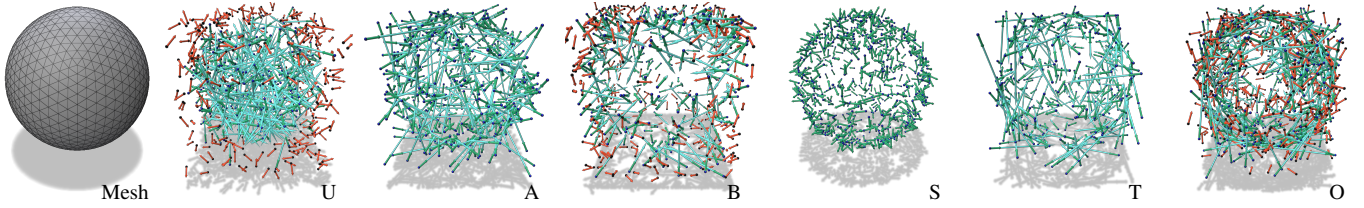

\newcommand{\ctwc}{0.077\textwidth}
\begin{figure*}[ht]
	\centering
	\includegraphics[width=\ctwc]{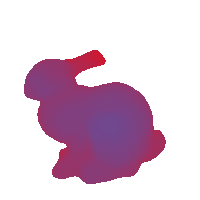}
	\includegraphics[width=\ctwc]{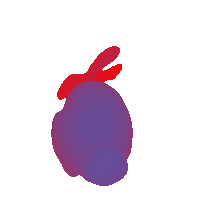}
	\includegraphics[width=\ctwc]{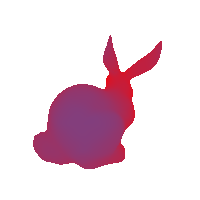}
	\includegraphics[width=\ctwc]{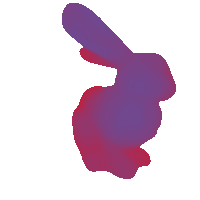} 
	\includegraphics[width=\ctwc]{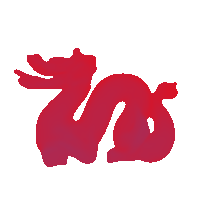}
	\includegraphics[width=\ctwc]{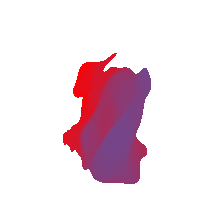}
	\includegraphics[width=\ctwc]{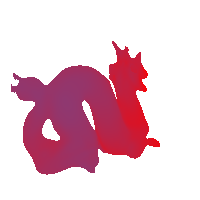}
	\includegraphics[width=\ctwc]{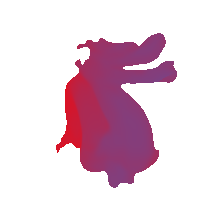}
	\includegraphics[width=\ctwc]{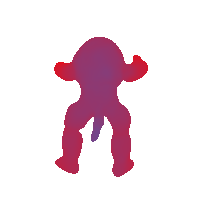}
	\includegraphics[width=\ctwc]{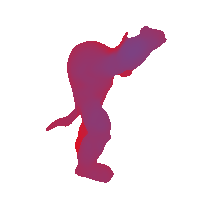}
	\includegraphics[width=\ctwc]{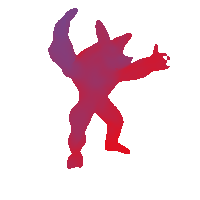}
	\includegraphics[width=\ctwc]{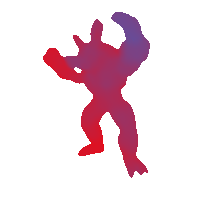}
	\includegraphics[width=\ctwc]{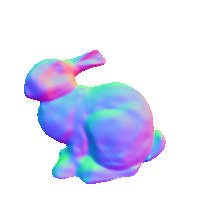}
	\includegraphics[width=\ctwc]{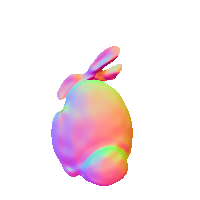}
	\includegraphics[width=\ctwc]{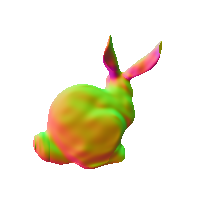}
	\includegraphics[width=\ctwc]{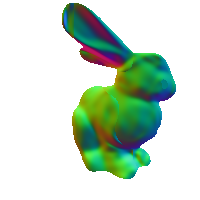} 
	\includegraphics[width=\ctwc]{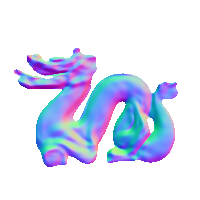}
	\includegraphics[width=\ctwc]{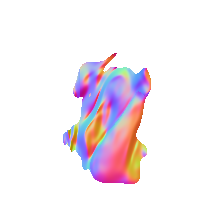}
	\includegraphics[width=\ctwc]{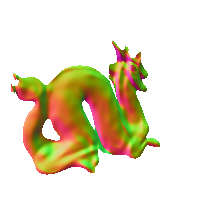}
	\includegraphics[width=\ctwc]{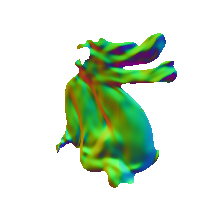}
	\includegraphics[width=\ctwc]{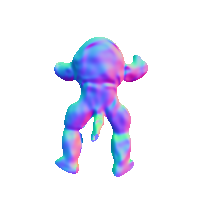}
	\includegraphics[width=\ctwc]{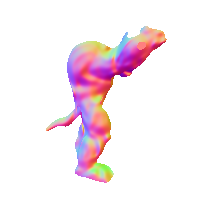}
	\includegraphics[width=\ctwc]{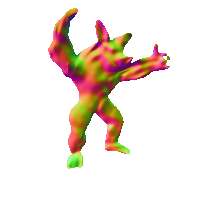}
	\includegraphics[width=\ctwc]{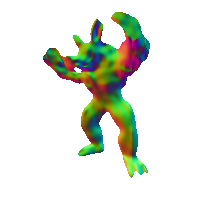}
	\includegraphics[width=\ctwc]{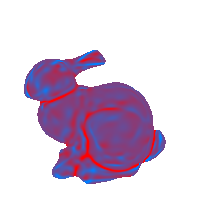}
	\includegraphics[width=\ctwc]{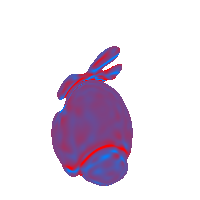}
	\includegraphics[width=\ctwc]{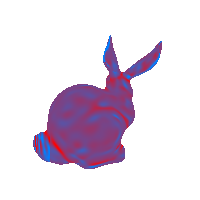}
	\includegraphics[width=\ctwc]{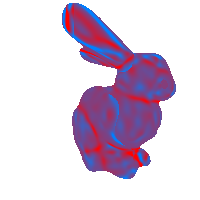} 
	\includegraphics[width=\ctwc]{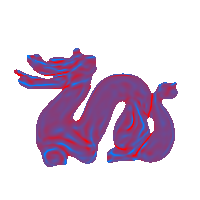}
	\includegraphics[width=\ctwc]{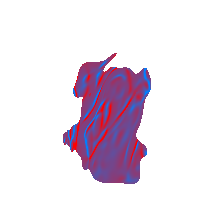}
	\includegraphics[width=\ctwc]{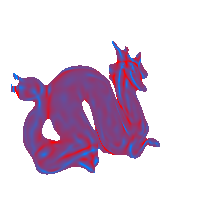}
	\includegraphics[width=\ctwc]{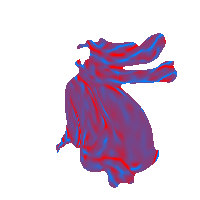}
	\includegraphics[width=\ctwc]{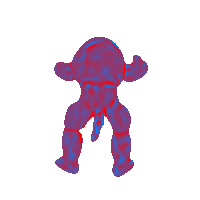}
	\includegraphics[width=\ctwc]{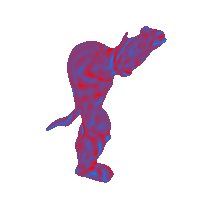}
	\includegraphics[width=\ctwc]{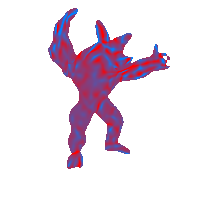}
	\includegraphics[width=\ctwc]{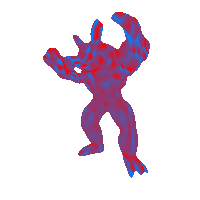}    
	\caption{
		Renders of DDF fits to shapes. %
		Rows: depth, normals, and mean curvature. %
		Each quantity is directly computed from the learned field, using network derivatives at the query oriented point $(p,v)$ per pixel. 
	}
	\label{fig:singobjfits}
\end{figure*}

\section{Learning DDFs}
\label{sec:app:learning}

\subsubsection{Mesh Data Extraction}
\label{pddf:app:sec:dataablation}

Given a mesh, %
we can obtain visibility $\xi$ and depth $d$
by ray-casting
from any %
$(p,v)$.
In total, we consider six types of data samples %
(see Fig.\ \ref{fig:datatypes}):
\textit{uniform} (U) random $(p,v)$; 
\textit{at-surface} (A), where 
$\xi(p,v)=1$;
\textit{bounding} (B), where $p\in\partial\mathcal{B}$ and $v$ points to the interior of $\mathcal{B}$;
\textit{surface} (S), where $p\in S$ and $v\sim\mathcal{U}[\mathbb{S}^2]$;
\textit{tangent} (T), 
where $v$ is in the tangent space of $q(p,v)\in S$;
and 
\textit{offset} (O), 
which offsets $p$ from T-samples
along
$n(p,v)$ by a small value. 

\begin{table*}
	\centering
	{
		\tabcolsep3.9pt
		\begin{tabular}{c|cccccc|cccccc}
			\multirow{2}{*}{} & 
			\multicolumn{6}{c|}{Minimum Distance Error ($L_1; \times 10$) $\downarrow$} &
			\multicolumn{6}{c}{Visibility Error (BCE) $\downarrow$} \\ 
			& $\mathcal{L}_{d,1}$-U & $\mathcal{L}_{d,1}$-A & $\mathcal{L}_{d,1}$-B & $\mathcal{L}_{d,1}$-O & $\mathcal{L}_{d,1}$-T & $\mathcal{L}_{d,1}$-S & %
			$\mathcal{L}_{\xi}$-U & $\mathcal{L}_{\xi}$-A & $\mathcal{L}_{\xi}$-B & $\mathcal{L}_{\xi}$-O & $\mathcal{L}_{\xi}$-T & $\mathcal{L}_{\xi}$-S \\\hline
			U & \tred{0.58} & 0.79 & 0.18 & 0.49 & 0.75 & 0.47 & 
			\tred{2.11} & 0.03 & 0.07 & \tpink{0.56} & 0.14 & 0.05
			\\
			A & 0.48 & \tpink{0.82} & \tpink{0.20} & 0.49 & 0.81 & 0.54 &  
			0.20 & 0.05 & 0.07 & 0.49 & \tpink{0.18} & \tpink{0.07}
			\\
			B & 0.37 & 0.67 & \tpink{0.20} & 0.46 & 0.73 & 0.58 &
			0.20 & 0.03 & \tpink{0.10} & 0.50 & 0.15 & 0.06
			\\
			O & 0.39 & 0.67 & 0.18 & \tred{0.56} & 0.70 & 0.59 &
			\tpink{0.28} & 0.01 & \tred{0.11} & \tred{1.71} & 0.03 & 0.03
			\\
			T & 0.39 & 0.70 & 0.19 & 0.45 & \tpink{0.84} & {0.65} & 
			0.19 & \tred{0.09} & 0.06 & 0.32 & \tred{0.48} & \tpink{0.07}
			\\
			S & \tpink{0.55} & \tred{0.95} & \tred{0.23} & \tpink{0.51} & \tred{0.89} & \tred{1.43} &
			0.14 & \tpink{0.06} & 0.06 & 0.42 & 0.17 & \tred{0.48}
			\\\hline
			-- & 0.45 & 0.75 & 0.19 & 0.50 & 0.77 & \tpink{0.67} &
			0.23 & 0.04 & 0.08 & \tpink{0.56} & 0.15 & \tpink{0.07}
		\end{tabular}
	}
	\caption[Data Type Ablations for DDF Fitting]{
		Data type ablation results
		(see \S\ref{datatypeablation}).
		Rows: type of data (i.e., $(p,v)$ sample type) ablated.
		Columns: errors on held-out data (left: min.\ distance loss, but computed with $L_1$ instead of $L_2$; right: binary cross-entropy-based visibility loss).
		Each column computes the error on a different sample type (e.g., $\mathcal{L}_{d,1}$-A is the minimum distance error on A-type data), via 25K held-out samples per type.
		Per loss type, the \tred{red} numbers are the \textit{worst} (highest) error cases; the \tpink{pink} numbers are the \textit{second-worst} error cases.
		Each scenario uses 100K samples of each data type, except for the ablated one;
		the ``--'' case uses 83,333 samples of all six data types 
		(to control for the total dataset size).
		Usually,
		performance on a data type is worst or second-worst when that type is ablated,
		suggesting
		the utility of different samples,
		particularly for hard types (e.g., S and T).
	}
	\label{pddf:tab:dataablation}
\end{table*}

\subsubsection{Loss Functions}
\label{sec:app:loss}

Our optimization objectives are defined per oriented point 
$(p,v)$. 
We denote $\xi$, $n$, and $d$ 
as the ground truth visibility, surface normal, and depth values, 
and let $\widehat{\xi}$, $\widehat{d}_i$, and $w_i$ denote the network predictions.
Recall $i^* = \arg\max_{j} w_j$ is the maximum likelihood PDDF index.

The \textit{minimum distance loss} 
trains the  
highest probability depth component:
$%
\mathcal{L}_d = \xi| \widehat{d}_{i^*} - d |^2.
$%
The \textit{visibility objective}, $L_\xi = \mathrm{BCE}(\xi,\widehat{\xi})$, is the binary cross entropy  between the visibility prediction and the ground truth.
A first-order \textit{normals loss}, %
$
\mathcal{L}_n = -\xi| n^T \widehat{n}_{i^*}(p,v) |,
$ 
uses Property \hyperref[property2]{II} to match surface normals to the underlying shape, 
via $ \nabla_p \widehat{d}_{i^*} $.
A \textit{Directed Eikonal regularization}, 
based on Property \hyperref[property1]{I}, is given by
\begin{equation}
	\mathcal{L}_{\mathrm{DE}} = 
	\gamma_{\mathrm{E},d}
	\sum_i \xi\left[ \nabla_p \widehat{d}_i v + 1 \right]^2
	+
	\gamma_{\mathrm{E},\xi}
	[\nabla_p \widehat{\xi} v]^2
	,
\end{equation}
applied on the visibility and each delta component of $d$, analogous to prior SDF work
(e.g., \cite{gropp2020implicit,lin2020sdf,yang2021deep,bangaru2022differentiable,yu2022monosdf}).

Finally, we utilize two {weight field regularizations}, 
which encourage 
(1) low entropy PDDF outputs 
(to prevent $i^*$ from switching unnecessarily),
and
(2) the maximum likelihood delta component to \textit{transition} (i.e., change $i^*$) 
when a discontinuity is required:
$ 
\mathcal{L}_W = \gamma_V \mathcal{L}_V + \gamma_T \mathcal{L}_T 
$. 
The first is a \textit{weight variance loss}: 
$\mathcal{L}_V = \prod_i w_i$.
The second is a \textit{weight transition loss}:
$
\mathcal{L}_T = %
\max( 0, \varepsilon_T - |\nabla_p w_1 n| )^2,
$
where $\varepsilon_T$ is a hyper-parameter controlling the desired transition speed. %
Since $K=2$, using $w_1$ alone is sufficient to enforce changes along the normal. %
Note that $\mathcal{L}_T$ is \textit{only} applied to oriented points that we wish to undergo a transition 
(i.e., where a discontinuity is desired, 
as illustrated in Fig.\ \ref{fig:discontillus} and \ref{fig:prob_illus}), 
namely surface (S) and tangent (T) data. 
The complete PDDF shape-fitting loss is then
\begin{equation} \label{eq:singlefitall}
	\mathfrak{L}_S = \gamma_d \mathcal{L}_d +
	\gamma_\xi \mathcal{L}_\xi +
	\gamma_n \mathcal{L}_n +
	\mathcal{L}_{\mathrm{DE}} +
	\mathcal{L}_W.
\end{equation}

%% file: applications.tex
\section{Applications}

\subsection{Single Field Fitting}
\label{sec:results:singlefieldfitting}

Fig.\ \ref{fig:singobjfits} shows single-object fits,
via PDDF renderings with a \textit{single network evaluation per pixel}.
Normals and curvatures are obtained 
using only one or more backward passes for the same oriented point used in the single forward pass.
We implement the PDDFs with SIREN \cite{sitzmann2020implicit},
as it allows for higher-order derivatives and has previously proven effective
(e.g., \cite{chan2021pi,jo2021cg}).
(See Supp.~\S\ts{\ref{appendix:singlefits}}{B} for details.)

\subsubsection{Data-type Ablation}
\label{datatypeablation}
We perform a small-scale ablation experiment to discern the importance of each data sample type.
In particular, we consider six scenarios on a single shape (the Stanford Bunny), in each of which we train with 100K samples of each type \textit{except one}, which is removed.
We consider one other scenario that has the same number of total points, but \textit{no} single type is ablated (i.e., 83,333 samples per type).
We then measure the depth and visibility prediction error on 25K held-out samples of each data type, including the ablated one.

Results are shown in Table \ref{pddf:tab:dataablation}.
In most cases, the worst or second-worst errors on a given data type are incurred by models without access to that type. 
One anomaly is ablating S-type data, which damages performance across all sample types. This may be due to difficulties in knowing when to transition between weight field components.
Another outlier is $\mathcal{L}_\xi$-A, where ablating T-type data is the most damaging (removing A-type is third); we suspect this is because T-type samples are effectively the hardest subset of A-type data, and hence have an outsized impact upon removal.
Further, A-type data has more overlap with others (e.g., U and B). 

Finally, we remark that our data sampling strategies provide an inductive bias. For instance, B-type data will be most important when rendering far from the shape, while more A-type samples upweight visibile vs.\ non-visible parts of the scene.
Overall, while this is a small-scale (single-shape) analysis, it does suggest that each data type has information that the others cannot fully replace, especially U, O, S, and T.

\subsubsection{Internal Structure and Composition}
\label{sec:internalstructure}
\begin{figure}
	\centering	
	\includegraphics[width=0.95\linewidth]{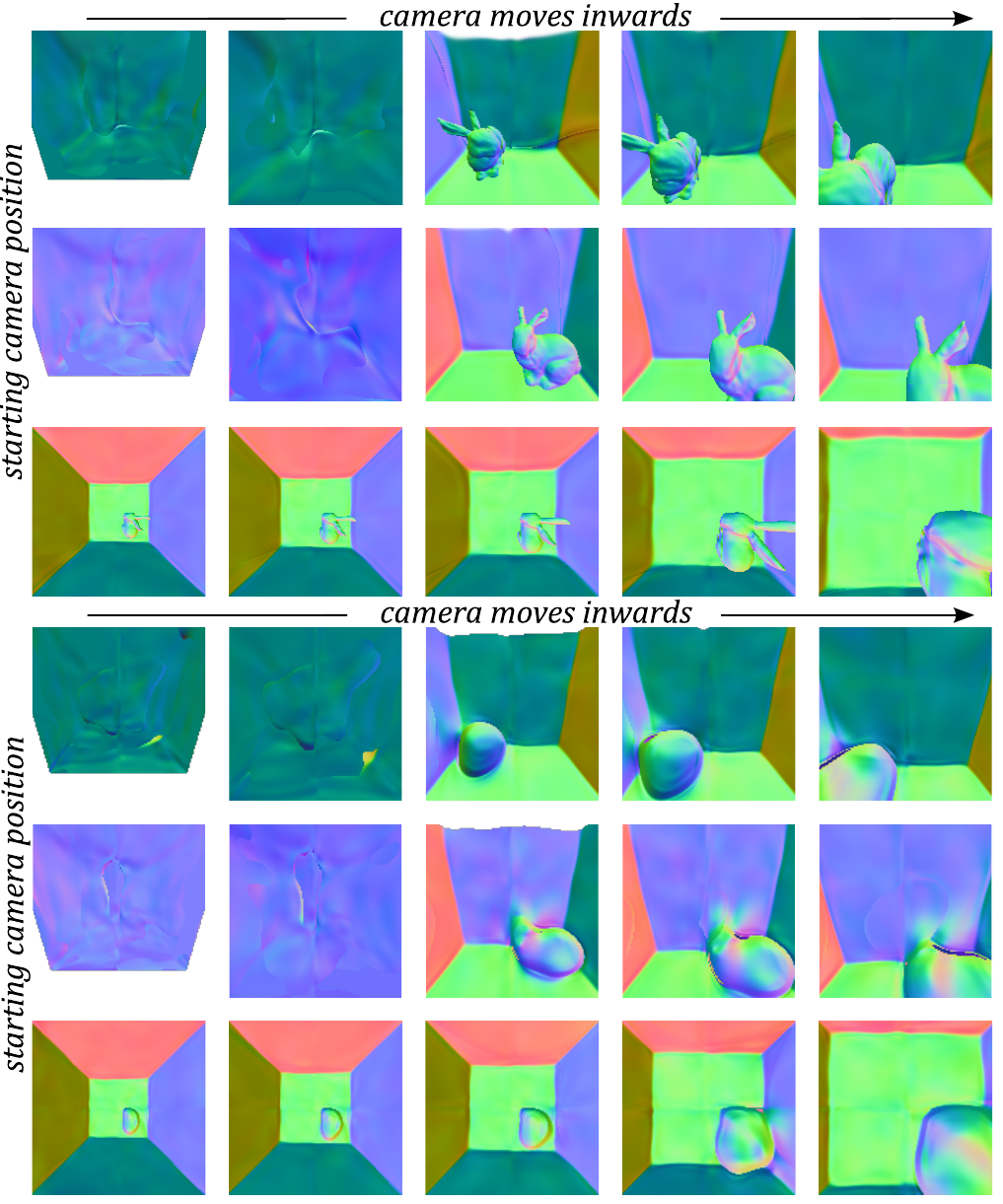}
	\caption{Example of internal structure rendering and composition. 
		Colours correspond to DDF surface normals 
		(as in Fig.\ \ref{fig:singobjfits}), 
		from the DDF. %
		The top insets compose two smaller DDFs, 
			while the bottom set uses a single larger monolithic one.
	}
	\label{fig:simpleroom}
\end{figure}

We discuss two additional modelling capabilities of DDFs: 
(i) internal structure representation
and
(ii) compositionality.
The first refers to the ability of our model to handle multi-layer surfaces:
we are able to place a camera inside a scene, within or between multiple surfaces, along a given direction.
This places our representation in contrast with recent works 
	\cite{zobeidi2021deep,sitzmann2021light}, 
	which do not model internal structure.
The second lies in the ease with which we can combine multiple DDFs, 
which is useful for manipulation without retraining
and
scaling to more complex scenes.
Our approach is inspired by prior work on soft rendering \cite{liu2019soft,gao2020learning}.
Formally, given a set of $N$ DDFs 
$ \zeta = \{ T^{(i)}, \xi^{(i)}, d^{(i)}, \mathcal{B}^{(i)} \}_{i=1}^N$,
where $T^{(i)}$ is a transform on oriented points converting world to object coordinates for the $i$th DDF 
(e.g., scale, rotation, and translation),
we can aggregate the visibility and depth fields 
into a single combined DDF.
For visibility of the combination of objects, we ask that \textit{at least} one surface is visible, implemented as:
\begin{equation}
	\xi_\zeta(p,v) = 1 - \prod_k (1 - \xi^{(k)}(T^{(k)}(p,v))).
\end{equation}
For depth, we want the closest visible surface to be the final output. 
One way to perform this is via a linear combination
\begin{equation}
	d_\zeta(p,v) = 
	\sum_k a_\zeta^{(k)}(p,v)\, d^{(k)}(T^{(k)}(p,v)),
\end{equation}
where $a_\zeta^{(k)}$ are computed via visibility and distance:
\begin{equation}
	a_\zeta(p,v) = \mathrm{Softmax}
	\left(  \left\{
	\frac{ \eta_T^{-1} \xi^{(k)}(T^{(k)}(p,v))}{ 
		\varepsilon_s + d^{(k)}(T^{(k)}(p,v)) }
	\right\}_k\,
	\right),
\end{equation}
with temperature $\eta_T$ and maximum inverse depth scale $\varepsilon_s$ as hyper-parameters.
This upweights contributions when distance is small, but visibility is high.
We exhibit these capabilities in Fig.\ \ref{fig:simpleroom}, 
which consists of two independently trained DDFs 
(one fit to five planes, forming a simple room, and
the other to the bunny mesh), %
where we simulate a camera starting outside the scene and entering the room.
For comparison,
to show the improved scaling of composing DDFs, we also attempted to fit the same scene using the single-object fitting procedure above. 
For fairness, we doubled the number of data samples extracted, as well as the size of each hidden layer.
In comparison to the top inset of Fig.\ \ref{fig:simpleroom}, this naive approach struggles to capture some high frequency details, though we suspect this could be mitigated to some extent by better sampling procedures.

\newcommand{\ctwcb}{0.0605\textwidth} %
\begin{figure}
	\centering
	\includegraphics[width=\ctwcb]{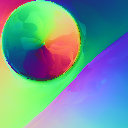}\hfill%
	\includegraphics[width=\ctwcb]{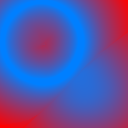}\hfill%
	\includegraphics[width=\ctwcb]{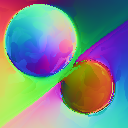}\hfill%
	\includegraphics[width=\ctwcb]{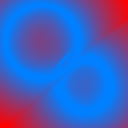}\hfill%
	\includegraphics[width=\ctwcb]{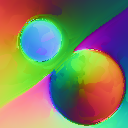}\hfill%
	\includegraphics[width=\ctwcb]{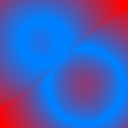}\hfill%
	\includegraphics[width=\ctwcb]{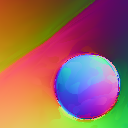}\hfill%
	\includegraphics[width=\ctwcb]{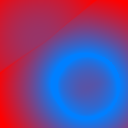}%
	\\[0.25mm]%
	\includegraphics[width=\ctwcb]{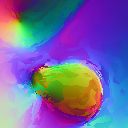}\hfill%
	\includegraphics[width=\ctwcb]{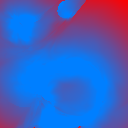}\hfill%
	\includegraphics[width=\ctwcb]{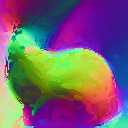}\hfill%
	\includegraphics[width=\ctwcb]{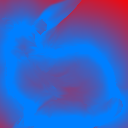}\hfill%
	\includegraphics[width=\ctwcb]{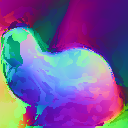}\hfill%
	\includegraphics[width=\ctwcb]{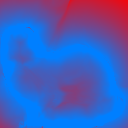}\hfill%
	\includegraphics[width=\ctwcb]{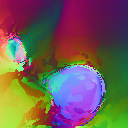}\hfill%
	\includegraphics[width=\ctwcb]{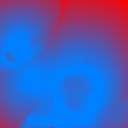}\hfill%
	\caption{
		The MDF, $v^*$, and its UDF (i.e., $d(p,v^*(p))$), in odd and even columns respectively, for two scenes (rows).
		For
		$v^*$, colours are 3D components; for the UDF, blue to red means near to far distances.
		Each image is a spatial slice in $z$, 
		where adjacent MDF-UDF pairs have the same $z$.
		Notice the colour change in $v^*$ as the slice moves through the shape, due to the closest surface switching from the front to the back of the shape (and thus flipping $v^*$).
		Some difficulties are also visible, (e.g., near surface intersections, where $v^*$ is ill-defined). 
	}
	\label{fig:udf}
\end{figure}

\subsubsection{UDF Extraction}
\label{sec:app:udfextract}
As noted in Property \hyperref[property4]{IV},  
one can extract a UDF from a DDF.
In particular, we optimize a field 
$v^* : \mathcal{B} \rightarrow \mathbb{S}^2 $, 
such that $\mathrm{UDF}(p) = d(p, v^*(p))$.
This \textit{Minimal Direction Field} (MDF), $v^*$, points to the closest point on $S$.\footnote{Discontinuities in $v^*$ occur at surfaces as before, but also on the medial surface of $S$ in $\mathcal{B}$. At such positions, there are multiple valid values of $v^*$.} 
We obtain it by optimizing an objective that prefers high visibility and low depth for a given $v^*(p)$.
Unlike directly fitting a UDF, this requires handling local minima for $v^*$ and non-visible (low $\xi$) directions. 
(See Fig.\ \ref{fig:udf} for visualizations and Supp.~\S\ts{\ref{appendix:udf}}{C} for details.)
When the normals exist, %
$v^*(p) = -n(p,v^*(p))$, in the notation of Property \hyperref[property2]{II}.
Recent work has highlighted the utility of UDFs over SDFs %
\cite{venkatesh2020dude,chibane2020neural,liu2023neudf};
in the case of DDFs, 
extracting a UDF or $v^*$ may provide useful auxiliary information for some tasks.

\subsubsection{Efficiency}
\label{sec:main:renderingefficiency}
To illustrate the gain in rendering efficiency, we compare to an existing differentiable sphere tracer, DIST \cite{liu2020dist}, on DeepSDF \cite{park2019deepsdf} models. Both representations rendered a $1024^2$ depth image.
The implicit field architectures are roughly comparable: DeepSDF used a ReLU-based MLP with eight hidden layers, while our PDDF used a SIREN MLP \cite{sitzmann2020implicit} (i.e., sine nonlinearity) with seven hidden layers, both with hidden dimensionalities all equal to 512. Per pixel, DIST must sphere trace along each camera ray, while the PDDF need only call the network once.
We find that DIST takes 5.3 seconds per render, while the DDF takes ${<}0.01$ seconds, a roughly ${\sim}$100$\times$ speedup. While this simplified setting does not completely characterize the tradeoff, 
it does suggest that DDFs obtain an advantage in efficiency.
Further, the runtime of sphere tracers depends on both the shape and the camera: for instance, simply moving the camera closer to the shape, which reduces the opportunities for certain runtime optimizations, increases rendering time by ${\sim}$1.8$\times$. 
In contrast, DDFs are unaffected by either element. See 
Supp.~\S\ts{\ref{sec:supp:renderingefficiency}}{B-C} 
for details.

\subsection{Single-Image 3D Reconstruction (SI3DR)}
\label{sec:app:si3dr}

\begin{figure}
	\includegraphics[width=0.99\linewidth]{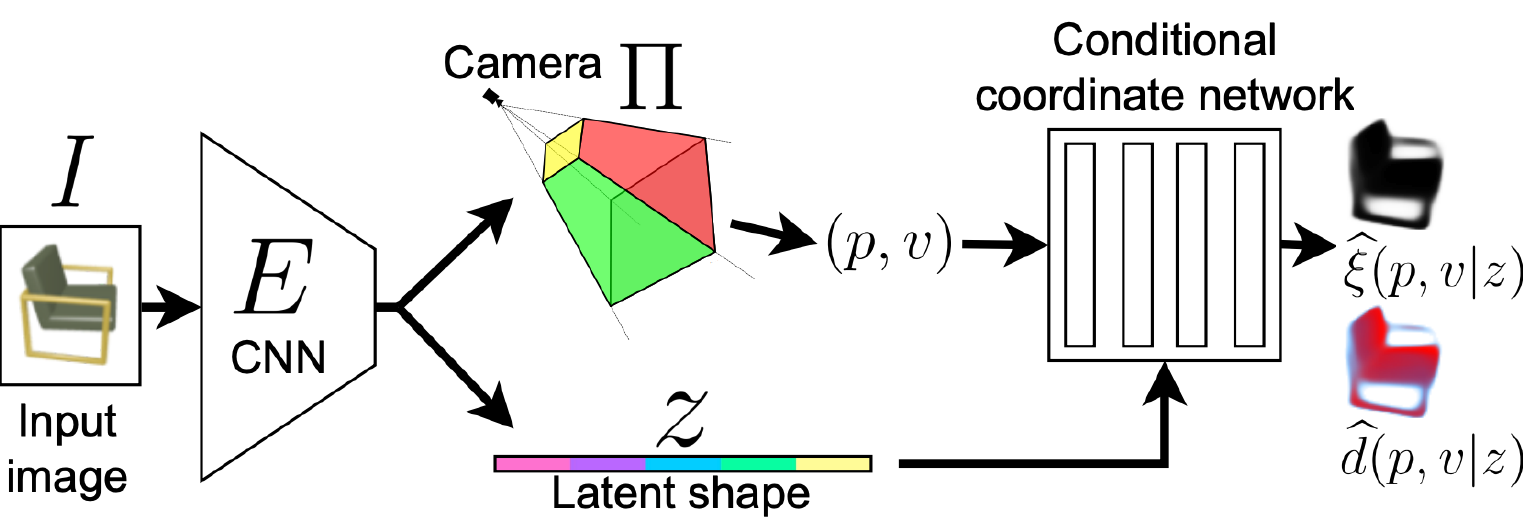}
	\caption{
		CPDDF-based SI3DR.
		A camera and latent shape are predicted from an image.
		The latent vector conditions a CPDDF, 
		which can render geometry
		from any viewpoint. 
	}\label{fig:si3drdesc}
\end{figure}

\begin{table*}[t] %
	\centering
		\begin{tabular}{r|ccccc | cccc | c c}
			& \multicolumn{5}{c|}{DDF} & 
			\multicolumn{4}{c|}{PC-SIREN} & \multirow{2}{*}{P2M \cite{wang2018pixel2mesh,wang2020pixel2mesh}} & \multirow{2}{*}{3DR \cite{choy20163d}} \\
			& $\Pi_g$-L & $\Pi_g$-S & $\widehat{\Pi}_\nabla$-S & $\widehat{\Pi}$-L  & $\widehat{\Pi}$-S  
			& $\Pi_g$-L  & $\Pi_g$-S  & $\widehat{\Pi}$-L  & $\widehat{\Pi}$-S &  &  \\\hline 
			$D_C$  $ \downarrow$ & 0.300 & 0.346 & 0.629 & 0.730 & 0.787  & 0.339 & 0.364 &  0.814 &  0.842 & 0.452 & 1.057 \\ 
			$F_{\tau}$  $ \uparrow$ & 68.95 & 61.65 & 53.25 & 57.98 & 51.36 & 69.44 & 64.76 & 56.74 & 52.65 & 64.45 & 39.83  \\ 
			$F_{2\tau}$  $\uparrow$ & 83.15 & 78.65 & 70.41 & 72.83 & 68.41 &       81.94 &	79.91 & 70.47 & 68.49 & 78.65 & 57.76
			\\ 
		\end{tabular} %
\caption[Single Image 3D Reconstruction (SI3DR) with PDDFs]{
	Single-image 3D reconstruction results.
	Columns: %
	L/S refer to sampling 5000/2466 points (2466 being used in P2M), $\Pi_g$/$\widehat{\Pi}$ means using the true versus predicted camera, 
	and $\widehat{\Pi}_\nabla$ denotes camera position correction using gradient descent.
	Metrics: $D_C$ is the Chamfer distance ($\times 1000$) and $F_{\tau}$ is the F-score ($\times 100$) at threshold $\tau = 10^{-4}$.
	PC-SIREN is our matched-architecture baseline;
	Pixel2Mesh (P2M) %
	and 
	3D-R2N2 (3DR) %
	are baselines using different shape modalities 
	(numbers from \cite{wang2020pixel2mesh}).
	Note: scenarios using $\Pi_g$ (i.e., evaluating in canonical object coordinates) are not directly comparable to P2M or 3DR,
	but serve to isolate pose vs.\ shape error.
	Overall, DDF-derived PCs 
	(1) perform similarly to directly learning to output a PC and 
	(2) underperform P2M overall, but outperform it in terms of shape quality when camera prediction error is excluded.
}
\label{tab:si3dr:avgs}
\end{table*}

We next utilize DDFs for single-image 3D reconstruction.
Given a colour image $I$, 
we predict the underlying latent shape $z_s$ 
and camera ${\Pi}$ that gave rise to the image, via an encoder $ E(I) = (\widehat{z}_s, \widehat{\Pi}) $ (see Fig.~\ref{fig:si3drdesc}). 
For decoding, we use a \textit{conditional} PDDF (CPDDF), which computes depth $\widehat{d}(p,v|z_s)$ and visibility $\widehat{\xi}(p,v|z_s)$. 
We use three loss terms: 
(a) shape DDF fitting in canonical pose $\mathfrak{L}_S$ (eq.\ \ref{eq:singlefitall}),
(b) camera prediction $\mathcal{L}_\Pi = ||\Pi_g - \widehat{\Pi}||_2^2$,
and 
(c) mask matching $ \mathcal{L}_M = \mathrm{BCE}(I_\alpha, \mathcal{R}_\xi(z_s|\widehat{\Pi})) $, where $I_\alpha$ is the input alpha channel and $ \mathcal{R}_\xi $ renders the DDF visibility.
The full objective is 
$\mathcal{L}_\mathrm{SI3DR} = \gamma_{R,S}\mathfrak{L}_S + \gamma_{R,\Pi}\mathcal{L}_\Pi  + \gamma_{R,M} \mathcal{L}_M $.
We implement $E$ as two ResNet-18 networks \cite{he2016deep}, while the CPDDF is a modulated SIREN \cite{mehta2021modulated}.
For evaluation,
camera extrinsics are either the predicted $\widehat{\Pi}$ or ground-truth $\Pi_g$ (to separate shape and camera errors).
See Supp.~\S\ts{\ref{appendix:si3dr}}{D} for details.

\subsubsection{Explicit Sampling}  
SI3DR evaluation commonly uses point clouds (PCs).
Thus, we present a simple PC extraction method, though it cannot guarantee uniform sampling over the shape $S$. 
Recall that $q(p,v) = p + d(p,v)v \in S$, if $\xi(p,v) = 1$. 
Thus, we sample $p\sim \mathcal{U}[\mathcal{B}]$, and wish to project them onto $S$.
Then, $v$ should be chosen to avoid $\xi(p,v) = 0$.
Hence, for each $p$, we sample $V(p) = \{v_i(p)\sim\mathcal{U}[\mathbb{S}^2]\}_{i=1}^{n_v}$ and ``compose'' them to estimate $\widehat{v}^{\,*}(p)$ (as in \S\ref{sec:internalstructure} and \S\ref{sec:app:udfextract}, but without optimization), giving $q(p,\widehat{v}^{\,*}(p))$ as a point on $S$.
Repeating this $N_H$ times (starting from $p\leftarrow q$) also helps, if depths are less accurate far from $S$.
We set $n_v=128$ and $N_H = 3$ (see Supp.~\S\ts{\ref{appendix:si3dr:nh1}}{D-C} for ablation with $N_H=1$).

\subsubsection{Baselines}  
Our primary baseline alters the \textit{shape representation}, while keeping the architecture and training setup as similar as possible. Specifically, it uses the same encoders as the DDF and an almost identical network for the decoder (changing only the input and output dimensionalities), but altered to output PCs directly (denoted PC-SIREN).
In particular, the decoder is an implicit shape mapping $f_b : \mathbb{R}^3 \rightarrow \mathbb{R}^3$, which takes $p\sim\mathcal{U}[-1,1]^3$ as input and directly returns $q = f_b(p) \in S$. 
Training uses the Chamfer distance $D_C$ 
and $\mathcal{L}_\Pi$.
We also compare to
the mesh-based Pixel2Mesh (P2M) \cite{wang2018pixel2mesh,wang2020pixel2mesh} and voxel-based 3D-R2N2 (3DR) \cite{choy20163d}.

\begin{table}
\centering
{ \setlength{\tabcolsep}{3pt}
\begin{tabular}{clccc}
	\makecell[c]{Comparative \\ Baseline} & 
	\makecell[c]{CPDDF \\ Setting}    &
	\makecell[c]{CPDDF \\ $D_C$} & %
	\makecell[c]{Baseline \\ $D_C$}   &
	Diff.~$\uparrow$ \\\toprule %
	\multirow{2}*{ \makecell[c]{PC-SIREN \\ Id.\ Ar.\ Baseline} } & 
	$ - \nabla_\Pi $ / $\widehat{\Pi}$ &
	0.787 & 0.842 & \textcolor{ForestGreen}{+0.055} \\
	& 
	$ + \nabla_\Pi $ / $\widehat{\Pi}$ &
	0.629 & 0.842 & \textcolor{ForestGreen}{+0.213} \\[2mm]	
	\makecell[c]{3DR \cite{choy20163d} \\ Voxel Baseline} & 
	$ + \nabla_\Pi $ / $\widehat{\Pi}$ &
	0.629 & 1.057 & \textcolor{ForestGreen}{+0.428} \\[2mm]
	\multirow{2}*{ \makecell[c]{P2M \cite{wang2018pixel2mesh} \\ Mesh Baseline} } &		
	$ + \nabla_\Pi $ / $\widehat{\Pi}$ &
	0.629 & 0.451 & \textcolor{BrickRed}{-0.178} \\
	& 
	$ + \nabla_\Pi $ / $\Pi_g$ &
	0.346 & 0.451 & \textcolor{ForestGreen}{+0.105} \\
\end{tabular}
}
\caption[Comparative Evaluation Summary of CPDDF-Based SI3DR]{
	We compare CPDDFs to a field with identical architecture (Id.\ Ar.), which outputs PCs,
	and two explicit-shape-based baselines,
	with modality-specialized architectures
	(vs.\ our use of generic networks).
	The Chamfer distance $D_C$ compares to the GT shape.
	``CPDDF Setting'' records: 
	(i) 
	silhouette-based optimization 
	(i.e., ``$ - \nabla_\Pi$'' vs.\ ``$ + \nabla_\Pi$'')
	and
	(ii) evaluation with
	the predicted vs.\ the GT camera 
	(i.e., ``$\widehat{\Pi}$'' vs.\  ``$\Pi_g$'').
	The $\Pi_g$ scenario is not fair,
	(as it uses the GT $\Pi$),
	but shows that much error is in pose, 
		not shape, prediction.
}\label{tab:si3dr:summarytable}
\end{table}

\begin{table}
	\centering
	\begin{tabular}{c|cccc}
		$\,$ & \makecell{Image \\ GAN} & \makecell{CPDDF-based \\ GAN} & \makecell{ CGR \\ \cite{aumentado2020cycle} } & \makecell{Image \\ VAE} \\\hline
		\rule{0pt}{0.5\normalbaselineskip} 
		FID \cite{heusel2017gans} $\downarrow$  &  15                     &   27            &   120 & 194 \\ 
		3D?      &  No  & Yes & Yes & No \\ 
	\end{tabular}	
	\caption[CPDDF-based Unpaired 2D-3D Generative Modelling Results]{
		Evaluation of our CPDDF-based 3D-aware generative model.
		We compared to a 3D-\textit{un}aware image GAN 
		(with the same critic),
		the Cyclic Generative Renderer (CGR) \cite{aumentado2020cycle}, which samples textured meshes and uses a differentiable rasterizer,
		and a 3D-\textit{un}aware VAE.
		(See \S\ref{sec:genmodel}).			
	}
	\label{pddf:gentable}
\end{table}

\newcommand{\alw}{0.274\textwidth}
\newcommand{\alh}{0.47in}
\newcommand{\aalh}{0.485in}
\newcommand{\alhp}{0.32in}
\newcommand{\aalhp}{0.323in}
\newcommand{\aalhps}{0.42in}
\newcommand{\cch}{0.30}
\newcommand{\ach}{0.01}
\newcommand{\cuthch}{0.11}
\newcommand{\cuthchp}{0.25}
\begin{figure*}
	\adjincludegraphics[height=\aalh,trim={ {\ach\width} {\cuthch\height} {\ach\width}  {\cuthch\height}},clip]{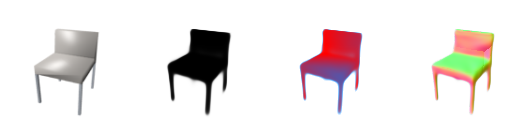}
	\adjincludegraphics[height=\alh,trim={ {\cch\width} {\cuthch\height} {\cch\width}  {\cuthch\height}},clip]{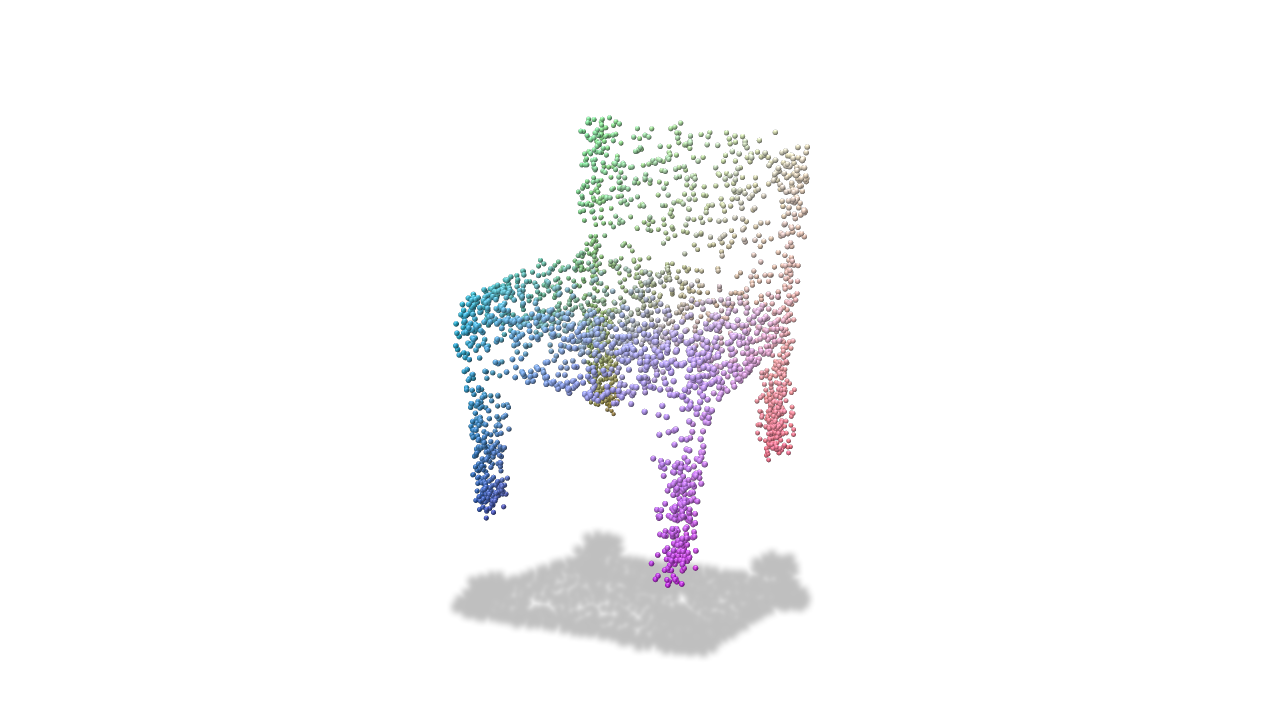}\hfill
	\adjincludegraphics[height=\alh,trim={ {\cch\width} {\cuthch\height} {\cch\width}  {\cuthch\height}},clip]{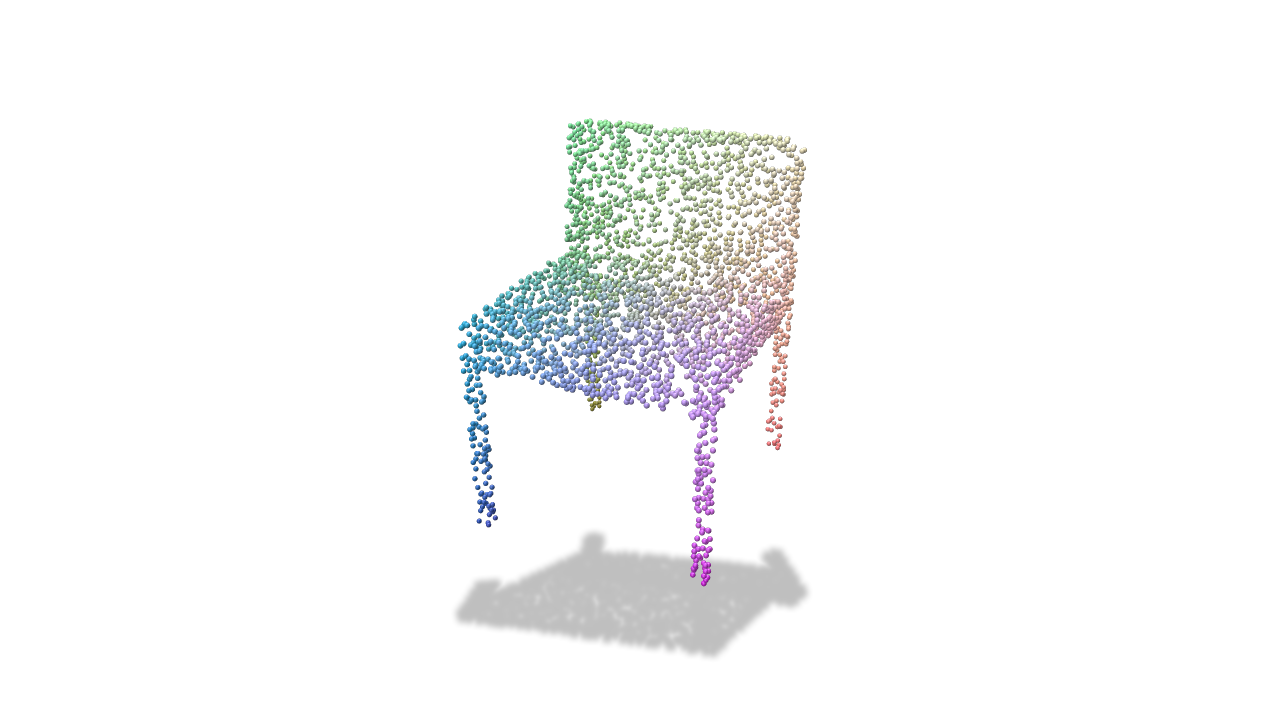}\hfill %
	\adjincludegraphics[height=\aalh,trim={ {\ach\width} {\cuthch\height} {\ach\width}  {\cuthch\height}},clip]{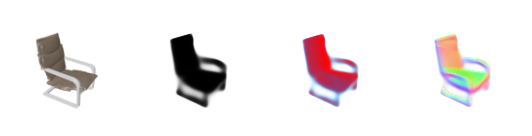}
	\adjincludegraphics[height=\alh,trim={ {\cch\width} {\cuthch\height} {\cch\width}  {\cuthch\height}},clip]{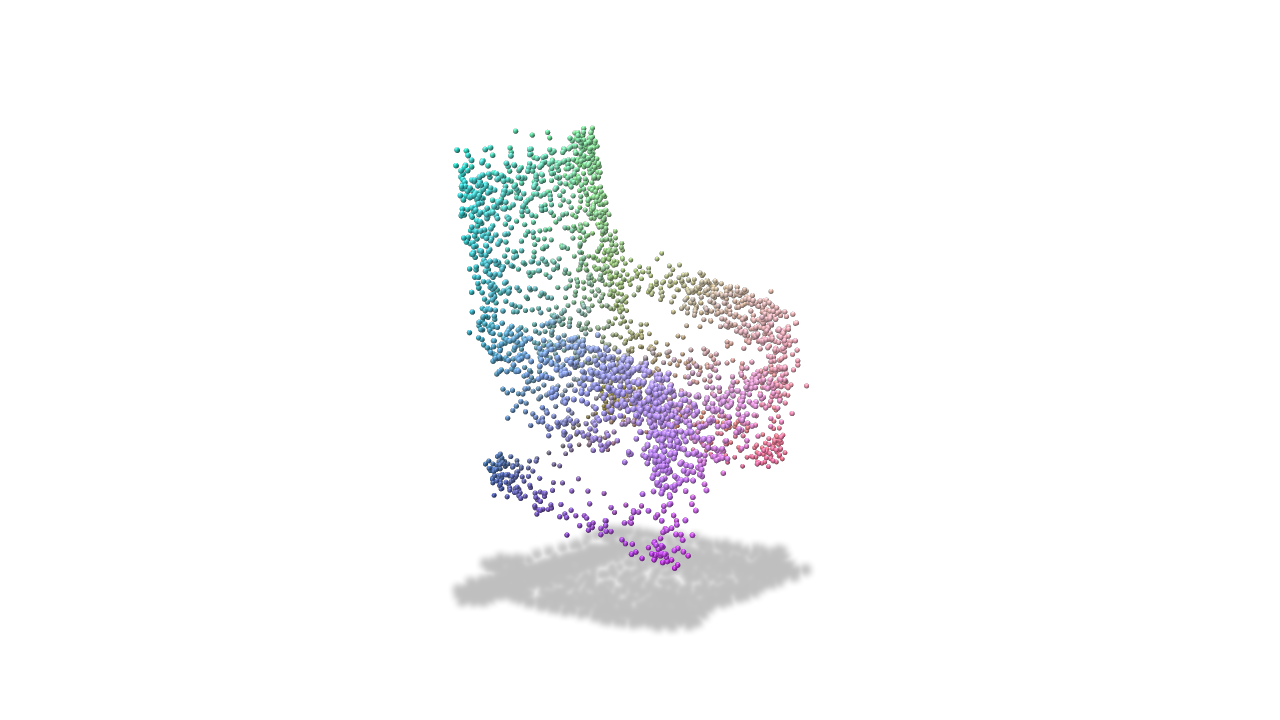}\hfill
	\adjincludegraphics[height=\alh,trim={ {\cch\width} {\cuthch\height} {\cch\width}  {\cuthch\height}},clip]{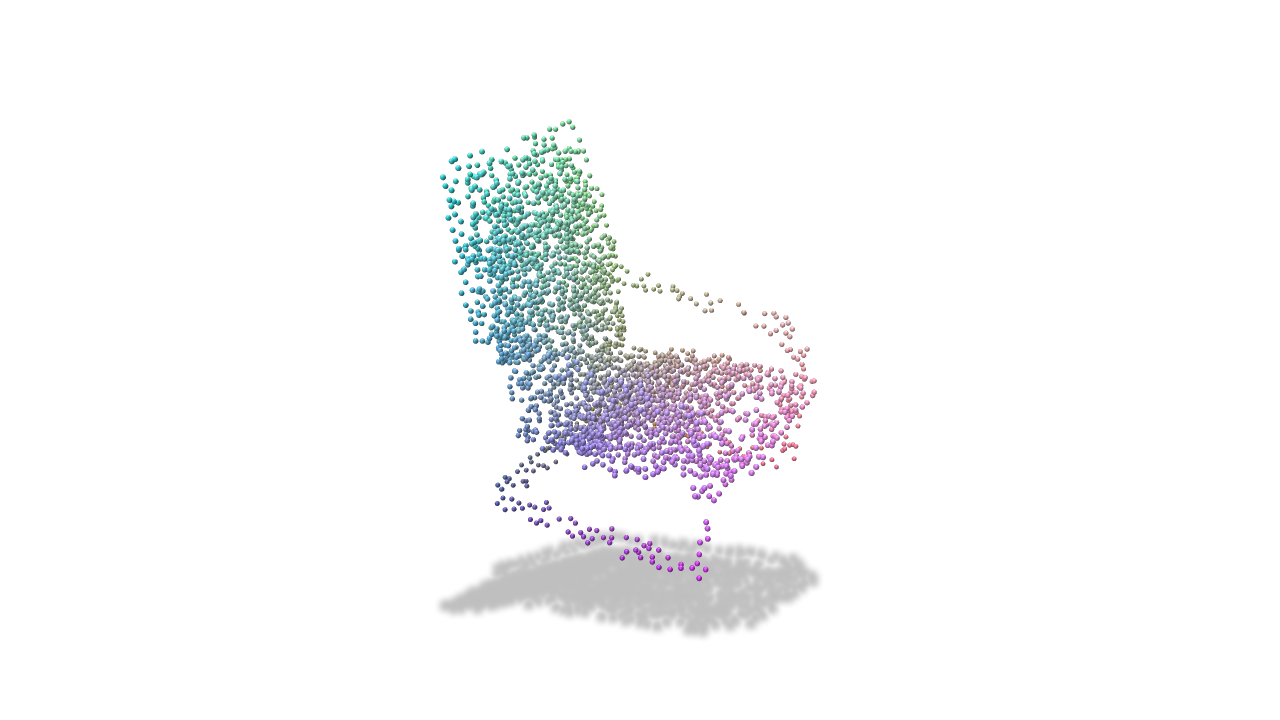} \\
	\adjincludegraphics[height=\aalh,trim={ {\ach\width} {\cuthch\height} {\ach\width}  {\cuthch\height}},clip]{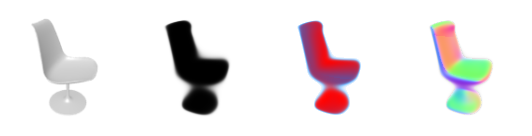}
	\adjincludegraphics[height=\alh,trim={ {\cch\width} {\cuthch\height} {\cch\width}  {\cuthch\height}},clip]{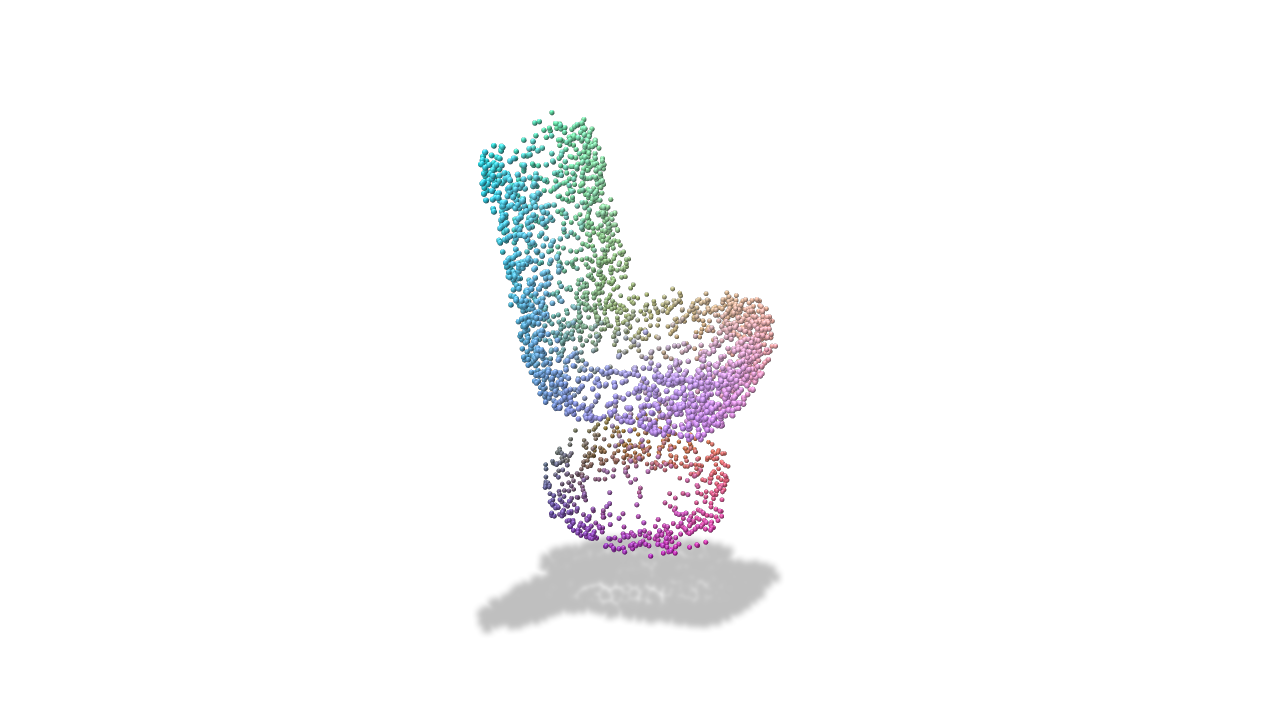}\hfill
	\adjincludegraphics[height=\alh,trim={ {\cch\width} {\cuthch\height} {\cch\width}  {\cuthch\height}},clip]{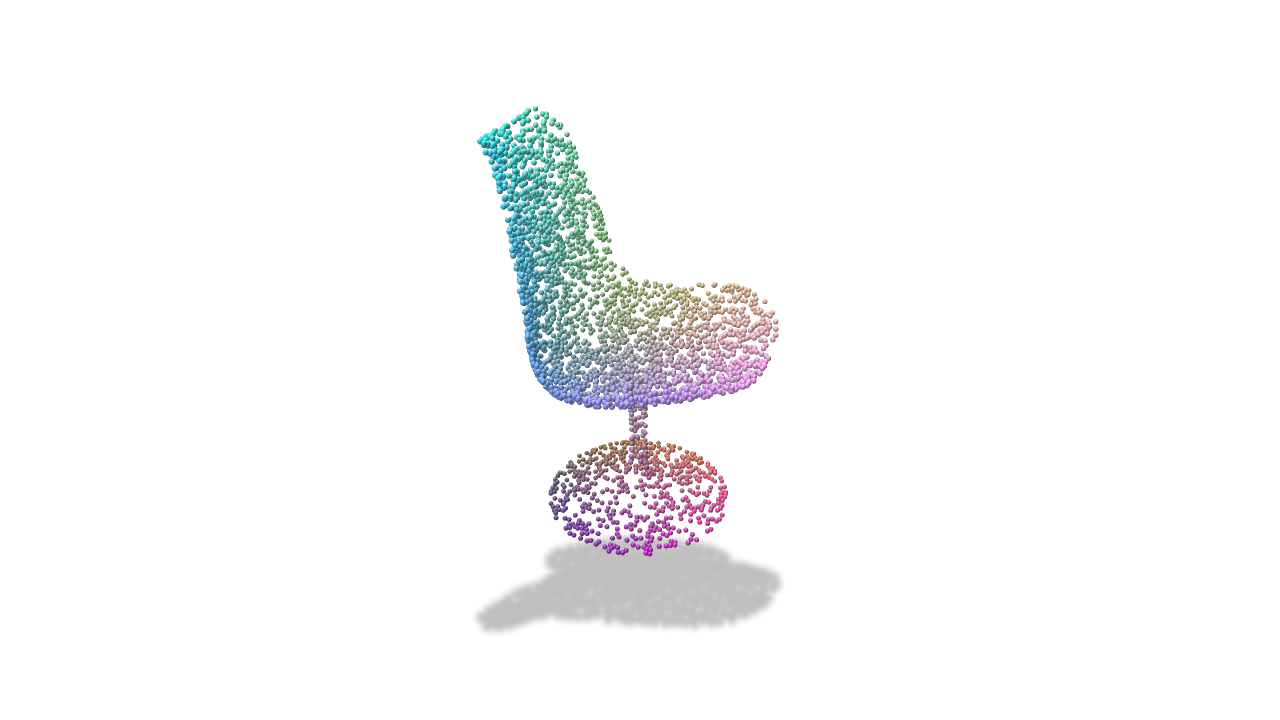}\hfill %
	\adjincludegraphics[height=\aalh,trim={ {\ach\width} {\cuthch\height} {\ach\width}  {\cuthch\height}},clip]{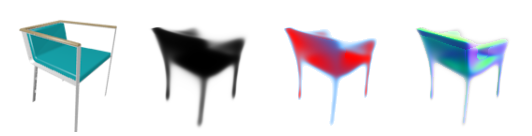}
	\adjincludegraphics[height=\alh,trim={ {\cch\width} {\cuthch\height} {\cch\width}  {\cuthch\height}},clip]{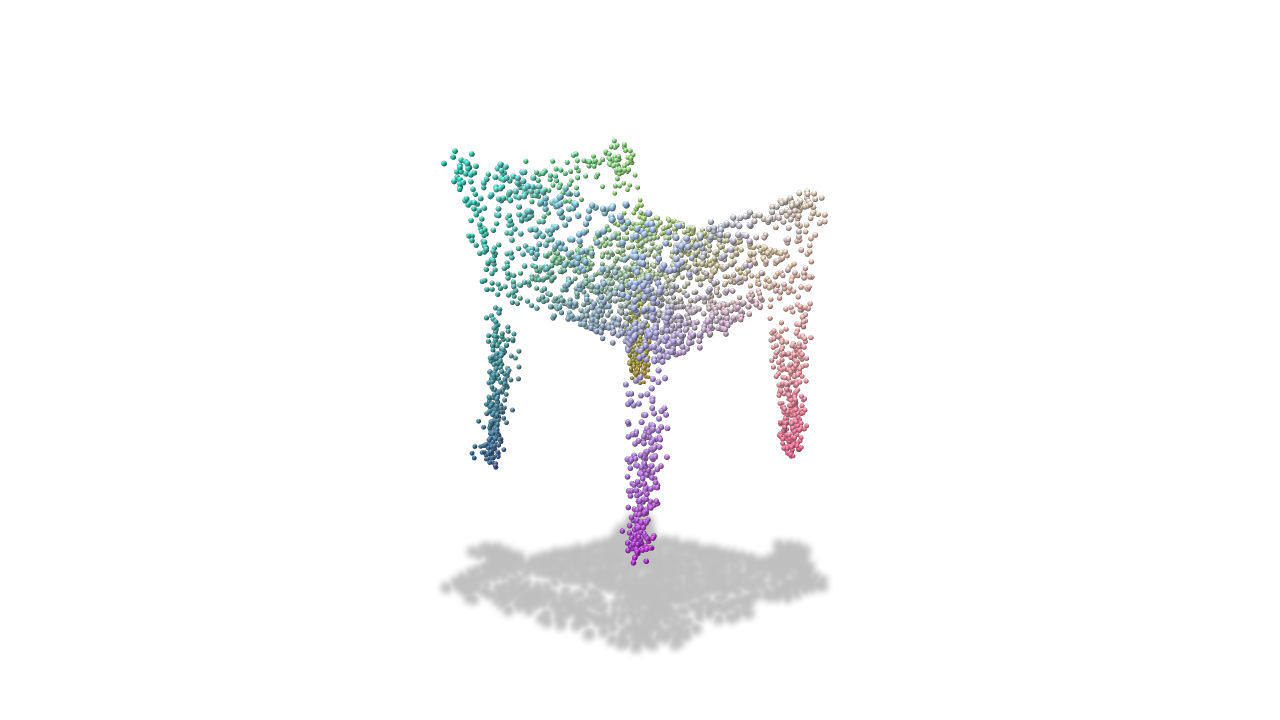}\hfill
	\adjincludegraphics[height=\alh,trim={ {\cch\width} {\cuthch\height} {\cch\width}  {\cuthch\height}},clip]{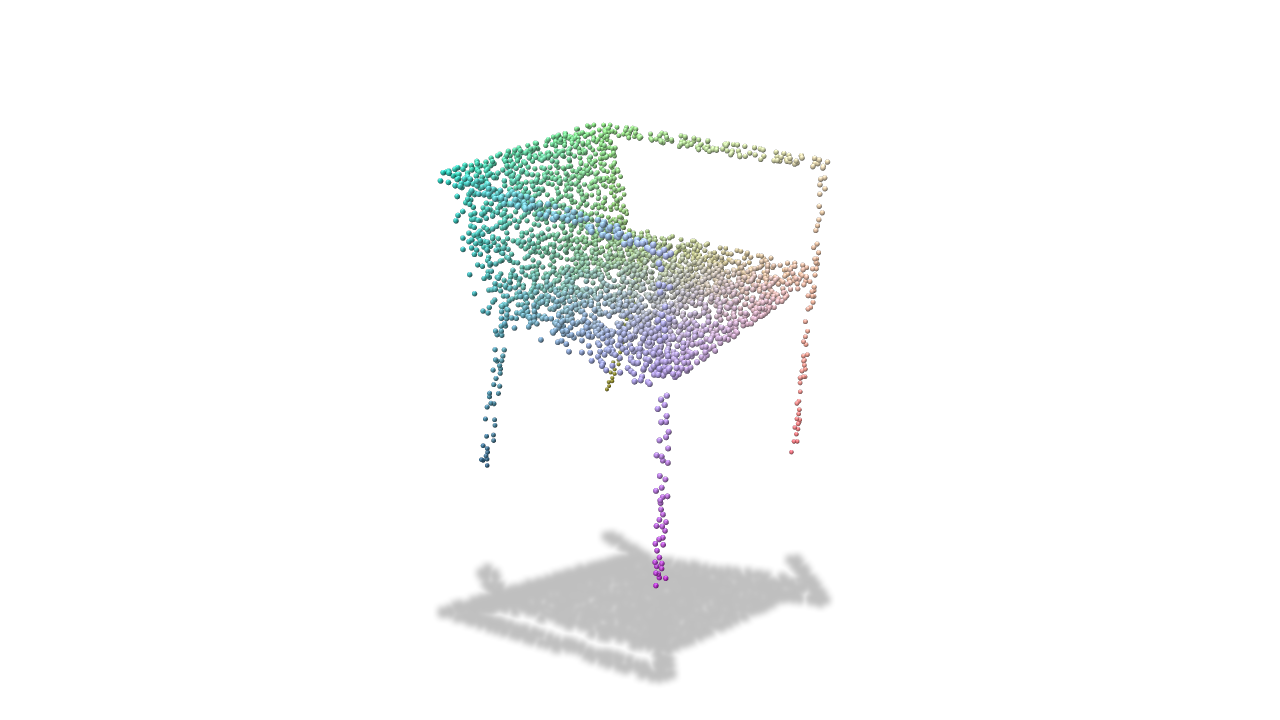} \\
	\adjincludegraphics[height=\aalhp,trim={ {\ach\width} {\cuthchp\height} {\ach\width}  {\cuthchp\height}},clip]{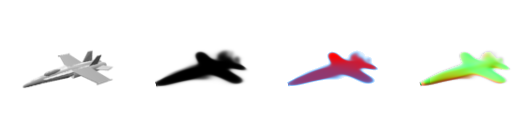}\hfill
	\adjincludegraphics[height=\alhp,trim={ {\cch\width} {\cuthchp\height} {\cch\width}  {\cuthchp\height}},clip]{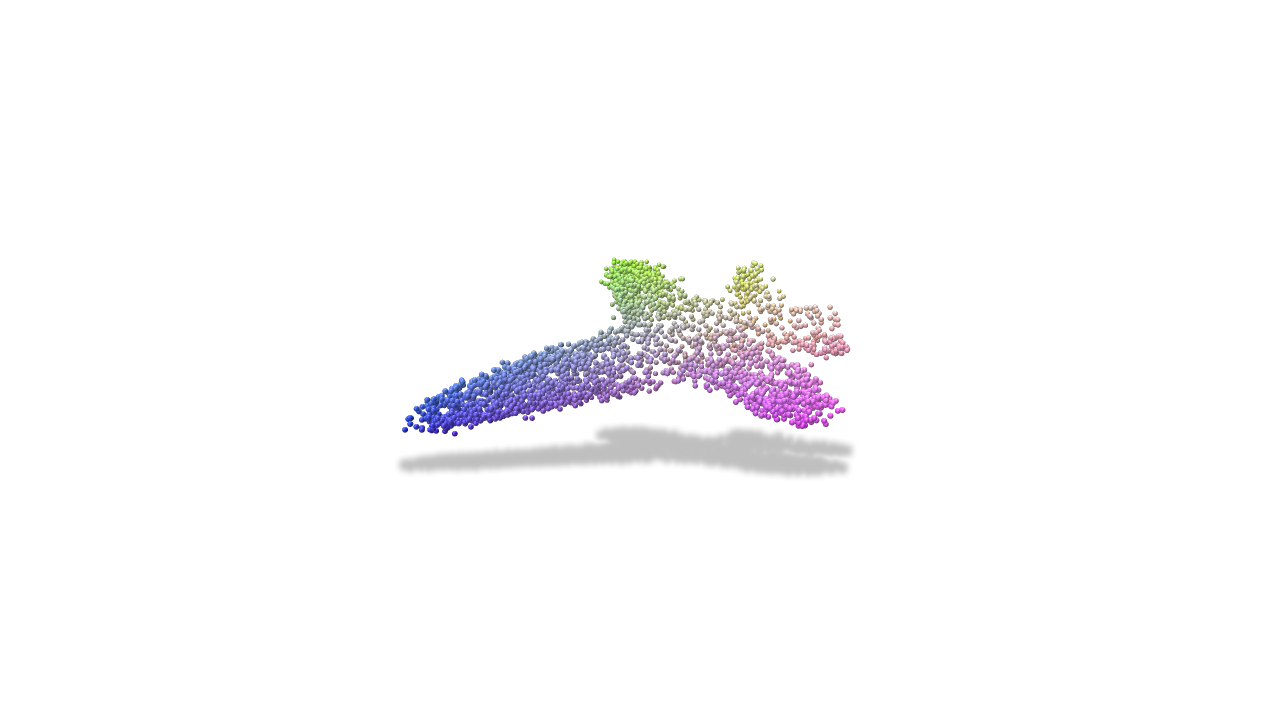}\hfill
	\adjincludegraphics[height=\alhp,trim={ {\cch\width} {\cuthchp\height} {\cch\width}  {\cuthchp\height}},clip]{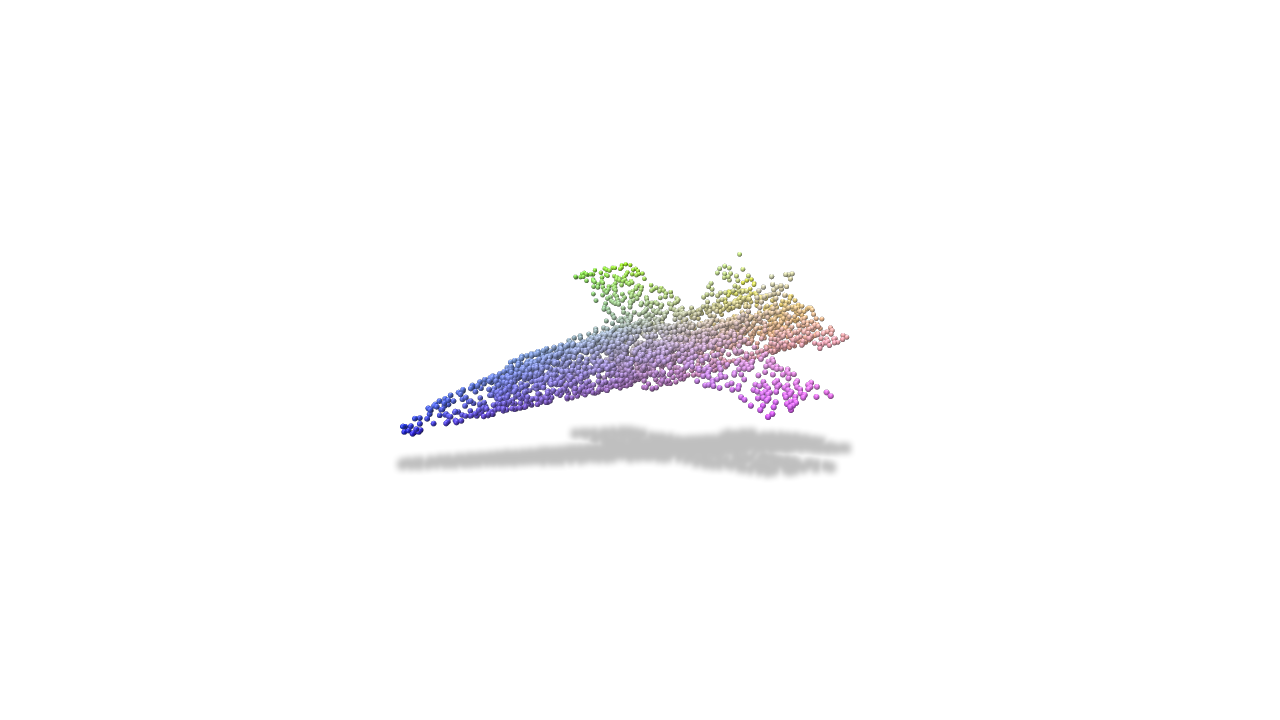}\hfill %
	\adjincludegraphics[height=\aalhp,trim={ {\ach\width} {\cuthchp\height} {\ach\width}  {\cuthchp\height}},clip]{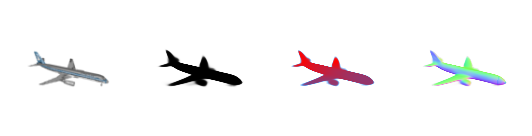}\hfill
	\adjincludegraphics[height=\alhp,trim={ {\cch\width} {\cuthchp\height} {\cch\width}  {\cuthchp\height}},clip]{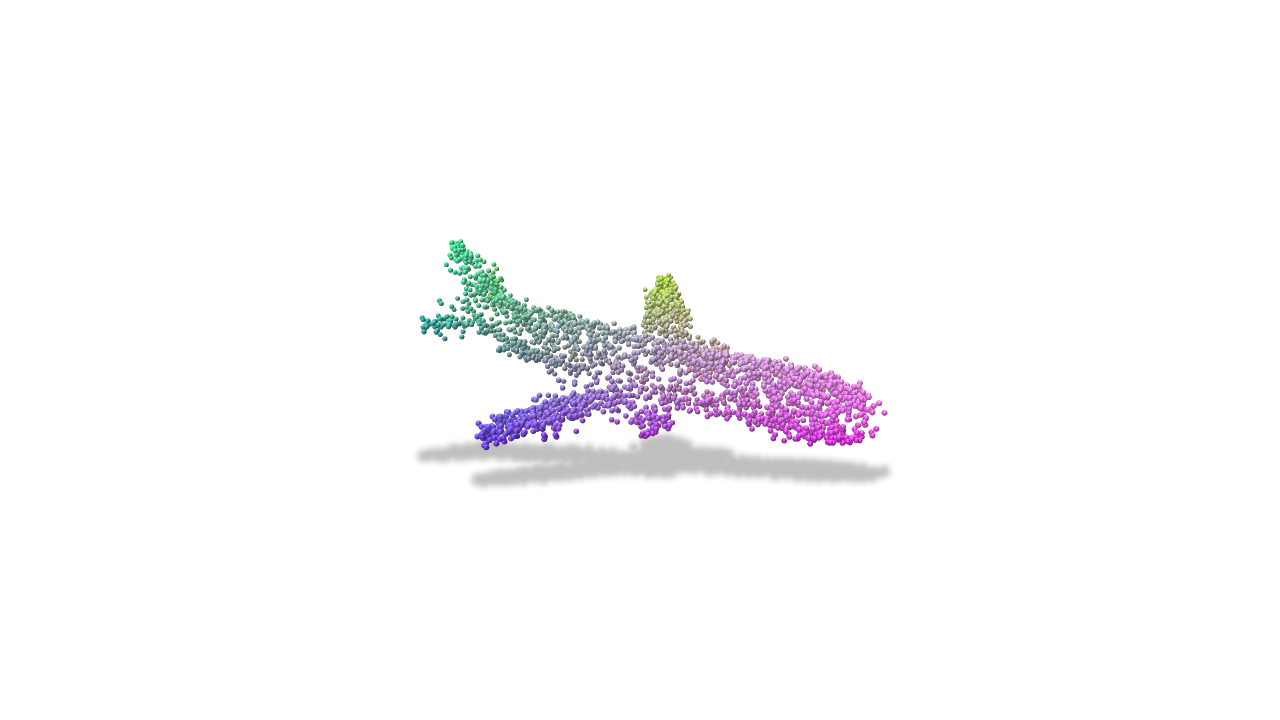}\hfill
	\adjincludegraphics[height=\alhp,trim={ {\cch\width} {\cuthchp\height} {\cch\width}  {\cuthchp\height}},clip]{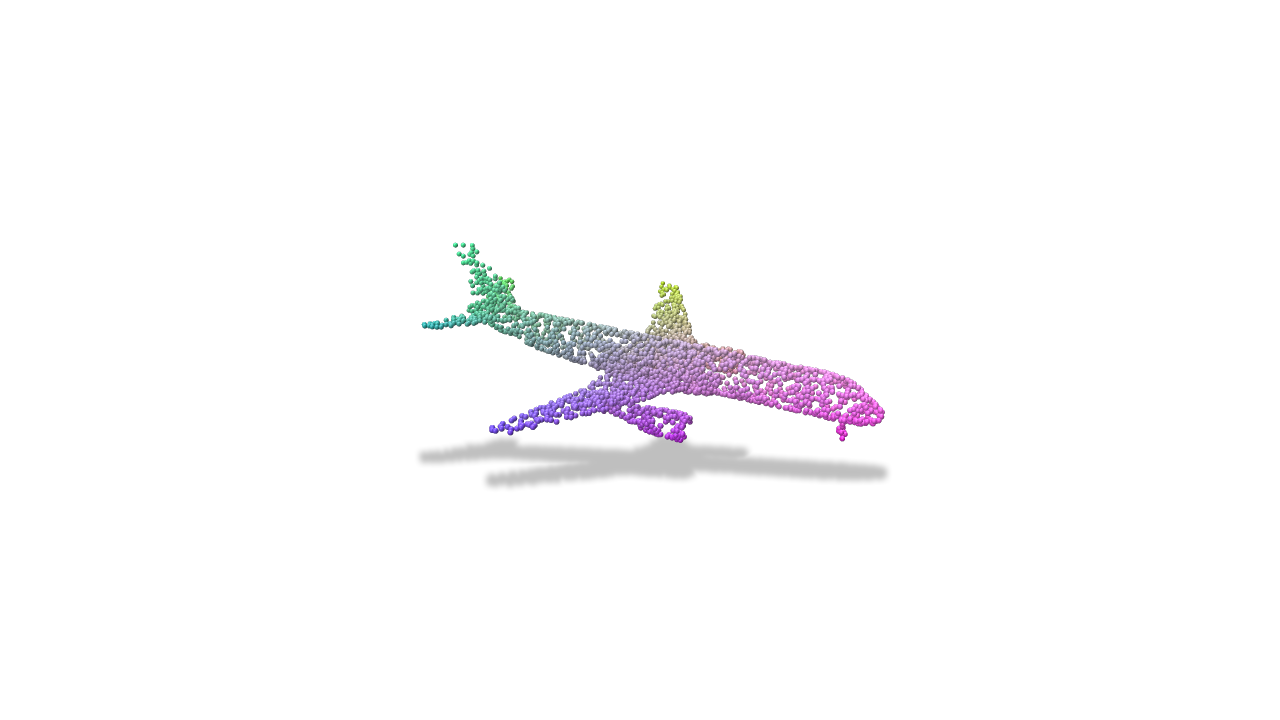} \\
	\adjincludegraphics[height=\aalhp,trim={ {\ach\width} {\cuthchp\height} {\ach\width}  {\cuthchp\height}},clip]{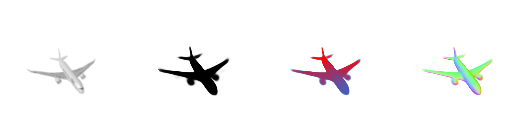}\hfill
	\adjincludegraphics[height=\alhp,trim={ {\cch\width} {\cuthchp\height} {\cch\width}  {\cuthchp\height}},clip]{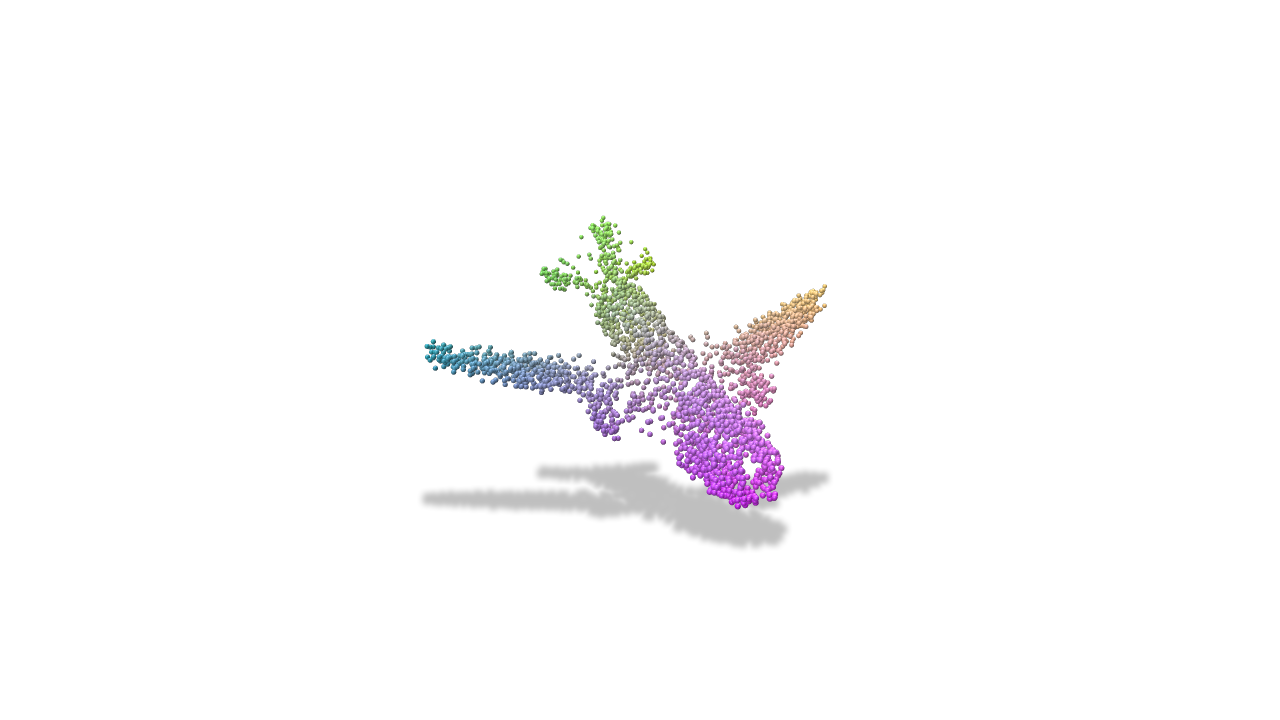}\hfill
	\adjincludegraphics[height=\alhp,trim={ {\cch\width} {\cuthchp\height} {\cch\width}  {\cuthchp\height}},clip]{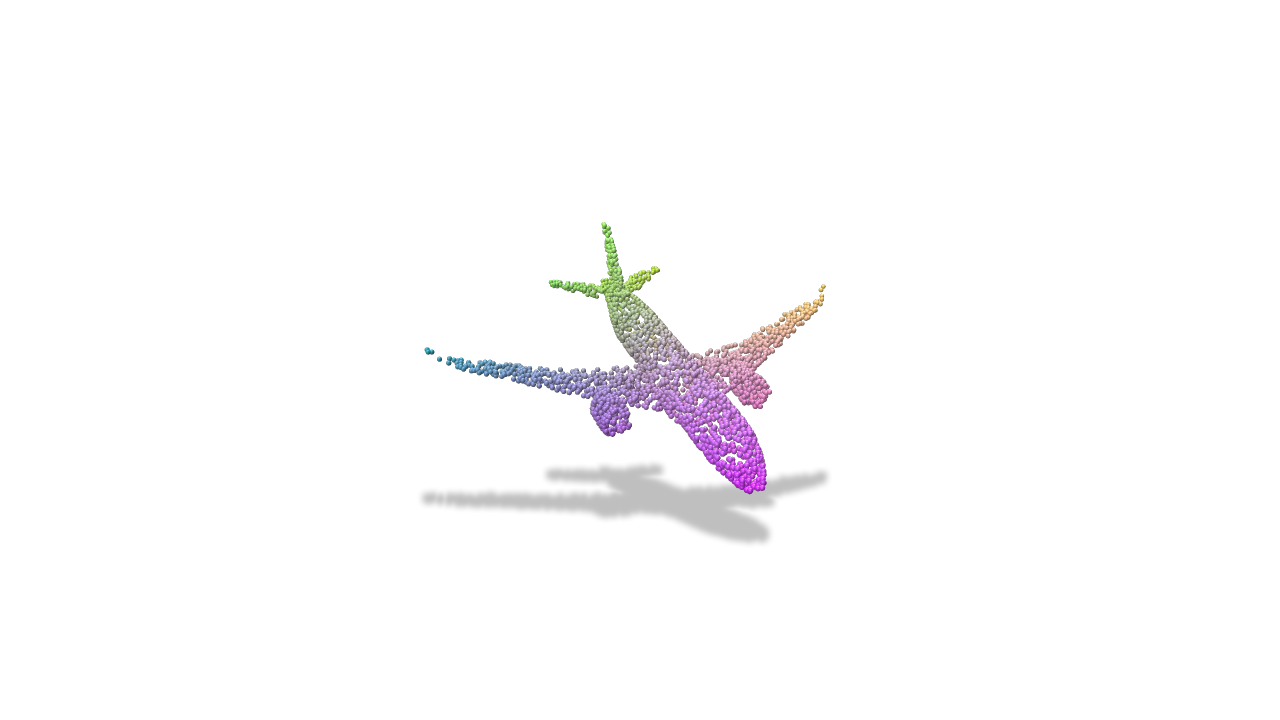}\hfill %
	\adjincludegraphics[height=\aalhp,trim={ {\ach\width} {\cuthchp\height} {\ach\width}  {\cuthchp\height}},clip]{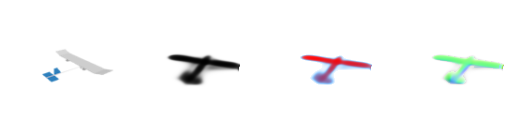}\hfill
	\adjincludegraphics[height=\alhp,trim={ {\cch\width} {\cuthchp\height} {\cch\width}  {\cuthchp\height}},clip]{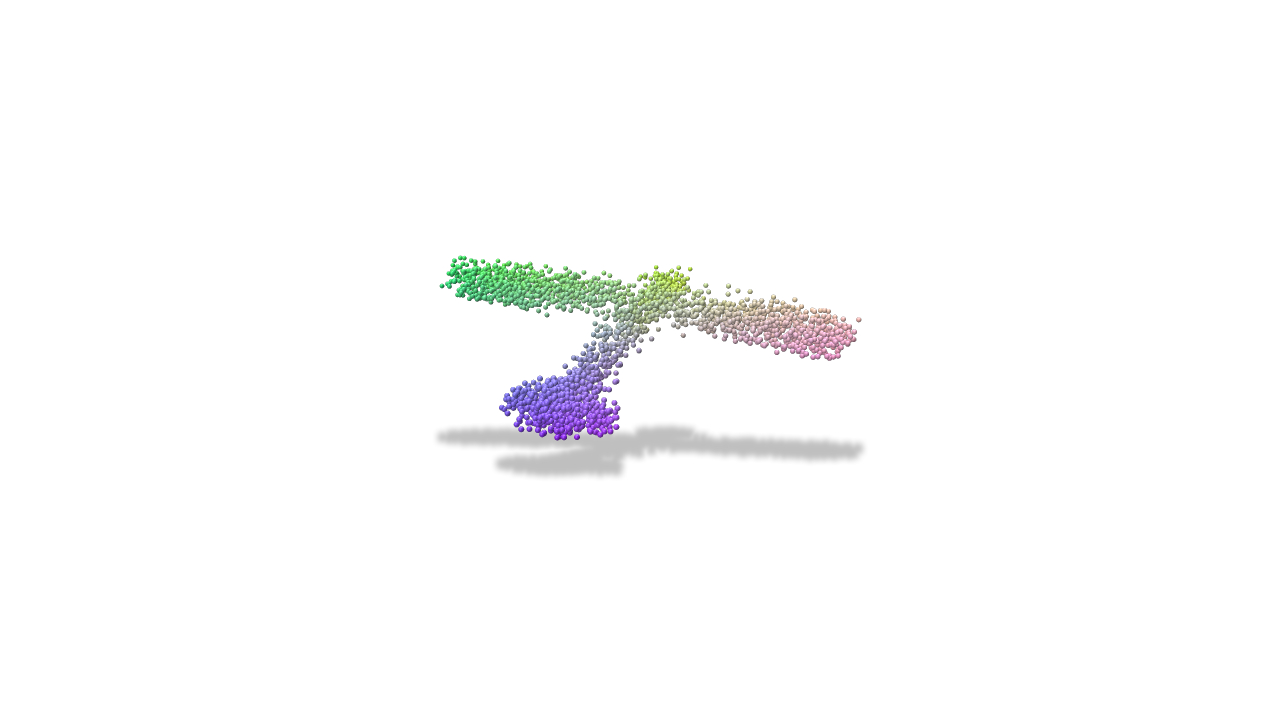}\hfill
	\adjincludegraphics[height=\alhp,trim={ {\cch\width} {\cuthchp\height} {\cch\width}  {\cuthchp\height}},clip]{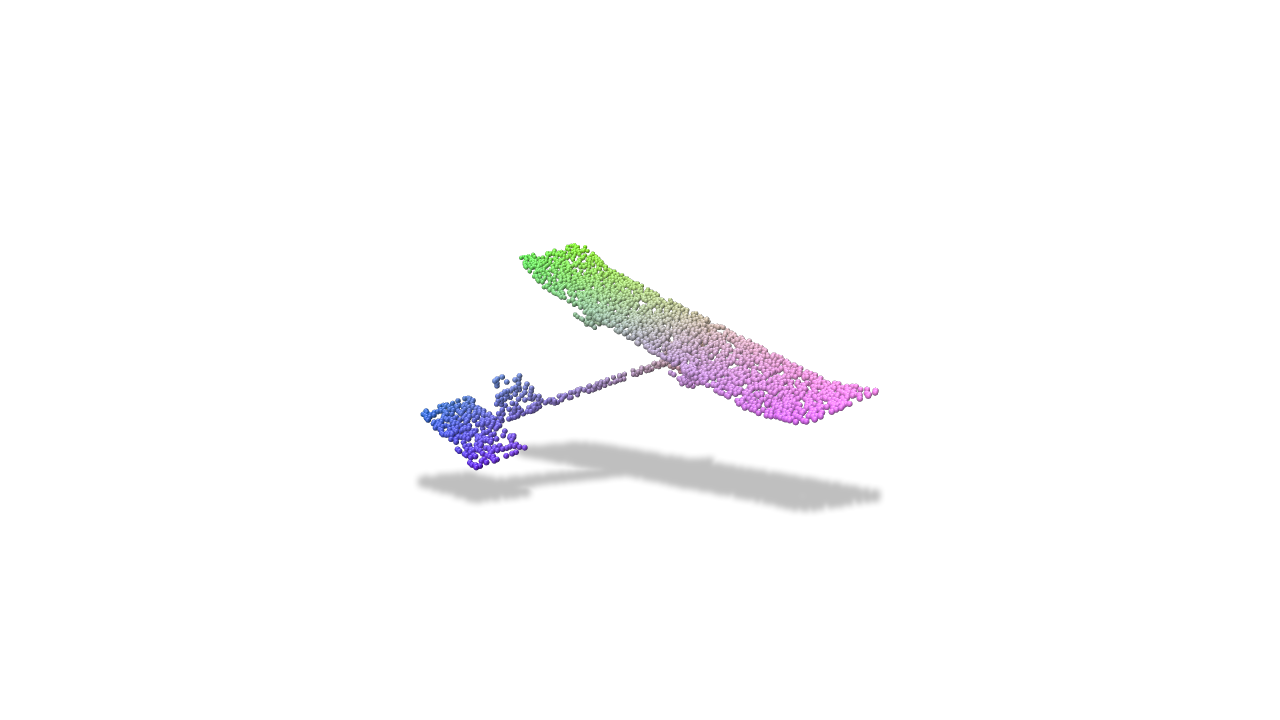} 
	\adjincludegraphics[height=\aalh,trim={ {\ach\width} {\cuthch\height} {\ach\width}  {\cuthch\height}},clip]{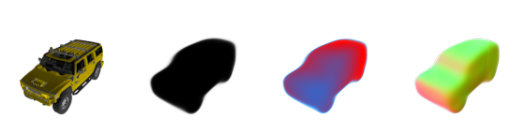}
	\adjincludegraphics[height=\alh,trim={ {\cch\width} {\cuthch\height} {\cch\width}  {\cuthch\height}},clip]{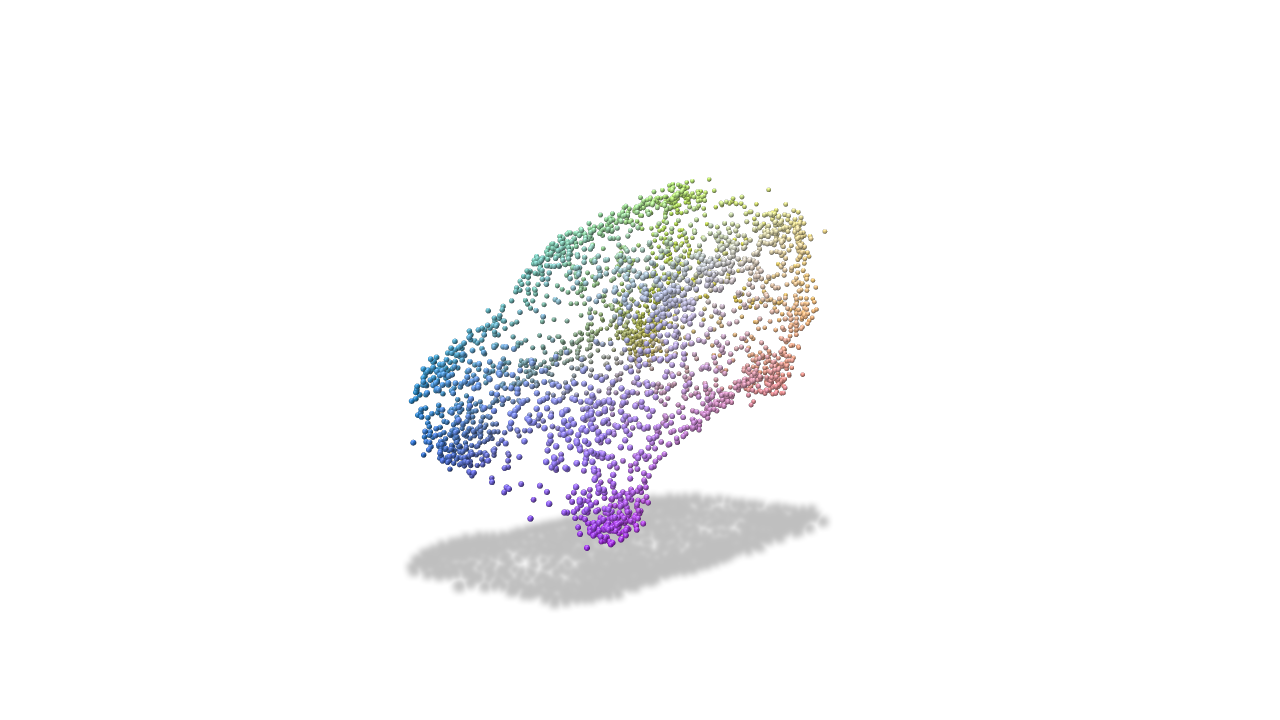}\hfill
	\adjincludegraphics[height=\alh,trim={ {\cch\width} {\cuthch\height} {\cch\width}  {\cuthch\height}},clip]{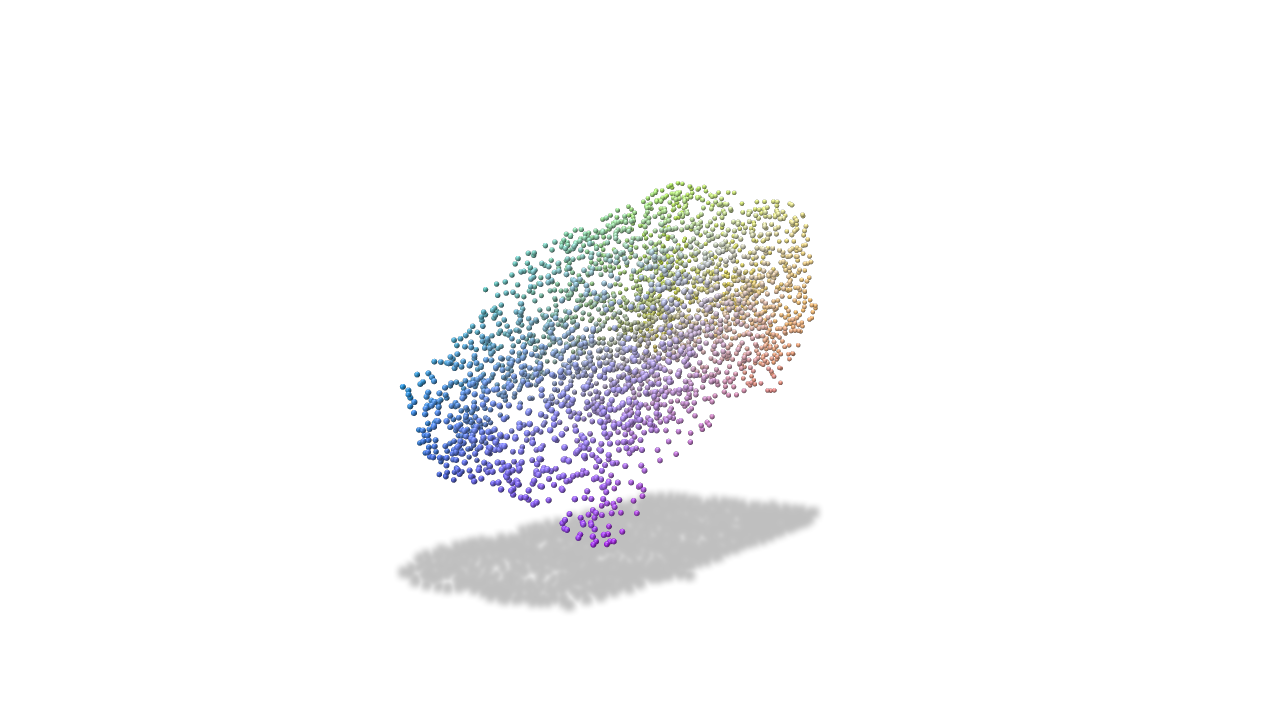}\hfill %
	\adjincludegraphics[height=\aalh,trim={ {\ach\width} {\cuthch\height} {\ach\width}  {\cuthch\height}},clip]{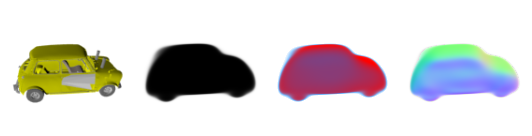}
	\adjincludegraphics[height=\alh,trim={ {\cch\width} {\cuthch\height} {\cch\width}  {\cuthch\height}},clip]{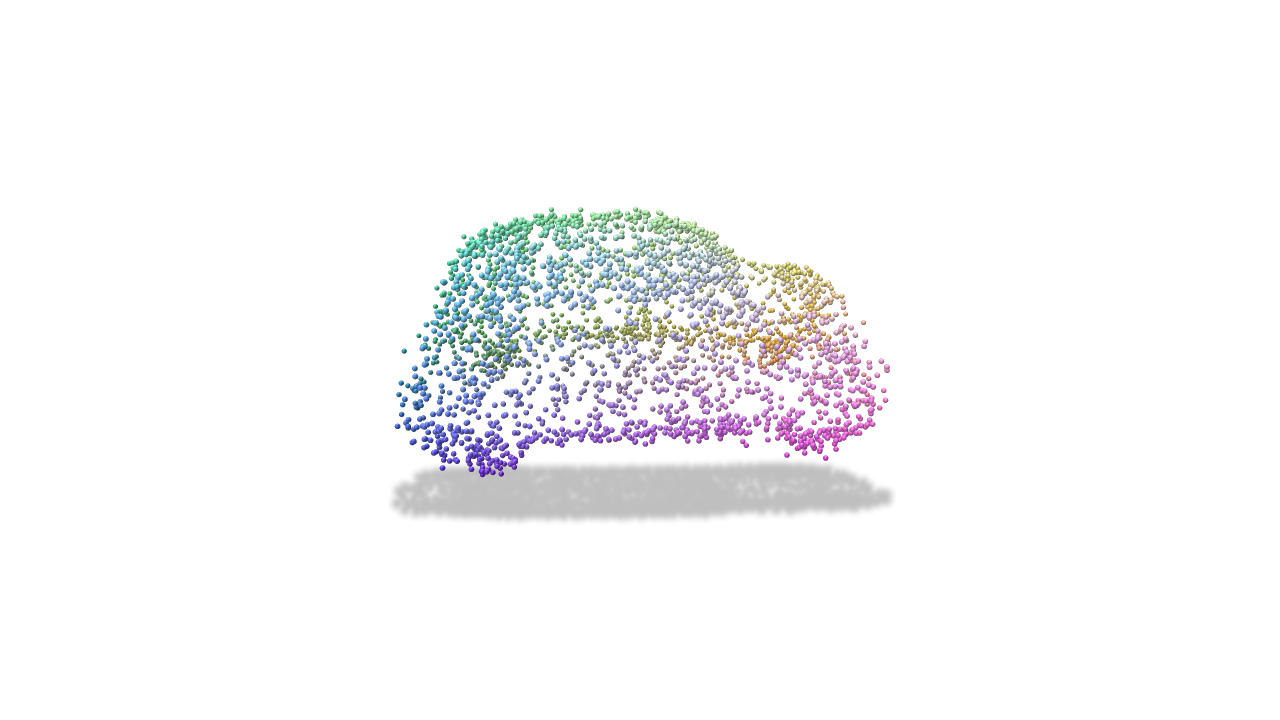}\hfill
	\adjincludegraphics[height=\alh,trim={ {\cch\width} {\cuthch\height} {\cch\width}  {\cuthch\height}},clip]{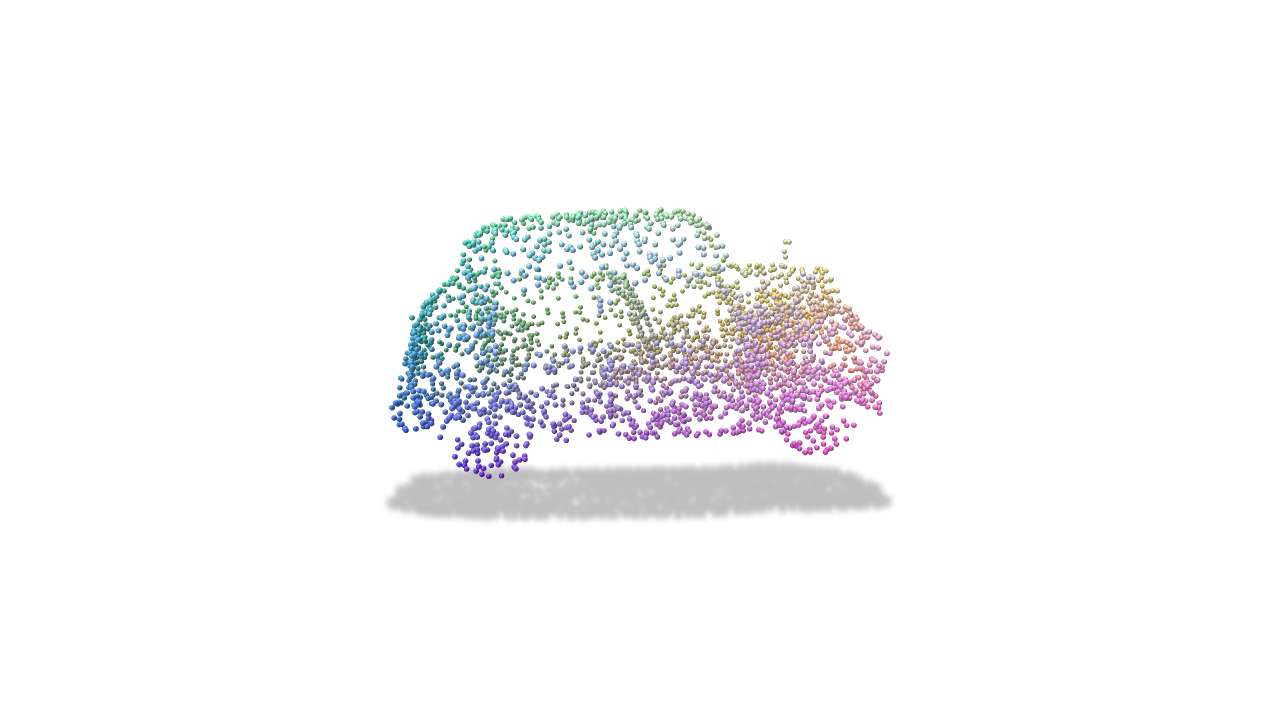} \\
	\adjincludegraphics[height=\aalh,trim={ {\ach\width} {\cuthch\height} {\ach\width}  {\cuthch\height}},clip]{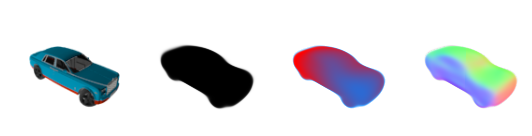}
	\adjincludegraphics[height=\alh,trim={ {\cch\width} {\cuthch\height} {\cch\width}  {\cuthch\height}},clip]{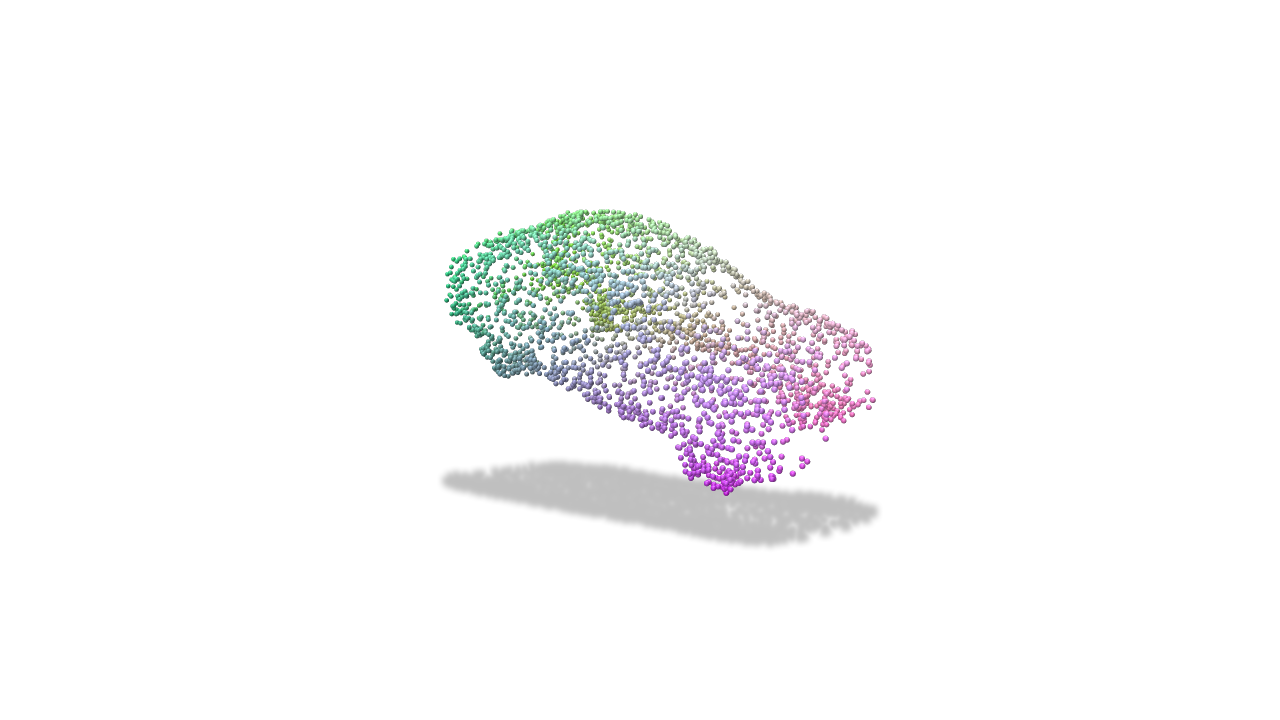}\hfill
	\adjincludegraphics[height=\alh,trim={ {\cch\width} {\cuthch\height} {\cch\width}  {\cuthch\height}},clip]{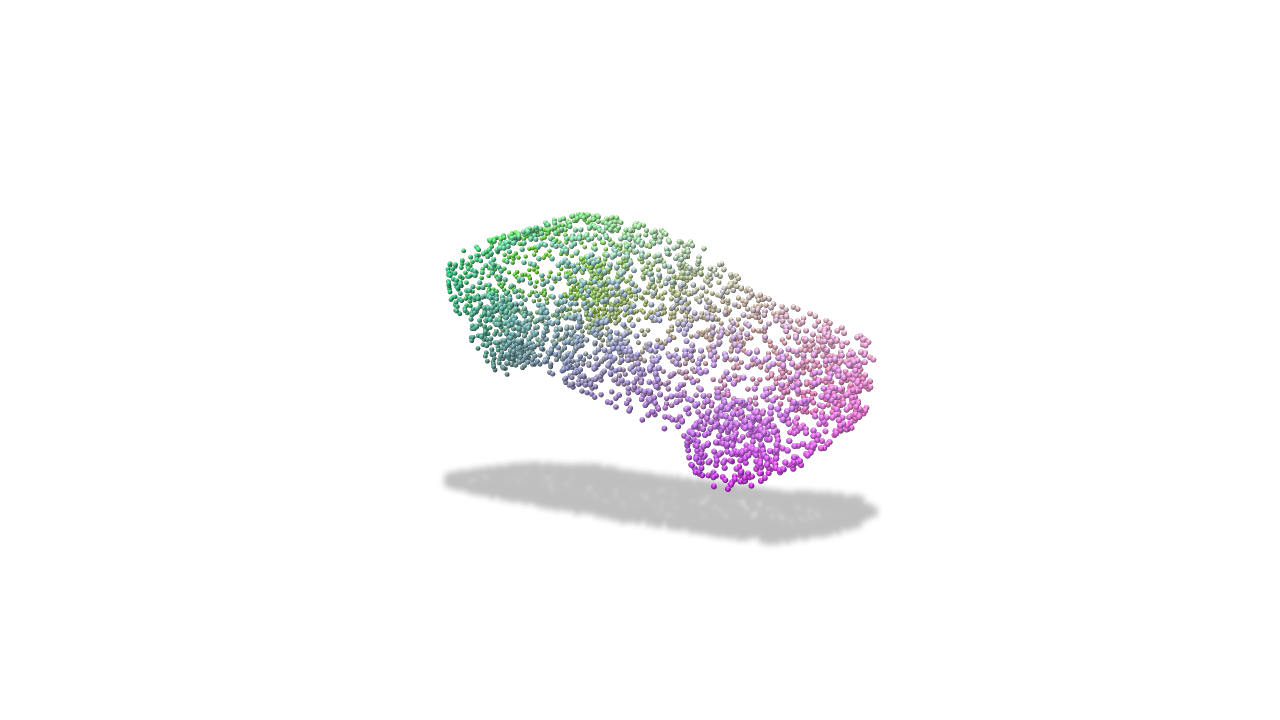}\hfill %
	\adjincludegraphics[height=\aalh,trim={ {\ach\width} {\cuthch\height} {\ach\width}  {\cuthch\height}},clip]{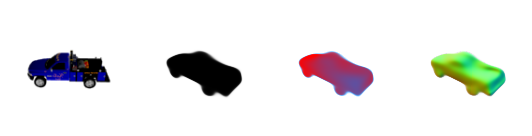}
	\adjincludegraphics[height=\alh,trim={ {\cch\width} {\cuthch\height} {\cch\width}  {\cuthch\height}},clip]{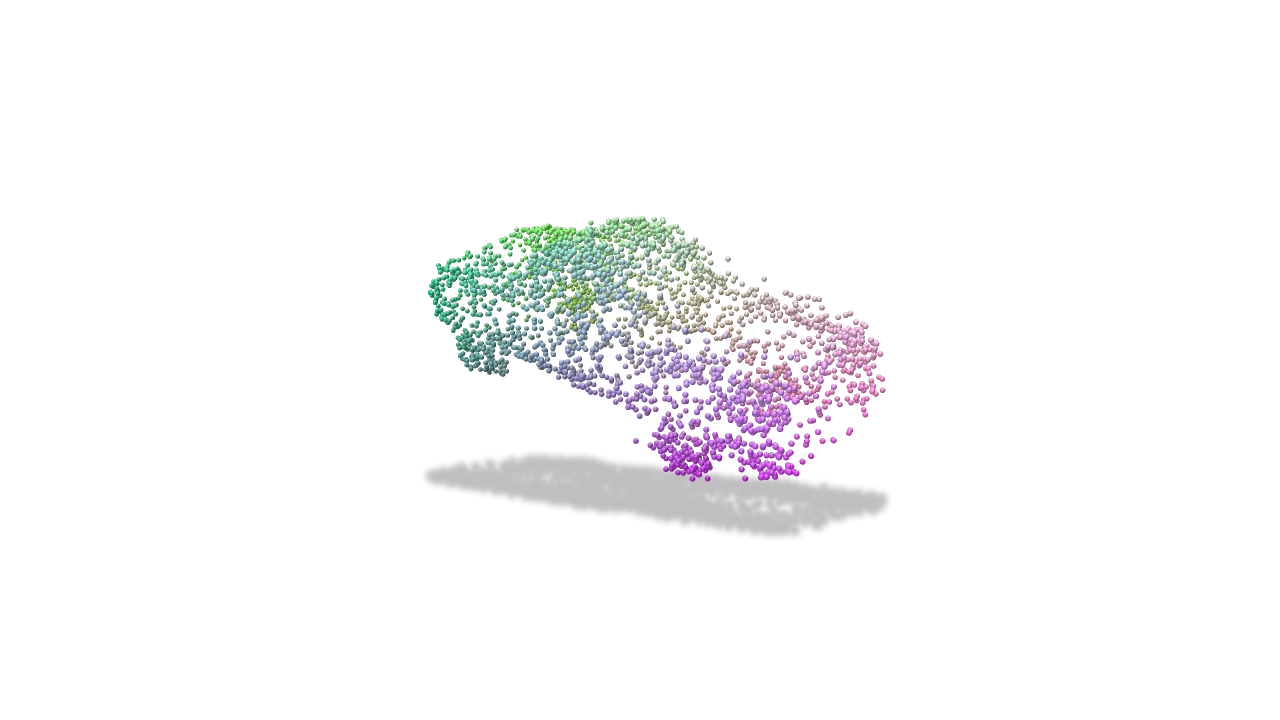}\hfill
	\adjincludegraphics[height=\alh,trim={ {\cch\width} {\cuthch\height} {\cch\width}  {\cuthch\height}},clip]{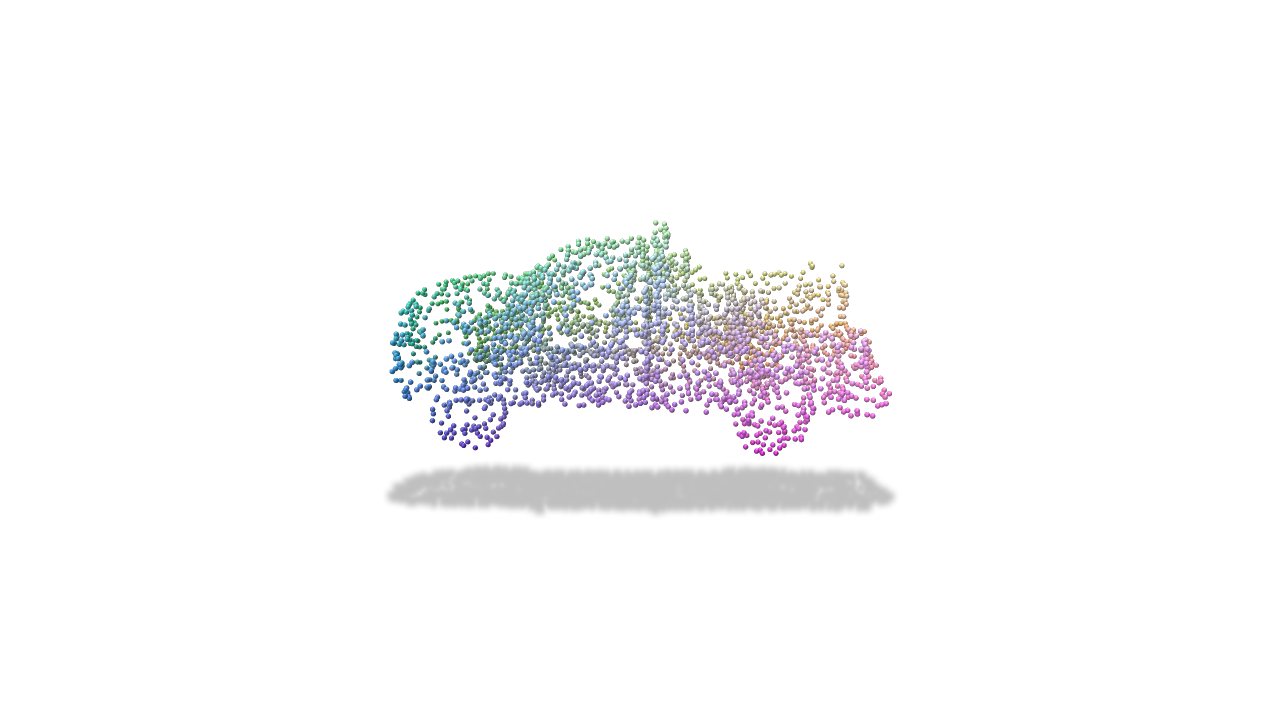}
	\caption{
		Single-image 3D reconstruction visualizations on held-out test data. 
		Columns represent 
		(i) the input RGB image, 
		(ii) the visibility $\widehat{\xi}$, 
		(iii) the depth $\widehat{d}$, 
		(iv) the normals $\widehat{n}$, 
		(v) the sampled point cloud (PC) from the DDF, and 
		(vi) a sample from the ground-truth PC.
		Quantities (ii-v) are all differentiably computed from the CPDDF and $\widehat{\Pi}$, per point or pixel, without post-processing.
		PC colours denote 3D coordinates.
		A high-error example is in the lower-right of each category.
	}
	\label{fig:si3drvis}
\end{figure*}

\subsubsection{Results}
We consider cars, planes, and chairs from ShapeNet \cite{shapenet2015}, using the data from %
\cite{choy20163d} (as in \cite{wang2018pixel2mesh}).
In Tables \ref{tab:si3dr:avgs} and \ref{tab:si3dr:summarytable},
we show DDFs perform comparably to the architecture-matched PC-SIREN baseline. 
See Fig.\ \ref{fig:si3drvis} 
and Supp.~Fig.~\ts{\ref{app:fig:si3drvis}}{18}
for visualizations. %
Generally, the inferred DDF shapes correctly reconstruct most inputs, including thin structures like chair legs, regardless of topology.
The most obvious errors are in pose estimation, but
the DDF can also sometimes output ``blurry'' shape parts when it is uncertain 
(e.g., for relatively rare shapes).
However, results can be improved 
by correcting $\widehat{\Pi}$ to $\widehat{\Pi}_\nabla = \argmin_{\Pi} \mathcal{L}_M $, via gradient descent 
(starting from $\widehat{\Pi}$)
on the input image alpha channel.
While explicit modalities 
can be differentiably rendered 
(e.g., \cite{kato2018neural,tulsiani2017multi,eldar}), 
DDFs can do so by construction,
without additional heuristics or learning.
Further,  %
(i) the DDF sampling procedure is not learned,
(ii) our model is not trained with $D_C$ (on which it is evaluated),
and 
(iii) DDFs are more versatile than PCs, 
enabling higher-order geometry extraction, built-in rendering, and  PC extraction.
Thus, changing from PCs to DDFs can enrich the representation without quality loss.

Compared to the other baselines,
DDFs with predicted $\widehat{\Pi}$ underperform P2M, but outperform 3DR.
Results with the ground-truth $\Pi_g$ indicate much of this error is due to camera prediction, 
though this is not directly comparable to P2M or 3DR. %
With $\Pi_g$, 
we are predicting in the object-centric,
rather than camera-centric, frame.
While each frame 
has benefits and downsides \cite{shin2018pixels,tatarchenko2019single},
in our case it is useful to 
separate shape vs.\ camera error.
Our scores with $\Pi_g$ suggest DDFs can infer shape at similar quality levels to existing work, despite the naive architecture and sampling strategy. 
We remark that we do not expect DDFs to 
directly compare to 
highly tuned, specialized models at the state-of-the-art.
Instead, we show that DDFs
can achieve good performance 
(especially for shape alone),
even using simple off-the-shelf components (ResNets and SIREN MLPs), without losing versatility. %

\begin{figure}[t] %
	\centering
	\begin{minipage}{0.49\textwidth}
		\centering
		\adjincludegraphics[width=0.99\textwidth,trim={{.14\width} {.19\height} {.14\width}  {.19\height}},clip]{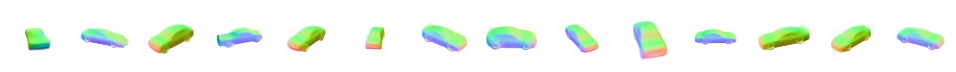}\hfill%
		\adjincludegraphics[width=0.99\textwidth,trim={ {.14\width} {.19\height} {.14\width}  {.19\height}},clip]{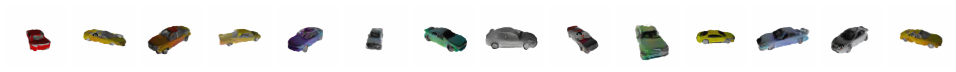}\hfill%
		\adjincludegraphics[width=0.99\textwidth,trim={ {.07\width} {.19\height} {.21\width}  {.19\height}},clip]{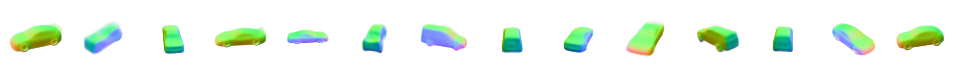}\hfill%
		\adjincludegraphics[width=0.99\textwidth,trim={ {.07\width} {.19\height} {.21\width}  {.19\height}},clip]{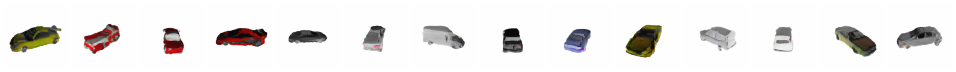}\hfill%
	\end{minipage}\hfill%
	\begin{minipage}{0.48\textwidth} %
		\centering
		\adjincludegraphics[width=0.99\textwidth,trim={ {.04\width} {.21\height} {.02\width}  {.21\height}},clip]{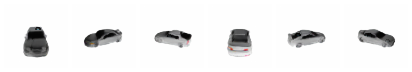}\hfill%
		\adjincludegraphics[width=0.99\textwidth,trim={ {.04\width} {.21\height} {.02\width}  {.21\height}},clip]{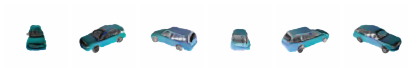}\hfill%
	\end{minipage} %
	\caption{
		Above: CPDDF-based VAE and image GAN samples.
		Below: views with fixed texture and shape. 
	}
	\label{fig:gangens}
\end{figure}

\subsection{Generative Modelling with Unpaired Data} %
\label{sec:genmodel}
We also apply CPDDFs to 3D-aware generative modelling,
using 2D-3D unpaired data 
(see, e.g., \cite{aumentado2020cycle,kaya2020self,miyauchi2018shape,zhu2018visual}). 
This takes advantage of 3D model data, 
yet avoids requiring paired data.
We utilize a two-stage approach: %
(i) a CPDDF-based variational autoencoder (VAE) \cite{kingma2014auto} on 3D shapes,
then
(ii) a generative adversarial network (GAN) \cite{goodfellow2014generative}, 
which convolutionally translates CPDDF-derived surface normal renders into colour images. %
Briefly, the VAE trains a PointNet encoder \cite{qi2017pointnet} and 
CPDDF decoder, while
the GAN performs cycle-consistent image-to-image translation \cite{CycleGAN2017} (from normals to RGB).
Fig.~\ref{fig:gangens} displays some example samples, 
	while Table~\ref{pddf:gentable} shows quantitative results.
While this underperforms a 2D image GAN 
with the same critic, 
it still outperforms samples from image VAEs or GAN-based textured low-poly 3D mesh renders \cite{aumentado2020cycle} in image quality.
See Supp.~\S\ts{\ref{appendix:genmodel}}{E} for details.

\section{Path Tracing with Intrinsic Appearance DDFs}
\label{sec:pathtracing}

\begin{figure*}
	\includegraphics[width=0.49\textwidth]{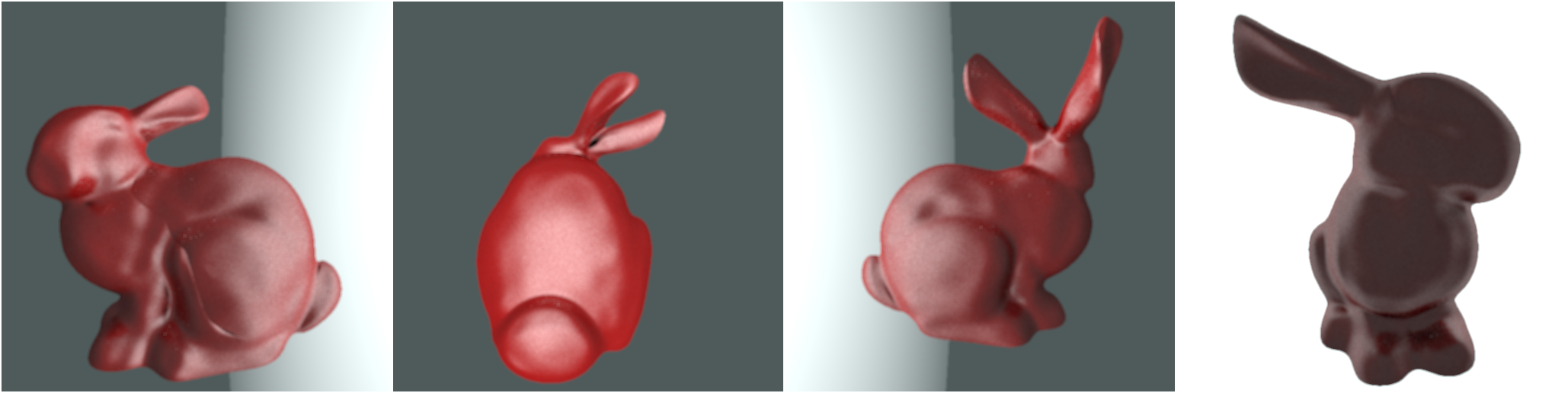} \hfill
	\includegraphics[width=0.49\textwidth]{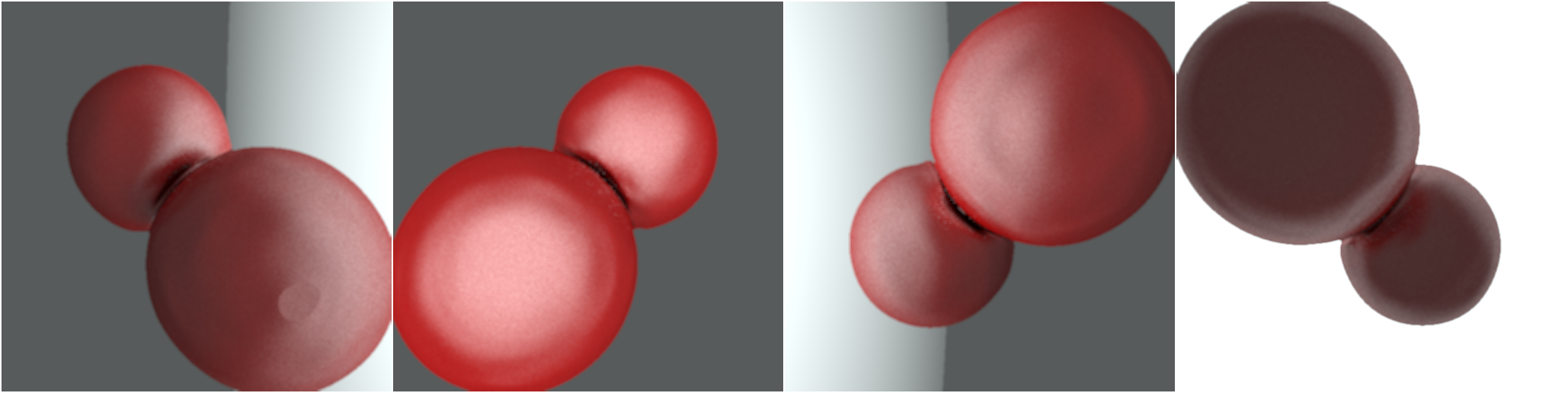} \\
	\includegraphics[width=0.49\textwidth]{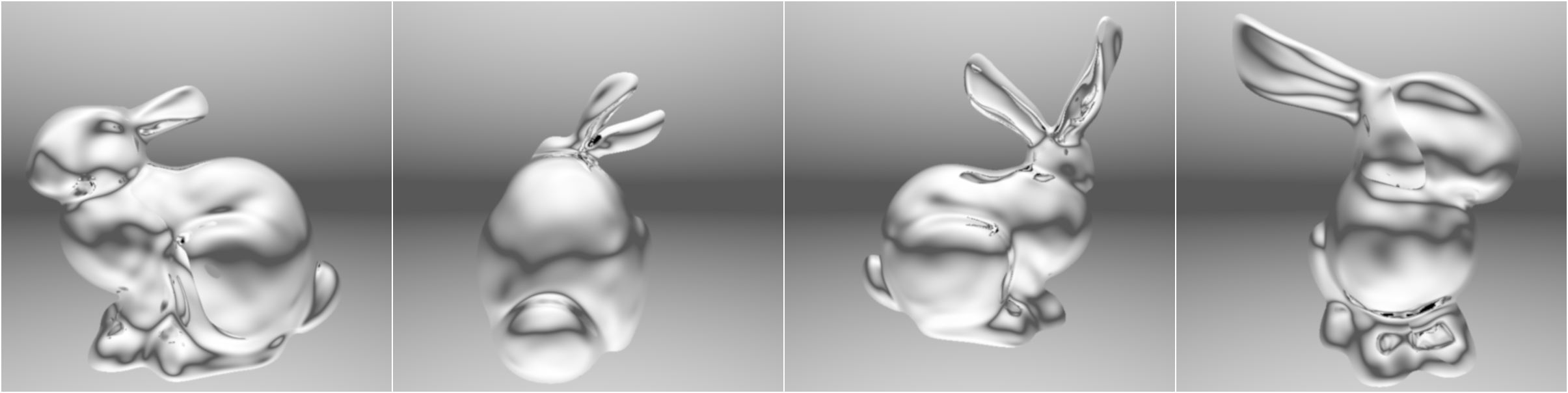} \hfill
	\includegraphics[width=0.49\textwidth]{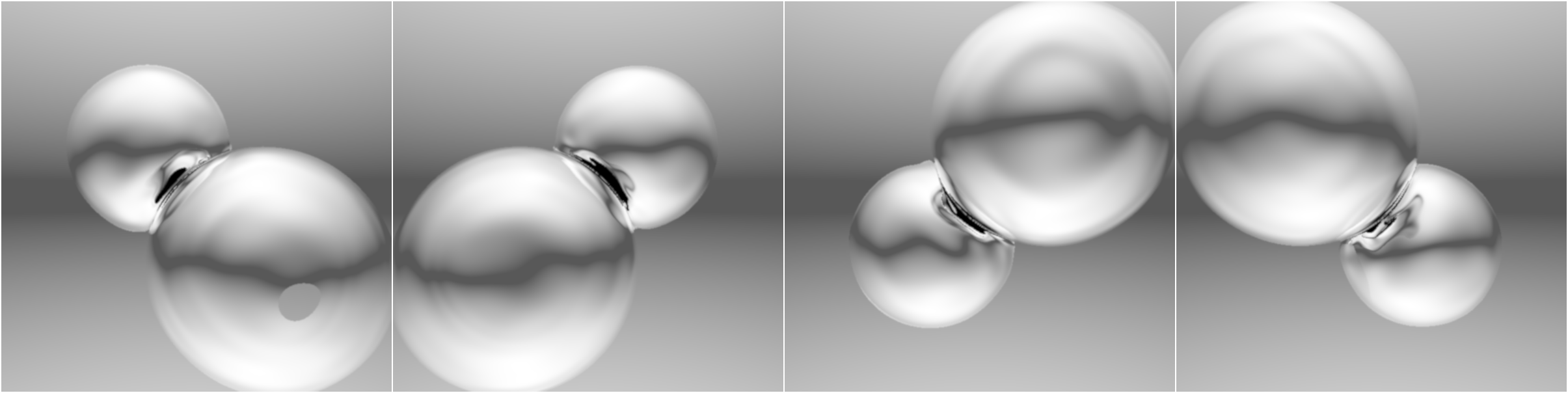} \\
	\includegraphics[width=0.49\textwidth]{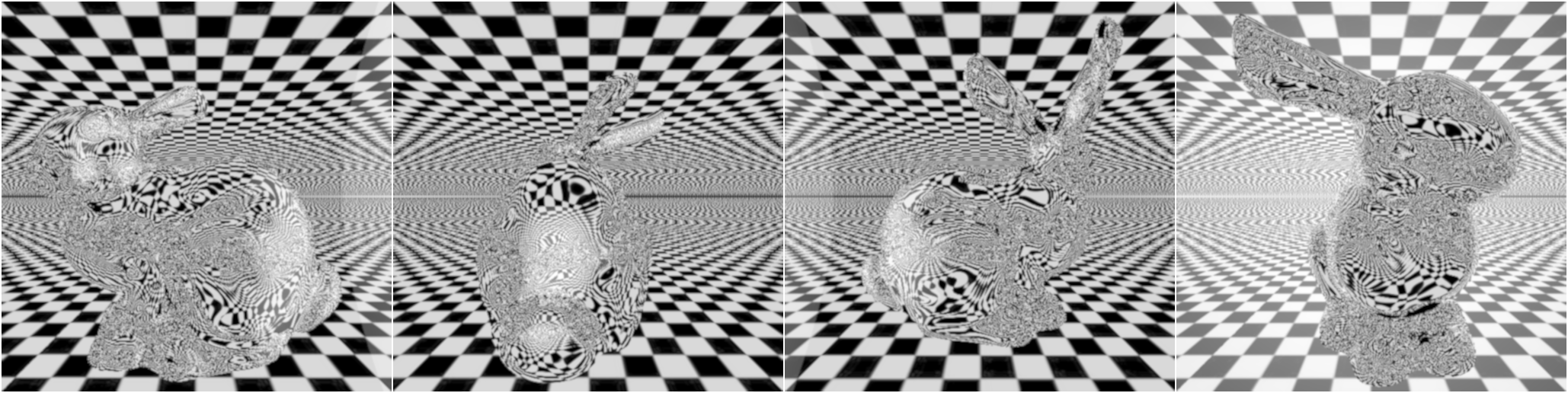} \hfill
	\includegraphics[width=0.49\textwidth]{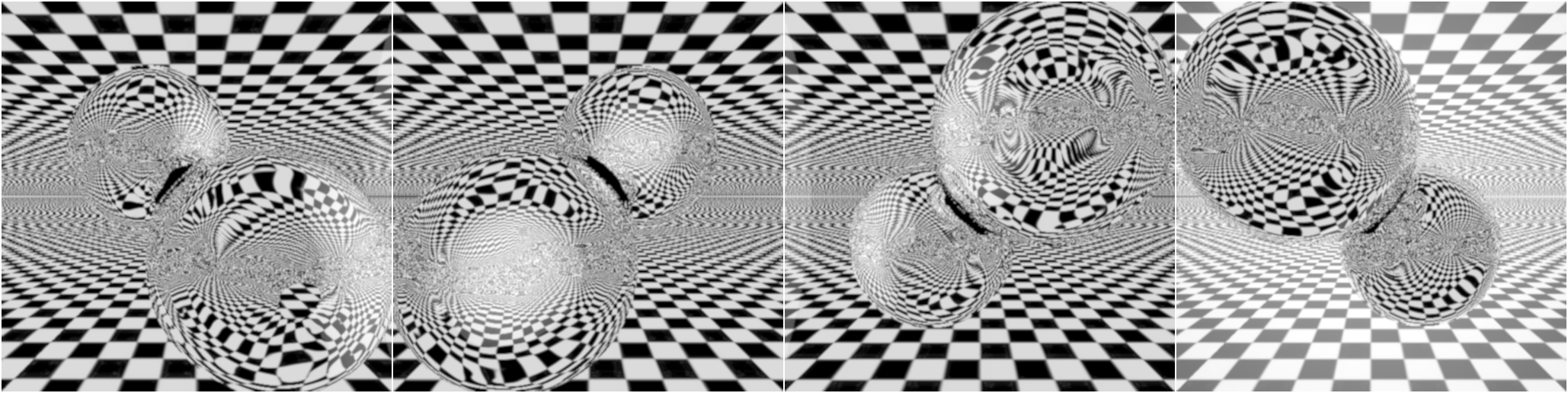} \\%
	\includegraphics[width=0.49\textwidth]{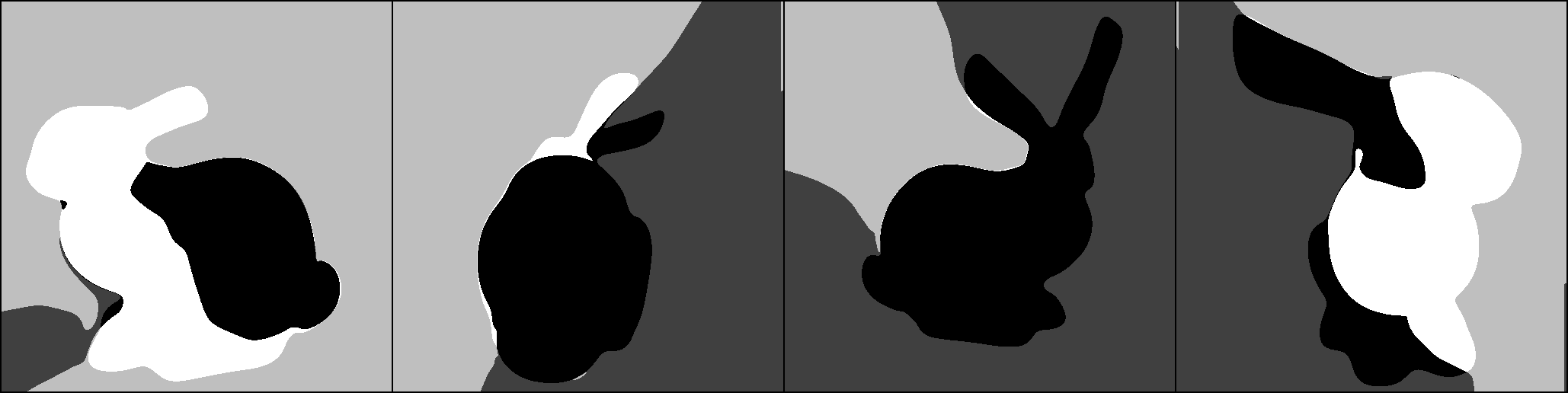} \hfill
	\includegraphics[width=0.49\textwidth]{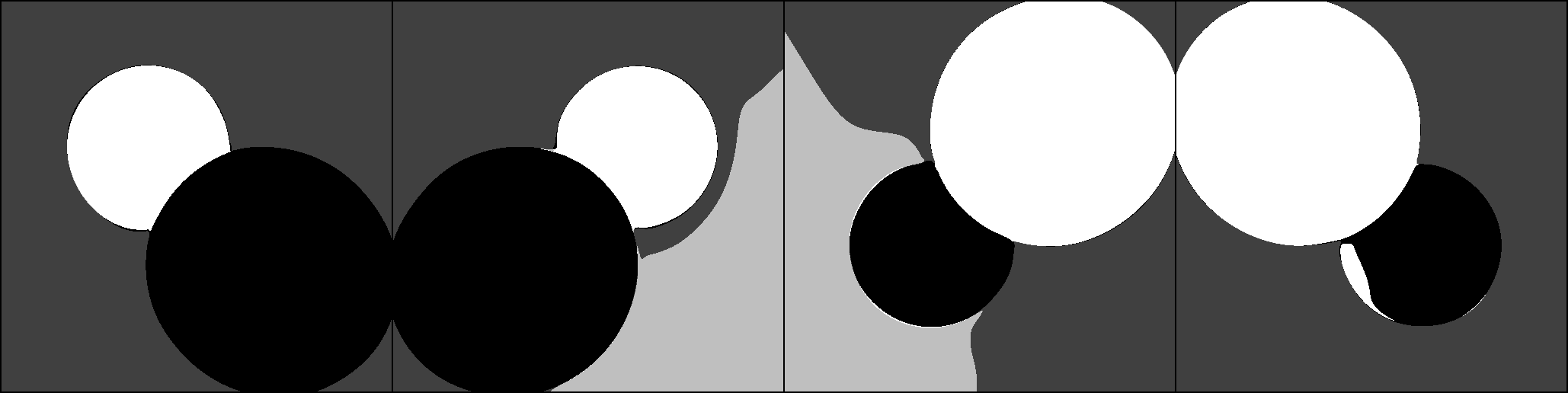}
	\caption{
		Examples of path-traced PDDFs under various lighting and material conditions.
		To better illustrate the lighting conditions,
		for rays that do not intersect the shape (i.e., $\xi = 0$), 
		we return the environmental lighting associated to that direction.
		Rows 1 and 2-3 show glossy and mirror materials, respectively,
		while row 4 shows the weight field,
		as in Fig.~\ref{fig:prob_illus}.
		The contribution of multibounce lighting can be seen, 
		for instance, in the bunny's ear reflected off its back 
		(row 2, column 3).
		See also Supp.\ Fig.~\ts{\ref{fig:pt:glow}}{25} 
		for ``glowing'' 3D shapes (i.e., with non-zero $M_L$), where the geometry itself can be a light source.
	} \label{fig:pathtraced}
\end{figure*}

A natural interpretation of a DDF is as the field of all possible depth images for a given scene.
However, this extends to \textit{any} ray (or camera) start point.
In particular, this includes \textit{inter-surface} rays: 
	a DDF provides the ``next'' surface that the ray 
	would intersect. 
Importantly, this corresponds to a fundamental operation when modelling
	the physics of \textit{light}: surface inter-reflections. 
In other words, DDFs can also be viewed as modelling
the set of all light paths through a scene,
where each path segment corresponds to a DDF call.
Thus,
any light path is equivalent to 
recursively applying a DDF.
Given this connection to light transport, 
we therefore explore how to integrate DDFs
into a path tracing framework.

Such an approach confers several benefits.
First, %
	a major focus in computer vision has been the disentanglement of images into constituent components
	(e.g., \cite{barrow1978recovering,adelson1996perception}): 
	geometry (shading), intrinsic appearance (reflectance), and illumination (lighting).
A DDF, combined with a reflectance and lighting model, naturally fits within this decomposition.
Second, many shape representations are not as  amenable to light transport.
For instance, the standard {NeRF} \cite{mildenhall2020nerf} acts as 
a radiance-emitting ``cloud of particles''. 
As such, appearance remains entangled.
In contrast, the interpretation of recursive DDF calls
		as inter-surface light transport 
	is naturally applicable to path tracing.

As a proof of concept, 
we consider a scenario with simplified materials and lighting,
to demonstrate path tracing of a DDF.
	(see Supp.~\S\ts{\ref{pddf:appendix:light}}{F} for details).
We augment a PDDF $(\xi,d)$ with two additional models.
The first is the \textit{material appearance}, 
	given by a Bidirectional Reflectance Distribution Function (BRDF) 
	$f_B : \real^3 \times \mathbb{S}^2 \times \mathbb{S}^2 \rightarrow \real^{|\mathcal{C}|}_+$,
	where $\mathcal{C}$ is the colour channel set,
	and an importance sampler
	$\Psi : \real^3 \times \mathbb{S}^2 \rightarrow \mathbb{P}[\mathbb{S}^2]$,
	where $\mathbb{P}[\mathbb{S}^2]$ is the set of distributions on $\mathbb{S}^2$,
	which stochastically decides how light bounces from a surface.
The second is the \textit{lighting model}, 
	which includes an emission field
	(i.e., glow),
	$M_L : \real^3 \times \mathbb{S}^2 \rightarrow \mathcal{C}$, 
and
an environment light, 
	$E_L : \mathbb{S}^2 \rightarrow \mathcal{C}$, 	
	which represents incoming radiance from a faraway source.
The combined model, which we call the intrinsic appearance DDF (IADDF) is 
$(\xi,d,f_B,\Psi,M_L,E_L)$.

We integrate the IADDF into a simple path tracing algorithm.
Briefly, for each pixel, 
we cast a ray into the scene, finding the scene point $q$ and normal $n$ via the DDF. 
Based on $\Psi$, we sample new outgoing directions from 
$q$, 
and use the DDF to find the next surface in that direction.
These ``bounces'' are the segments of a light path, being built backwards from the eye to the light source.
This continues until the ray misses any geometry (i.e., $\xi = 0$),
returning $E_L$ in that direction.
The final pixel irradiance corresponds to this value 
(attenuated by the material $f_B$ through the bounces),
plus any emissions (from $M_L$) encountered along the way. 
All surface intersections and normals are computed from the DDF.

We display example renders in Fig.~\ref{fig:pathtraced}.
As this is only a proof-of-concept of the forward rendering process,
	we implement $(f_B,\Psi,M_L,E_L)$ as analytic functions,
	while $(\xi,d)$ are fit to single shapes (see \S\ref{sec:results:singlefieldfitting}).
We utilize simple glossy and mirror materials 
	to showcase the accuracy of the DDF geometry
	(as accurate depths, visibilities, and normals are all 
		necessary for visual plausibility).
We see that not only are the ``first-order'' values accurate 
	(i.e., those generated by a standard initial ray-casting through the image pixels), 
	but the values computed at the surface (for a bounce) are sufficiently accurate for the tracing process to work.
However, certain pathologies in the underlying {PDDF} geometry are apparent: 
	the weight field seam that manifests as a ``cut''
		(e.g., column one),
	difficulty with occlusion
		(e.g., ``black spots'' where the two spheres meet),
	and inaccuracies in the normals
	(e.g., waviness in the reflected ``horizon'' on the second-row spheres).
Nevertheless, this appears to be a promising approach to combining
	disentangled appearance modelling with differentiable path tracing.
We also emphasize that \textit{internal structure modelling} 
(see also \S\ref{sec:internalstructure}), 
which is ignored by similar contemporary methods 
(e.g., \cite{sitzmann2021light,zobeidi2021deep}),
is essential --
without it, inter-surface bounces cannot be computed.

%% file: theory.tex
\newcommand{\bo}{\mathcal{B}}
\newcommand{\boe}{\mathcal{B}_\varepsilon}
\newcommand{\mbo}{$\mathcal{B}$}
\newcommand{\mxi}{$\xi$}

\begin{figure}
	\centering
	\includegraphics[width=0.95\linewidth]{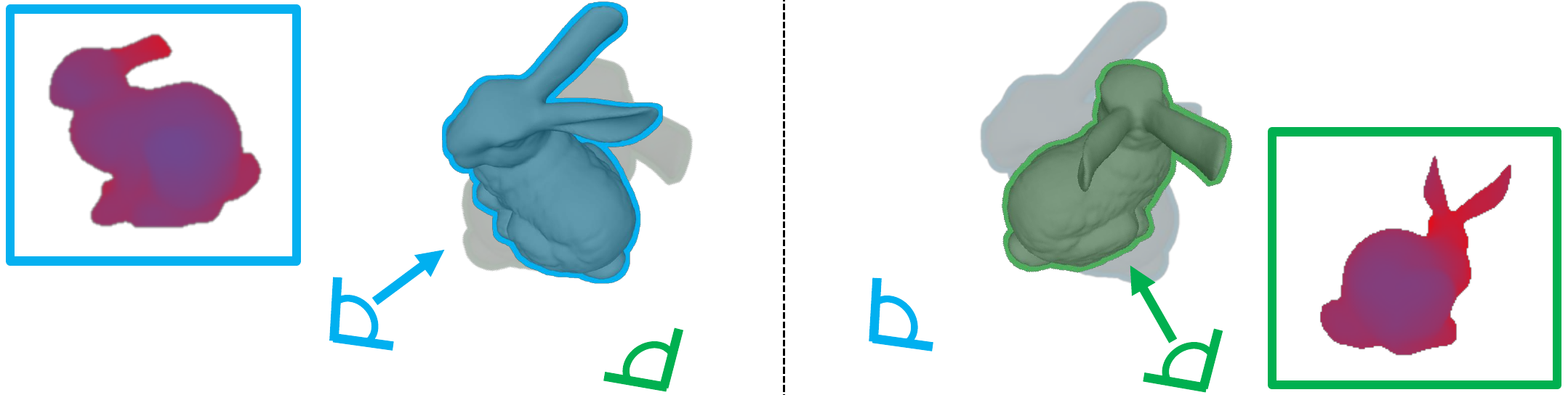}
	\caption{
		\tempp{An illustration of a view-\textit{in}consistent DDF. Due to its 5D nature, depth-rendering the same field from different viewpoints can produce incompatible geometries.}
	} \label{fig:illusincon}
\end{figure}

\section{Theory: The Geometry of View Consistency}
\label{sec:theory}

The defining characteristic of a DDF is its five-dimensional nature.
While one naturally associates 3D shapes with 3D constructs, 
	the computational advantages of DDFs stem fundamentally 
	from the increase in dimensionality.
However, the tradeoff in this case is the lack of theoretical guarantees of \textit{view consistency} (VC).
Specifically, unlike representations that are inherently 3D, 
	DDFs require regularization to ensure predicted surface points 
	do not vary with view direction.
Thus, for a DDF implemented with a neural network,
	the question of how to enforce VC is of great practical importance.
In particular, 
	can we {guarantee} such an implementation is actually VC?

More precisely, consider an unconstrained differentiable 5D field.
Assuming noisy initialization, before fitting, such a field will 
	be akin to a cloud of noise, without sensible structure across viewpoints. 
\tempp{Fig.~\ref{fig:illusincon} also provides an example of inconsistent geometry.}
Clearly, such entities do \textit{not} represent any reasonable notion of a 3D shape.  
Under what conditions, then, can we say such a field \textit{does} represent a 3D shape?
	
In this section, we answer these questions through
a comprehensive theoretical investigation of 
(i) when a DDF is VC and 
(ii) show that this is conceptually equivalent to the DDF being an accurate representation of some 3D shape.
We show that a straightforward set of constraints on the field,
	which can be checked locally 
	(i.e., properties based on pointwise evaluations in 5D),
are sufficient to guarantee consistency. 
Due to space constraints, 
	we present a more complete story, with additional details and full proofs, 
	in Supp.~\S\ts{\ref{suppmat:theory}}{G} and \S\ts{\ref{suppmat:proofs}}{H}.

\subsection{Preliminaries and Notation}

We begin by defining the basic setting, including the domain of our fields, the definition of a shape, and the oriented points that characterize our ray-based geometric representation. 

\begin{mydef}{Domain}{pddf:prelim:domain} 
Let $\bo \subset \real^3$ be a convex and compact domain,
	where $\partial \bo$ is smooth with outward-facing surface normals.
Further, we define the 5D product space $\Gamma = \bo \times \mathbb{S}^2$.
\end{mydef}

We primarily operate in the ``interior'' of $\bo$, written $\boe$:
\begin{mydef}{$\varepsilon$-Domain}{pddf:prelim:edomain} 
For $\varepsilon > 0$,
let $\boe$ be the $\varepsilon$-domain of $\bo$, defined via 
\begin{equation}
	\boe = \left\{ p \in \bo \,\middle\vert\, 
	\min_{b\in\partial\bo} || p - b || \geq \varepsilon \right\}.
\end{equation}
As for $\bo$, we define
 $\Gamma_\varepsilon = \boe \times \mathbb{S}^2$.
\end{mydef}

The role of $\boe$ is to enable us to define a notion of 
``away from the boundary''
(i.e., $\bo\setminus\boe$ is a thin $\varepsilon$-width shell around the edges of the domain).
Our shapes will thus reside in $\boe\subset\real^3$,
but our field will operate in 5D ray-space (i.e., $\Gamma$).

\begin{mydef}{Oriented Points and Rays}{pddf:prelim:orienrays}
	We denote an \textit{oriented point} via
	$\tau = (p,v) \in \Gamma$. %
	Any oriented point induces a 3D \textit{ray}: $r_\tau(t) = p + tv,\, t\geq 0$.
	In a slight abuse of notation, we may refer to the $\tau$ and $r_\tau$ forms of such 5D elements interchangeably. 
\end{mydef}

Fundamentally, our interest is in representing 3D shapes, 
which are defined simply as follows.

\begin{mydef}{Shapes}{pddf:prelim:shape}
	We define a \textit{shape} to be a compact set $S\subset \bo$.
\end{mydef}

Usually, we will be interested in shapes 
$S \subset \boe$,
for $\varepsilon > 0$.
Note that using $\boe$ does not strongly constrain the shape: for example, one can simply use the bounding sphere of the given point set, and then inflate it enough to satisfy the $\varepsilon$ condition.

Next, for notational simplicity, we consider restrictions of functions to a fixed ray.

\begin{mydef}{Along-Ray Functions}{pddf:prelim:alongray}
	We define the ``along-ray'' form of a function
	$ g : \Gamma \rightarrow \mathcal{X} $,
	which maps into a set $\mathcal{X}$, to be:
	\begin{equation}
		f_g(s\mid \tau) := g(p + sv, v) = g(r_\tau(s), v),
	\end{equation}
	where $s \geq 0$ and $\tau = (p,v) \in \Gamma$.
\end{mydef}

Finally, we denote intersections of rays and point sets via:
\begin{mydef}{Intersecting Rays and Point Sets}{pddf:prelim:intersections}
	Consider $S\subseteq \bo$ and $\tau\in\Gamma$.
	
	\textbullet~\textit{Intersected Points}.
	Let $S_\tau\subseteq S$ denote the points in $S$ intersected by $r_\tau$
	(i.e., $q\in S_\tau$ iff $\,\exists\; t\geq 0$ such that $r_\tau(t) = q$).
	For clarity, we may write $[S]_\tau$ as well.
	
	\textbullet~\textit{Intersecting Rays}.
	Let $\mathcal{I}_S \subseteq \Gamma$ be the set of rays 
	that intersect $S$ %
	(i.e., $\tau \in \mathcal{I}_{S}$ if $\exists\; q\in S$ such that $r_\tau(s) = q$ for some $s\geq 0$).
	For $q\in\bo$, we denote $ \mathcal{I}_{q} := \mathcal{I}_{\{q\}} $.
\end{mydef}

\subsection{View Consistency for Visibility Fields}

First, we consider when a visibility field (VF) alone will be VC. We do so by defining an \textit{un}constrained field, and then devising constraints that can ensure VC.

\begin{mydef}{BOZ Field}{pddf:vis:bozf:main} \index{BOZ field}
	Let $\xi : \Gamma \rightarrow \{ 0, 1 \}$. 
	We restrict $\xi$ to have an open zero set; i.e.,
	$\forall \; \tau \in \Gamma$, if $\xi(\tau) = 0$, then $\exists\; \varepsilon > 0$ such that $\xi(\widetilde{\tau}) = 0 \;\forall\; \widetilde{\tau} \in B_{\varepsilon}(\tau) \cap \Gamma $, where $B_{\varepsilon}(\tau)$ is the open ball centered at $\tau$ of radius $\varepsilon$.
	We call any such \underline{b}inary field, with an \underline{o}pen \underline{z}ero set, a BOZ field.
\end{mydef}

Without further conditions, a BOZ field does not have an obvious connection to 3D shapes or their multiview silhouettes (i.e., most BOZ fields do not represent a consistent 3D point set from all viewpoints).
Hence, our goal is to understand \textit{when} (or under what conditions) a BOZ field acts as a continuous representation of the silhouettes of some coherent shape
(i.e., when it assigns every ray a binary indication 
	as to whether a shape point exists along that ray or not).
Thus, we next define three conditions that we will show impose consistency on $\xi$.  
We will find a close analogy between these constraints and the ones needed for the depth field within the DDF.

Define 
$\mathcal{O}[V,U] := \{ \tau=(p,v) \in\Gamma \,|\, p\in V,\, r_\tau(s) \notin U \,\forall\, s\geq 0 \}$ 
as the rays \textit{starting} in $V$ that \textit{miss} $U$.
Hence, $\mathcal{O}[\bo\setminus\boe, \boe]$ are the ``outward rays''
from $\bo\setminus\boe$.
\begin{mycdef}{BC$_\xi$}{pddf:vis:bcxi}
	A BOZ field $\xi$ satisfies the 
	\textit{Non-Visible Boundary Condition},
	denoted BC$_\xi$, if
	$\xi(\tau) = 0 \;\forall\; \tau\in 
	\mathcal{O}[\bo\setminus\boe, \boe]$.
\end{mycdef}
In words, BC$_\xi$ demands that any rays that (i) start close to the boundary and (ii) do not intersect the inner domain $\boe$ must be non-visible. 
Intuitively, if $\xi$ were to represent a shape $S\subset\boe$, then
BC$_\xi$ prevents $S$ from being visible on rays 
that start outside of $\boe$ and ``look away'' from it.

\begin{mycdef}[]{DE$_\xi$}{pddf:vis:dexi}
	A BOZ field $\xi$ satisfies the \textit{Directed Eikonal Constraint} DE$_\xi$ 
	if
	$\xi$ is always non-increasing along a ray
	(i.e., $ f_\xi(s_1|\tau) \leq f_\xi(s_2|\tau) \;\forall\; \tau \in \Gamma, s_1 > s_2$).
\end{mycdef}
Intuitively, DE$_\xi$ demands that visibility can only ever go from ``seeing'' to ``not seeing'' along a ray
(i.e., one should not suddenly see a new point in the shape become visible).

The last constraint requires characterizing field behaviour at specific special points.
In particular, we are interested in $q\in\bo$ where $\xi$ ``flips'' along some direction $v\in\mathbb{S}^2$.
So, let us first consider the 
field value as one approaches $q$ 
along $v$:
\begin{align}
	\mathcal{F}_{\xi,v}[\mathrm{Op},\varsigma](q) &= 
	\lim_{\epsilon\downarrow 0} 
	\underset{s\in(0,\epsilon)}{\mathrm{Op}}
	\xi(q + \varsigma sv, v), 
\end{align}
where $\mathrm{Op}\in \{ \inf, \sup \}$ 
and $\varsigma\in\{ -1, +1 \}$.
Let
$ \mathcal{F}_{\rightarrow,\xi,v}(q) :=
	\mathcal{F}_{\xi,v}[\sup,-1](q) $
and
$ \mathcal{F}_{\leftarrow,\xi,v}(q) :=
	\mathcal{F}_{\xi,v}[\inf,1](q) $. 
Intuitively, $\mathcal{F}_{\rightarrow,\xi,v}$ 
is the maximum value as one moves forward along a ray, 
while $\mathcal{F}_{\leftarrow,\xi,v}$ 
is the minimum as one moves backward.
A ``flip'' therefore occurs when these two do not match at a point.
In particular, a ``one-to-zero flip'' along $v$ occurs when
\begin{equation}
	\mathcal{C}_{\mathrm{Flip}}(q|\xi,v) :=
	\left( \mathcal{F}_{\rightarrow,\xi,v}(q) = 1 \right)
	\,\land\, 
	\left( \mathcal{F}_{\leftarrow,\xi,v}(q) = 0 \right).
	\label{eq:flip}
\end{equation}
Our field constraint uses this concept as follows:

\begin{mycdef}{IO$_\xi$}{pddf:vis:ioxi}
	A BOZ field $\xi$ satisfies the \textit{Isotropic Opaqueness Constraint} (IOC) at $q\in\bo$ iff
	the following holds:
	\begin{align}
		\big(				
		\exists\; v\;\mathrm{s.t.}\;
		\mathcal{C}_{\mathrm{Flip}}(q|\xi,v)
		\big)
		\,\implies\, &
		\big(
		\xi(\tau) = 1
		\;\forall\;\tau \in \mathcal{I}_q 
		\big), \nonumber
	\end{align}
	where $\mathcal{I}_q $ is the set of rays in $\Gamma$
	that intersect $q$.
	An \textit{IO field} (IO$_\xi$) satisfies the IOC everywhere.	
\end{mycdef}

In words, if there is \textit{any} direction 
along which the field flips from one to zero,
then the point at which that occurs must be ``isotropically opaque''
(i.e., from any direction, any ray that hits that point must produce a visibility value of one).
Intuitively, IO$_\xi$ says that 
such ``one-to-zero`` discontinuities in $\xi$ are special. 
As we shall see, they act similar to surface points.
In particular, for any point $q\in\bo$, 
if even a single direction $v$ exists such that 
a one-to-zero discontinuity occurs along $v$ at $q$, 
then \textit{any} ray that goes through $q$ must also be visible.    

\begin{figure*}
	\centering
	\includegraphics[width=0.97\textwidth]{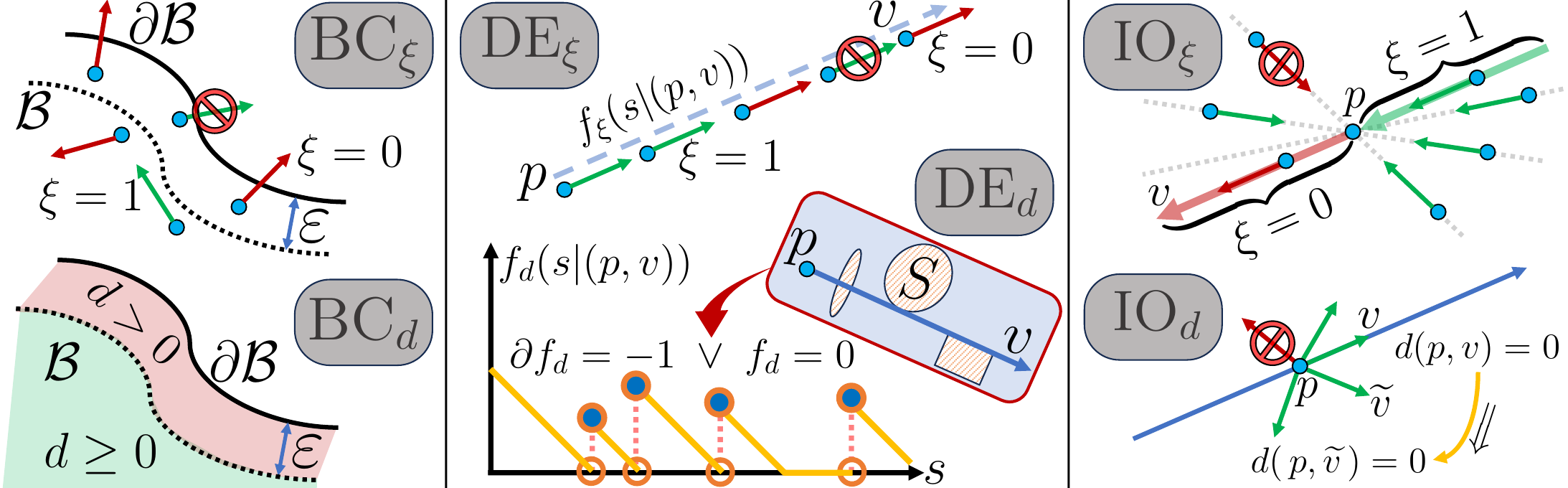}
	\caption{
		The fundamental constraints that link visibility and depth fields to being shape representations.
		The boundary conditions (BC) demand that 
		$d$ does not predict geometry too close 
		to $\partial\bo$ and that 
		$\xi=0$ whenever it looks ``outward''.
		For the directed Eikonal (DE) constraints,
		we ask that 
		$f_d$ decreases at unit rate %
		and 
		$f_\xi$ never increases. 
		Finally, isotropic opaqueness (IO) asks that 
		zeroes in $d$ are zeroes 
		from all directions,
		while one-to-zero flips in $\xi$ 
		must be visible from all directions.  
	} \label{fig:theory:constraints}
\end{figure*}

With these three constraints 
(see Fig.~\ref{fig:theory:constraints} for an illustration),
we can now define a subset of BOZ fields, 
called \textit{visibility fields} (VFs),
which are intimately tied to shapes.

\begin{mydef}{Visibility Field (VF)}{pddf:vis:visibilityfields:main}
	We define a Visibility Field to be a BOZ field such that
	IO$_\xi$, BC$_\xi$, and DE$_\xi$ are satisfied.	
\end{mydef}

\tempp{
Conceptually, VFs 
(i) bound visible geometry within a finite domain,
(ii) prevent new geometry from ``appearing out-of-nowhere'' 
	(when it should have been visible earlier), and 
(iii) enforce that ``flip positions'' (Eq.~\ref{eq:flip}) are visible from all viewpoints. 
Such discontinuities are effectively \textit{surface points}, where visibility ``flips'' because one has moved \textit{through} the surface. 
These special positions link the 5D field to a 3D shape.
}

\begin{mydef}{Positional Discontinuities of VFs}{pddf:vis:posdiscont:main}
	The positional discontinuities of a VF $\xi$ are defined by
	\begin{align}
		Q_\xi :=
		\closure & %
		\left(
		\left\{ 
		q \in \bo
		\;\middle\vert\;
		\exists\; v\;\mathrm{s.t.}\; 
			\mathcal{C}_{\mathrm{Flip}}(q|\xi,v)
		\right\}
		\right), %
	\end{align}  
	where $\closure$ is the closure operator on sets.
\end{mydef}
Intuitively, given a VF $\xi$, the set $Q_\xi$
simply contains all of the one-to-zero discontinuities in $\xi$.
More specifically, %
if there exists some direction $v$ such that the field $\xi$
flips from one to zero 
(i.e., from ``seeing something'' to ``seeing nothing'') 
at $q\in\bo$,
then $q\in Q_\xi$.
In Supp.~Lemma~\ts{\ref{lma:pddf:vis:posdiscont}}{G.2}, 
we show that $Q_\xi \subseteq \boe$ is a shape.
Hence, the discontinuities $Q_\xi$
are a good candidate for a potential shape that
$\xi$ is implicitly representing.
First, 
we need to define what it means for a shape $S$
to \textit{induce} a $\xi$-field 
(i.e., for $\xi$  to ``exactly'' represent $S$).\footnote{See Supp.~Def.~\ts{\ref{df:pddf:vis:uppercont}}{G.17} for an explanation of notation at discontinuities.}

\begin{mydef}{Shape-Induced Binary Fields}{pddf:vis:sibfs:main}
	A BOZ field $\xi$ is \textit{induced} by a shape $S\subseteq\boe$ iff
	\begin{equation}
		\xi(\tau) = 1 \;\iff\; \tau \in \mathcal{I}_{S}.
	\end{equation}
	where
	$\mathcal{I}_{S} \subseteq\Gamma$ is the set of rays 
	that intersect $S$.
	Further, given a shape $S$, its induced field $\xi$ is unique.
\end{mydef}

Notice that the induction relation is an ``iff'', 
meaning $\xi(\tau) = 0$ implies that \textit{no} $q\in S$
can be present along the ray $r_\tau$.
In addition, importantly, while the $S$-induced field $\xi$ is unique,
the inducing shape for a given $\xi$ is \textit{not} necessarily unique.
For instance, imagine a sphere with another shape (say, another sphere) inside it -- such a shape will induce the same VF as using the outermost sphere alone for the inducement. 
Given this definition,
our first result in this section shows that any shape-induced field 
satisfies the constraints of a VF.
\begin{restatthm}{Induced Binary Fields Are VFs}{pddf:vis:sibfavf:main}
	Let $S\subseteq \boe$ be a shape, with an induced field $\xi$.
	Then $\xi$ is a Vf 
	(i.e., IO$_\xi$, BC$_\xi$, and DE$_\xi$ hold).
	Also, the positional discontinuities of $\xi$ satisfy $Q_\xi \subseteq S$.
\end{restatthm}

So far, 
we have shown that any shape can be used to generate or induce 
a VF (in the sense of Def.~\ref{df:pddf:vis:visibilityfields:main}).
This defines a notion of ``shape representation'' for such fields.

\begin{mydef}{Shape {\Xirepname}s as VC VFs}{pddf:vis:shaperep}
	A VF $\xi$ is \textit{View Consistent} (VC) iff there exists a shape $S$ 
	such that $S$ induces $\xi$.
	We can therefore call  
	any such $\xi$ a \textit{shape} \textit{{\xirepname}} for the point set $S$.
\end{mydef} \index{shape indicator}
Note we call it a shape indicator instead of a representation because we cannot necessarily reconstruct $S$ from $\xi$ (i.e., it would be an incomplete representation). Instead, we call it an {\xirepname}, because it ``indicates'' whether or not a shape exists along a given ray
(i.e., acts as a per-ray indicator function). 
We next show that the local constraints of a visibility field
are sufficient for it to be a representation of some shape.
\begin{restatthm}{Every VF is a Shape \Xirepname}{pddf:vis:evfias:main}
	Let $\xi$ be a visibility field. 
	Then there exists a shape $S$ such that $\xi$ is induced by $S$.
	Further, $Q_\xi$ induces $\xi$.
\end{restatthm}

So far, we have shown a close duality between visibility fields (VFs)
and shape indicators (i.e., between binary fields with specific local constraints on the field, namely IO$_\xi$, DE$_\xi$, and BC$_\xi$, and binary fields constructed from a given point set).
However, it is clear that many inducing point sets can induce the same field.
We can therefore ask for a ``minimal example'' over an equivalence class of inducing point sets
(i.e., given many shapes $S$ that all induce the same $\xi$, which is the ``simplest''?), which we answer in the following corollary.

\begin{restatcor}{Minimal Characterization of VFs}{pddf:vis:minimalcharvis}
	Let $\mathfrak{S}[\xi]$ 
		be the equivalence class of $\xi$-inducing shapes
		(i.e., $ \mathfrak{S}[\xi] = \{ S \mid S\subseteq\bo \;\text{induces}\; \xi \}$).
	Then, $Q_\xi \in \mathfrak{S}[\xi]$ is the \textit{smallest} closed point set among all such inducers.
\end{restatcor}

Hence, $Q_\xi$ is special among shapes that induce $\xi$,
in that it forms a subset of any other shape that also induces $\xi$.
As an aside,  Supp.~\S\ts{\ref{appendix:theory:visfields}}{G-C}
provides another interpretation. %
For any $\xi$-inducing $S$, %
the subset of $S$ %
that is ``directly visible'' from $\partial\bo$
is equivalent to $Q_\xi$.
In other words, to get the minimal inducing points from $S$, we need not construct $\xi$; instead, we can ``look inwards'' from $\partial \bo$ and find the ``observable'' points.

In summary, %
(i) ``inducement'' defines how a shape $S$ gives rise to a VF $\xi$,
	which acts as its indicator along rays;
(ii) the constraints that define a VF $\xi$ 
	(i.e., IO$_\xi$, DE$_\xi$, BC$_\xi$) 
	are sufficient to guarantee that 
	\textit{some} point set must exist that induces $\xi$
	(i.e., every VF is the indicator of some shape); and
(iii) the positional discontinuities $Q_\xi$ are the 
	\textit{minimal} closed subset of $S$ that induces a given $\xi$.
Note, %
in Supp.~\S\ts{\ref{appendix:theory:sfsvhvf}}{G-C1}, 
we survey work in
shape-from-silhouette (e.g., \cite{laurentini1997many,richards1987inferring,cheung2005shape}),
visual hulls (e.g., \cite{laurentini1994visual}),
and space carving \cite{kutulakos2000theory}.

Nevertheless, questions still arise regarding VFs and their representational capacity.
One aspect of shapes ``missed'' by VFs is \textit{internal structure}:
shape points that are completely surrounded by other shape points cannot be preserved by the field.
In other words, given a shape $S$ with internal structure, 
any $S$-induced field will not be able to recover such structure
(hence the existence of $\mathfrak{S}[\xi]$). %
One can see this by noting that $Q_\xi$
will not include such internal structure.
These points, however, fully encode the \textit{field} (as they induce it).
Combining a VF with a distance field, as we will do next, enables a complete shape representation, including internal structure.

\subsection{View Consistency for Directed Distance Fields}

We now move on to the theory of DDFs, where we will combine a visibility field (VF) with a depth field.
Analogous to the previous section, 
we define view consistency (VC) through a notion of ``shape representation''; 
i.e., whether or not some shape (point set) exists that induces the DDF.
In this case, the VF will be used to control where 
(in $\Gamma$) 
the distance field needs to be constrained.\footnote{
	Note: in Supp.~\S\ts{\ref{appendix:theory:simpleddfs}}{G-B}, we also discuss a way to define distance fields without a VF, which can be constrained similarly.
} 
Specifically, starting from a VF $\xi$ (with its associated constraints), on visible rays ($\xi=1$), we derive constraints on a depth field $d$ that are analogous to those of the VF. 
With two additional constraints that ensure the compatibility between $\xi$ and $d$, we then show the resulting DDF $(\xi,d)$ is a VC shape representation. 

Analogous to the BOZ field (Def.~\ref{df:pddf:vis:bozf:main}),
we start with the following definition of a general non-negative field, which is generally not sufficiently constrained to be represent a shape.

\begin{mydef}{NNBC Field}{pddf:simple:nnbc} \index{NNBC field}
	An \textit{NNBC field} on $\Gamma$ is a \underline{n}on-\underline{n}egative, \underline{b}ounded scalar field that is piece-wise \underline{c}ontinuously differentiable along rays, 
	written $ d : \Gamma \rightarrow \real_{\geq 0} $.
\end{mydef} 

In general, NNBC fields need not have an obvious connection to any shapes.
We treat such fields as \textit{putative}  representations for shapes -- the goal is to understand the conditions under which such fields are ``equivalent to'' some shape. 
Such constrained fields should generate view-consistent depths. %

\begin{mycdef}{BC$_d$}{pddf:simple:bcd}
	An NNBC field $d$ satisfies the \textit{Positive Boundary Condition} BC$_d$ iff
	$\inf_{v\in\mathbb{S}^2} d(p,v) > 0 \;\forall\; p\in \bo\setminus\boe.$		
\end{mycdef}

Recall that $\bo\setminus\boe$ is the ``outer shell'' of 
$\bo$. %
In other words, BC$_d$ demands that $d$ cannot have any zeroes close to $\partial\bo$. 

Next, we denote the infimum approached by $d$ along $v$ via
\begin{equation}
	\mathcal{A}_{d}(q,v) = \lim_{\epsilon \downarrow 0} \inf_{s\in (0,\epsilon)}
	d(q - sv, v). 
	\label{theory:dinf}
\end{equation}

\begin{mycdef}{DE$_d$}{pddf:simple:ded}
	An NNBC field $d$ satisfies the \textit{Directed Eikonal Constraint} DE$_d$ if
	$\partial_s f_d(s|\tau) = -1$, except at along-ray zeroes  
	$(q,v)$, such that  
	$\mathcal{A}_{d}(q,v) = 0$.
\end{mycdef}

The second constraint is on the derivative of $d$:
it says that, along any ray, $d$ must decrease linearly, at unit rate, unless $d=0$ (a value at which it may potentially stay).
\tempp{
Intuitively, $d$ must act like a distance function, along any ray.
}

Next, %
we denote the $\inf$ approached over \textit{all} $v$'s via
\begin{equation}
	\mathcal{A}_{d}(q) = \lim_{\epsilon \downarrow 0} 
	\inf_{ \substack{s\in (0,\epsilon)\\ v\in\mathbb{S}^2 } }
	d(q - sv, v). 
	\label{theory:dinfv}
\end{equation}

\begin{mycdef}{IO$_d$}{pddf:simple:iod}
	An NNBC field $d$ satisfies the \textit{Isotropic Opaqueness Constraint} at a point 
	$q\in\bo$ if
	\begin{align}
		& \left(
		\mathcal{A}_{d}(q) = 0
		\right)
		\,\implies\, %
\left(
		\mathcal{A}_{d}(q,v) = 0
		\;\forall\; v\in\mathbb{S}^2\right).    
		\label{eq:iod:main}
	\end{align}
	We say $d$ is isotropically opaque, denoted IO$_d$, if it satisfies the Isotropic Opaqueness Constraint $\forall\;q\in\bo$.
\end{mycdef}

Notice that IO$_d$ is \textit{stronger} than (i.e., it implies) the following constraint:
for any $(q,v)\in\Gamma$,
\begin{align}
	&\left(
	\exists\,v  \;|\; 
	\mathcal{A}_{d}(q,v)
	= 0
	\right)
	\,\implies\, %
	\left(
	\mathcal{A}_{d}(q,v)
	= 0
	\;\forall\; v %
	\right).
\end{align}
Namely, if $d$ approaches zero at some position $q$ from \textit{any} direction $v$, then it must approach zero at $q$ from \textit{all} directions.

Together, these field constraints enable us to define a \textit{simple DDF}, which we show in Supp.~\ts{\ref{appendix:theory:simpleddfs}}{G-B} satisfies certain theoretical consistency guarantees \textit{without} the need for a VF.

\begin{mydef}{Simple DDF}{pddf:simple:simpleddf} \index{simple DDF}
	We define a simple DDF to be an NNBC field such 
	that IO$_d$, BC$_d$, and DE$_d$ are satisfied.
\end{mydef}

Notice the close analogy between the constraints for simple DDFs and VFs. Fig.~\ref{fig:theory:constraints} illustrates all six fundamental constraints.
While we do not dwell on simple DDFs alone, they enable us to define the following special set of points.

\begin{mydef}{Positional Zeroes of Simple DDFs}{pddf:simple:zeroes:main}
	We denote the \textit{positional zeroes} of a Simple DDF as
	\begin{equation}
		Q_d = 
		\left\{ 
			p\in\bo \;\middle\vert\; \mathcal{A}_{d}(q) = 0	
		\right\}.
	\end{equation}
\end{mydef}

For simple DDFs alone, $Q_d$ defines the shape represented by the field in a formal sense (see Supp.~Thm.~\ts{\ref{th:pddf:simple:esdiasr}}{G.2}). 
However, the positional zeroes will be fundamental to full DDFs as well.

Notice that the simple DDF constraints do not involve a VF;
hence, we next begin to link them.
This actually relaxes some requirements on $d$, 
as only \textit{visible} rays are constrained.

\begin{mydef}{Visible and Non-Visible Ray Sets}{pddf:full:vnvrs}
	Let $\xi$ be a visibility field.
	We define the sets of visible and non-visible rays, respectively, as
	\begin{align}
		\visrays[\xi]    &= \{ \tau \in \Gamma \mid \xi(\tau) = 1 \} \\
		\nonvisrays[\xi] &= \{ \tau \in \Gamma \mid \xi(\tau) = 0 \}. 
	\end{align}
\end{mydef}

We next limit the simple DDF constraints to $\visrays[\xi]$.
Define $\mathcal{A}_{d,\xi}(q)$
as the \textit{visible} infimum of $d$ at $q$:
$$
\mathcal{A}_{d,\xi}(q) = 
\lim_{\epsilon \downarrow 0} 
\inf_{ \substack{s\in (0,\epsilon)\\ 
		v\in\mathbb{S}^2 \,\mid\, \xi(q,v) = 1 } }
d(q - sv, v).
$$
\begin{mydef}{$\xi$-Coherent Simple DDFs}{pddf:full:xicoherent}	
	Given a VF $\xi$,
	an NNBC field $d$ is a $\xi$-coherent simple DDF
	iff it satisfies %

		(IO$_{d,\xi}$) 
		\textit{Isotropic opaqueness on visible rays.}
		\begin{align*}
			\forall\;
			(q,v)\in\visrays[\xi]: & \;
			\mathcal{A}_{d,\xi}(q) = 0
			\implies 
			\mathcal{A}_d(q,v) = 0.
		\end{align*}
		(BC$_{d,\xi}$)
		\textit{Positive boundary condition on visible rays.}
		\begin{equation*}
			\big(
			(p,v)\in \visrays[\xi] \,\land\, 
			p\in \bo\setminus\boe
			\big)
			\,\implies\,
			d(p,v) > 0 .
		\end{equation*}
		
		(DE$_{d,\xi}$)
		\textit{Directed Eikonal constraint on visible rays.}
		\begin{equation*}
			(\tau\in\visrays[\xi]
			\,\land\,
			d(\tau) > 0)
			\,\implies\,
			\partial_s f_d(s\mid\tau)|_{s=0} = -1.
		\end{equation*}
\end{mydef}
In words, a $\xi$-coherent depth field $d$ satisfies the
IO, BC, and DE constraints
whenever $\xi = 1$ (i.e., it is a simple DDF on $\visrays[\xi]$).
Outside of the visible rays $\visrays[\xi]$, $d$ is essentially unconstrained.
These modified constraints allow us to define the following modification of the positional zeroes.

\begin{mydef}{Locally Visible Depth Zeroes}{pddf:full:lvdz:main}	
	Let $\xi$ be a visibility field and $d$ be $\xi$-coherent.
	Then the \textit{locally visible depth zeroes} (LVDZs) are given by
	\begin{equation}
		\lvz = 
		\closure
		\left(\left\{
		\widetilde{q} \in Q_d  
		\; \middle\vert \;
		\exists\, v %
		\;\mathrm{s.t.}\;
		\xi(\widetilde{q}, v) = 1
		\right\}\right),
	\end{equation}
	where 
	$\closure$ is the closure operator. 
\end{mydef}
The LVDZs thus include any depth zero
(i.e., $q\in Q_d$)
that is ``locally visible'' along some direction. 

\begin{figure*}
	\centering
	\includegraphics[width=0.98\textwidth]{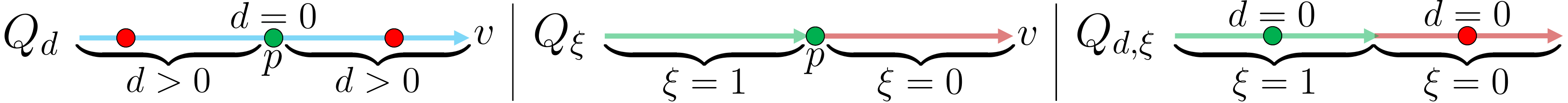}
	\caption{
		Illustration of the fundamental point sets of a DDF.
		The positional depth zeroes, $Q_d$ (Def.~\ref{df:pddf:simple:zeroes:main}), occur when $d(\tau) = 0$ for some $\tau=(p,v)$. 
		The positional visibility discontinuities, $Q_\xi$ (Def.~\ref{df:pddf:vis:posdiscont:main}), demarcate 
		where the field flips from one (``seeing something'') to zero (``seeing nothing'').
		The locally visible zeroes, $Q_{d,\xi}$ (Def.~\ref{df:pddf:full:lvdz:main}), mark positions that 
		(i) are zeroes of $d$ (i.e., in $Q_d$) and
		(ii) are ``visible'' along some direction, at an infinitesimal distance.
		See also Fig.~\ref{fig:theory:fps2}.
	} \label{fig:theory:fps}
\end{figure*}

\begin{figure}
	\centering
	\includegraphics[width=0.475\textwidth]{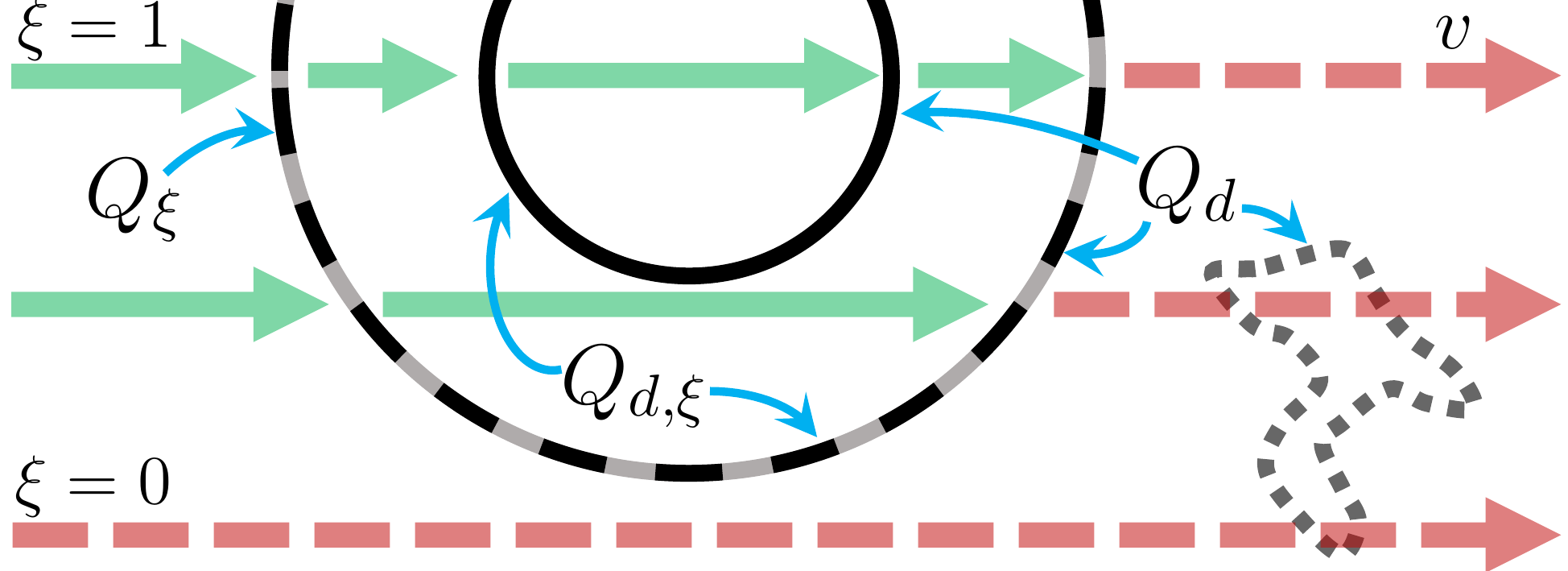}
	\caption{
		Example of the fundamental DDF point sets.
		The VF is shown with green/solid ($\xi=1$) and red/dashed ($\xi=0$) arrows.
		Additional \textit{non}-visible points
		are on the right (grey dots).
		All such points are part of $Q_d$; 
		however, $Q_\xi$ includes only the outer circle, 
		where $\xi$ flips.
		In contrast, 
		the LVDZs, $Q_{d,\xi}$, 
		encompass \textit{both} the inner and outer circle.
		For DDFs,
			$Q_\xi \subseteq Q_{d,\xi} \subseteq Q_d$ 
		(see Supp.~Lemma~\ts{\ref{lma:pddf:full:fundps}}{G.3}).
	} \label{fig:theory:fps2}
\end{figure}

We now have three ``fundamental point sets'' associated to a DDF,
	illustrated in Figs.~\ref{fig:theory:fps} and \ref{fig:theory:fps2}.
The LVDZs ($\lvz$) are analogous 
to the zeroes $Q_d$ for simple DDFs and the discontinuities $Q_\xi$ for VFs, defining \textit{which shape} is represented.

\begin{figure}
	\centering
	\includegraphics[width=0.475\textwidth]{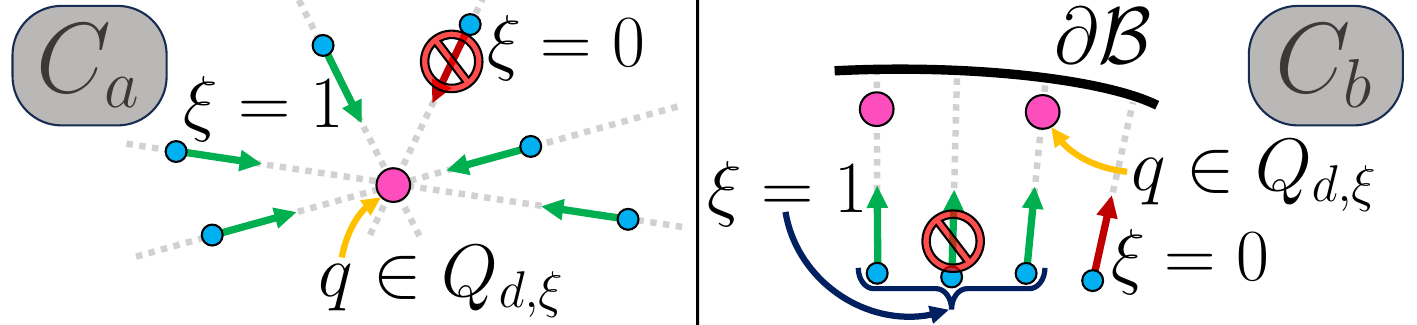}
	\caption{
		Illustration of inter-field compatibility (see Con.\ Def.~\ref{cdf:pddf:full:compat}), which
		ensures rays are visible iff they intersect the LVDZs
		(Def.~\ref{df:pddf:full:lvdz:main}).
		See Fig.~\ref{fig:theory:constraints} for intra-field constraints.
	} \label{fig:theory:compat}
\end{figure}

First, note that $\xi$-coherence merely enforces $d$ to be ``internally'' consistent (i.e., a simple DDF along visible rays).
Hence, it is \textit{not} sufficient to ensure $(\xi,d)$ actually represents a shape.
Two additional constraints are needed, which ensure $\xi$ and $d$ are compatible
(see Fig.~\ref{fig:theory:compat}).
Both rely on the LVDZs. %

\begin{mycdef}{Compatibility}{pddf:full:compat}	
	A visibility field $\xi$ and simple DDF $d$ are \textit{compatible} iff
	
	(a)
	LVDZs are isotropically opaque with respect to $\xi$:
	\begin{equation}
		q \in \lvz
		\implies
		\tau \in\visrays[\xi] \;\forall\; \tau\in\mathcal{I}_q.
	\end{equation}

	(b)
	Every visible ray must hit an LVDZ:
	\begin{equation}
		\tau \in\visrays[\xi] \implies [\bo]_\tau \cap \lvz \ne \varnothing.
	\end{equation}

\end{mycdef}

This is finally enough to define a full DDF, which we will show is a VC shape representation.

\begin{mydef}{Directed Distance Field (DDF)}{pddf:full:fullddf:main}	
	A BOZ field $\xi$ and an NNBC field $d$ form a 
	\textit{Directed Distance Field} $ F = (\xi,d) $ iff:
	\begin{enumerate}
		\item 
		$\xi$ is a visibility field.
		\item 
		$d$ is a $\xi$-coherent Simple DDF.
		\item 
		$\xi$ and $d$ are compatible.
	\end{enumerate}
\end{mydef}

The definition of a DDF can be intuited as combining a VF with a simple DDF that is constrained only along visible rays, 
along with two additional restrictions that force the two to be consistent.
The latter constraints hinge on the ``locally visible'' zeroes 
$q\in\lvz$
(i.e., at least one ray exists, 
along which $q$ is visible at an infinitesimal distance).
Such zeroes are (i) always visible and (ii) every visible ray must intersect at least one.

Following the previous section on VFs, 
we next define an approach to ``field induction'': 
given a shape $S$, how can we construct a DDF, 
consisting of a ``field pair'' $(\xi,d)$,
that appropriately represents $S$?
We then show that inducing a field in this manner, versus having a DDF with constraints on local field behaviour 
(i.e., Def.~\ref{df:pddf:full:fullddf:main})
are equivalent.

\begin{mydef}{Induced Field Pair}{pddf:full:inducedfieldpair}	
	Consider a shape $S \subseteq \boe$.
	By Def.~\ref{df:pddf:vis:sibfs:main},  %
	$S$ induces a unique visibility field, $\xi$.
	We also define an induced distance field $d$ to be any NNBC field that satisfies 
	\begin{equation}
		d(\tau) = \min_{q\in [S]_\tau} ||q - p||
		\;\forall\; 
		\tau\in\visrays[\xi].
		\label{eq22}
	\end{equation}
	Then, %
	any such pair $(\xi,d)$
	is an \textit{$S$-induced field pair}. %
\end{mydef}

This builds on the notion of shape-induced VFs 	
	(Def.~\ref{df:pddf:vis:sibfs:main}).
Recall, the induced visibility $\xi(\tau)$ is one iff $\tau$ intersects $S$. 
Hence, for a visible ray $\tau\in\visrays[\xi]$, 
$d$ is constrained to always predict the distance to 
the closest $q\in S_\tau$.
\tempp{
Thus, Eq.~\ref{eq22} simply follows the definition of a ray-based depth field from \S\ref{sec:ddf}.
}
Again, similar to Theorem~\ref{th:pddf:vis:sibfavf:main}, 
	when a shape $S$ induces a field pair $F$, 
	we find that $F$ follows the field requirements of DDFs 
	(Def.~\ref{df:pddf:full:fullddf:main}).

\begin{restatthm}{Induced Field Pairs are DDFs}{pddf:full:ifpaddfs}
	Let $S$ be a shape and $F=(\xi, d)$ be an $S$-induced field pair.
	Then $F$ is a DDF 
	that satisfies $\lvz = S$.	
\end{restatthm}

Thus, any field generated from a shape is a DDF.
We next want to show the converse: 
	any DDF is a VC representation of some shape.
\tempp{
	Intuitively, this would imply our field constraints (which do not refer to a predefined shape) are equivalent to deriving the field from a shape.
}
First, we require a notion of equivalence between fields that ignores non-visible rays.

\begin{mydef}{DDF Equivalence}{pddf:full:equivddf}	
	Consider two DDFs, $F_1 = (\xi_1,d_1)$ and $F_2 = (\xi_2,d_2)$.
	Then $F_1$ and $F_2$ are equivalent iff
	(i) $\xi_1(\tau) = \xi_2(\tau)\;\forall\; \tau\in\Gamma$
	and 
	(ii) $d_1(\tau) = d_2(\tau) \;\forall\; \tau\in\visrays[\xi_1]$.
\end{mydef}

This enables defining when a DDF is a shape representation.

\begin{mydef}{DDFs as Shape Representations}{pddf:full:shapereps}	
	A DDF $F$ is view consistent (VC) iff it is equivalent to a DDF that has been induced by a shape $S$.
	I.e., $F$ is VC iff $\exists$ $S$ s.t.\ %
	$F$ is an $S$-induced field pair.
	In such a case, 
	we say that \textit{$F$ is a shape representation for $S$}.
\end{mydef}

Finally, 
analogously to Theorem~\ref{th:pddf:vis:evfias:main},
we can assert that the field constraints of Def.~\ref{df:pddf:full:fullddf:main}
are sufficient to guarantee that a DDF represents some shape.

\begin{restatthm}{Every DDF is View Consistent}{pddf:full:eddfiasr}
	Let $F=(\xi,d)$ be a DDF 
	(Def.~\ref{df:pddf:full:fullddf:main}). %
	Then $F$ must be VC 
	(i.e., $\exists$ a shape $S$ such that $F$ is an $S$-induced field pair).
	Further, $\lvz$ is a shape that induces $F$.
\end{restatthm}

This theorem finally links \textit{full} DDFs and view consistency.
A DDF can therefore represent a shape in two equivalent ways: 
(i) beginning with a point set and inducing a field pair from it, or 
(ii) starting from a visibility field and a distance field,
and enforcing per-field requirements (BC, IO, DE),
as well as compatibility.
We reiterate that 
Supp.~\S\ts{\ref{suppmat:theory}}{G} has a more complete exploration of these results, with proofs in Supp.~\S\ts{\ref{suppmat:proofs}}{H}.

The nature of the field constraints suggests an important use case for DDFs in practice.
Since field properties can be checked in a point-wise manner (e.g., DE$_d$ asks for along-ray derivatives to be $-1$), they are amenable to use as differentiable regularization losses.
This is similar to the Eikonal loss used in SDFs (e.g., \cite{gropp2020implicit,lin2020sdf,yang2021deep,bangaru2022differentiable,yu2022monosdf}), for example;
	indeed, resolving these constraints for DDFs is reminiscent of 
	solving a 5D partial differential equation,
	where BC and IO act like boundary conditions, 
		while DE needs to be solved across space.
Advances in optimization of implicit fields with differential constraints
	should make implementation of DDFs more viable.
In general, when designing a neural architecture for a DDF, 
	one can also check which theoretical properties are 
	guaranteed by the design and which may be violated.
Hence, we expect that our theoretical results will be useful 
	in the construction and analysis of any 5D geometry field,
	where view consistency needs to be enforced.
\tempp{
	In this work, we trained directly with 3D data, mitigating consistency issues; however, deriving fields from less constrained data, such as images, will require addressing such problems.
}

%% file: discussion.tex
\section{Discussion}

We have devised \textit{directed distance fields} (DDFs), a novel shape representation that maps oriented points to depth and visibility values, and a probabilistic extension for handling discontinuities. 
In contrast to NeRFs or U/SDFs, depth requires a single per-pixel forward pass, while normals use one further backward pass.
Efficient differentiable normals rendering is thus much easier for DDFs than for
for voxels, NeRFs, or occupancy fields.
Unlike meshes, DDFs are topologically unconstrained, and rendering is independent of shape complexity. 

We investigated a number of properties and use-cases for DDFs,
	including 3D data extraction, fitting shapes, UDF extraction, composition, generative modelling, and single-image 3D reconstruction.
While DDFs form a continuous field of depth images,
recursive calls can be interpreted as 
tracing inter-surface light paths.
Finally, we examined the theory of \textit{view consistency} (VC) for DDFs,
	as their 5D nature permits view-dependent geometric inconsistencies.
We showed that a small set of field constraints guarantee a given DDF is VC, and hence a proper representation for some 3D shape.

For future work, 
using DDFs for inverse graphics (e.g., for multiview stereo or novel view synthesis)
seems promising,
particularly via constraints based on our VC results,
as well as
handling volumetricity (not just surfaces), light transport, and translucency.
Finally, we hope to apply DDFs to other areas, %
	such as virtual reality and visuotactile perception,
	where collision modelling and geometric rendering are important.

%% file: bib_and_bio.tex
\bibliographystyle{IEEEtran}
\bibliography{defs,bib}

\newcommand{\tristanbio}{ 
received a BSc from McGill University in 2016 and an MSc from the University of Toronto (UofT) in 2018.
He is currently a PhD student in AI at UofT and works at the Samsung Artificial Intelligence Center in Toronto on problems in computer vision.
His research interests include 3D shape, disentangled representation learning, and generative modelling. 
}

\newcommand{\stavrosbio}{
is a Research Scientist at the Samsung AI Center in Toronto. He is a member of the Image Quality Enhancement team, working on problems related to super-resolution, denoising and obstruction removal. His research interests also include 2D/3D shape understanding, object recognition, and segmentation. Stavros received his degree in Electrical and Computer Engineering from the National Technical University of Athens, Greece, and his PhD in Applied Math and Computer Science from Universit\'e Paris-Saclay, in France.
}

\newcommand{\svenbio}{
received the B.A.Sc.\ degree in Systems Design Engineering from the University of Waterloo, in 1983, and the M.S.\ and Ph.D.\ degrees in Computer Science from the University of Maryland, in 1988 and 1991, respectively. He is Professor and past Chair of the Department of Computer Science at the University of Toronto, and is also Vice President and Head of the new Samsung Toronto AI Research Center, which opened in May, 2018. Prior to that, he was a faculty member at Rutgers University where he held a joint appointment between the Department of Computer Science and the Rutgers Center for Cognitive Science (RuCCS). His research interests revolve around the problem of shape perception in computer vision and, more recently, human vision. He has received the National Science Foundation CAREER award, the Government of Ontario Premiere's Research Excellence Award (PREA), and the Lifetime Research Achievement Award from the Canadian Image Processing and Pattern Recognition Society (CIPPRS). He was the Editor-in-Chief of the IEEE Transactions on Pattern Analysis and Machine Intelligence, from 2017-2021, currently serves on seven editorial boards, and is co-editor of the Morgan \& Claypool Synthesis Lectures on Computer Vision. He is a Fellow of the Institute of Electrical and Electronics Engineers (IEEE), a Fellow of the International Association for Pattern Recognition (IAPR), and an IEEE Golden Core Member.
}

\newcommand{\allanbio}{
received his B.Sc.\ in 1976 from the University of British Columbia, his Ph.D.\ in Applied Mathematics in 1980 from Caltech, and then moved to a postdoctoral position in the Mathematics Department at Stanford University. In 1982 he joined the faculty at the Department of Computer Science at the University of Toronto, becoming a full professor in 1991. Dr. Jepson was an Associate of the Canadian Institute of Advanced Research (CIFAR) for 1986 to 1989, for 2004 to 2009, and was a Scholar at CIFAR for 1989 to 1995. He was Chief Scientist at Samsung's AI Center in Toronto from Sept.\ 2018 to Dec.\ 2022.
}

\tr{}{
\ts{

\section*{Biographies}
{   
	\small
\noindent\textbf{Tristan Aumentado-Armstrong}%
	\tristanbio \\\\ 
\noindent\textbf{Stavros Tsogkas}%
	\stavrosbio \\\\
\noindent\textbf{Sven Dickinson}%
	\svenbio \\\\
\noindent\textbf{Allan Jepson}%
	\allanbio }

\vspace*{0.25in}

}{
\section*{Biography Section}

{
	\vspace{-33pt}
	\begin{IEEEbiographynophoto}{Tristan Aumentado-Armstrong}
		\tristanbio
	\end{IEEEbiographynophoto}%
	\begin{IEEEbiographynophoto}{Stavros Tsogkas}
		\stavrosbio
	\end{IEEEbiographynophoto}%
	\begin{IEEEbiographynophoto}{Sven Dickinson}
		\svenbio
	\end{IEEEbiographynophoto}%
	\begin{IEEEbiographynophoto}{Allan Jepson}
		\allanbio
	\end{IEEEbiographynophoto}%
}

\vfill %
} %
} %

%% file: supp-geomprops.tex
\section{Proofs of Geometric Properties}
\label{appendix:proofs}

{For} a shape $S$, 
we consider visible oriented points $(p,v)$ 
(i.e., $\xi(p,v) = 1$), 
such that $q(p,v) = p + d(p,v) v \in S$, unless otherwise specified.

\subsection{Property I: Directed Eikonal Equation}
\label{appendix:proofs:prop1}

First, note that for any visible $(p,v)$ 
(with $p\notin S$),
there exists an %
$\epsilon > 0$
such that any
$ \delta \in (\delta_0 - \epsilon, \delta_0 + \epsilon) $
satisfies 
$q(p,v) = q(p+\delta v,v) \in S$.
In such a case, by definition of the directed distance field (DDF),
$
d(p + \delta v, v) = d(p,v) - \delta.
$
Restrict $\delta$ to this open interval.
The directional derivative along $v$ with respect to position $p$ is then
\begin{align}
	\nabla_p d(p,v)^T v
	&=
	\lim_{\delta\rightarrow 0}
	\frac{d(p + \delta v, v) - d(p,v)}{ \delta } \\
	&=
	\lim_{\delta\rightarrow 0}
	\frac{d(p, v) - \delta - d(p,v)}{ \delta } \\
	&= 
	-1,
\end{align}
as required.

\subsubsection{Gradient Norm Lower Bound}

For any visible $(p,v)$, 
since $|\nabla_p d v| = ||\nabla_p d||_2 ||v||_2\, |\cos(\theta(\nabla_p d, v))| = 1$, 
where $\theta(u_1,u_2)$ denotes the angle between vectors $u_1$ and $u_2$,
we also see that 
$||\nabla_p d||_2 = 1 / |\cos(\theta(\nabla_p d, v))|$,
and thus 
$||\nabla_p d||_2 \geq 1$.

\subsubsection{Visibility Gradient}
Consider the same setup as in \ref{appendix:proofs:prop1}.
The visibility field satisfies a similar property: $\xi(p + \delta v, v) = \xi(p, v)$, as the visibility cannot change when  moving along the same view line (away from $S$).
Thus, we get $\nabla_p \xi(p,v)^Tv = 0$.

\subsection{Property II: Surface Normals}
\label{appendix:proofs:prop2}

Consider a coordinate system with origin $q_0\in S$,
with a frame given by
$(\widehat{i},\widehat{j},\widehat{k})$
where $\widehat{k} = n(q_0)$ and $\widehat{i},\widehat{j}\in\mathcal{T}_{q_0}(S)$ spans the tangent space at $q_0$.
Locally, near $q_0$, reparameterize $S$ in this coordinate system via
\begin{equation}
	S(x,y) = (x, y, f_S(x,y)), \label{app:eq:n0}
\end{equation}
where $f_S$ controls the extension of the surface in the normal direction.
Notice that 
\begin{equation}
	\partial_\alpha S|_{q_0} 
	= (\delta_{x\alpha}, \delta_{y\alpha}, \partial_\alpha f_S|_{q_0}),
\end{equation}
where $\alpha\in\{x,y\}$ and $\delta_{x\alpha}$ is the Kronecker delta,
but since
$\partial_x S|_{q_0} = \widehat{i}$ %
and
$\partial_y S|_{q_0} = \widehat{j}$ %
are in the tangent plane,
\begin{equation}
	\partial_\alpha f_S|_{q_0} = 0. \label{app:eq:n1}
\end{equation}

Consider any oriented position $(p,v)$ that points to $q(p,v)\in S$ on the surface.
Locally, the surface can be reparameterized in terms of $(p,v)$:
\begin{equation}
	q(p,v) = p + d(p,v)v \in S. \label{app:eq:n2}
\end{equation}
Yet, using $q = (q_x,q_y,q_z)$,
we can write this via Eq.\ \ref{app:eq:n0} as 
\begin{equation}
	S(q_x,q_y) = (q_x, q_y, f_S(q_x,q_y)),
\end{equation}
where $q_x$ is the component along the $x$ direction in local coordinates, 
which depends on $p$ and $v$.
In other words, the $z$ component of $q$ depends on $p$ via:
\begin{equation}
	q_z(p,v) = f_S(q_x(p,v), q_y(p,v)). \label{app:eq:n3}
\end{equation}

Let $(p_0,v_0)$ point to $q_0$ (i.e., $q_0 = q(p_0,v_0)$).
Then:
\begin{align*}
	\partial_{p_i} f_S|_{p_0}
	&=
	\underbrace{
		\partial_{q_x} f_S|_{q_0}
	}_0
	\partial_{p_i} q_x|_{p_0}
	+
	\underbrace{
		\partial_{q_y} f_S|_{q_0}
	}_0
	\partial_{p_i} q_y|_{p_0} \\
	&=
	0 \;\, \forall\; i\in\{x,y,z\},
\end{align*}
since $\partial_{q_\alpha} f_S|_{q_0} = 0$ using Eq.\ \ref{app:eq:n1}.

Derivatives with respect to position are then given by
\begin{align}
	\partial_{p_z} q_z|_{p_0}
	&= 1 + \partial_{p_z} d(p,v)|_{p_0} v_z
	= \partial_{p_z} f_S|_{p_0}
	= 0  \label{app:eq:nt1} \\
	\partial_{p_\alpha} q_z|_{p_0}
	&= \partial_{p_\alpha} d(p,v)|_{p_0} v_z
	= \partial_{p_\alpha} f_S|_{p_0}
	= 0,  \label{app:eq:nt2}
\end{align}
using Eq.\ \ref{app:eq:n2} and Eq.\ \ref{app:eq:n3},
with $\alpha\in\{x,y\}$.
Thus, using Eq.\ \ref{app:eq:nt1} and Eq.\ \ref{app:eq:nt2},
\begin{equation}
	\frac{\partial}{\partial(p_x,p_y,p_z)} d(p,v)|_{p_0} = (0, 0, -1/v_z),
\end{equation}
in local coordinates. 
Since the $z$ component here is along the direction of the surface normal, 
$n(q_0) = n(p_0,v_0)$,
this can be rewritten as 
\begin{equation}
	\nabla_p d(p,v)|_{p_0,v_0} = \frac{-1}{v_z} n(q_0)^T = \frac{-n(q_0)^T}{n(q_0)^T v} .
\end{equation}
For any $C^1$ surface $S$, and point $q_0\in S$ (that is intersected by visible oriented point $(p,v)$), we can always construct such a coordinate system,
so we can more generally write:
\begin{equation}
	\nabla_p d(p,v) = \frac{-n^T}{n^T v}.
\end{equation}

\subsection{Property III: Gradient Consistency}
\label{appendix:proofs:prop3}

Consider the same setup (coordinate system) and notation as in the Proof of Property II above (\S\ref{appendix:proofs:prop2}).
Since $v\in\mathbb{S}^2$, it suffices to consider an infinitesimal rotational perturbation of some initial view direction $v_0$:
\begin{equation}
	dR(t) = [\omega]_\times dt,
\end{equation}
where 
$[\omega]_\times$ is a skew-symmetric cross-product matrix 
of some angular velocity vector $\omega$,
so that
$\widetilde{v} = (I + dR(t))v_0$ 
and 
$dv = \widetilde{v} - v_0 =[\omega]_\times\, dt\, v_0 $
is the change in the view (given $dt$) and a velocity of 
$u := \partial_t v = [\omega]_\times v_0$.
The change in surface position,
$q(p,v(t)) = p + d(p,v(t)) v(t)$,
with respect to $t$ is then
\begin{align*}
	\partial_t q 
	&= v\partial_t d + d \partial_t v \\
	&= (\partial_v d \partial_t v)v + d\partial_t v \\
	&= (\partial_v d u)v + d u .
\end{align*}

The $z$ component, in local coordinates, is then
\begin{align*}
	\partial_t q_z|_{q_0} 
	&= \partial_t f_S(q_x,q_y)|_{q_0} \\
	&= \partial_{q_x} f_S|_{q_0} \partial_t q_x + 
	\partial_{q_y} f_S|_{q_0} \partial_t q_y \\
	&= 0,
\end{align*}
via Eq.\ \ref{app:eq:n1}.
Thus, $\partial_t q_z|_{q_0} 
= (\partial_v d u)v_z + d u_z
= 0$, meaning
\begin{equation}
	d u^Tn = -(\partial_v d u) v^Tn
\end{equation}
using 
$u_z = u^Tn$
and 
$v_z = v^Tn$ at $q_0$.
Recalling that $\partial_p d = -n^T / (n^Tv)$ and rearranging,
we get 
\begin{equation}
	\partial_v d u = -d \frac{n^T}{v^Tn}u = d \partial_p d u.
\end{equation}
Notice that $u = \partial_t v = [\omega]_\times v_0$ is orthogonal to $v_0$ and the arbitrary vector $\omega$. 
Hence, for any visible oriented point $(p,v)$ away from viewing the boundary of $S$ (i.e., where $n^Tv=0$) and for all $\omega\in\mathbb{R}^3$, we have
\begin{equation}
	\label{app:eq:crossprod}
	\nabla_v d [\omega]_\times v = d\, \nabla_v d [\omega]_\times v.
\end{equation}
This constrains the derivatives with respect to the view vector to be closely related to those with respect to position, along any directions not parallel to $v$.

\subsubsection{Alternative Expression}
Note that the inner product with $\delta_v$ primarily serves to restrict the directional derivative in valid directions of $v$.
A cleaner expression can be obtained with a projection operator, which removes directional components parallel to $v$.

First, let us define $d$ to normalize $v$: 
\begin{equation}
	\label{app:eq:altdv:normalized}
	d(p,v) := d\left( p, \frac{v}{||v||_2} \right), 
\end{equation}
for all $v\in\mathbb{R}^3 \setminus \{0\}$. 
Then, consider a perturbation along the view direction of size $|\delta| < 1$:
\begin{equation}
	\label{app:eq:altdv:zerochangea}
	d(p_0, v_0 + \delta v_0)
	= d(p_0, v_0(1+\delta))
	=d(p_0,v_0),
\end{equation}
with $||v_0|| = 1$ and
using Eq.\ \ref{app:eq:altdv:normalized} for the last step.
This means the directional derivative along $v$ must satisfy \begin{equation}
	\label{app:eq:altdv:zerochangeb}
	\partial_v d(p,v) v = 0.
\end{equation}
In the previous section, we showed that
$\partial_v d u = d\partial_p d u$ for all $u \perp v$.
Let $\mathcal{P}_v = I - vv^T$ be the orthogonal projection removing components parallel to $v$.
Then we can rewrite the result of the previous section
as $\partial_v d \mathcal{P}_v = d\partial_p d \mathcal{P}_v$.
But $\partial_v d \mathcal{P}_v = \partial_v d - \partial_v d vv^T = \partial_v d$ by Eq.\ \ref{app:eq:altdv:zerochangeb}.
Thus, we may write
\begin{equation}
	\partial_v d = d\, \partial_p d\, \mathcal{P}_v,
\end{equation}
which agrees with Eq.\ \ref{app:eq:crossprod}.

\subsection{Property V: Local Differential Geometry}
\label{appendix:proofs:prop5}

Given a visible oriented point $(p,v)$, 
we have shown that the surface normal $n(p,v)$ on $S$
(at $q = p + d(p,v)v$)
is computable from $\nabla_p d$ (see Supp.~\ref{appendix:proofs:prop2}).
Curvatures, however, require second-order information.

We first construct a local coordinate system via $n$, 
by choosing two tangent vectors at $q$: 
$t_x,t_y\in\mathcal{T}_q(S)$, 
where $||t_\alpha||_2=1$ and $t_x^T t_y=0$.
In practice, this can be done by sampling Gaussian vectors, 
and extracting an orthogonal tangent basis from them.
We can then reparameterize the surface near $q_0 = q(p_0,v_0)$ (with surface normal $n_0 = n(q_0)$) via
$S(x,y) = q_0 + xt_x + yt_y + f_S(x,y)n_0$,
where $x$ and $y$ effectively control the position on the tangent plane.
Alternatively, 
we can write $S(x,y) = q(p(u),v_0)$,
which parameterizes $S$ about the oriented point 
(or viewpoint) $p(u)$, 
where $u=(x,y)$, and $p(u) = p_0 + xt_x + yt_y$.
Notice that $p(u)$ is essentially a local movement of $p$ parallel to the tangent plane at $q_0\in S$ (which is ``pointed to'' by $(p_0,v_0)$). Note that $p_0$, $v_0$, and $q_0$ are fixed; only $u$ and $p(u)$ are varied.
Further, notice that the plane defined by $p(u)$ (in the normalized tangent directions $t_x$ and $t_y$) is parallel to the surface tangent plane at $q_0$ (and thus orthogonal to $n_0$), but \textit{not} necessarily perpendicular to $v$.

Using this local frame on $\mathcal{T}_{q_0}(S)$,
the first-order derivatives of the surface are
\begin{align}
	\partial_{i} S|_{u=0}
	&= \partial_{i} \left( p(u) + v_0 d(p(u),v_0) \right)|_{u=0} \\
	&= t_i + \partial_{i} d(p(u),v_0) |_{u=0} v_0 \\
	&= t_i + (\nabla_p d(p,v_0) t_i) v_0
\end{align}
where $i\in\{x,y\}$ and $p_j$ is the $j$th component of $p$.
Then the metric tensor (first fundamental form) is given by
\begin{align}
	g_{ij} 
	&= \partial_{i} S|_{u=0}^T \, \partial_j S|_{u=0} \\
	&= \delta_{ij} + c_ic_j + c_i v_0^Tt_j + c_j v_0^T t_i
	\label{appendix:eq:gfail},
\end{align}
where $i,j\in\{x,y\}$, $u=(x,y)$, 
$\delta_{ij}$ is the Kronecker delta function, and $c_k = \nabla_p d t_k$.
Notice that $\nabla_p d$ is parallel to $n$ 
(see Property \hyperref[property2]{II}), 
and thus orthogonal to $t_1$ and $t_2$.
Thus, $c_x = c_y = 0$, and so, 
in theory, $g_{ij} = \delta_{ij}$, 
but this may not hold for a learned DDF,
if one computes directly with Eq.\ \ref{appendix:eq:gfail}.
However, in practice, 
we use the theoretical value (where $g = I_2$ in local coordinates),
as deviations from it are due to error only (assuming correct normals).

Next, the shape tensor (second fundamental form) $\RomanII_{ij}$ can be computed from the second-order derivatives of the surface \cite{kreyszigdg} via:
\begin{align}
	\RomanII_{ij}
	&= n_0^T \partial_{ij} S|_{u=0} \\
	&= n_0^T \partial_{ij} ( p(u) + v_0 d(p(u),v_0) )|_{u=0} \nonumber\\
	&= n_0^T v_0 \partial_{ij}  d(p(u),v_0) |_{u=0}  \nonumber\\ 
	&= n_0^T v_0 \partial_i \sum_k \partial_{p_k} d(p(u),v_0) 
	\underbrace{[\partial_j p(u)]_k}_{[t_j]_k} |_{u=0}  \nonumber\\
	&= n_0^T v_0 \sum_k \partial_i \partial_{p_k} d(p(u),v_0) 
	[t_j]_k |_{u=0}  \nonumber\\
	&= n_0^T v_0 \sum_k [t_j]_k 
	\sum_\ell 
	\partial_{p_\ell} \partial_{p_k} d(p(u),v_0)
	[\partial_i p(u)]_\ell
	|_{u=0}  \nonumber\\
	&= n_0^T v_0 \sum_k [t_j]_k 
	\sum_\ell 
	\underbrace{
		\partial_{p_\ell,p_k} d(p(u),v_0)
	}_{ \mathcal{H}_p[d]_{k \ell} }
	[t_i]_\ell
	|_{u=0}  \nonumber\\
	&= \left( t_j^T \mathcal{H}_p[d] t_i \right) n_0^T v_0,
	\label{app:eq:curvderiv} 
\end{align}
where 
$i,j\in\{x,y\}$, 
$u=(x,y)$, 
$n_0 = n(q_0)$,
$[t_i]_k$ is the $k$th component of $t_i$,
$\mathcal{H}_p[d]$ is the Hessian of $d$ with respect to $p$ 
(at $u=0$),
and 
$p(u) = p_0 + x t_x + y t_y$.
Note that our parameterization of the surface using the DDF (i.e., $q(p(u),v_0)$) defines surface deviations in terms of $v_0$; the $n_0^T v_0$ term undoes this effect, rewriting the deviation in terms of $n$ instead.

Curvatures can then be computed via the shape tensor (see, e.g., \cite{kreyszigdg}).
Gaussian curvature is given by
$\mathcal{C}_K = \det(\RomanII)/\det(g)$, 
while mean curvature is written
$\mathcal{C}_H = \mathrm{tr}(\RomanII g^{-1})$.
Notice that these quantities can be computed for any visible $(p,v)$, 
using only $d(p,v)$ and derivatives of $d$ with respect to $p$ (e.g., with auto-differentiation).
Thus, the curvatures of any visible surface point can be computed using only local field information at that oriented point.

\subsection{Neural Depth Renderers as DDFs}
\label{appendix:neurren}
A natural question is how to parallelize a DDF.
Consider a function $f : \Lambda_\Pi \times \Lambda_S \rightarrow \mathcal{I}_{d,\xi}^m$, where $ \mathcal{I}_{d,\xi} = \mathbb{R}_+ \times [0,1]$, from continuous camera parameters (i.e., elements of $\Lambda_\Pi$) and some space of shapes $\Lambda_S$ to an $m$-pixel depth and visibility image.
Then $f$ is implicitly a conditional DDF, as each pixel value can be computed as $(d(p,v), \xi(p,v))$, where $p$ and $v$ are a fixed function of $\Pi\in\Lambda_\Pi$.
In other words, the camera intrinsics and extrinsics determine the $p$ and $v$ value for each pixel, with $p$ determined by the centre of projection (camera position) and $v$ determined by the direction corresponding to that pixel, from the centre of projection out into the scene.
We may thus regard the depth pixel value as a function of $p$ and $v$, via the camera parameters.
This holds regardless of the architecture of $f$. %
Thus, all the properties of DDFs hold in these cases as well: for example, for a depth image $I_d$, camera position $C_p$ and viewpoint $C_v$,  Property \hyperref[property2]{II} relates $\partial_{C_p} I_d$ to the surface normals image, while Property \hyperref[property3]{III} constrains $\partial_{C_v} I_d$.
We believe this point of view can improve the framework for differentiable neural rendering of geometric quantities (e.g., \cite{yan2016perspective,wu2017marrnet,tulsiani2017multi,nguyen2018rendernet}).

%% file: supp-nonptapps.tex
\section{Single-Entity DDFs}
\label{appendix:singlefits}

\subsection{Mesh Data Extraction}
\label{appendix:sec:datatypes}

We display visualizations of each data type in Fig. \ref{fig:datatypes}.
We briefly describe how each is computed (note that we only need to obtain $(p,v)$, after which $(\xi,d)$ can be obtained via ray-triangle intersection):
\begin{itemize}
	\item Uniform (U): simply sample $p\sim\mathcal{U}[\mathcal{B}]$ and $v\sim\mathcal{U}[\mathbb{S}^2]$.
	\item At-surface (A): start with $q_0\sim\mathcal{U}[S]$ (obtained via area-weighted sampling) and $v_0\sim\mathcal{U}[\mathbb{S}^2]$. Sample $p$ on the line between $q_0$ and its intersection with $\mathcal{B}$ along $v_0$. Set $v = -v_0$. Note that the final output data may not actually intersect $q_0$ (since one may pass through a surface when sampling $p$).
	\item Bounding (B): sample $p\sim\mathcal{U}[\partial \mathcal{B}]$ and $v\sim\mathcal{U}[\mathbb{S}^2]$, but restrict $v$ to point to the interior of $\mathcal{B}$. 
	\item Surface (S): simply use $p\sim\mathcal{U}[S]$ and then take $v\sim\mathcal{U}[\mathbb{S}^2]$.
	\item Tangent (T): the procedure is the same as for A-type data, except we enforce $v_0$ to lie in the tangent plane $\mathcal{T}_{q_0}(S)$.
	\item Offset (O): take a T-type $(p,v)$ and simply do $p\leftarrow p + \varsigma_O \epsilon_O n_0$, where $n_0$ is the normal at $q_0$, and we set $\epsilon_O = 0.05$ and sample $\varsigma_O \sim \mathcal{U}[\{-1,1\}]$.
\end{itemize}
Since we assume $\mathcal{B}$ is an axis aligned box (with maximum length of 2; i.e., at least one dimension is $[-1,1]$), sampling positions on it, or directions with respect to it, is straightforward. See also \S\ref{datatypeablation}, which examines the effect of ablating each data type.

\subsection{Single-Entity Fitting}
\label{appendix:sec:singleentityfitting}

Given a mesh, we first extract data of each type, obtaining $(p,v)$ tuples in the following amounts:
250K (A and U) and 125K (B, T, O, and S). 
Since rendering outside $\mathcal{B}$ uses query points on $\partial\mathcal{B}$, we bias the sampling procedure for T, A, and O data, 
such that 10\% of the $p$-values are sampled from $\partial\mathcal{B}$.
For each minibatch, we sample
6K (A and U) and 3K (B, T, O, and S) points, across data types. 
In addition to these, we sample an additional 1K 
uniformly random oriented points per minibatch,
on which we compute only regularization losses 
($\mathcal{L}_V$ and $\mathcal{L}_\mathrm{DE}$), 
for which ground truth values are not needed.

We note that not all loss terms are applied to all data types. As discussed in \S\ref{sec:app:learning}, the transition loss $\mathcal{L}_T$ is only applied to S and T type data (since we do not want spurious field discontinuities, due to the field switching between components, except when necessary).
As briefly noted in Property \hyperref[property2]{II}, the DDF gradient $\nabla_p d$ (and hence field-derived surface normals) are not well-defined when $n$ and $v$ are orthogonal.
The gradients are also not well-defined on S-type data (since $d$ is explicitly discontinuous for $p\in S$, and thus $\nabla_p d$ does not exist there). 
Hence, we do not apply the directed Eikonal regularization $\mathcal{L}_\text{DE}$ or the normals loss $\mathcal{L}_n$ to S, T, or O data.
Similarly, the weight variance loss $\mathcal{L}_V$ (designed to reduce weight field entropy) should not be applied to S, T, or O samples, as the weight field should be transitioning near those samples (and thus have a higher pointwise Bernoulli variance).
The remaining losses are applied on all data types.

We then optimize Eq.\ \ref{eq:singlefitall} via Adam \cite{kingma2014adam}
(learning rate: $10^{-4}$; $\beta_1 = 0.9$, $\beta_2 = 0.999$), 
using a reduction-on-plateau learning rate schedule 
(reduction by 0.9, minimum of 5K iterations between reductions).
We run for 100K iterations,
using 
$\gamma_d = 5$,
$\gamma_\xi = 1$,
$\gamma_n = 10$,
$\gamma_V = 1$,
$\gamma_\mathrm{E,d} = 0.05$, 
$\gamma_\mathrm{E,\xi} = 0.01$, and
$\gamma_T = 0.25$.
Note that we double $\gamma_d$ on A and U data.
The field itself is implemented as a SIREN with seven hidden layers, initialized with $\omega_0 = 1$, each of size 512, mapping $(p,v)\in\mathbb{R}^6 \rightarrow (\{d_i\}_{i=1}^K,\{w_i\}_{i=1}^K,\xi)\in\mathbb{R}^{2K+1}$ 
(note that the set of $w_i$'s has $K-1$ degrees of freedom).
We use $K=2$ delta components.
An axis-aligned bounding box is used for $\mathcal{B}$.
Since $w_2 = 1 - w_1$, we output only $w_1$ and pass it through a sigmoid non-linearity. 
We apply ReLU to all $d_j$ outputs, to enforce positive values, and sigmoid to $\xi$.
We use the implementation from \cite{aman_dalmia_2020_3902941} for SIREN.
In terms of data, we test on shapes from the Stanford Scanning Repository
\cite{stanfordscanningrepo} (data specifically from
\cite{turk1994zippered,curless1996volumetric,krishnamurthy1996fitting}).

\subsection{Rendering Efficiency}
\label{sec:supp:renderingefficiency}
As noted in the main paper (\S\ts{\ref{sec:main:renderingefficiency}}{V-A4}), 
we tried to use approximately comparable MLP-based architectures. 
Both models were run on an NVIDIA GeForce RTX 2080 Ti.
We used the default settings of DIST \cite{liu2020dist} for rendering, including no aggressive marching, and used 50 marching steps, the lowest value we found in the released code.
To mitigate the effect of the choice of shape and the camera position, we ran DIST on five shapes, each with two azimuths (90 degrees apart) and at two distances from the shape (0.25 and 2.5). 
At the far distance, DIST renders in 
5.34 seconds (standard deviation 0.35); 
at the close distance, DIST takes 
9.78 seconds (standard deviation 1.58), 
due to differences in which heuristics can be applied to optimize runtime.
In contrast, the DDF renders in 0.006 seconds, with negligible difference across shapes or cameras, since we call the MLP once regardless of whether or where the shape exists along that particular ray.

\section{UDFs and MDFs}
\label{appendix:udf}

\subsection{MDF Extraction}
\label{appendix:udf:vstarextract}

We obtain $v^*$ via a fitting procedure, with a forward pass similar to composition 
(see \S\ref{sec:results:singlefieldfitting}).
Starting from an already trained (P)DDF, 
we define a small new SIREN network $g_v$
(five hidden layers, each size 128),
which maps position to a set of $K_c$ \textit{candidate directions}, 
such that
$g_v:\mathbb{R}^3\rightarrow\mathbb{S}^{2\times K_c}$
Given a position $p$ and candidates $\{v^*_i\}_{i=1}^{K_c} = g_v(p)$,
we can compute 
$ \zeta_{v^*} = \{ d(p,v^*_i(p)), \xi(p,v^*_i(p)) \}_{i=1}^{K_c} $.
To obtain a UDF depth estimate, we use 
\begin{equation}
	\widehat{\mathrm{UDF}}(p) = 
	\sum_i \omega_{\zeta_{v^*}}^{(i)}(p) \, d(p,v^*_i(p)),
\end{equation}
where the weights $\omega_{\zeta_{v^*}}$ are based on those from our explicit composition method:
\begin{equation}
	\omega_{\zeta_{v^*}}(p)
	=
	\mathrm{Softmax}
	\left(  
	\left\{
	\frac{ 
		\eta_{T}^{-1} \xi(p,v^*_i(p))
	}{ 
		\varepsilon_s + d(p,v^*_i(p)) 
	}
	\right\}_i\,
	\right).
\end{equation}
We use the same $\eta_T$ and $\varepsilon_s$ as for composition.
Note that this formulation essentially takes the candidate directions, computes their associated distances, and then (softly) chooses the one that has lowest distance value while still being visible. 
We can also obtain $v^*(p)$ via these weights, using
\begin{align}
	\widetilde{v}^*(p) &= 
	\sum_i \omega_{\zeta_{v^*}}^{(i)}(p) \, v^*_i(p) \\
	{v}^*(p) &= 
	\frac{\widetilde{v}^*(p)}{||\widetilde{v}^*(p)||_2} ,
\end{align}
Note that, in the ideal case, $v^*(p) = -\nabla_p \mathrm{UDF}(p)^T$, 
meaning we could compute it with a backward pass,
but we found such an approach was noisier in practice.

To train $g_v$ to obtain good candidates, we use the following loss: 
\begin{align}
	\mathcal{L}_{v^*}
	&= \frac{1}{K_c}\sum_i d(p,v^*_i(p)) - \xi(p,v^*_i(p)) \label{app:eq:vstartermone} \\
	&\;\;\; + \frac{2\tau_n}{K_c^2 - K_c} \sum_i \sum_{j \ne i} v^*_i(p)^T v^*_j(p) \\
	&\;\;\; + \frac{\tau_d}{K_c} \sum_i \left[ v^*_i(p)^T n(p,v^*_i(p)) + 1 \right]^2
\end{align}
where the 
first sum encourages obtaining the direction with minimum depth that is visible, 
the second prevents collapse of the candidates to a single direction (e.g., local minima), and 
the third encodes alignment with the local surface normals (similar to the directed Eikonal regularizer; see Supp.~\ref{appendix:vstarsurfacenormals}).
Note that the first sum (Eq.\ \ref{app:eq:vstartermone}) includes a $d$ term, which linearly penalizes longer distances, and a $-\xi$ term, which pushes the visibility (i.e., probability of surface existence in the direction $v_i^*(p)$, from position $p$) to be high.
We use $K_c = 5$, $\tau_n = 5\times 10^{-3}$, and $\tau_d = 0.1$.
Optimization was run for $10^4$ iterations
using Adam 
(LR: $10^{-4}$; $\beta_1 = 0.9$, $\beta_2 = 0.999$).
Each update used 4096 points, 
uniformly drawn from the bounding volume 
($p\sim\mathcal{U}[\mathcal{B}]$).

\subsection{Surface Normals of the MDF $v^*$}
\label{appendix:vstarsurfacenormals}
Let $v^*$ be defined as in \S\ref{sec:app:udfextract} and choose $p$ such that $v^*(p)$ is not multivalued (i.e., neither on $S$ nor on the medial surface of $S$).
Recall that, by definition, 
\begin{equation}
	\mathrm{UDF}(p) = d(p,v^*(p)) \label{udf:1}
\end{equation}
and
\begin{equation}
	n(p,v) = n(q(p,v)) = \frac{\nabla_p d(p,v)^T}{||\nabla_p d(p,v)||}
\end{equation}
where $n(p,v)^T v < 0$ (see Property \hyperref[property2]{II}).
Then, by Eq.\ \ref{udf:1}, 
\begin{align*}
	\nabla_p \mathrm{UDF}(p) 
	&= \nabla_p d(p,v^*(p)) \\
	&= ||\nabla_p d(p,v^*(p))|| \,
	\frac{\nabla_p d(p,v^*(p))}{||\nabla_p d(p,v^*(p))||} \\
	&= ||\nabla_p \mathrm{UDF}(p)||\; n(p,v^*(p))^T  \\
	&= n(p,v^*(p))^T,
\end{align*}
via the S/UDF Eikonal equation in the last step.

Finally, by the directed Eikonal property (\hyperref[property1]{I}),
we have:
\begin{equation}
	\nabla_p \mathrm{UDF}(p) v^*(p) =
	\nabla_p d(p,v^*(p)) v^*(p) =
	-1,
\end{equation}
meaning $\nabla_p \mathrm{UDF}(p) = -v^*(p)^T$.
Indeed, recall that the gradient of a UDF has 
(1) unit norm (i.e., satisfies the Eikonal equation) and 
(2) points away from the closest point on $S$,
which is precisely the definition of $-v^*(p)$.

\input{supp-si3dr}

\input{supp-ugen}

%% file: supp-si3dr.tex
\begin{figure*}
	\adjincludegraphics[height=\aalh,trim={ {\ach\width} {\cuthch\height} {\ach\width}  {\cuthch\height}},clip]{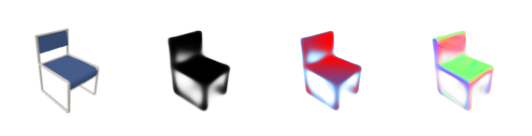}
	\adjincludegraphics[height=\alh,trim={ {\cch\width} {\cuthch\height} {\cch\width}  {\cuthch\height}},clip]{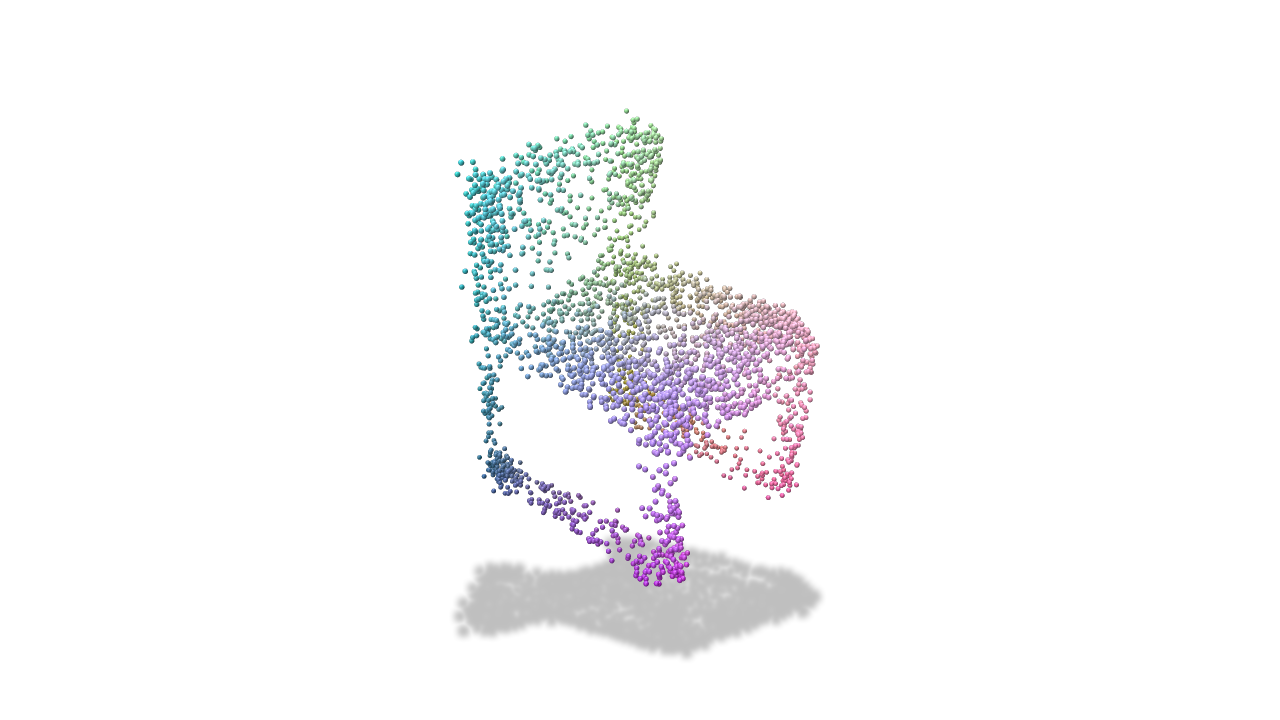}\hfill
	\adjincludegraphics[height=\alh,trim={ {\cch\width} {\cuthch\height} {\cch\width}  {\cuthch\height}},clip]{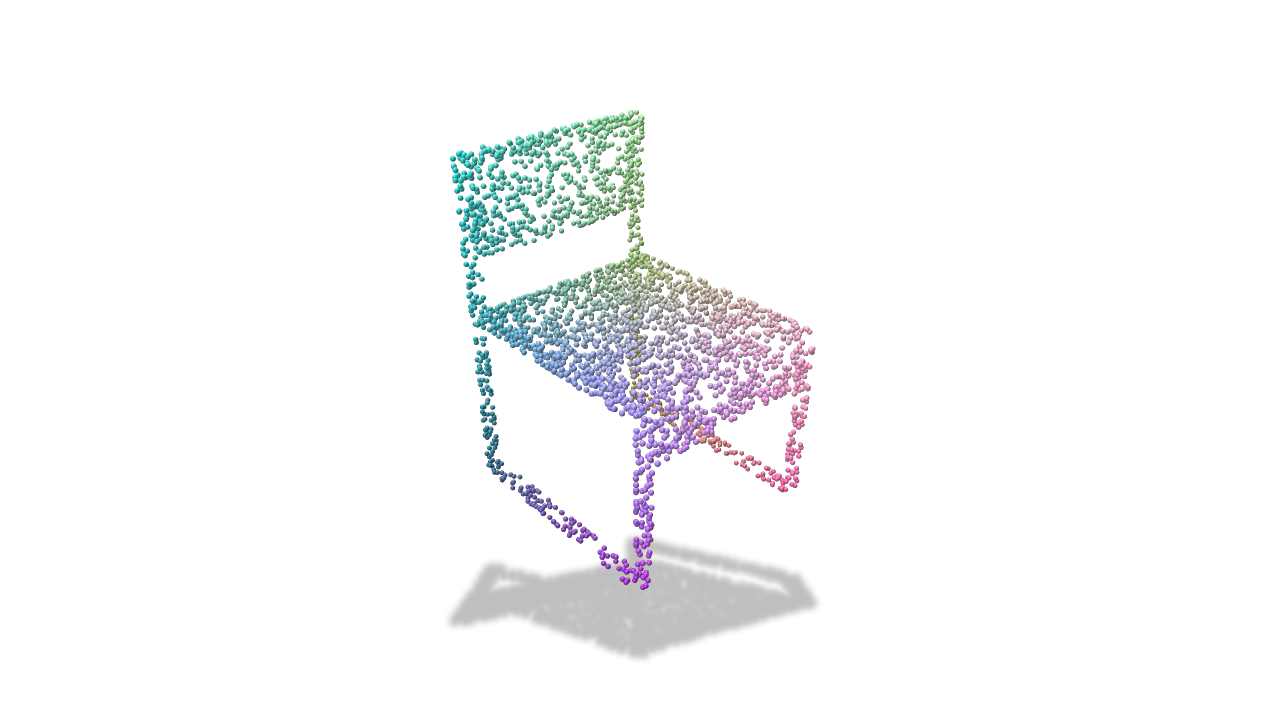}\hfill %
	\adjincludegraphics[height=\aalh,trim={ {\ach\width} {\cuthch\height} {\ach\width}  {\cuthch\height}},clip]{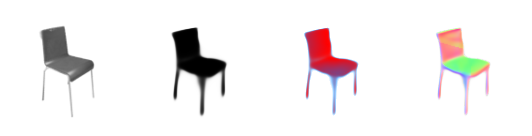}
	\adjincludegraphics[height=\alh,trim={ {\cch\width} {\cuthch\height} {\cch\width}  {\cuthch\height}},clip]{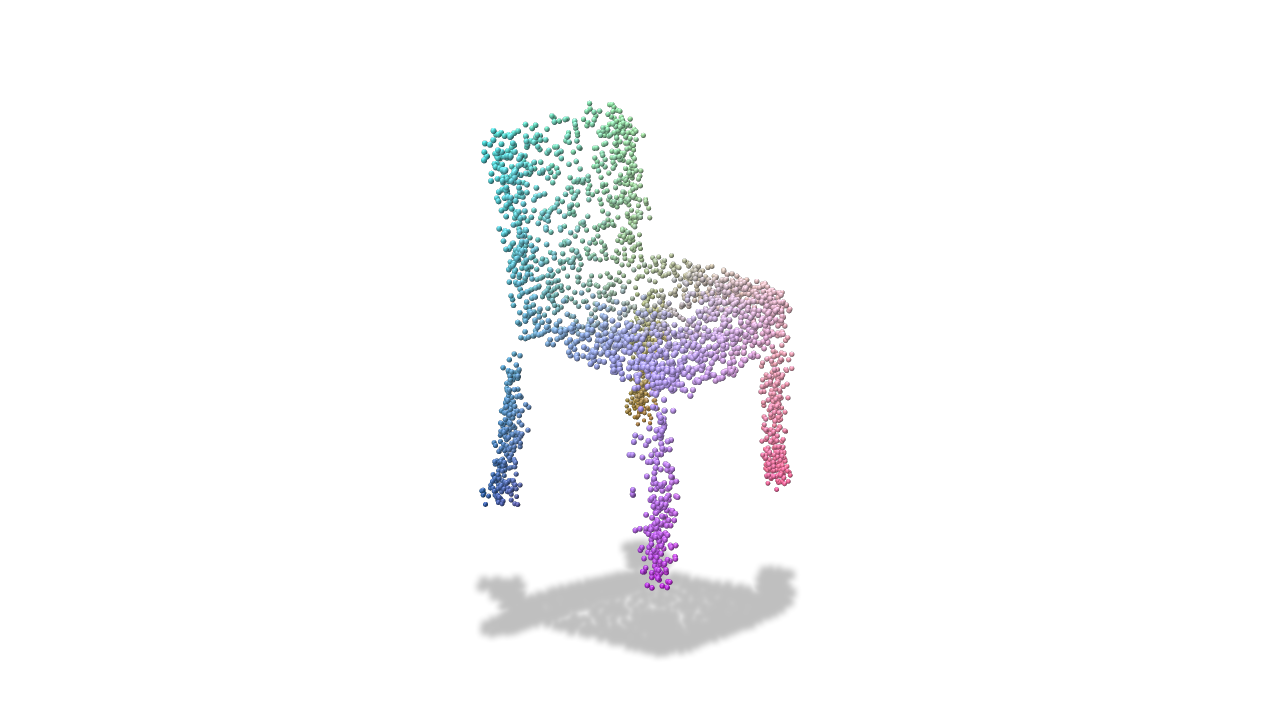}\hfill
	\adjincludegraphics[height=\alh,trim={ {\cch\width} {\cuthch\height} {\cch\width}  {\cuthch\height}},clip]{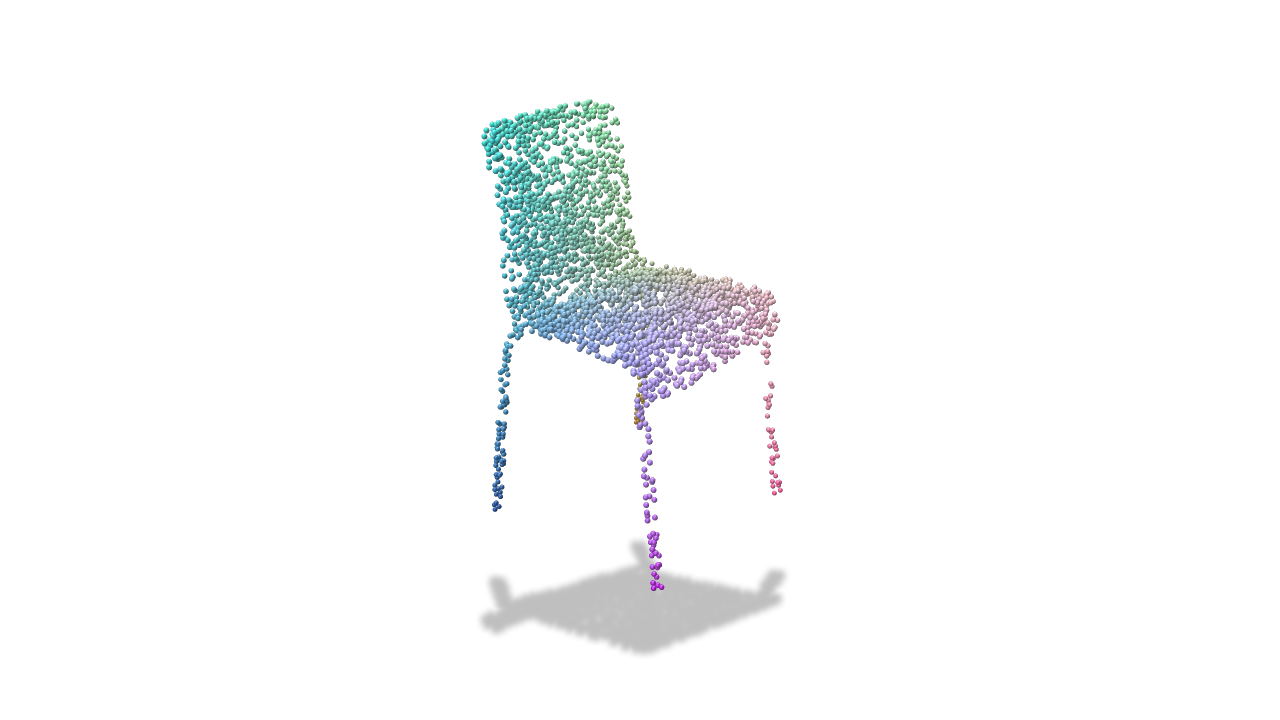} \\
	\adjincludegraphics[height=\aalh,trim={ {\ach\width} {\cuthch\height} {\ach\width}  {\cuthch\height}},clip]{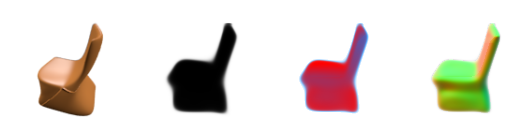}
	\adjincludegraphics[height=\alh,trim={ {\cch\width} {\cuthch\height} {\cch\width}  {\cuthch\height}},clip]{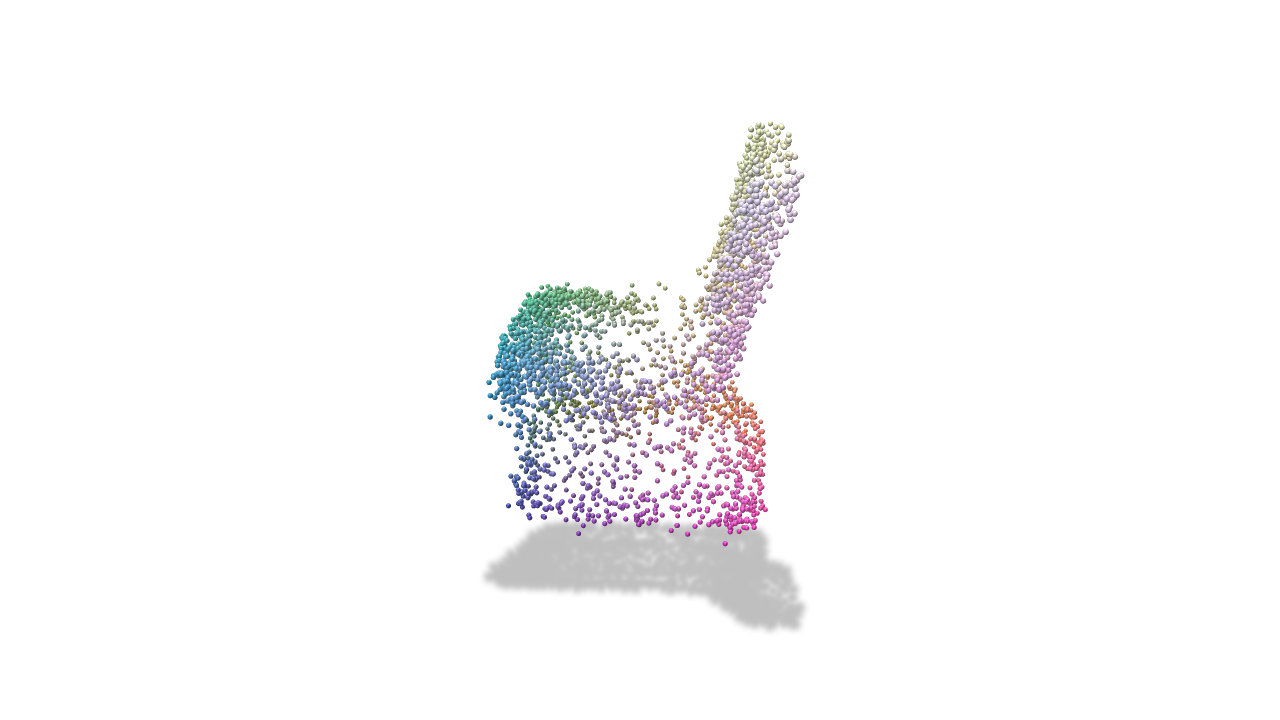}\hfill
	\adjincludegraphics[height=\alh,trim={ {\cch\width} {\cuthch\height} {\cch\width}  {\cuthch\height}},clip]{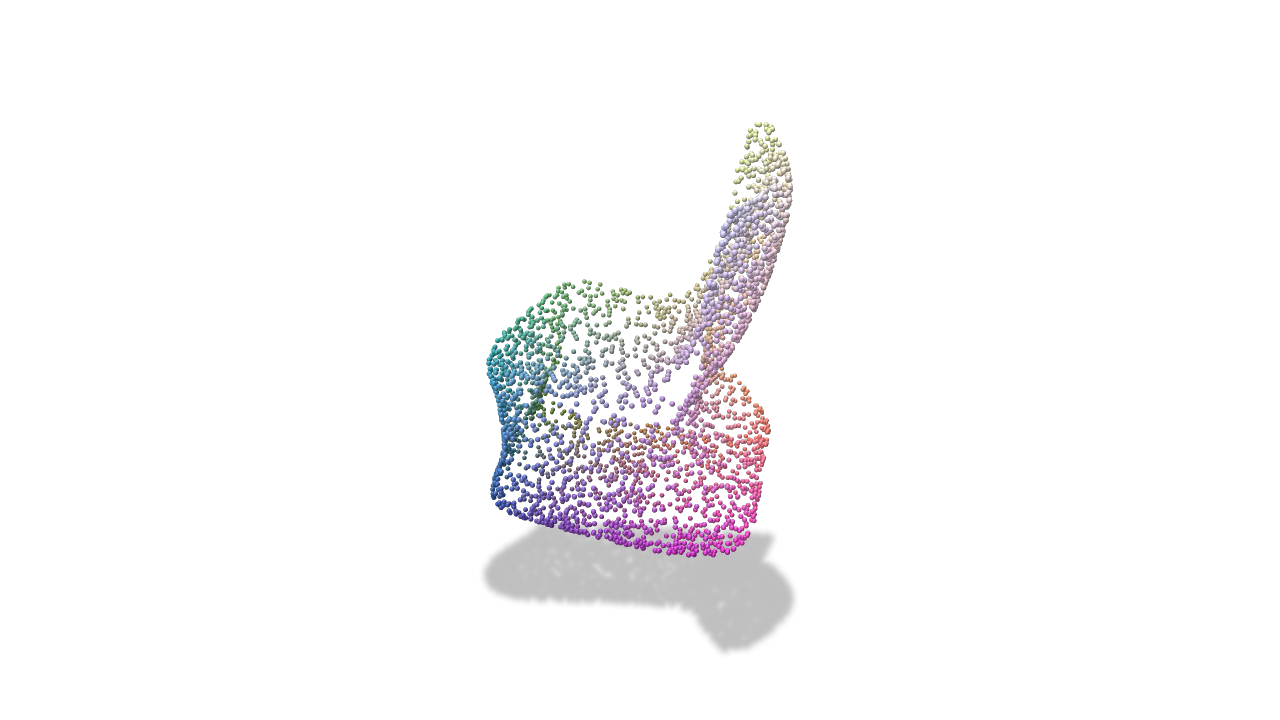}\hfill %
	\adjincludegraphics[height=\aalh,trim={ {\ach\width} {\cuthch\height} {\ach\width}  {\cuthch\height}},clip]{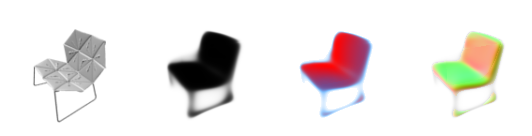}
	\adjincludegraphics[height=\alh,trim={ {\cch\width} {\cuthch\height} {\cch\width}  {\cuthch\height}},clip]{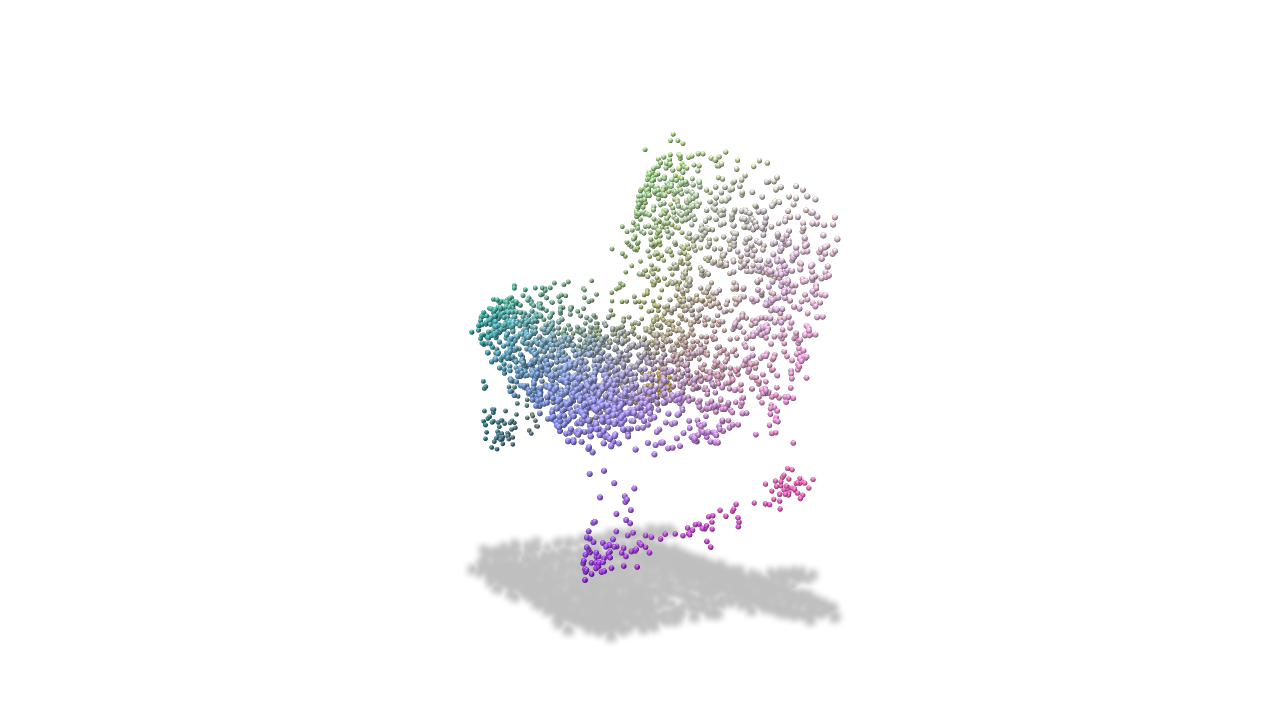}\hfill
	\adjincludegraphics[height=\alh,trim={ {\cch\width} {\cuthch\height} {\cch\width}  {\cuthch\height}},clip]{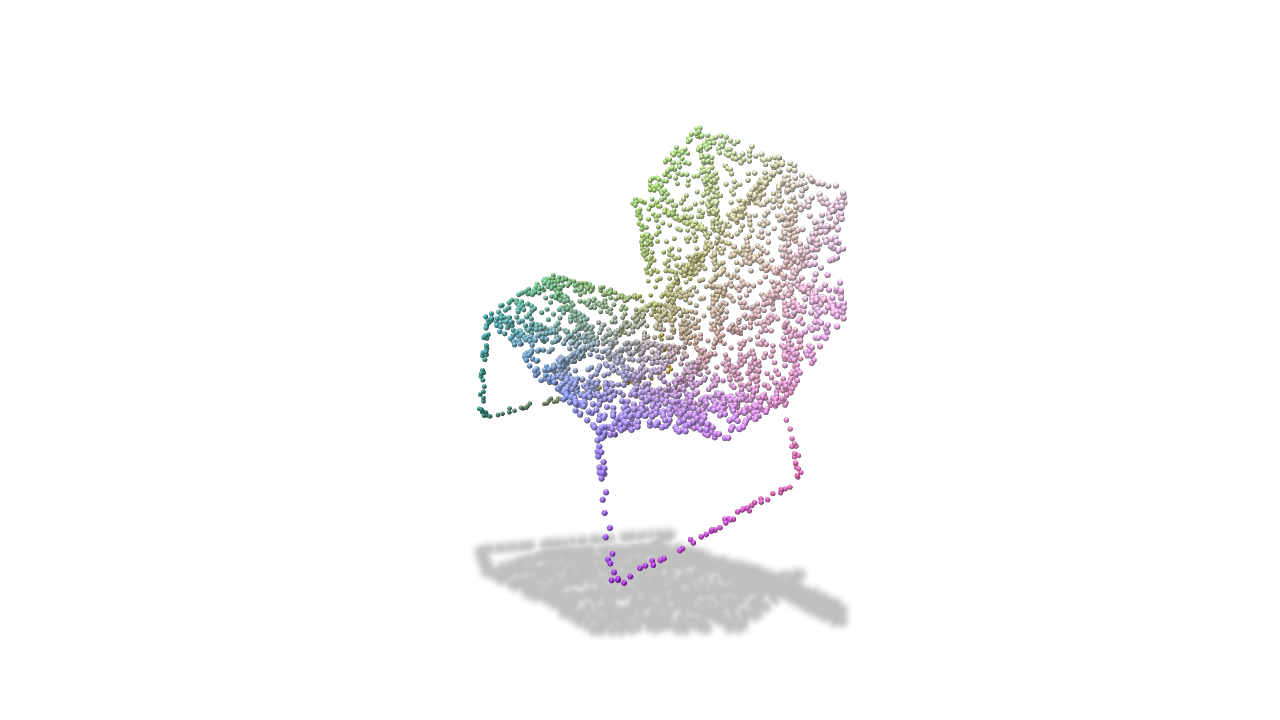} \\
	\adjincludegraphics[height=\aalh,trim={ {\ach\width} {\cuthch\height} {\ach\width}  {\cuthch\height}},clip]{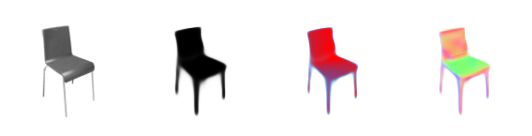}
	\adjincludegraphics[height=\alh,trim={ {\cch\width} {\cuthch\height} {\cch\width}  {\cuthch\height}},clip]{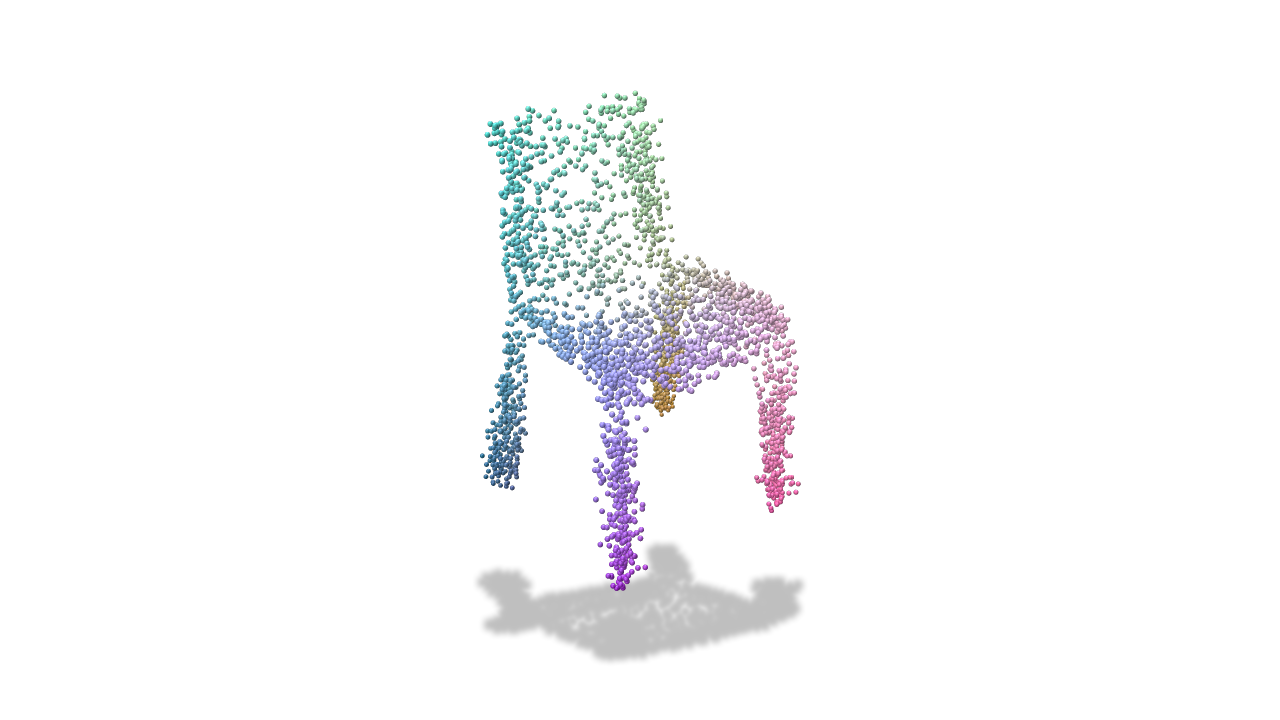}\hfill
	\adjincludegraphics[height=\alh,trim={ {\cch\width} {\cuthch\height} {\cch\width}  {\cuthch\height}},clip]{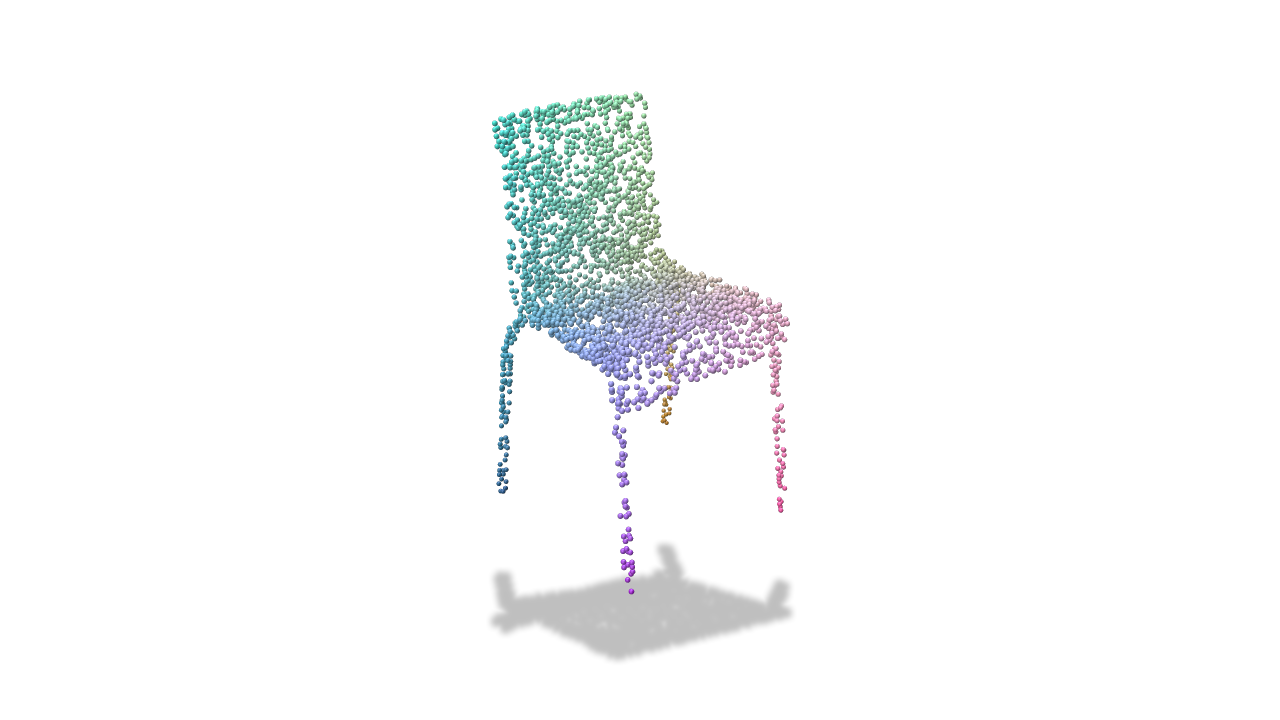}\hfill %
	\adjincludegraphics[height=\aalh,trim={ {\ach\width} {\cuthch\height} {\ach\width}  {\cuthch\height}},clip]{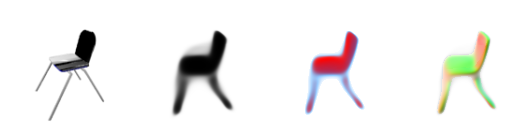}
	\adjincludegraphics[height=\alh,trim={ {\cch\width} {\cuthch\height} {\cch\width}  {\cuthch\height}},clip]{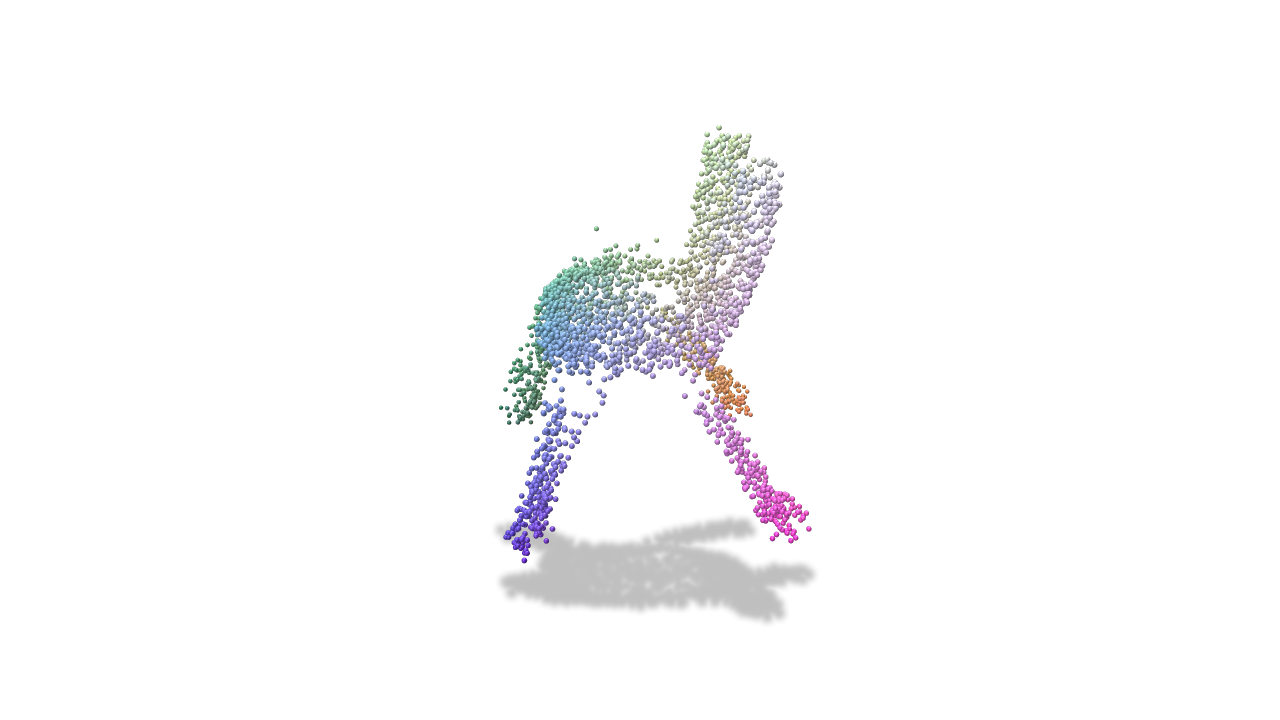}\hfill
	\adjincludegraphics[height=\alh,trim={ {\cch\width} {\cuthch\height} {\cch\width}  {\cuthch\height}},clip]{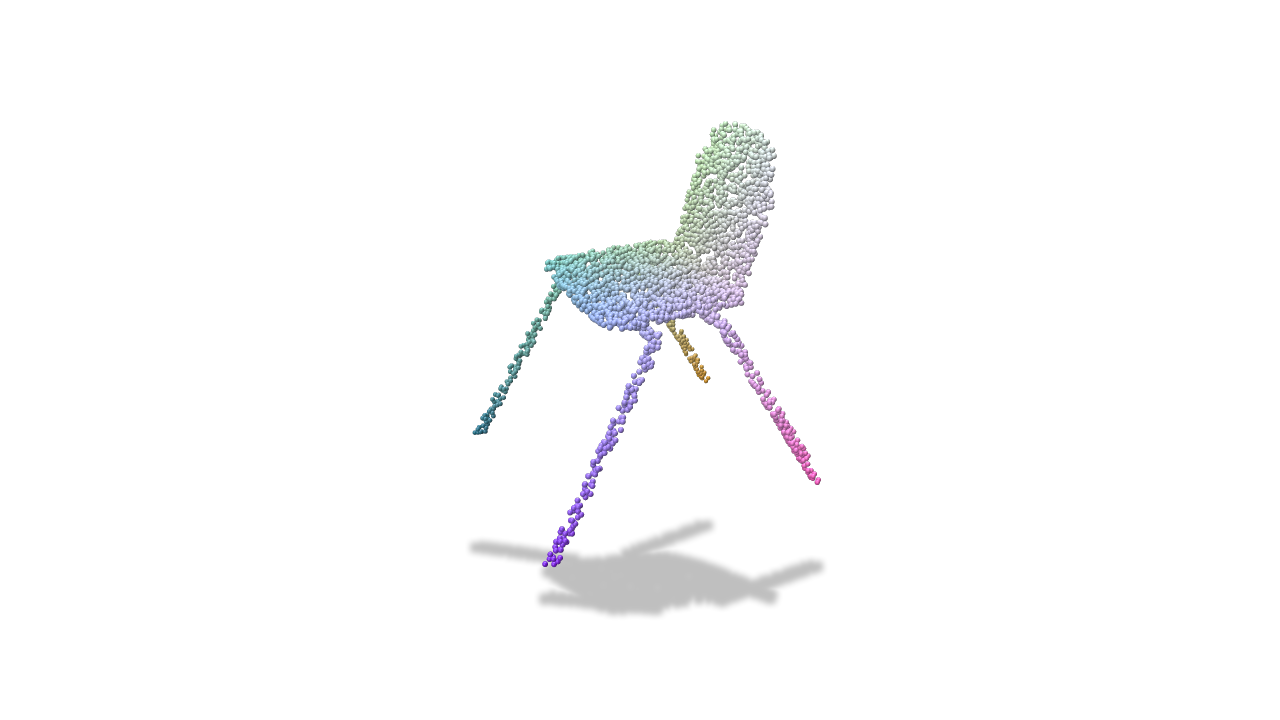} \\
	\adjincludegraphics[height=\aalh,trim={ {\ach\width} {\cuthch\height} {\ach\width}  {\cuthch\height}},clip]{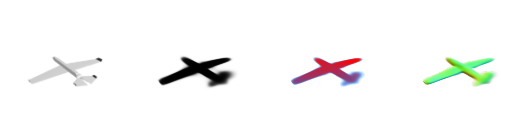}
	\adjincludegraphics[height=\alh,trim={ {\cch\width} {\cuthch\height} {\cch\width}  {\cuthch\height}},clip]{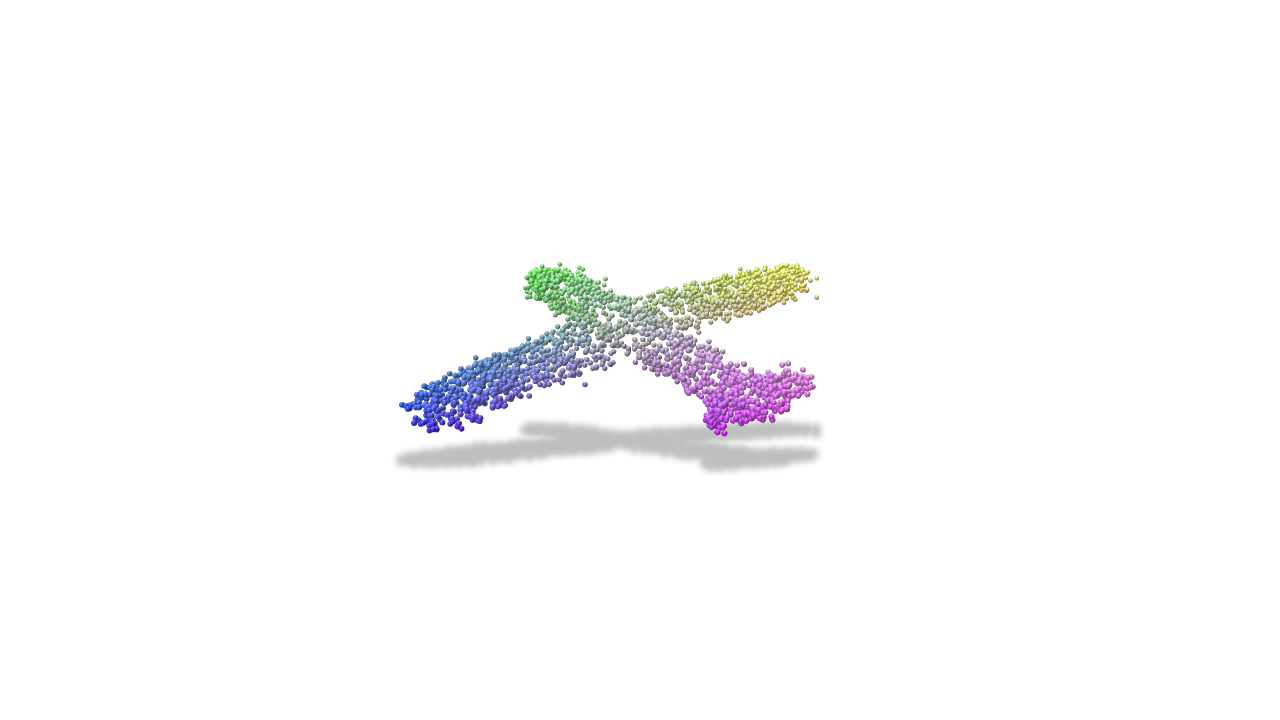}\hfill
	\adjincludegraphics[height=\alh,trim={ {\cch\width} {\cuthch\height} {\cch\width}  {\cuthch\height}},clip]{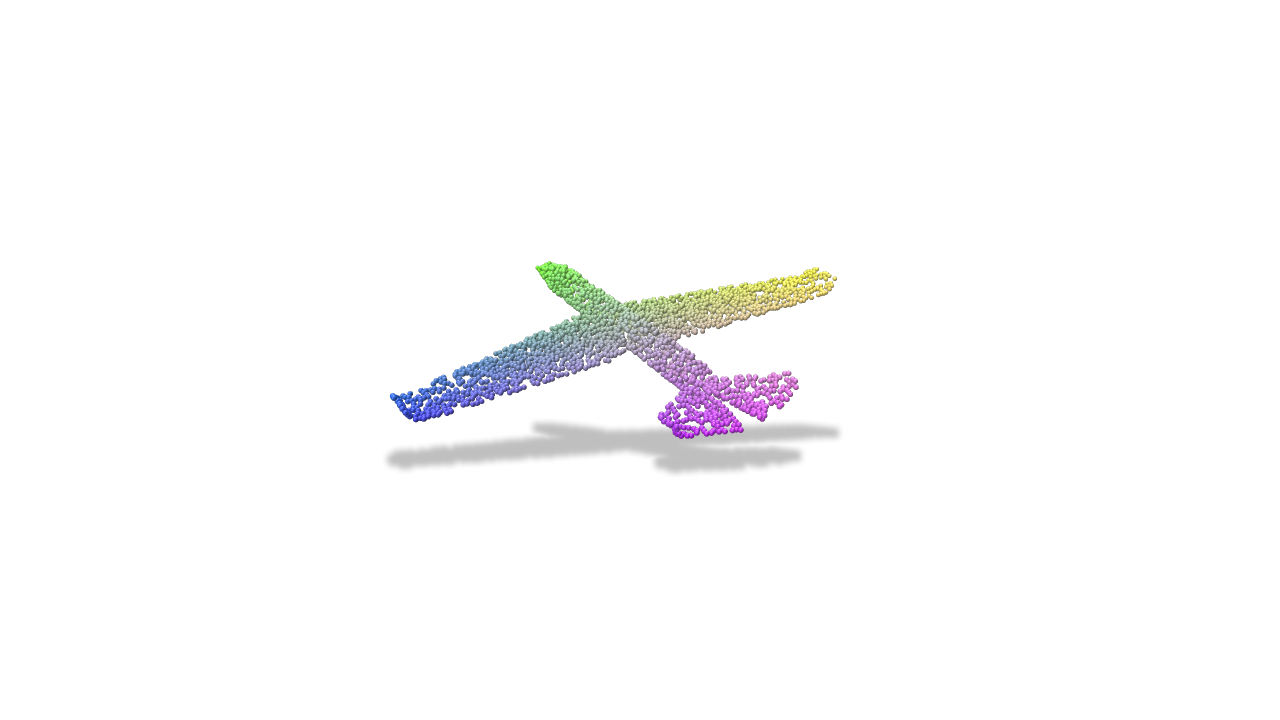}\hfill %
	\adjincludegraphics[height=\aalh,trim={ {\ach\width} {\cuthch\height} {\ach\width}  {\cuthch\height}},clip]{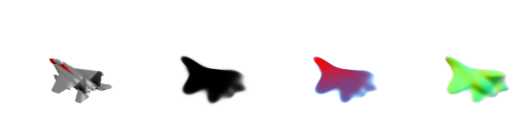}
	\adjincludegraphics[height=\alh,trim={ {\cch\width} {\cuthch\height} {\cch\width}  {\cuthch\height}},clip]{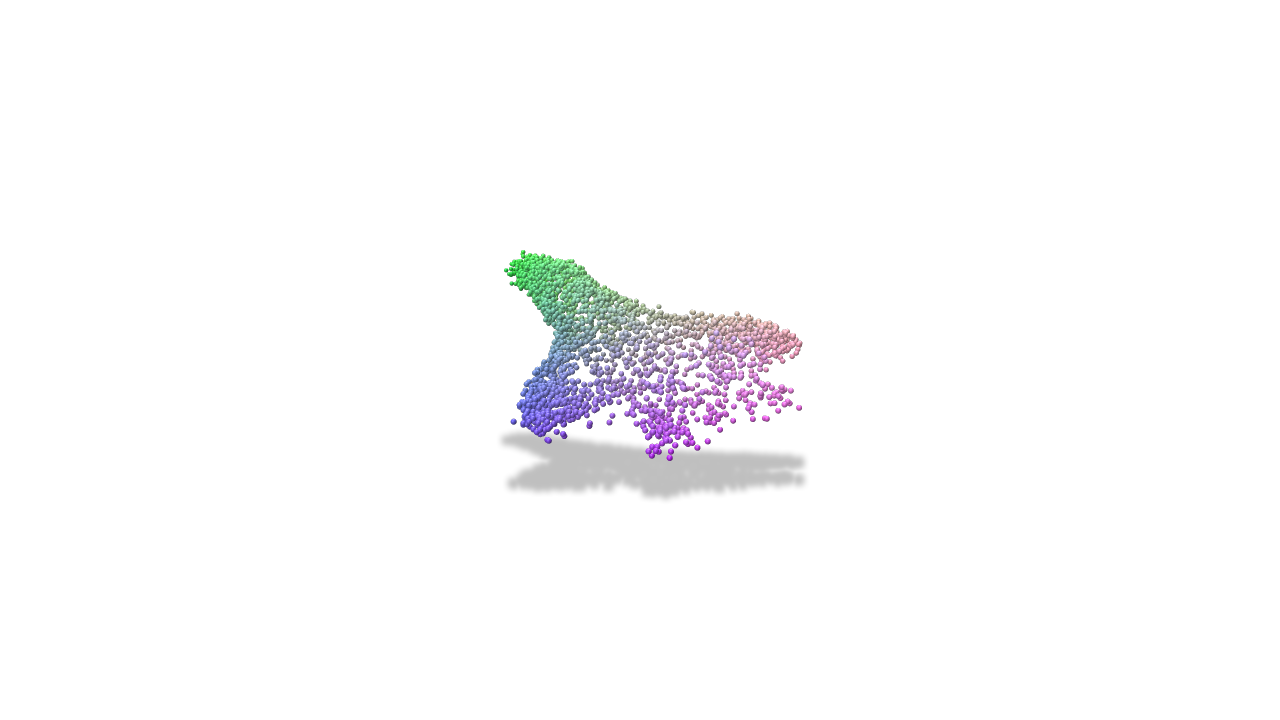}\hfill
	\adjincludegraphics[height=\alh,trim={ {\cch\width} {\cuthch\height} {\cch\width}  {\cuthch\height}},clip]{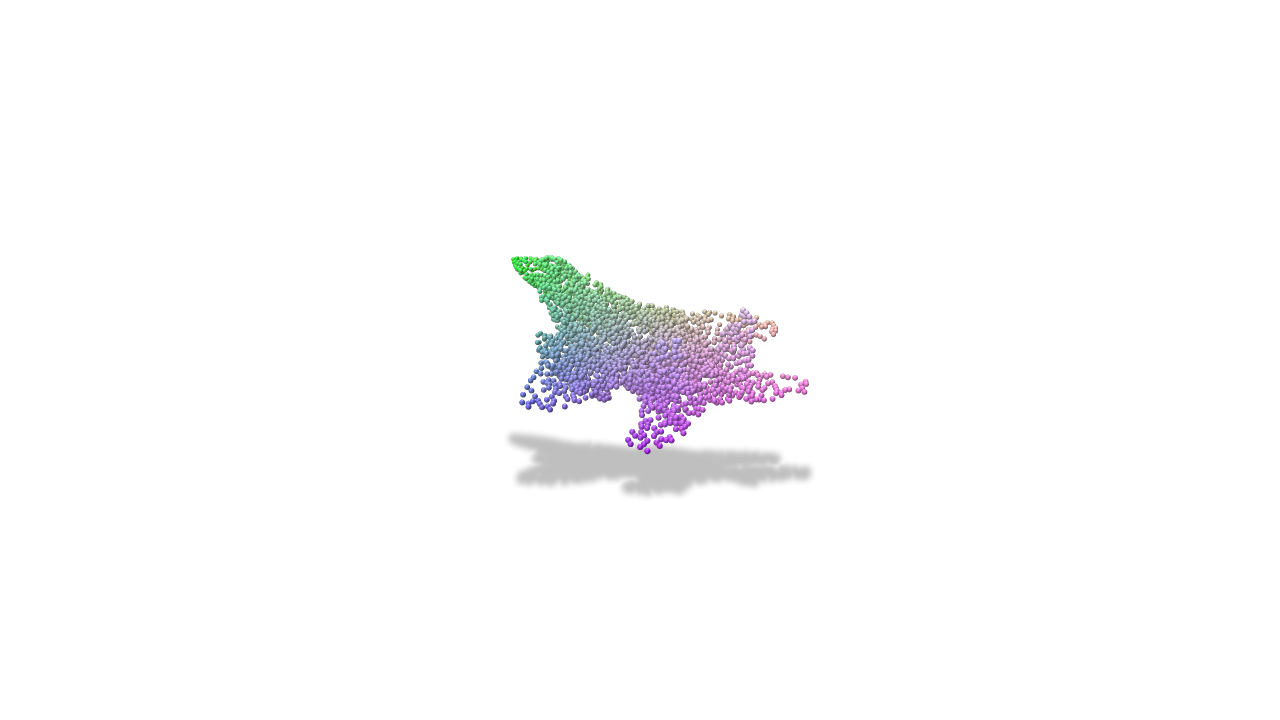} \\
	\adjincludegraphics[height=\aalh,trim={ {\ach\width} {\cuthch\height} {\ach\width}  {\cuthch\height}},clip]{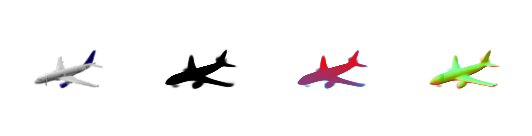}
	\adjincludegraphics[height=\alh,trim={ {\cch\width} {\cuthch\height} {\cch\width}  {\cuthch\height}},clip]{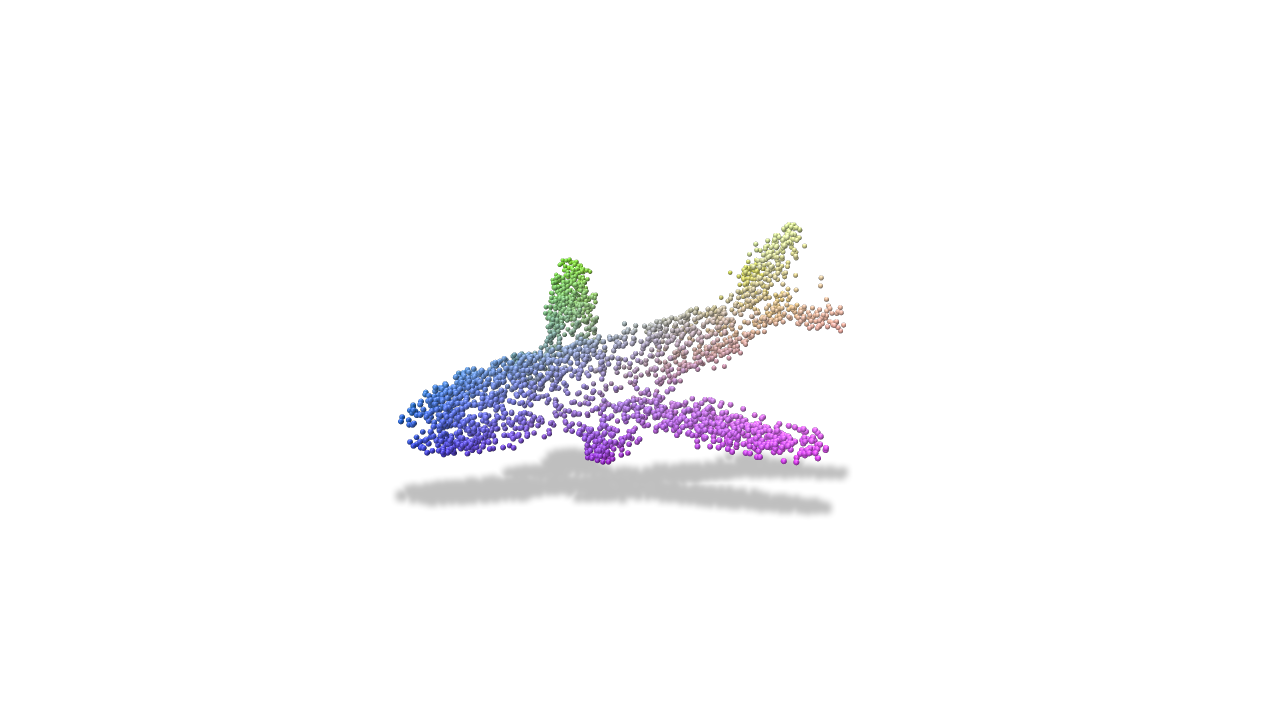}\hfill
	\adjincludegraphics[height=\alh,trim={ {\cch\width} {\cuthch\height} {\cch\width}  {\cuthch\height}},clip]{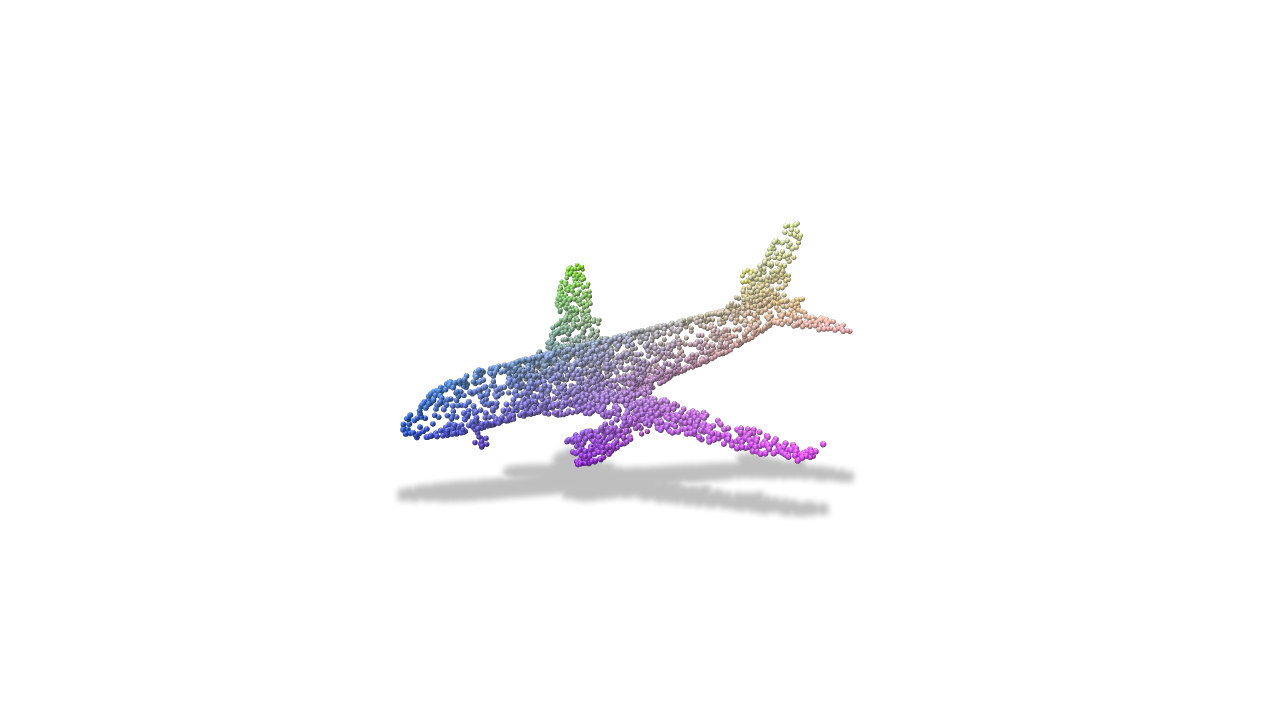}\hfill %
	\adjincludegraphics[height=\aalh,trim={ {\ach\width} {\cuthch\height} {\ach\width}  {\cuthch\height}},clip]{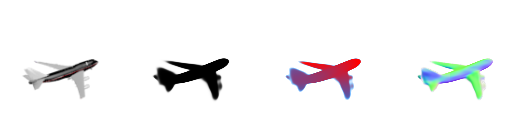}
	\adjincludegraphics[height=\alh,trim={ {\cch\width} {\cuthch\height} {\cch\width}  {\cuthch\height}},clip]{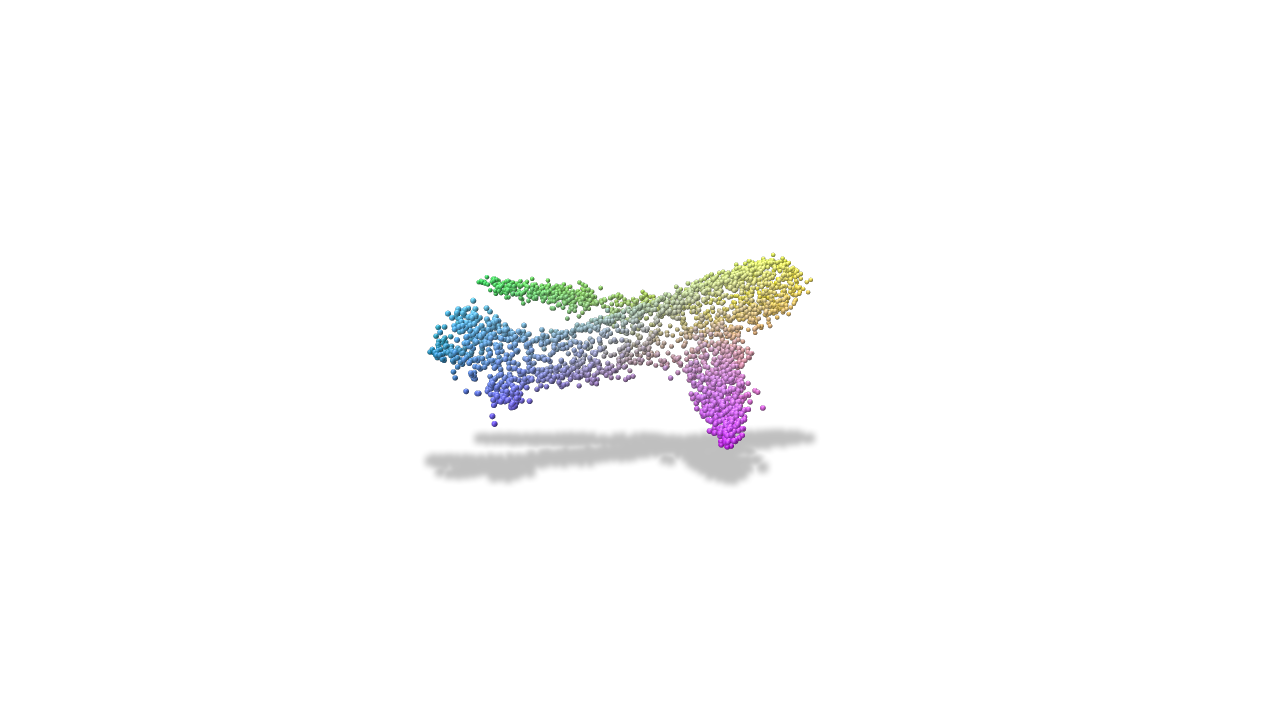}\hfill
	\adjincludegraphics[height=\alh,trim={ {\cch\width} {\cuthch\height} {\cch\width}  {\cuthch\height}},clip]{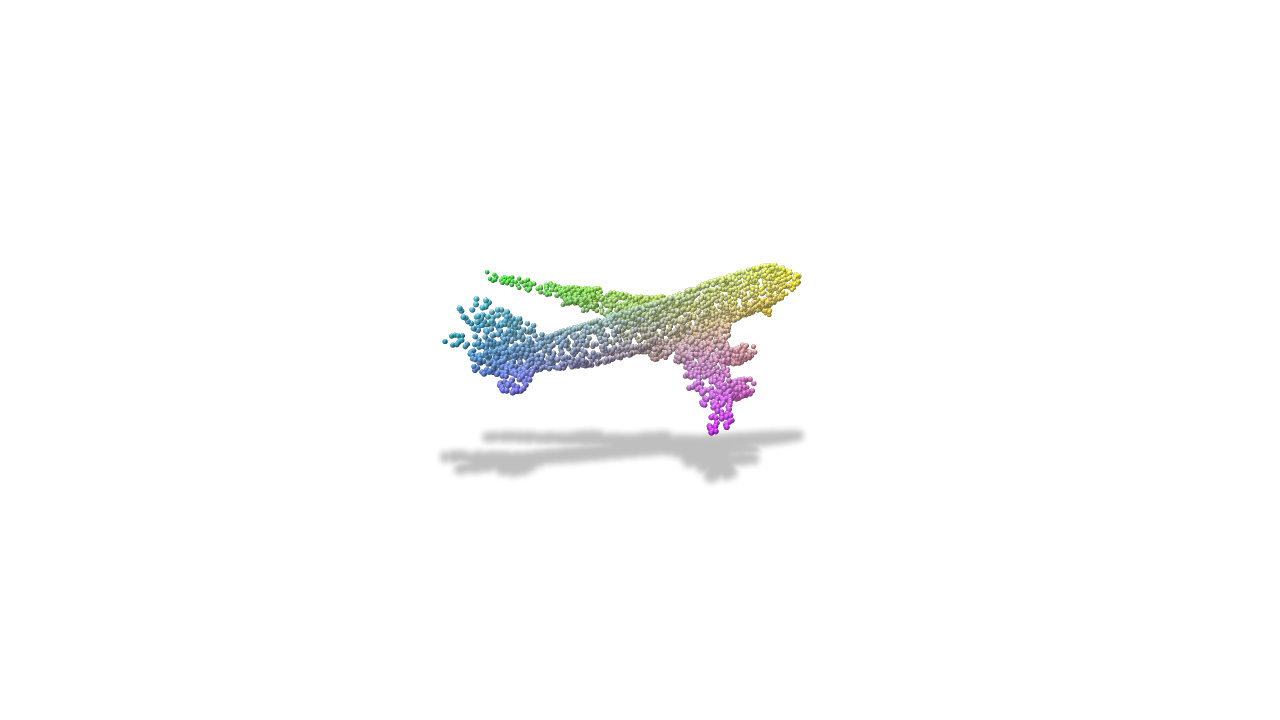} \\
	\adjincludegraphics[height=\aalh,trim={ {\ach\width} {\cuthch\height} {\ach\width}  {\cuthch\height}},clip]{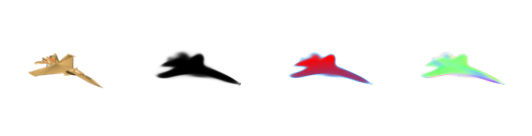}
	\adjincludegraphics[height=\alh,trim={ {\cch\width} {\cuthch\height} {\cch\width}  {\cuthch\height}},clip]{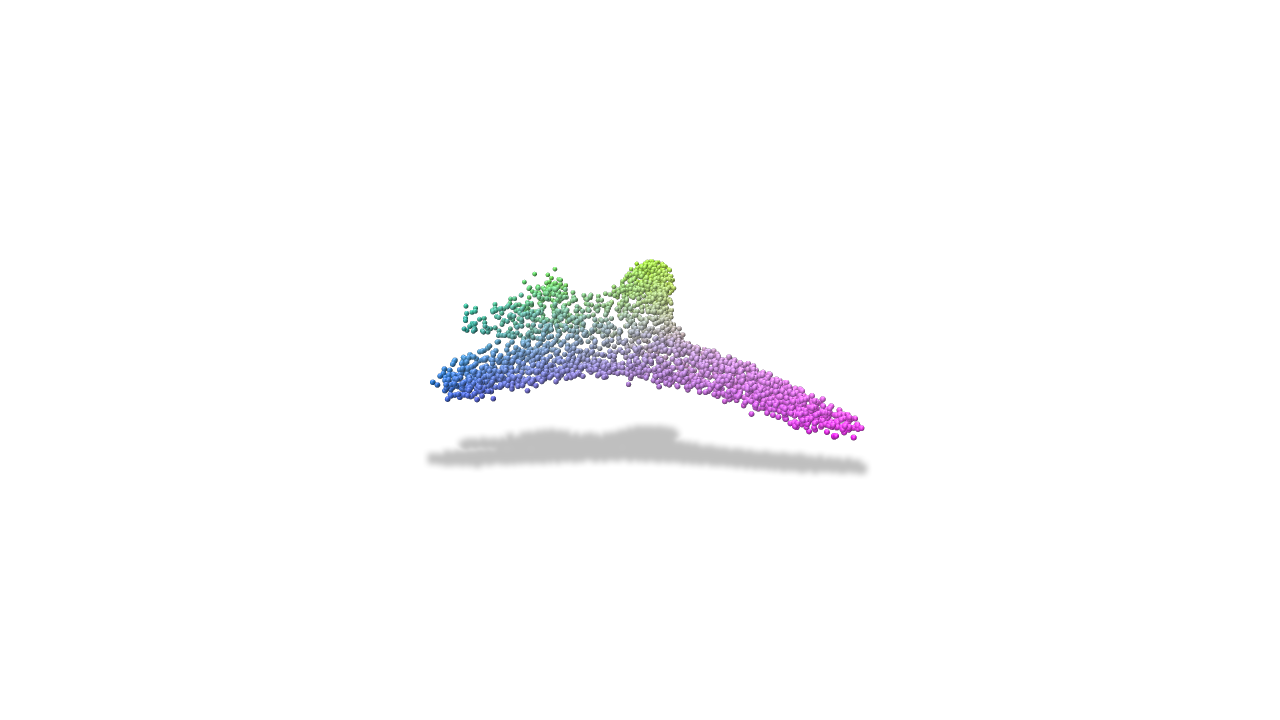}\hfill
	\adjincludegraphics[height=\alh,trim={ {\cch\width} {\cuthch\height} {\cch\width}  {\cuthch\height}},clip]{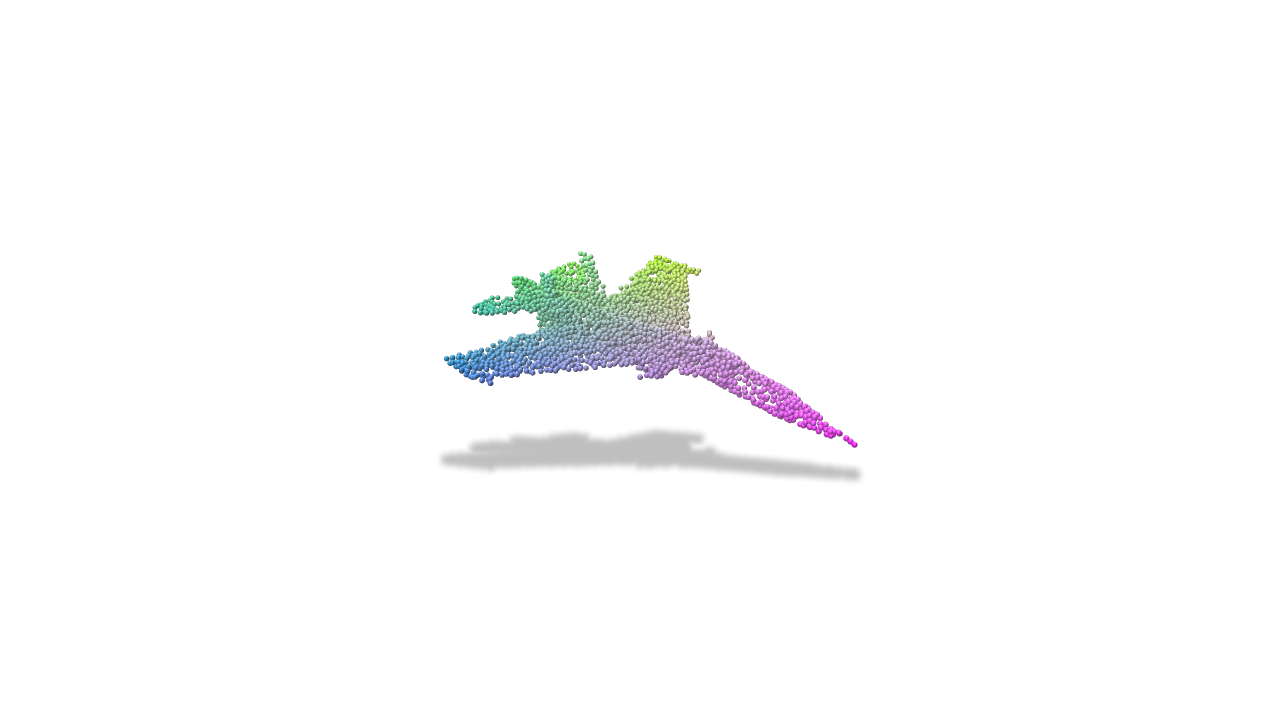}\hfill %
	\adjincludegraphics[height=\aalh,trim={ {\ach\width} {\cuthch\height} {\ach\width}  {\cuthch\height}},clip]{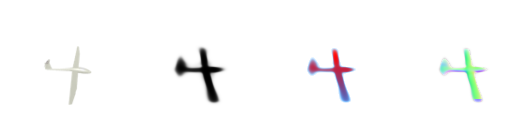}
	\adjincludegraphics[height=\alh,trim={ {\cch\width} {\cuthch\height} {\cch\width}  {\cuthch\height}},clip]{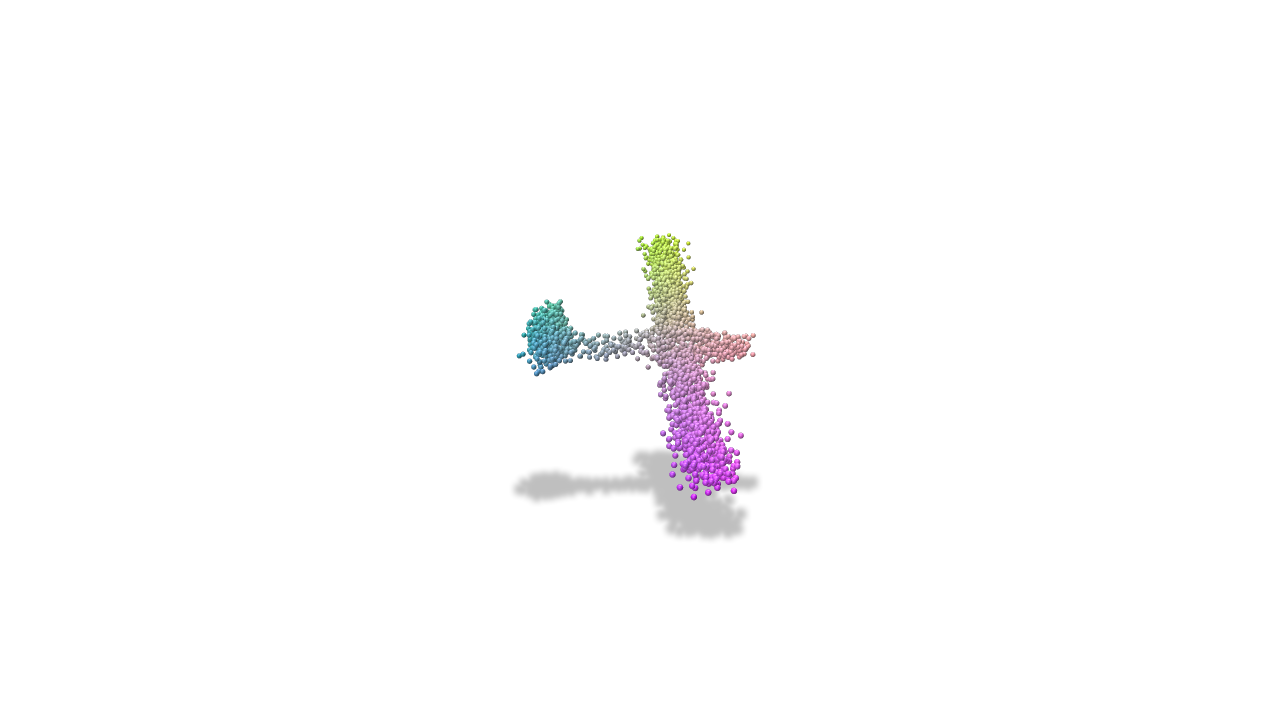}\hfill
	\adjincludegraphics[height=\alh,trim={ {\cch\width} {\cuthch\height} {\cch\width}  {\cuthch\height}},clip]{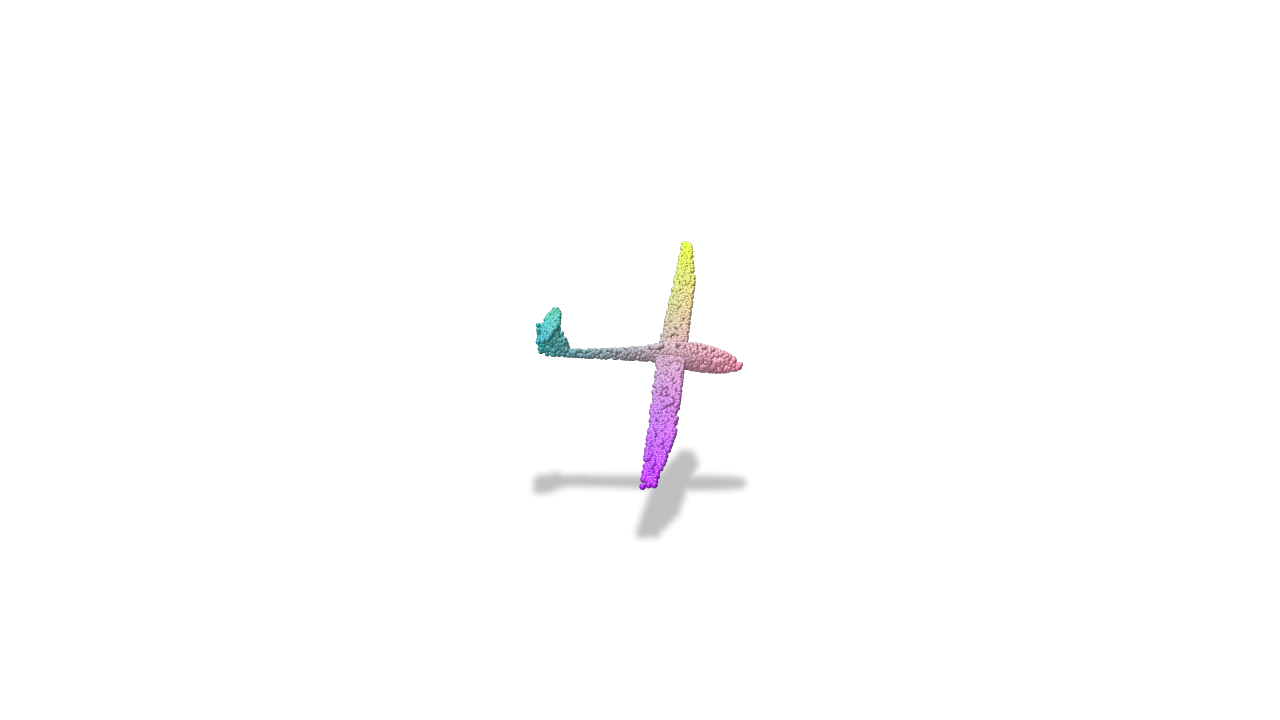} 
	\adjincludegraphics[height=\aalh,trim={ {\ach\width} {\cuthch\height} {\ach\width}  {\cuthch\height}},clip]{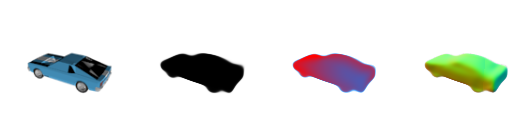}
	\adjincludegraphics[height=\alh,trim={ {\cch\width} {\cuthch\height} {\cch\width}  {\cuthch\height}},clip]{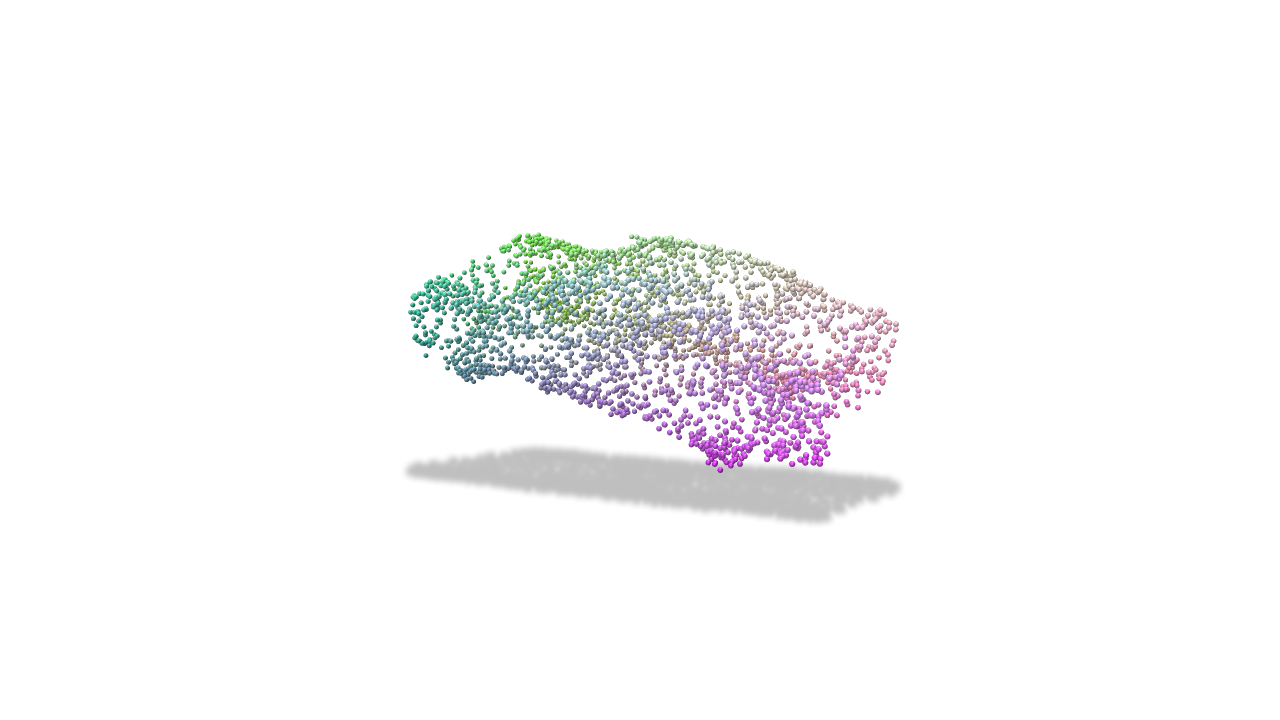}\hfill
	\adjincludegraphics[height=\alh,trim={ {\cch\width} {\cuthch\height} {\cch\width}  {\cuthch\height}},clip]{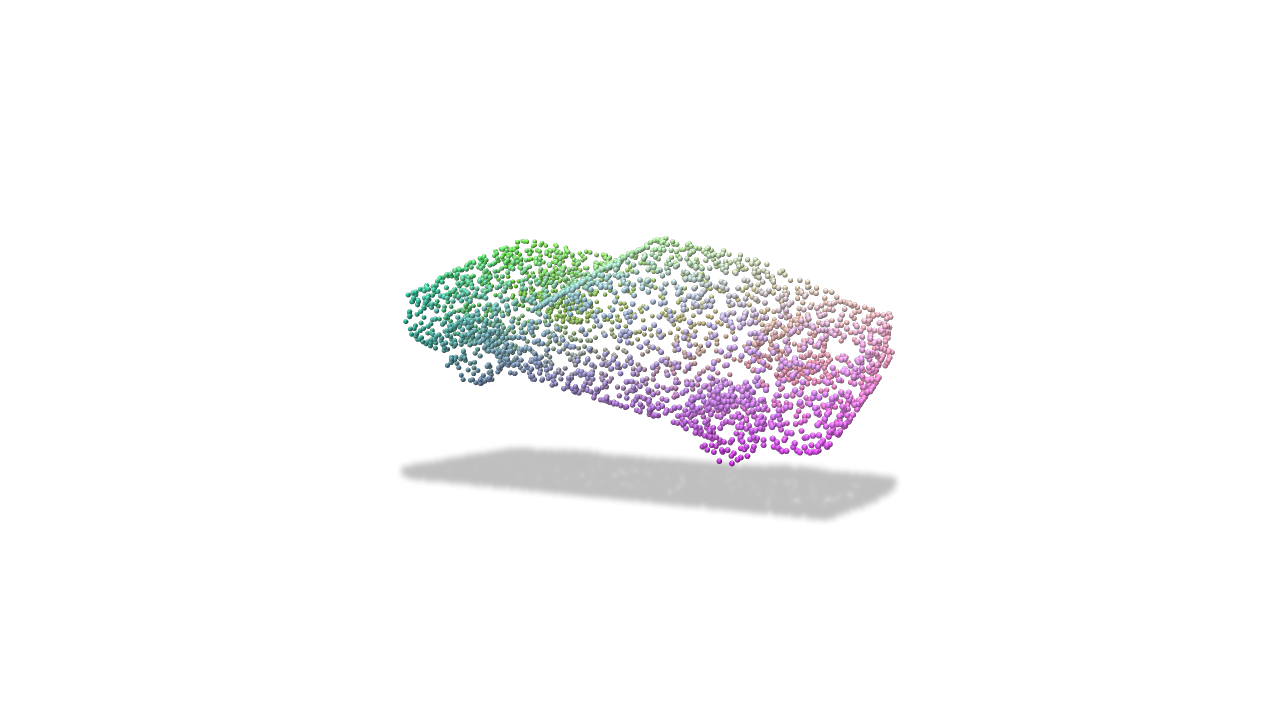}\hfill %
	\adjincludegraphics[height=\aalh,trim={ {\ach\width} {\cuthch\height} {\ach\width}  {\cuthch\height}},clip]{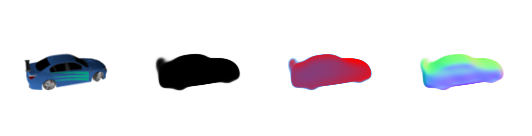}
	\adjincludegraphics[height=\alh,trim={ {\cch\width} {\cuthch\height} {\cch\width}  {\cuthch\height}},clip]{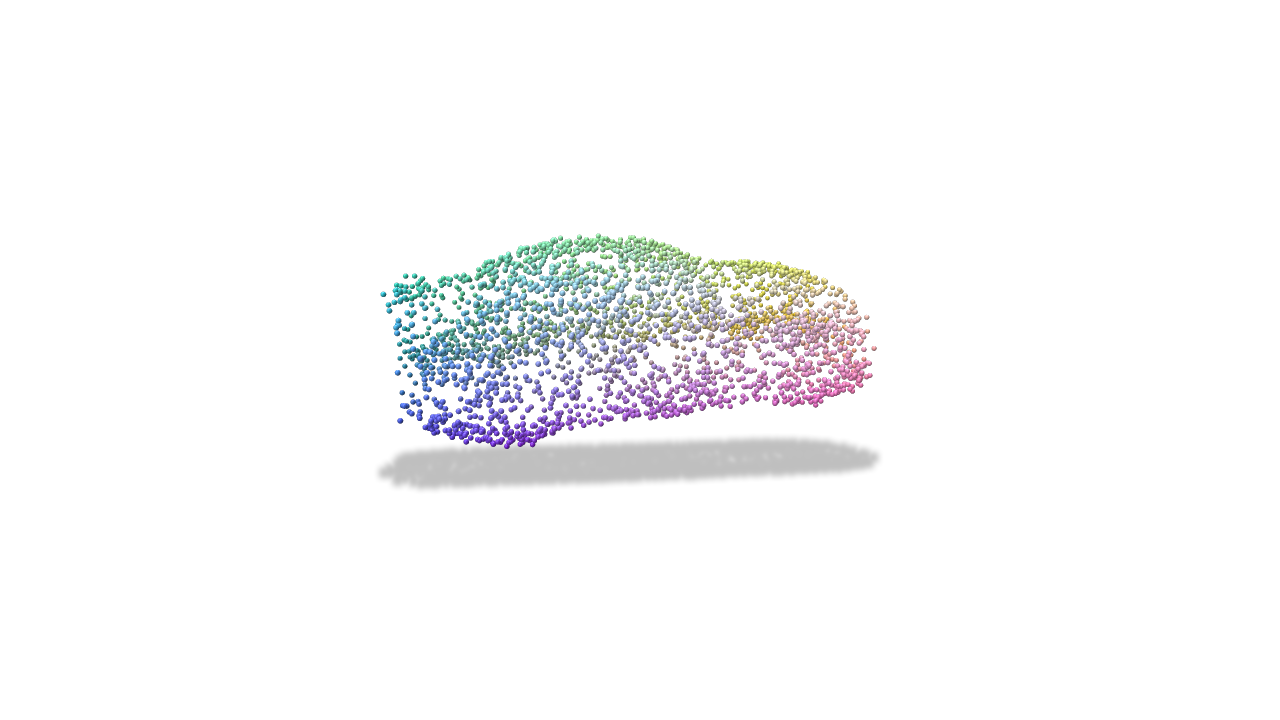}\hfill
	\adjincludegraphics[height=\alh,trim={ {\cch\width} {\cuthch\height} {\cch\width}  {\cuthch\height}},clip]{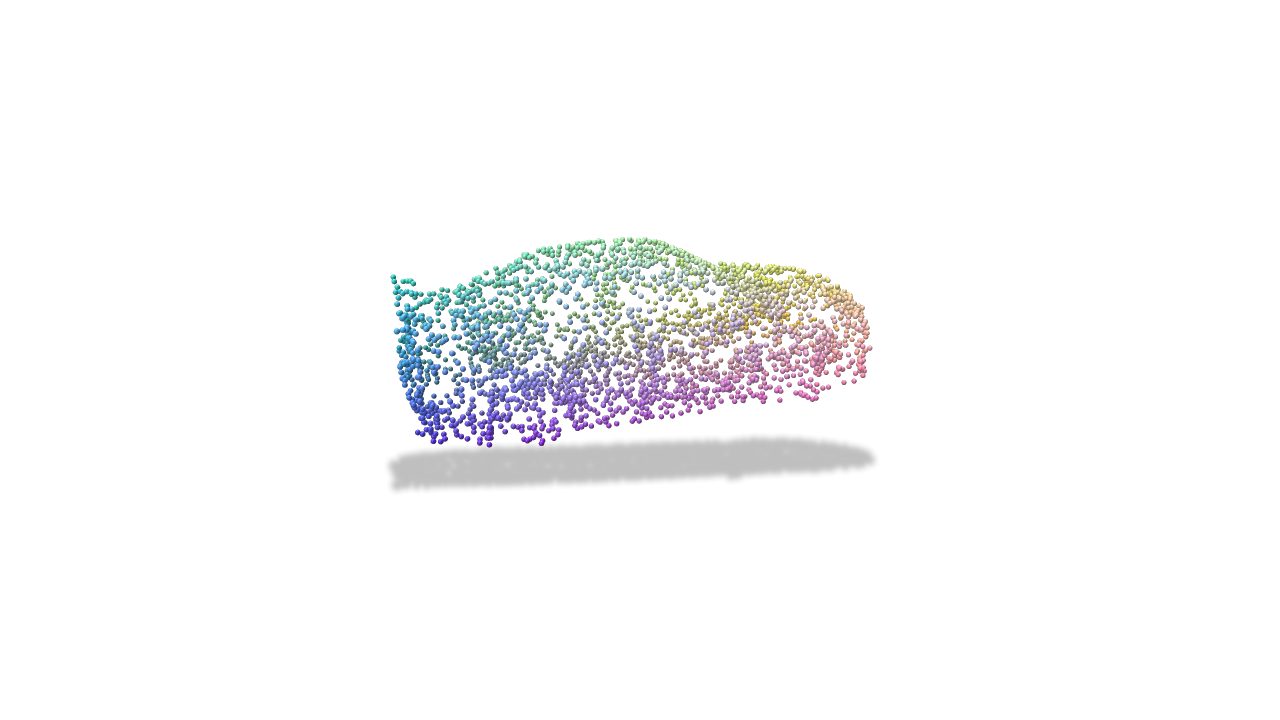} \\
	\adjincludegraphics[height=\aalh,trim={ {\ach\width} {\cuthch\height} {\ach\width}  {\cuthch\height}},clip]{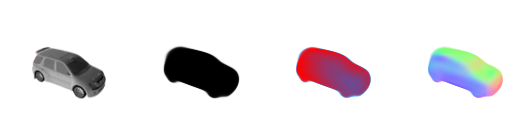}
	\adjincludegraphics[height=\alh,trim={ {\cch\width} {\cuthch\height} {\cch\width}  {\cuthch\height}},clip]{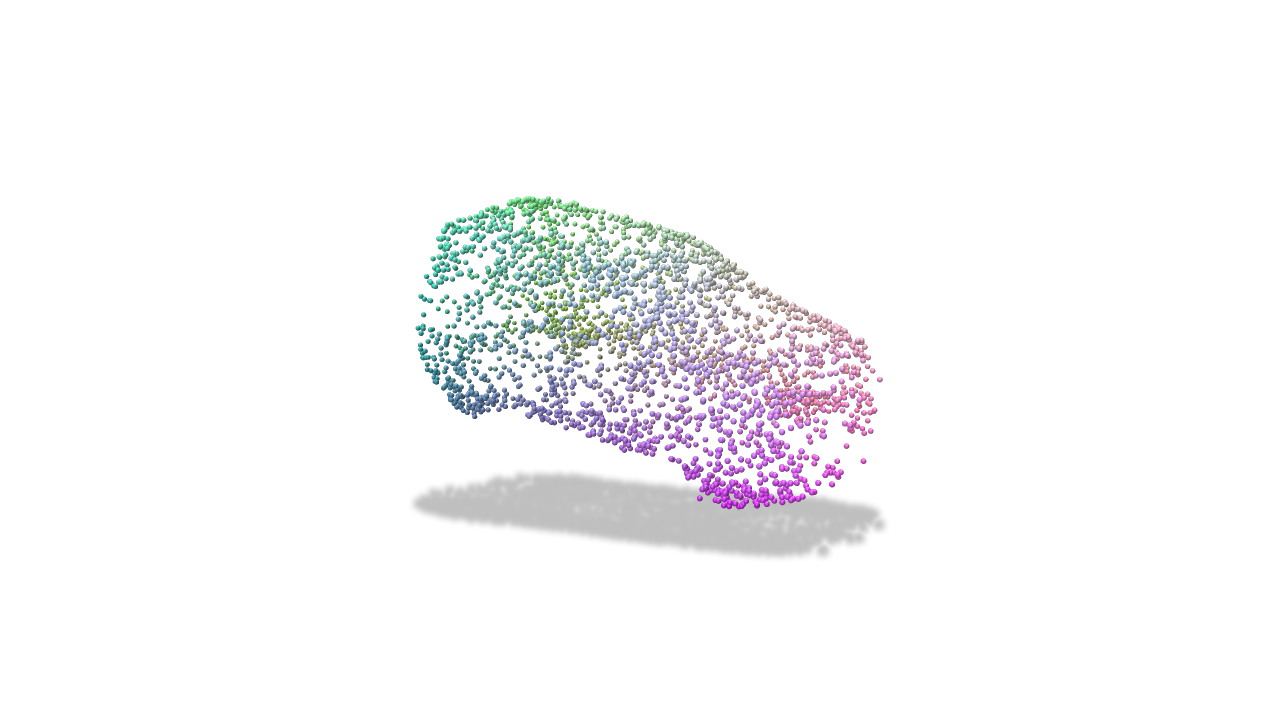}\hfill
	\adjincludegraphics[height=\alh,trim={ {\cch\width} {\cuthch\height} {\cch\width}  {\cuthch\height}},clip]{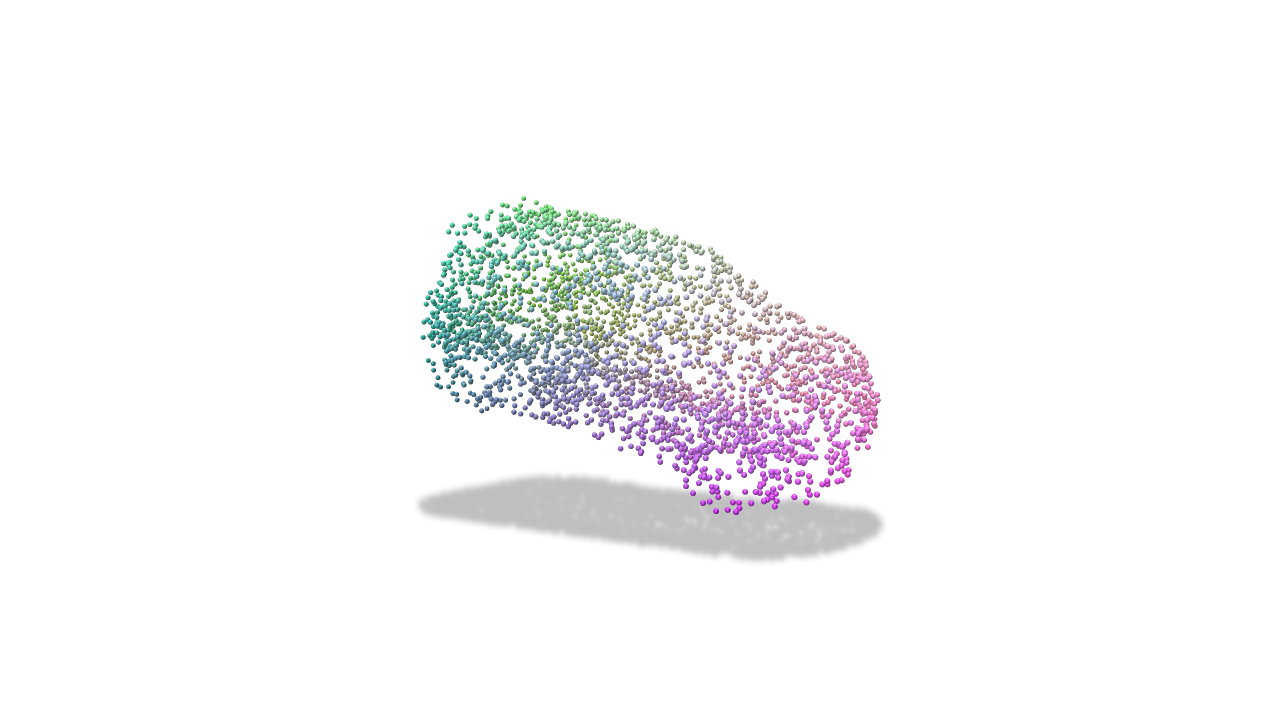}\hfill %
	\adjincludegraphics[height=\aalh,trim={ {\ach\width} {\cuthch\height} {\ach\width}  {\cuthch\height}},clip]{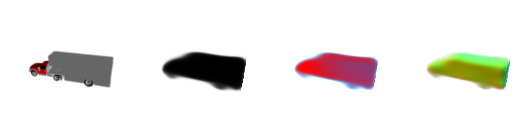}
	\adjincludegraphics[height=\alh,trim={ {\cch\width} {\cuthch\height} {\cch\width}  {\cuthch\height}},clip]{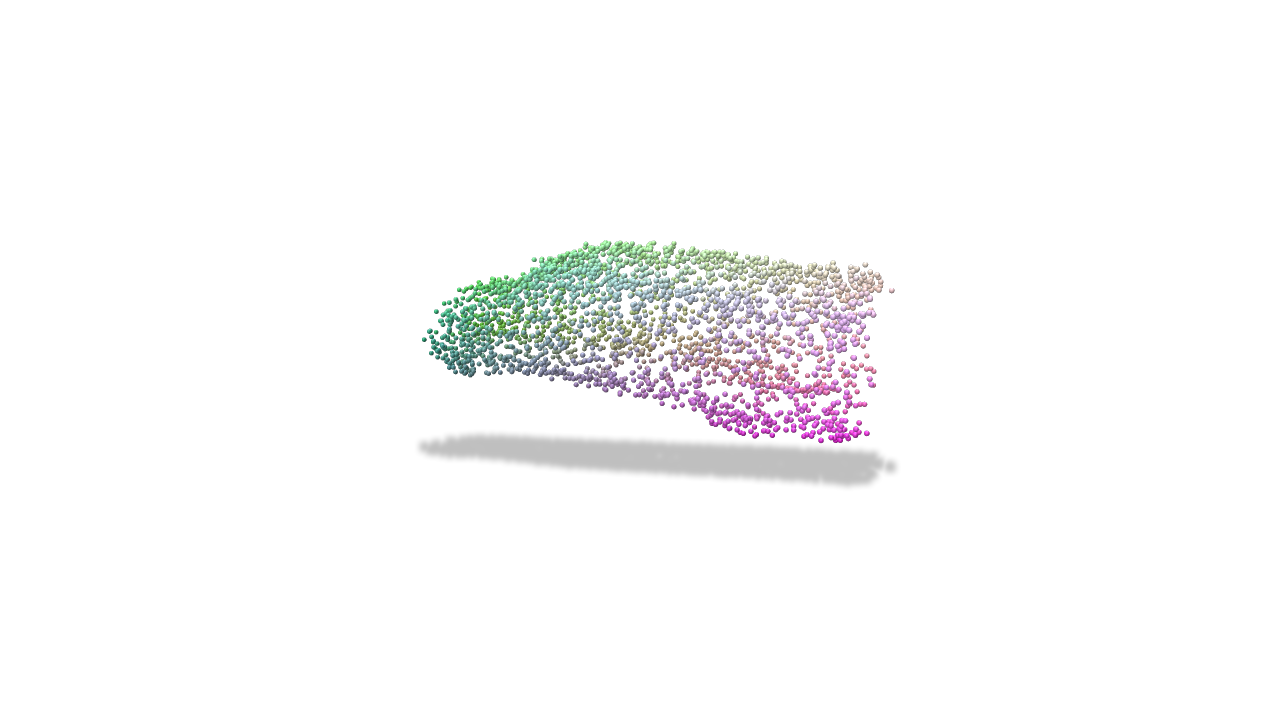}\hfill
	\adjincludegraphics[height=\alh,trim={ {\cch\width} {\cuthch\height} {\cch\width}  {\cuthch\height}},clip]{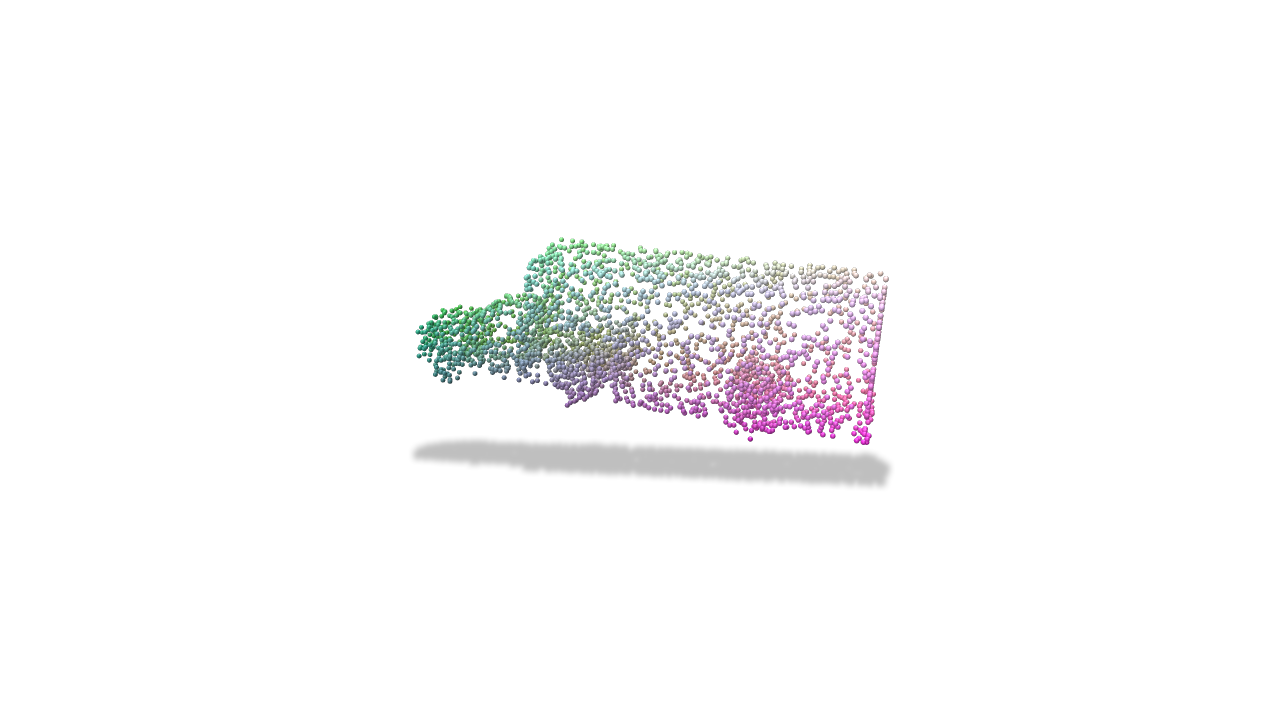} \\
	\adjincludegraphics[height=\aalh,trim={ {\ach\width} {\cuthch\height} {\ach\width}  {\cuthch\height}},clip]{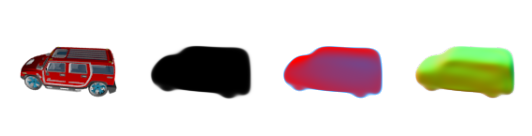}
	\adjincludegraphics[height=\alh,trim={ {\cch\width} {\cuthch\height} {\cch\width}  {\cuthch\height}},clip]{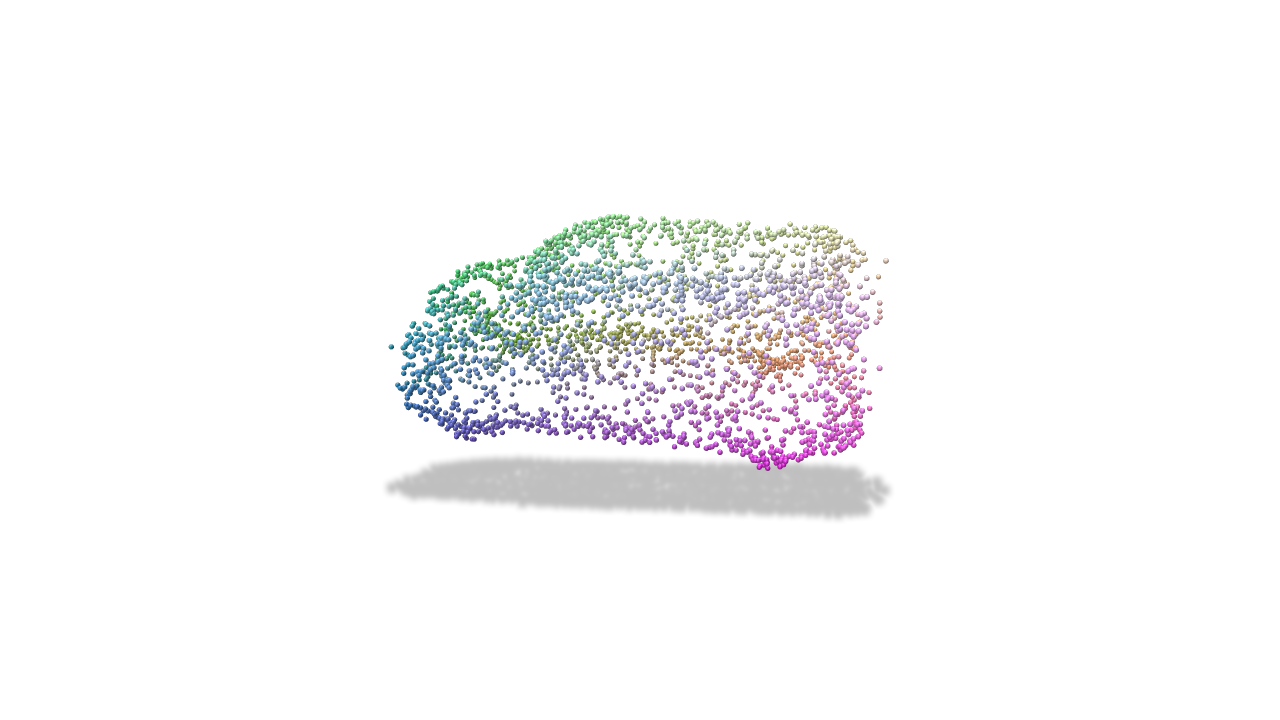}\hfill
	\adjincludegraphics[height=\alh,trim={ {\cch\width} {\cuthch\height} {\cch\width}  {\cuthch\height}},clip]{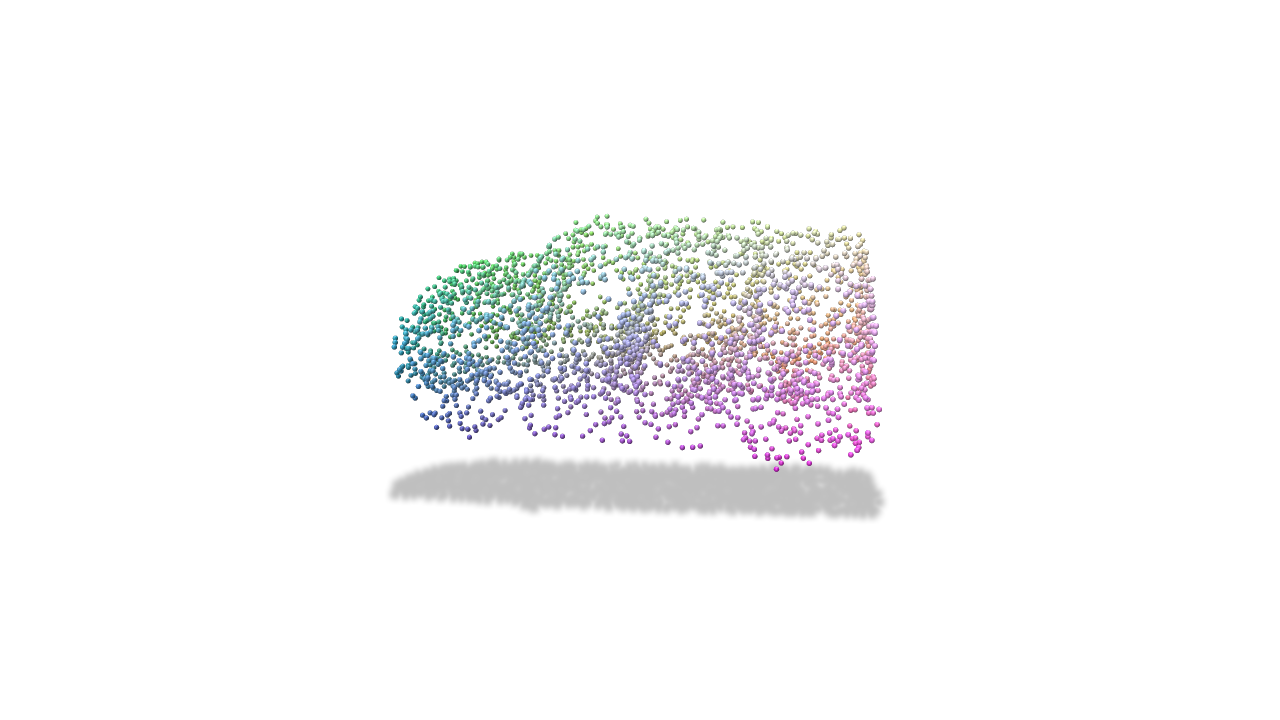}\hfill %
	\adjincludegraphics[height=\aalh,trim={ {\ach\width} {\cuthch\height} {\ach\width}  {\cuthch\height}},clip]{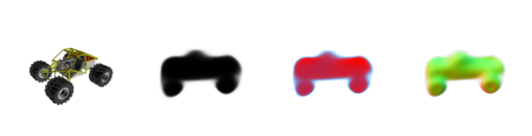}
	\adjincludegraphics[height=\alh,trim={ {\cch\width} {\cuthch\height} {\cch\width}  {\cuthch\height}},clip]{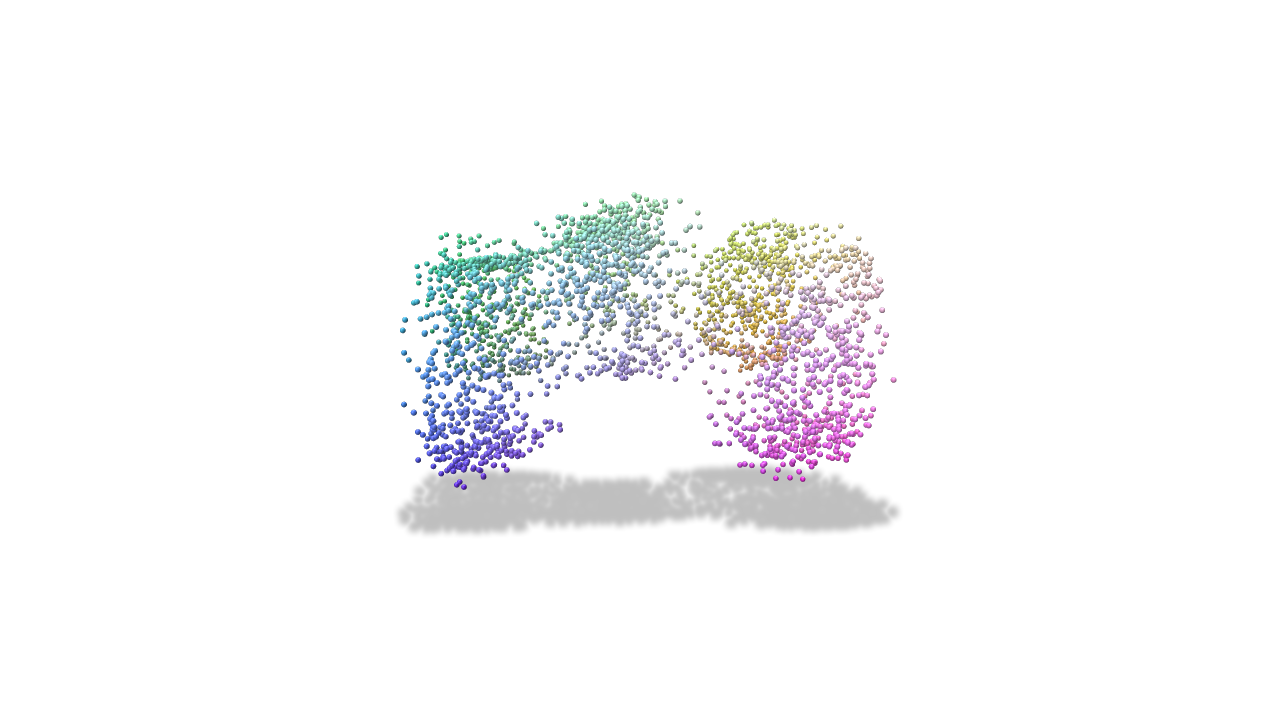}\hfill
	\adjincludegraphics[height=\alh,trim={ {\cch\width} {\cuthch\height} {\cch\width}  {\cuthch\height}},clip]{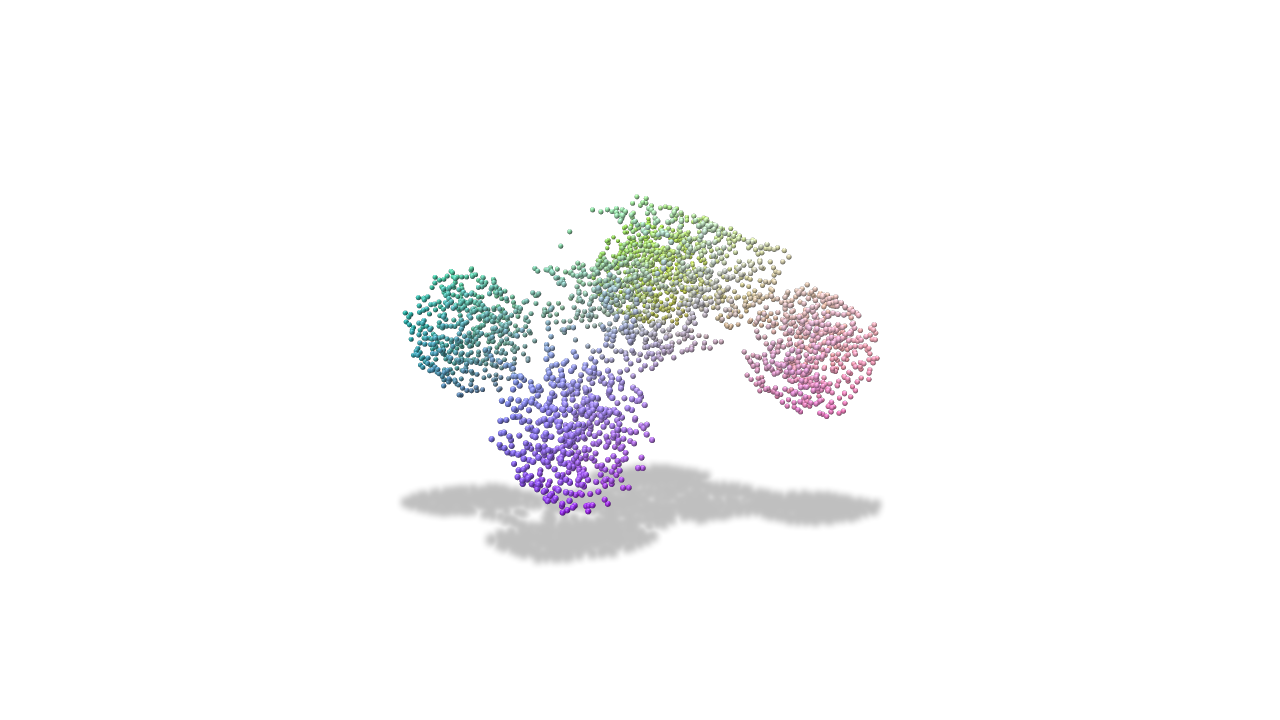}\\
	\caption{
		Single-image 3D reconstruction visualizations on held-out data. 
		Per inset, columns represent 
		(i) the input RGB image, 
		(ii) the visibility $\widehat{\xi}$, 
		(iii) the depth $\widehat{d}$, 
		(iv) the normals $\widehat{n}$, 
		(v) the sampled point cloud (PC) from the DDF, and 
		(vi) a sample from the ground-truth PC.
		Quantities (ii-v) are all differentiably computed directly from the CPDDF and $\widehat{\Pi}$, per point or pixel (i.e., no post-processing needed).
		PC colours denote 3D coordinates.
		A high-error example is in the lower-right of each category.
		See also Fig.\ \ref{fig:si3drvis} and Supp.~\S\ref{app:sec:si3draddvis}.
	}
	\label{app:fig:si3drvis}
\end{figure*}

\section{Single-Image 3D Reconstruction}
\label{appendix:si3dr}

\subsection{Model Architecture and Training Details}

\textbf{Architecture and Optimization.}
Our CPDDF is a modulated SIREN \cite{mehta2021modulated} with layer sizes \verb|(512, 512, 512,| \verb|256, 256,| \verb|256, 256)|. 
Note that we use a softplus activation instead of ReLU, when multiplying the modulator to the intermediate features.
The encoders (one ResNet-18 for the camera, and another for inferring the latent conditioning shape vector) take 128 $\times$ 128 RGBA images as input. 
However, when rendering the visibility mask for $\mathcal{L}_M$, we output 64 $\times$ 64 images.
We set $\gamma_{R,S} = 1$, $\gamma_{R,\Pi} = 5$, and $\gamma_{R,M} = 10$.
We also used small weight decays on the camera and shape predictors ($10^{-6}$ and $10^{-3}$, respectively). %
Training ran for 100K iterations with AdamW \cite{loshchilov2018decoupled} and a batch size of 32. 
We used $\dim(z_s) = 512$ and a SIREN initializer of $\omega_0 = 1$. 

\textbf{Shape-fitting Loss.}
We use the same settings as the single-shape fitting experiments, with slight modifications:
$\gamma_d = 5$,
$\gamma_\xi = 10$,
$\gamma_n = 10$,
$\gamma_V = 1$,
$\gamma_\mathrm{E,d} = 0.1$, 
$\gamma_\mathrm{E,\xi} = 0.25$, 
and
$\gamma_T = 0.1$.
An additional loss on the variance of $\xi$ is also applied 
(to reduce the visibility entropy at each point, which leads to fuzzy renders):
\begin{equation}
	\mathcal{L}_{V,\xi} = \gamma_{V,\xi} \xi(p,v)[ 1 - \xi(p,v) ],
\end{equation}
where we set
$\gamma_{V,\xi} = 0.25$.
This encourages less blurriness in the shape output due to uncertainty in the visibility field.
For minibatches, we sample 1.2K (A and U) and 300 (S, B, T, and O) oriented points per shape, with each minibatch containing 32 shapes.

\textbf{Camera Loss.}
The camera fitting loss $\mathcal{L}_\Pi$ utilizes a camera representation only involving extrinsic position (the camera is assumed to be looking at the origin). 
In particular, we use the azimuth, elevation, and radius representation of the camera position.
Before computing the $L_2$ loss, we z-score normalize each element, based on the training set statistics.
We also restrict the predicted camera $\widehat{\Pi}$ to be within the range of the parameters observed in the dataset.

\textbf{Rendering Scale Factor.}
We note that an additional scale factor is needed for rendering DDFs for ShapeNet. 
Since ShapeNet shapes are normalized with an instance-dependent measure (the bounding box diagonal), one needs to know the scale to reproduce the output image.
This is an issue as our CPDDF always outputs a shape in $[-1,1]^3$, with training data normalized to have longest axis-aligned bounding box length equal to two.
Further, it creates an ambiguity (with respect to the output image) with the camera position (radius from the shape).
At train time, before rendering to compute $\mathcal{L}_M$, 
we use the ground-truth scale factor. 
At test time, we estimate it by sampling a point cloud and measuring the diagonal.

\textbf{Data Extraction.}
We use the ShapeNet-v1 \cite{shapenet2015} data and splits from Pixel2Mesh \cite{wang2018pixel2mesh,wang2020pixel2mesh}, with the renders by Choy et al.\  \cite{choy20163d}.
For DDF training, 
per data type,
we sample 20K (A and U) and 10K (S, B, T, and O) training samples per shape,
using the watertight form of the meshes (via \cite{Kaolin}), 
decimated to 10K triangles.
These samples are only used for training, not evaluation, and are in a canonical aligned pose.
We set the maximum offset size for O-type data as $\epsilon_O = 0.02$; remaining parameters are the same as those used for single shapes (see \S\ref{appendix:singlefits}).

\textbf{Explicit Sampling Details.}
We also remark that our explicit sampling algorithm slightly oversamples points initially (when requesting a point set of size $n_p$, we actually sample $(1 + \varepsilon_p) n_p$ via $p \sim \mathcal{U}[\mathcal{B}]$).
The final point cloud, however, is sorted by visibility (i.e., by $\xi(p,\widehat{v}^{\,*}(p))$) and only the top $n_p$ points are returned.
We used $\varepsilon_p = 0.1$ in all experiments.
This is to prevent outputting non-visible points.

\textbf{PC-SIREN Baseline Details.}
Recall that the PC-SIREN is our architecture-matched baseline, with identical encoders and a nearly identical decoder architecture to the DDF one. 
Here, the decoder is a mapping $f_b : \mathbb{R}^3 \rightarrow \mathbb{R}^3$, which is trained to compute $f_b(p)\in S$ from $p\sim[-1,1]^3$, but uses an identical set of SIREN hidden layers as the DDF. 
A set of sampled random points can thus be mapped into a point cloud of arbitrary size.
The baseline has 25,645,574 parameters, while the DDF-based model has 25,647,367 parameters.
The camera loss $\mathcal{L}_\Pi$ is unchanged and the mask-matching loss $\mathcal{L}_M$ is not used.
The shape-fitting loss $\mathfrak{L}_S$ is replaced with a standard Chamfer loss $D_C$ \cite{chamfer}, computed with 1024 points per shape with a batch-size of 32. The remaining aspects of training remain the same.

\subsection{Table of Detailed Results}
\label{appendix:si3dr:detresults}

We show complete results across all categories in Table \ref{tab:si3dr}.

\begin{table*}[t] %
	{
		\tabcolsep3.1pt 
		\centering
		\resizebox{\textwidth}{!}{%
			\begin{tblr}{cr|ccccc | cccc | c c}
				&  & \multicolumn{5}{c|}{DDF} & 
				\multicolumn{4}{c|}{PC-SIREN} & \multirow{2}{*}{P2M} & \multirow{2}{*}{3DR} \\
				&  & $\Pi_g$-L & $\Pi_g$-S & $\widehat{\Pi}_\nabla$-S & $\widehat{\Pi}$-L  & $\widehat{\Pi}$-S  
				& $\Pi_g$-L  & $\Pi_g$-S  & $\widehat{\Pi}$-L  & $\widehat{\Pi}$-S &  &  \\\hline 
				& $D_C$  $ \downarrow$ & 0.459 & 0.512 & 0.823 & 0.855  & 0.919 
				& 0.431 & 0.465 & 0.876 & 0.915 
				& 0.610 & 1.432 \\ 
				Chairs & $F_{\tau}$ $ \uparrow$ & 55.47 & 48.40 & 42.28 & 47.51 & 41.08 
				& 62.25 & 56.36 & 50.56 & 45.57 
				& 54.38 & 40.22 \\ 
				& $F_{2\tau}$ $ \uparrow$ & 72.82 & 67.75 & 60.39 & 63.81 & 58.98 
				& 77.38 & 74.56 & 65.43 & 62.76 
				& 70.42 & 55.20 \\ \hline[dashed] 
				& $D_C$  $ \downarrow$ & 0.210 & 0.239 & 0.673 & 0.793 & 0.836 
				& 0.215 & 0.227 & 0.829 & 0.844 
				& 0.477 & 0.895 \\ 
				Planes & $F_{\tau}$  $ \uparrow$ & 80.46 & 76.62 & 63.32 & 63.75 & 60.54 
				& 81.49 & 80.11 & 63.64 & 62.34 
				& 71.12 & 41.46 \\ 
				& $F_{2\tau}$  $ \uparrow$ & 90.05 & 88.55 & 76.16 & 74.96 & 73.47 
				& 89.71 & 89.18 & 74.76 & 74.18 
				& 81.38 & 63.23 \\ \hline[dashed]
				& $D_C$  $ \downarrow$ & 0.231 & 0.288 & 0.390 & 0.541 & 0.606 
				& 0.371 & 0.400 & 0.737 & 0.768 
				& 0.268 & 0.845 \\ 
				Cars & $F_{\tau}$  $ \uparrow$ & 70.91 & 59.93 & 54.16 & 62.69 & 52.47 
				& 64.57 & 57.82 & 56.01 & 50.04
				& 67.86 & 37.80 \\ 
				& $F_{2\tau}$  $ \uparrow$ & 86.57 & 79.66 & 74.68 & 79.71 & 72.78 
				& 78.72 & 76.00 & 71.22 & 68.52 
				& 84.15 & 54.84 \\ \hline
				& $D_C$  $ \downarrow$ & 0.300 & 0.346 & 0.629 & 0.730 & 0.787  & 0.339 & 0.364 &  0.814 &  0.842 & 0.452 & 1.057 \\ 
				Avg & $F_{\tau}$  $ \uparrow$ & 68.95 & 61.65 & 53.25 & 57.98 & 51.36 & 69.44 & 64.76 & 56.74 & 52.65 & 64.45 & 39.83  \\ 
				& $F_{2\tau}$  $\uparrow$ & 83.15 & 78.65 & 70.41 & 72.83 & 68.41 &       81.94 &	79.91 & 70.47 & 68.49 & 78.65 & 57.76
				\\ 
		\end{tblr}}
	}
	\caption[Single Image 3D Reconstruction (SI3DR) with PDDFs]{
		Single-image 3D reconstruction results.
		Rows: ShapeNet categories and performance metrics.
		Columns: %
		L/S refer to sampling 5000/2466 points for evaluation (2466 being the output size of P2M), $\Pi_g$/$\widehat{\Pi}$ denote using the true versus predicted camera for evaluation (the former case removing camera prediction error), and $\widehat{\Pi}_\nabla$ 
		test-time camera correction from the predicted position using gradient descent.
		Metrics: $D_C$ is the Chamfer distance ($\times 1000$), $F_{\tau}$ is the F-score ($\times 100$) at threshold $\tau = 10^{-4}$.
		PC-SIREN is our matched-architecture baseline;
		Pixel2Mesh (P2M) \cite{wang2018pixel2mesh,wang2020pixel2mesh} 
		and 
		3D-R2N2 (3DR) \cite{choy20163d} 
		are baselines using different shape modalities 
		(numbers from \cite{wang2020pixel2mesh}).
		Note that scenarios using $\Pi_g$ (effectively evaluating shapes in canonical object coordinates) are not directly comparable to P2M or 3DR.
		Overall, DDF-derived PCs 
		(1) perform similarly to directly learning to output a PC and 
		(2) underperform P2M overall, but outperform it in terms of shape quality when camera prediction error is excluded.
	}
	\label{tab:si3dr}
\end{table*}

\subsection{Ablation with $N_H=1$}

\label{appendix:si3dr:nh1}

Recall that $N_H$ is the number of times to ``cycle'' the points (projecting them towards the surface via the DDF) when sampling an explicit point cloud shape from a DDF (see \S\ref{sec:app:si3dr}).
We show results with $N_H=1$ in Table \ref{tab:si3dr1h}.
In most cases, it is slightly worse than using $N_H = 3$, by 1-3 $F$-score units;
occasionally, however, it is  marginally better: on Planes-$\widehat{\Pi}$, it has slightly lower $D_C$, though this does not translate to better $F$-score.

\begin{table} %
	\centering
	\begin{tabular}{cr|cccc}
		&  & \multicolumn{4}{c}{DDF}  \\
		&  & $\Pi_g$-L & $\Pi_g$-S  & $\widehat{\Pi}$-L  & $\widehat{\Pi}$-S  \\\hline 
		& $D_C$  $ \downarrow$ & 0.477 & 0.532 & 0.861 & 0.928  \\ 
		Chairs & $F_{\tau}$ $ \uparrow$ & 54.37 & 47.26 & 46.81 & 40.35  \\ 
		& $F_{2\tau}$ $ \uparrow$ & 71.62 & 66.33 & 63.09 & 58.09  \\\hline 
		& $D_C$  $ \downarrow$ & 0.201 & 0.231 & 0.748 & 0.799  \\ 
		Planes & $F_{\tau}$  $ \uparrow$ & 80.69 & 76.71 & 63.86 & 60.48   \\ 
		&      $F_{2\tau}$  $ \uparrow$ & 90.23 & 88.49 & 75.30 & 73.55  \\ \hline
		& $D_C$  $ \downarrow$ & 0.235 & 0.309 & 0.545 & 0.628  \\ 
		Cars & $F_{\tau}$  $ \uparrow$ & 68.21 & 57.36 & 59.84 & 49.98  \\ 
		& $F_{2\tau}$  $ \uparrow$ & 83.99 & 76.41 & 76.87 & 69.32  \\ 
	\end{tabular}
	\caption{
		Single-image 3D reconstruction results with 
		``single-hop'' point sampling
		($N_H = 1$).
	}
	\label{tab:si3dr1h}
\end{table}

\subsection{Additional Visualizations}
\label{app:sec:si3draddvis}

Some additional visualizations are shown in Fig.\ \ref{app:fig:si3drvis}. 
For highly novel inputs, we also observe that, sometimes, the network does not adapt well to the shape (e.g., see the chairs example in the second row and second column).
While much error is due to the incorrectly predicted camera, the DDF outputs can also be a bit blurrier, especially when it is uncertain about the shape.
This can occur on thin structures, which are hard to localize (e.g., the chair legs in either row one and column one, or row three column two), or atypical inputs (e.g., row three, column two of the cars examples), where the network does not have enough examples to obtain high quality geometry.
One can also see some non-uniform densities from our point cloud sampling algorithm (e.g., concentrations of points on the chair legs in column two, or on the wheels of several examples of cars).

%% file: supp-ugen.tex
\section{Generative Modelling}
\label{appendix:genmodel}

\begin{figure}
	\centering
	\begin{tikzpicture}
		\node[xshift=0cm,yshift=0cm,draw,rounded corners] (PN) at (-0.0,0.0) {$(P,N)$};
		\node[xshift=0cm,yshift=0cm,draw,circle,fill={rgb,255:red,255; green,128; blue,128},inner sep=1.3pt] (zs1) [right=0.4in of PN] {$z_s$};
		\node[xshift=0cm,yshift=0cm,draw,rounded corners] (d1) [right=0.4in of zs1] {$d(p,v|z_s)$};
		\draw [-latex,thick] (PN.east) -- node[right,above] {$E$} (zs1.west);
		\draw [-latex,thick,densely dashed] (zs1.east) -- node[right,above] { } (d1.west);
	\end{tikzpicture}\\[3mm]
	\begin{tikzpicture}
		\node[xshift=0cm,yshift=0cm,draw,circle,fill={rgb,255:red,160; green,203; blue,250},inner sep=0.9pt] (pi) at (0,0) {$\Pi$};
		\node[xshift=0cm,yshift=0cm,draw,circle,fill={rgb,255:red,255; green,128; blue,128},inner sep=1.3pt] (zs) [below left=0.25in and 0.25in of pi] {$z_s$};
		\path let \p1=(pi), \p2=(zs) in node[] (h1) at (\x1,\y2) { };
		\node[xshift=0cm,yshift=0cm,draw,rounded corners] (In) [right=0.4in of h1] {$I_n$};
		\node[xshift=0cm,yshift=0cm,draw,circle,fill={rgb,255:red,85; green,181; blue,98},inner sep=1.1pt] (zt) [above right=0.2in and 0.25in of In] {$z_T$};
		\path let \p1=(zt), \p2=(zs) in node[draw,circle,inner sep=0pt] (h2) at (\x1,\y2) { $+$ };
		\node[xshift=0cm,yshift=0cm,draw,rounded corners] (It) [right=0.4in of h2] {$I_T$};
		\draw [thick,densely dashed] (zs.east) --  node[right,above] {} (h1.center) ;
		\draw [-latex,thick,densely dashed] (h1.center) --  node[right,above] { $\nabla_p d$ } (In.west) ;
		\draw [-latex,thick] (h2.east) --  node[right,above] { $f_T$ } (It.west) ;
		\draw [thick,densely dashed] (pi.south) --  node[right,above] {} (h1.center) ;
		\draw [-latex,thick] (zt.south) -- node[] {} (h2.north) ;
		\draw [-latex,thick] (In.east) -- node[] {} (h2.west) ;
	\end{tikzpicture}
	\caption{
		Two-stage unpaired generative modelling architecture. 
		\textit{Upper inset}: 
		VAE formulation mapping a point cloud $P$ and associated normals $N$ to latent shape vector $z_s$ via PointNet encoder $E$, and decoding into depth values via conditioning a PDDF. 
		\textit{Lower inset}: 
		latent shape $z_s$ and camera $\Pi$ are randomly sampled and used to render a surface normals image $I_n$, via the derivatives of the learned conditional PDDF 
		(see Property \hyperref[property2]{II}).
		The normals map $I_n$ is then concatenated ($\oplus$) with a sampled latent texture $z_T$, and used to compute the final RGB image $I_T = f_T(I_n,z_T)$.
		Coloured circles indicate random variables sampled from a particular distribution; dashed lines indicate computation with multiple forward passes (per point or pixel). 
	}
	\label{fig:ugenarch}
\end{figure}
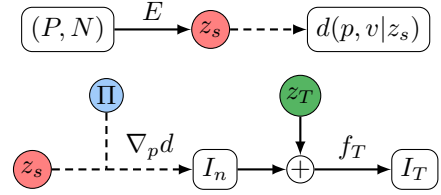

\begin{figure*}%
	\centering
	\adjincludegraphics[width=0.99\textwidth,trim={{.0\width} 0 {.0\width}  0},clip]{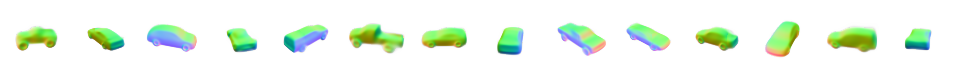}    
	\adjincludegraphics[width=0.99\textwidth,trim={{.0\width} 0 {.0\width}  0},clip]{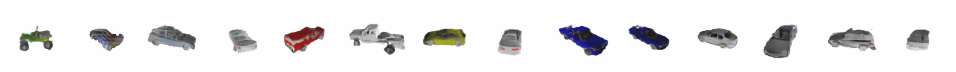}
	\adjincludegraphics[width=0.99\textwidth,trim={{.0\width} 0 {.0\width} 0},clip]{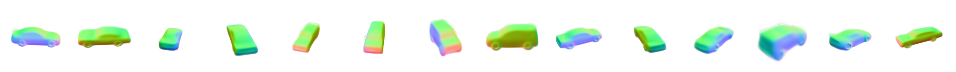}
	\adjincludegraphics[width=0.99\textwidth,trim={{.0\width} 0 {.0\width}  0},clip]{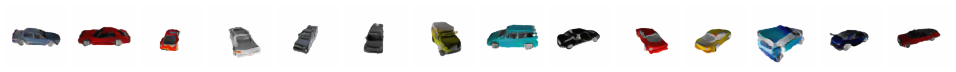}
	\caption{
		Additional example samples from the ShapeVAE and translational image GAN.
	}
	\label{fig:gangenssuppa}
\end{figure*}

\begin{figure}%
	\centering
	\adjincludegraphics[width=0.235\textwidth,trim={{.0\width} 0 {.0\width}  0},clip]{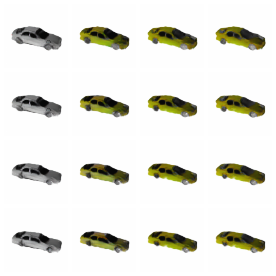}
	\hfill
	\adjincludegraphics[width=0.235\textwidth,trim={ {.0\width} 0 {.0\width}  0},clip]{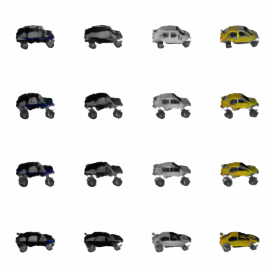}\\
	\adjincludegraphics[width=0.235\textwidth,trim={{.0\width} 0 {.0\width}  0},clip]{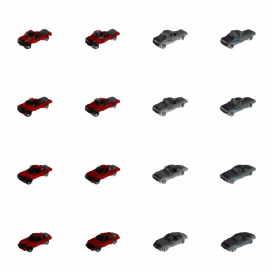}
	\hfill
	\adjincludegraphics[width=0.235\textwidth,trim={ {.0\width} 0 {.0\width}  0},clip]{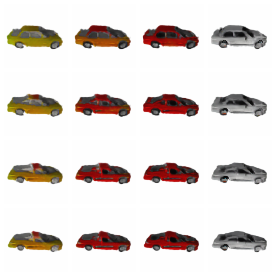}\\
	\caption{
		Example interpolations from the ShapeVAE and translational image GAN.
	}
	\label{fig:gangenssuppb}
\end{figure}

See Fig.\ \ref{fig:ugenarch} for a diagram of our two-stage 3D-to-2D modality translation architecture.
See also Fig.\ \ref{fig:gangenssuppa} and \ref{fig:gangenssuppb} for additional sample visualizations (as in Fig.\ \ref{fig:gangens}).

\subsection{Shape VAE}
The first stage learns a VAE \cite{rezende2014stochastic,kingma2014auto}
on 3D shape, in order to 
(i) obtain a latent shape variable $z_s$ that is approximately Gaussian distributed, and 
(ii) learn a CPDDF using $z_s$ that can encode a shape in a form that is easily and efficiently rendered, yet still encodes higher order shape information.
We learn a PointNet encoder \cite{qi2017pointnet} $E$ 
that maps a point cloud with normals $(P,N)$ 
to a latent shape vector $z_s = E(P,N)$.
The conditional PDDF (CPDDF) then acts as the decoder, 
computing depth values as $d(p,v|z_s)$ for a given input.
To implement conditional field computations, 
we use the modulated SIREN approach \cite{mehta2021modulated}.
This can be trained by the following $\beta$-VAE loss
\cite{higgins2016beta}:
\begin{equation}
	\mathfrak{L}_{\mathrm{VAE}} = 
	\mathfrak{L}_S + \beta \mathcal{L}_\mathrm{KL},
\end{equation}
where $\mathcal{L}_\mathrm{KL}$ is the standard KL divergence loss between a Gaussian prior and the approximate VAE posterior, 
and $\mathfrak{L}_S$ acts analogously to the reconstruction likelihood.
We use ShapeNet cars \cite{shapenet2015} to fit the Shape VAE.
Our goal is to show that we can rapidly train a model that can sample renders of surface normal geometry (from a unified 3D model), which is either difficult (or computationally costly) for most implicit approaches, 
or it uses a model directly on 2D surface normal images (missing a unified 3D shape model).
These images could be useful for other downstream models, which may expect 2D input, but still propagate gradients to an underlying 3D object model.

\textbf{Data.}
Data is extracted from 1200 randomly chosen shapes from ShapeNet-v1 cars \cite{shapenet2015}, 
sampling 60k (A and U) and 30K (S, B, T, and O) oriented points.
We downsample shapes to 10K triangles before extraction.
For minibatches, we sample 1.2K (A and U) and 300 (S, B, T, and O) oriented points per shape, with each minibatch containing 32 shapes.
We also sample point clouds $P$ (with normals $N$) of size 1024 to send to the PointNet at each minibatch.

\textbf{Architectures.}
The PointNet encoder follows the standard classification architecture 
\cite{qi2017pointnet},
with four \verb|Conv1D-BatchNorm-ReLU| blocks 
(sizes: 64, 64, 128, 1024), 
followed 
by max pooling and a multilayer perceptron with \verb|Linear-BatchNorm-ReLU| blocks (two hidden layers of size 512; dropout probability 0.1).
No point transformers are used.
The final output is of size $2\mathrm{dim}(z_s)$.
We then compute the approximate variational posterior $q(z_s|P,N) = \mathcal{N}(z_s|\mu(P,N),\Sigma(P,N))$ 
with two networks for mean $\mu$ and diagonal log-variance matrix $\Sigma$, each structured as 
\verb|Linear-ReLU-Linear| 
(in which all layers are of dimensionality $\mathrm{dim}(z_s)$; note that each posterior parameter network takes half of the vector output from the PointNet encoder as input).

For the decoder,
we use eight layers \verb|(512,| \verb|512,| \verb|512, 512, 256, 256, 256, 256)| (with the modulated SIREN \cite{mehta2021modulated}).
We set $\mathrm{dim}(z_s) = 400$ and $\mathrm{dim}(z_T) = 64$.

\textbf{Training.}
We run for 100K iterations, using 
$\beta = 0.05$,
$\gamma_d = 5$,
$\gamma_\xi = 10$,
$\gamma_n = 10$,
$\gamma_V = 1$,
$\gamma_\mathrm{E,d} = 0.1$, 
$\gamma_\mathrm{E,\xi} = 0.1$, 
$\gamma_{V,\xi} = 0.1$,
and
$\gamma_T = 0.1$.
Adam is used for optimization 
(learning rate $10^{-4}$; $\beta_1 = 0.9$, $\beta_2 = 0.999$).

\subsection{Image GAN}

After training a Shape VAE, 
the CPDDF decoder can be used to render a surface normals image $I_n$
(see \S\ref{sec:app:rendering} and Fig.\ \ref{fig:singobjfits}).
To perform generation,
we first sample latent shape $z_s\sim\mathcal{N}(0,I)$
and
camera $\Pi$, 
which includes extrinsics (position and orientation) and focal length,
following with normals map rendering to get $I_n$.
We then use a convolutional network $f_T$ to obtain the RGB image $I_T$, 
based on the residual image-to-image translation architecture 
from CycleGAN \cite{CycleGAN2017}.
This is done by sampling $z_T\sim\mathcal{N}(0,I)$
and
computing $I_T = f_T(I_n,z_T)$,
where
one concatenates $z_T$ to each pixel of $I_n$ before processing.
Notice that $z_s$, $z_T$, and $\Pi$ are independent, while the final texture (appearance) depends directly on $z_T$, and indirectly on $z_s$ and $\Pi$, through $I_n$.

For training,
a non-saturating GAN loss \cite{goodfellow2014generative} 
with a zero-centered gradient penalty for the discriminator \cite{roth2017stabilizing,mescheder2018training} 
is used 
(as in \cite{niemeyer2020giraffe,chan2021pi}).
To ensure that $f_T$ preserves information regarding 3D shape and latent texture, we use two consistency losses:
\begin{equation}
	\mathcal{L}_\mathrm{C,S} = \mathrm{MSE}(I_n, \widehat{I}_n)
	\;\;\;\&\;\;\;
	\mathcal{L}_\mathrm{C,T} = \mathrm{MSE}(z_T, \widehat{z}_T),
\end{equation}
where $\mathrm{MSE}$ is the mean squared error, 
$\widehat{I}_n = g_T(I_T)$ (with $g_T$ having identical architecture to $f_T$),
and 
$\widehat{z}_T = h_T(I_T)$ 
(with $h_T$ implemented as a ResNet-20 \cite{he2016deep,Idelbayev18a}).
The first loss, $\mathcal{L}_\mathrm{C,S}$,
encourages the fake RGB image $I_T$ to retain the 3D shape information from $I_n$ through the translation process
(i.e., implicitly forcing the image resemble the input normals)
while the second loss, $\mathcal{L}_\mathrm{C,T}$,
does the same for the latent texture 
(implicitly strengthening textural consistency across viewpoints).
The final loss for the GAN image generator is
\begin{equation}
	\mathfrak{L}_{\mathrm{GAN}} = 
	\mathcal{L}_\mathrm{A} + 
	\gamma_{C,S} \mathcal{L}_\mathrm{C,S} + 
	\gamma_{C,T} \mathcal{L}_\mathrm{C,T}, 
\end{equation}
where $\mathcal{L}_\mathrm{A}$ is the adversarial loss for the generator and
the last two terms enforce consistency
(see also \cite{miyauchi2018shape,kaya2020self,aumentado2020cycle}).
We fit to the dataset of ShapeNet renders from Choy et al.\ \cite{choy20163d}. 
Note that information on the correspondence with the 3D models is not used (i.e., the images and shapes are treated independently).

\textbf{Generation process.}
Recall that our generation process can be written
$I_T = G(z_s,z_T,\Pi) = f_T(I_n(z_s,\Pi)\oplus z_T)$, 
where $I_n(z_s,\Pi)$ denotes the CPDDF normals render 
(see \S\ref{sec:app:rendering}) and $\oplus$ refers to concatenating $z_T$ to every pixel of $I_n$.
For latent sampling, 
$z_s,z_T\sim\mathcal{N}(0,I)$, while the camera $\Pi$ is sampled from the upper hemisphere above the object, oriented toward the origin at a fixed distance and with a fixed focal length.
The image size was set to $64\times 64$.

\textbf{Networks.}
The translation network $f_T$ exactly follows the architecture from CycleGAN \cite{CycleGAN2017} consisting of residual blocks 
(two downsampling layers, then six resolution-retaining layers, followed by two upsampling layers), using the code from \cite{cyclegancode}.
The normals consistency network, $g_T$, has identical architecture to $f_T$, 
while the texture consistency network, $h_T$, is a ResNet-20 \cite{he2016deep,Idelbayev18a}.
We utilize the convolutional discriminator implementation from Mimicry \cite{lee2020mimicry}, based on DCGAN \cite{radford2015unsupervised}.

\textbf{Training.}
Our image GAN is trained in the standard alternating manner, 
using two critic training steps for every generator step.
The non-saturating loss \cite{goodfellow2014generative} was used, along with a zero-centered gradient penalty \cite{roth2017stabilizing,mescheder2018training} (with a weight of 10 during critic training).
We used the following loss weights:
$\gamma_{C,S} = 1$ %
and
$\gamma_{C,T} = 1$ %
For optimization, we use Adam 
(learning rate $10^{-4}$; $\beta_1 = 0.0$, $\beta_2 = 0.9$)
for 100K iterations 
(with the same reduce-on-plateau scheduler as in Supp.~\ref{appendix:singlefits}).

\textbf{2D GAN Comparison.}
As mentioned in the paper, we trained a convolutional GAN with the same loss and critic architecture using the Mimicry library \cite{lee2020mimicry}. 
We evaluated both our model and the 2D GAN with Frechet Inception Distance (FID), using torch-fidelity \cite{obukhov2020torchfidelity} with 50K samples.

%% file: supp-pathtrace.tex
\section{DDF-based Path Tracing Details}
\label{pddf:appendix:light}

\input{pt}

\subsection{Technical Details}
\label{appendix:light:techdetails}

\begin{figure*}
	\centering
	\includegraphics[width=0.495\textwidth]{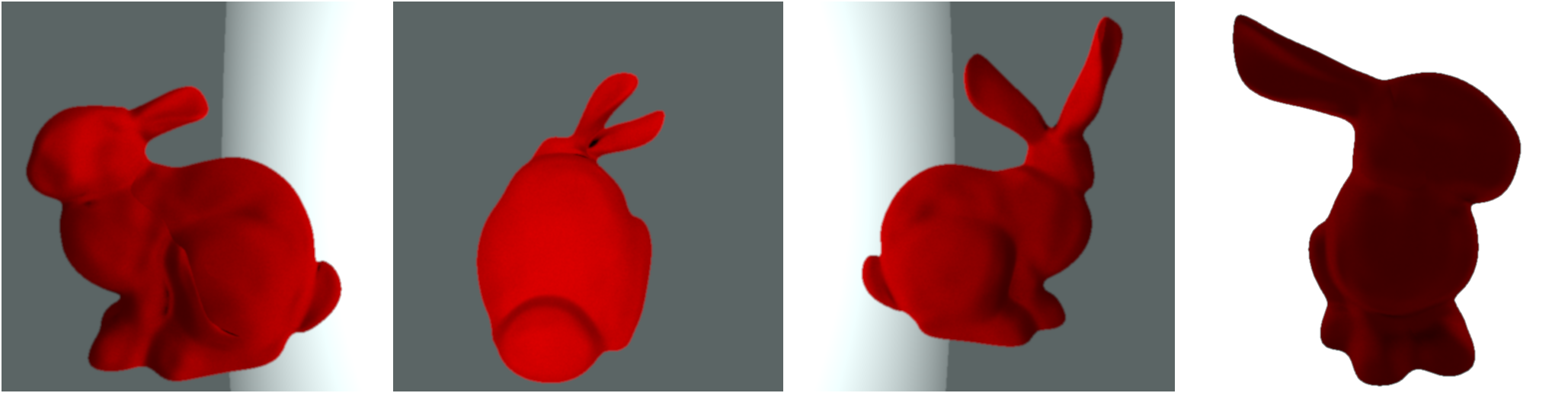} \hfill
	\includegraphics[width=0.495\textwidth]{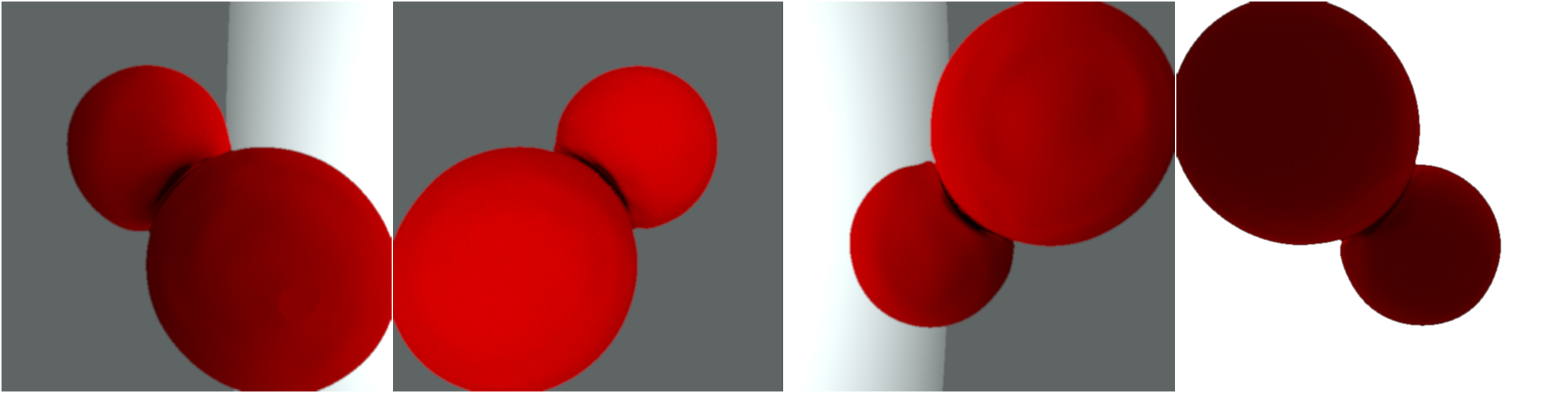} \\
	\includegraphics[width=0.495\textwidth]{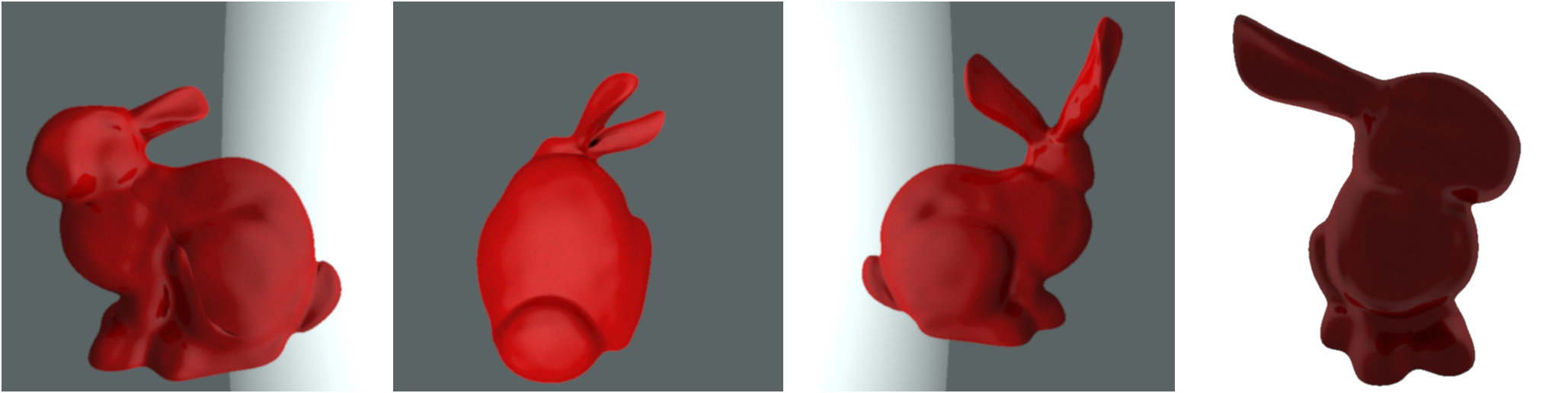} \hfill 
	\includegraphics[width=0.495\textwidth]{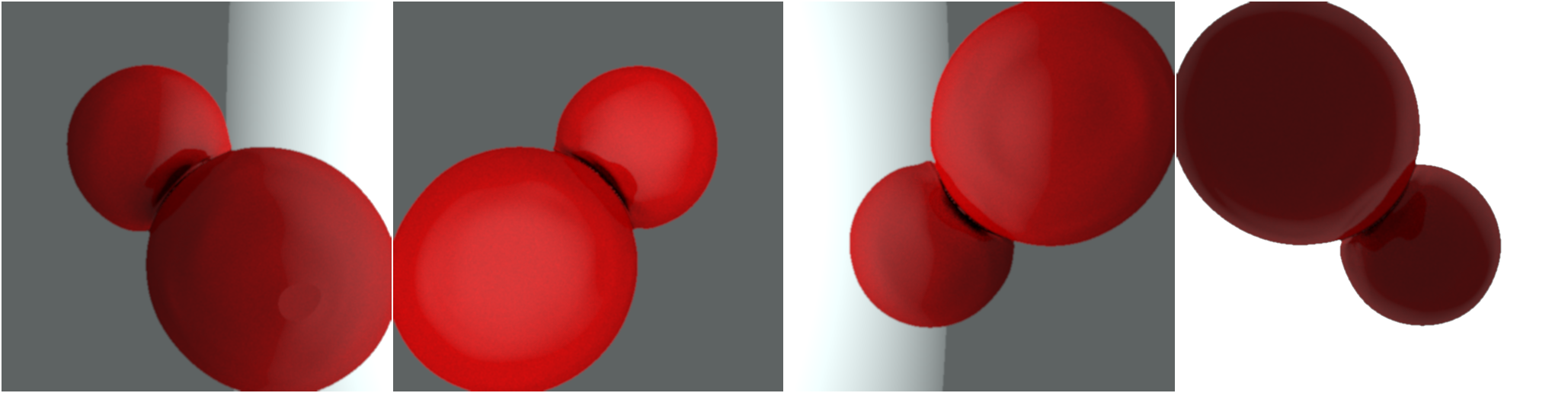} \\
	\includegraphics[width=0.495\textwidth]{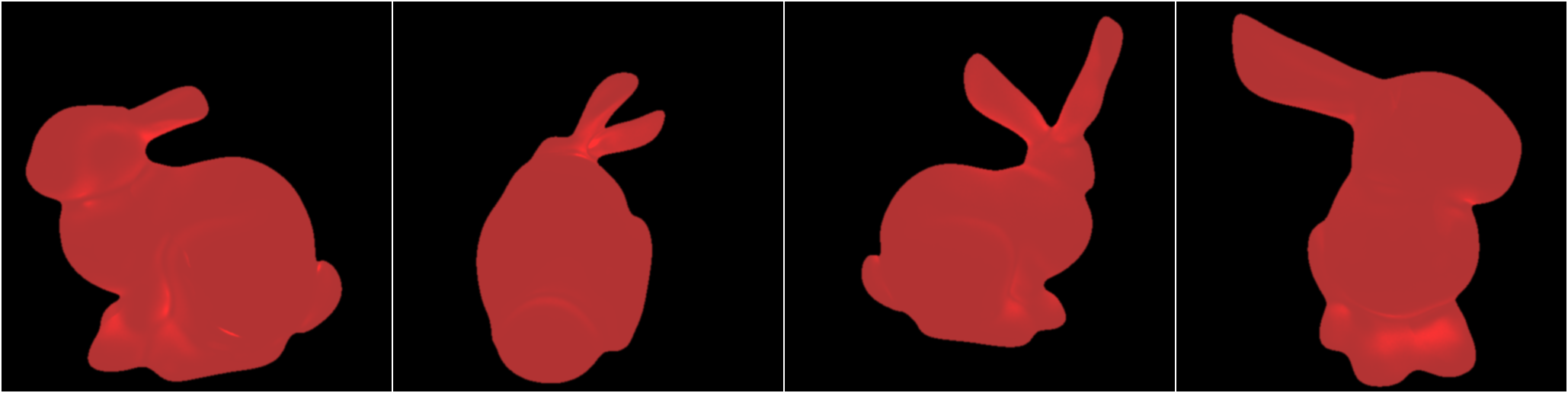} \hfill
	\includegraphics[width=0.495\textwidth]{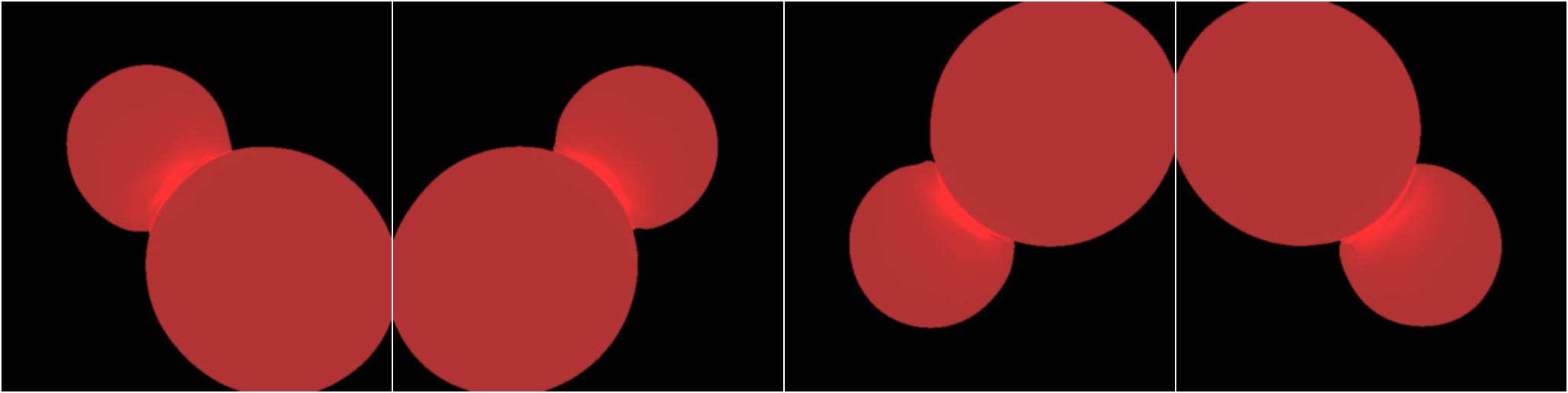}
	\caption{
		Examples of path-traced DDFs, as in Fig.~\ref{fig:pathtraced}.
		Top row: Lambertian material.
		Second row: glossy material with $\alpha = 3$ and $\eta_L = 0.25$.
		Notice the ``sheen'' added by the mirror term, which injects the glossiness.
		Third row:
		no external lighting is used. Instead, a simple ``glow'' (constant $M_L > 0$) is emitted by the geometry; hence the areas of the shape that would normally be shadowed due to self-occlusion are actually highlighted.
	} \label{fig:pt:glow}
\end{figure*}

\subsubsection{Hyper-Parameters and Implementation Details}
\label{pddf:appendix:light:imp}
The fitting process was the same as in \S\ref{sec:results:singlefieldfitting}, 
except we use double the number of extracted samples 
(as well as 500K W-type samples; see \S\ref{pddf:app:pathtracing}).
In all cases, we used number of bounces $n=3$, 
\acrshort{mc} samples\footnote{Note: the mirror material only requires one sample (though, for instance, non-mirror materials encountered \textit{after} a mirror could lead to requiring more).} $m=256$,
and post-processing Gaussian blur $\sigma=1$ (images are $512^2$),
except for the glossy material 
with $\alpha = 3$, 
which used $m=512$, 
and $\alpha = 1$, 
which used $m=1024$ and $\sigma=3$.
Fig.~\ref{fig:pathtraced}, row one, used
$\alpha=1$ and $\eta_L = 0.75$.

\subsubsection{Projected Gaussian Importance Sampler}
\label{pddf:appendix:light:material}

Recall the simple glossy material defined in Eq.~\ref{pddf:eq:glossy},
specifically the importance sampler $\Psi$:
\begin{align}
	\Psi_\mathrm{G}(q,\omega_o\,|\,n) 
	= &
	\eta_L \mathbb{U}[ \Omega[n(q)] ]
	\; + \nonumber \\ &
	(1 - \eta_L)\, \mathcal{N}_{\mathrm{Proj}}(\omega_r(-\omega_o,n), \Sigma_p(\alpha)).
\end{align}
Two operations need to be implemented: density evaluation and sampling. %

We use a basic simplification to compute the inverse density, by assuming we know the origin of a sample:
\begin{align}
	\Psi_G(v\,|\, q,\omega_o,n)^{-1} = &
	\frac{ \indic{v\sim \mathbb{U}[ \Omega[n(q)] ]} }{ \eta_L }
	P^{-1}_\mathbb{U}(v)
	\;+ \nonumber \\ &
	\frac{ \indic{v\sim \mathcal{N}_\mathrm{Proj}} }{ 1 - \eta_L }
P_{\mathcal{N}_\mathrm{Proj}}^{-1}(v \,|\, \omega_r, \Sigma_p(\alpha)),
\end{align}
where $ P_\mathbb{U} $ and $ P_{\mathcal{N}_\mathrm{Proj}} $ are density functions, 
$\omega_r = \omega_r(-\omega_o,n)$, and
$\indic{\cdot}$ is the indicator function. 
We discuss the actual density used by $ P_{\mathcal{N}_\mathrm{Proj}} $ later.

For obtaining a single sample from this combined material (i.e., $ v \sim \Psi_\mathrm{G}(q,\omega_o\,|\,n)  $), 
we first choose which mixture mode (specular or Lambertian) to pick:
$ b\sim \mathcal{B}(\eta_L) $, where $\mathcal{B}(\eta_L)$ is the Bernoulli distribution.
If $b = 0$, we sample uniformly, from $ \mathbb{U}[ \Omega[n(q)] ] $.
Otherwise, if $b = 1$, we take a Gaussian sample $\widetilde{\zeta} \sim \mathcal{N}(\omega_r, \Sigma_p(\alpha)) $
and then normalize it via $ v = \widetilde{\zeta} / || \widetilde{\zeta}||_2$.

To compute the density $ P_{\mathcal{N}_\mathrm{Proj}} $, we use an approximation.
We assume that we know $\zeta = \widetilde{\zeta} - \omega_r$, and then compute
$ P_{\mathcal{N}_\mathrm{Proj}}(v \,|\, \omega_r, \Sigma_p(\alpha)) 
\approx C( P_\mathcal{N}(\zeta\,|\, 0, \Sigma_p(\alpha)) ) $.
We use a simple clamping function, $C$, to keep the density numerically stable.

The key parameter that controls the glossiness of the material is the Phong specular exponent $\alpha$, 
which is related to the ``spread'' of the bounce directions about the reflection direction.
This affects the density through the noise covariance
$ \Sigma_p(\alpha) = f_\alpha \exp( - \kappa_0 \alpha ) I_3 $,
where $f_\alpha = 3.5\times 10^{-\alpha}$ and $\kappa_0 = 1$.
We remark that this ad hoc formulation means that one should not interpret $\alpha$ the same way as in other Phong models.

In terms of differentiability, the reparameterization trick can be used to optimize $\alpha$, 
since we are using a diagonal Gaussian (as in most \acrshort{vae}s).
However, notice that back-propagation to $\eta_L$ through the sampling process is non-trivial;
since it controls which mixture component ($b$) to sample,
it describes a fundamentally discrete random indicator function (i.e., Bernoulli variable).
Techniques for handling discrete random variables in stochastic computation graphics
(e.g., continuous relaxations \cite{jang2016categorical,maddison2016concrete} 
or score functions \cite{grathwohl2017backpropagation}) 
would be necessary for obtaining derivatives.

%% file: pt.tex
\subsection{Path Tracing with Intrinsic Appearance DDFs}	
\label{pddf:app:pathtracing}

\begin{figure} %
	\centering
	\includegraphics[width=0.95\linewidth]{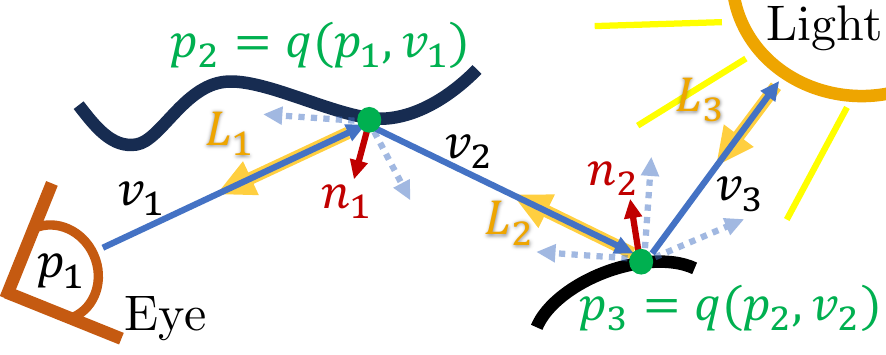}
	\caption{
		Illustration of our simple DDF-based path tracing algorithm.
		To simulate the pixel value obtained in a given direction, 
		we use recursively query the DDF (i.e., the mapping $q$).
		Upon encountering a light source, the light value is propagated back to the eye
		through the sequence of rays.
		The entire process is differentiable, including the normals computation.
	} \label{fig:lightbouncing}
\end{figure}

We depict the path tracing process in Fig.~\ref{fig:lightbouncing}.
We describe the path tracing process, material models, and lighting specifiers of the Intrinsic Appearance DDF (IADDF) in \S\ref{pddf:light:iaddf}, \S\ref{pddf:light:pathtracing}, and \S\ref{pddf:light:materials}
We also describe pseudocode of the algorithms in 
Figs.~\ref{fig:algolight:pt} and \ref{fig:algolight:trace}.
Additional implementation details are in \S\ref{appendix:pt:postproc} and \S\ref{appendix:light:techdetails}.
Finally, we present additional discussion in
\S\ref{pddf:light:adddisc}.

\subsection{The Intrinsic Appearance DDF (IADDF)}
\label{pddf:light:iaddf}

Recall the classical notion of intrinsic images,
which separated a 2D image into 
depth, ``orientation'' (or surface normals), reflectance, and illumination \cite{barrow1978recovering}\footnote{
	Given that we consider the material reflectance of an entity 
	to be its \textit{in}trinsic appearance,
	we could also reasonably call the illumination-derived image component,
	coming from the external environment, its \textit{ex}trinsic appearance.
}.
In this section, we show how to define a scene description 
covering these pieces,
by combining a \acrshort{ddf} with simple \acrshort{inr}s for appearance.
In particular, 
taking advantage of the light transport interpretation of \acrshort{ddf}s,
we factorize rendered image colour into fields for intrinsic appearance and lighting.
This combined formulation can be thought of as an 
``intrinsic image field'',
because it acts as a forward model for the set of all {intrinsic images},
whether of geometry or appearance 
(i.e., it can continuously render such intrinsic images, from any viewpoint).
We therefore denote the combined model an \textit{Intrinsic Appearance DDF} 
(\acrshort{iaddf}),
as it incorporates a model of lighting-aware intrinsic appearance along with the geometric information held by the \acrshort{ddf}.
In the context of vision, maintaining an \acrshort{iaddf} model of the scene 
automatically provides colour constancy as a side-effect, 
simply by replacing image appearance with intrinsic appearance.
The \acrshort{iaddf} consists of the following fields:
\begin{itemize}
	\item 
	The geometry field, $\mathcal{G}$, modelled by a \acrshort{ddf}, consisting of
	\begin{itemize}
		\item The distance field 
		$d : \real^3 \times \mathbb{S}^2 \rightarrow \real_+$. 
		\item The visibility field 
		$\xi : \real^3 \times \mathbb{S}^2 \rightarrow [0,1]$.		
	\end{itemize}
	\item 
	The material appearance model, $\mathcal{M}$. 
	We consider a simple approach consisting of a continuous
	Bidirectional Reflectance Distribution Function (BRDF)
	and a directional sampling function, which provides guidance on bounce directions (e.g., for the sake of computational efficiency when rendering or learning). $\mathcal{M}$ includes\footnote{We use $\mathcal{C}$ to denote the radiance emitted from a position (commonly relegated to, e.g., a simple RGB colour value).}
	\begin{itemize}
		\item The BRDF 
		$f_B : \real^3 \times \mathbb{S}^2 \times \mathbb{S}^2 \rightarrow \real^{|\mathcal{C}|}_+$. 
		\item The importance sampler\footnote{
			We remark that $\Psi$ depends on the local surface normal $n$ through the spatial position $q$ at which it is evaluated. (We make this indirect dependence explicit in Figs.\ \ref{fig:algolight:pt} and \ref{fig:algolight:trace}).
		} 
		$\Psi : \real^3 \times \mathbb{S}^2 \rightarrow \mathbb{P}[\mathbb{S}^2]$,
		where $\mathbb{P}[\mathbb{S}^2]$ is the set of probability distributions on $\mathbb{S}^2$. 	
	\end{itemize}	
	\item 
	The lighting model, $\mathscr{L}$. 
	To simplify our setting, we consider two fields: 
	one for entities locally emitting light and
	another for measuring the contribution of the external environment.
	Thus, in our case, $\mathscr{L}$ consists of
	\begin{itemize}
		\item The emission field, 
		which handles objects directly emitting light (e.g., glowing), 
		$M_L : \real^3 \times \mathbb{S}^2 \rightarrow \mathcal{C}$. 
		\item The environment light, which represents light ``incoming'' to the scene from some faraway source,
		$E_L : \mathbb{S}^2 \rightarrow \mathcal{C}$. 	
	\end{itemize}		
\end{itemize}
In our simplified scenario, 
the \acrshort{iaddf} is simply the tuple 
$(\mathcal{G},\mathcal{M},\mathscr{L}) = (d,\xi,f_B,\Psi,M_L,E_L)$.	

\subsection{IADDF Path Tracing}
\label{pddf:light:pathtracing}

Given an \acrshort{iaddf}, 
we have disentangled models 
for the scene geometry, lighting, and intrinsic appearance.
However, unlike other neural or differentiable rendering methods,
the presence of an illumination model now necessitates 
some form of simulated light transport.
In particular, we need to solve the classical rendering equation 
\cite{kajiya1986rendering,immel1986radiosity}, 
which can be formalized in the recursive integral
\begin{equation}
	\mathcal{L}(q,\omega_o)
	= \mathcal{L}_M(q,\omega_o)
	+ \int_{ \Omega[n(q)] } 
	f_B(q,\omega_i,\omega_o)\, \mathcal{L}_i(q,\omega_i) 
	\, d\omega_i^\perp,
\end{equation}
where 
$q$ is a surface point,
$n(q)$ is the surface normal at $q$,
$\Omega[n(q)]$ is the hemi-sphere about $n$,
$\mathcal{L}(q,\omega_o)$ is the outgoing radiance from $q$ in direction $\omega_o$,
$\mathcal{L}_M(q,\omega_o)$ is the emitted radiance (from $q$ in direction $\omega_o$),
$f_B(q,\omega_i,\omega_o)$ is the BRDF 
(which computes how irradiance from $\omega_i$ is 
converted to radiance along $\omega_o$),
$\mathcal{L}_i(q,\omega_i)$ is the incoming irradiance to $q$ from direction $\omega_i$,
and
$d\omega_i^\perp$ is the projected solid angle measure
(where %
$d\omega_i^\perp = | \omega_i\cdot n(q) | \,d\omega_i$ 
under Lambert's cosine law,
foreshortening the area element from oblique viewpoints)
\cite{fascione2019path,jain1995machine}.
In our simple setting, the recursion and connection to \acrshort{ddf}s becomes more clear by rewriting the incoming radiance to
\begin{equation}
	\mathcal{L}_i(q,\omega_i) = 
	\begin{cases}
		\mathcal{L}(q - d(q,-\omega_i)\omega_i, - \omega_i) 
		& \;\mathrm{ if }\; \xi(q,-\omega_i) > 0.5 \\
		\mathcal{L}_E(q,-\omega_i) 
		& \;\mathrm{otherwise}
	\end{cases},
\end{equation}
where $(d,\xi)$ are the \acrshort{ddf}
and $\mathcal{L}_E(q,\omega)$ is the light from the environment
falling on $q$ from direction $\omega$.
(e.g., \cite{hadadan2021neural}).
In other words, 
the incoming radiance at a point is either 
(i) the outgoing radiance reflected off of another surface 
(i.e., inter-reflections) or
(ii) the external light from the environment
(when another surface is not visible).\footnote{
	We remark that there is an abuse of notation and terminology.
	In particular,
	$\mathcal{L}$ and $\mathcal{L}_M$
	should be radiant \textit{densities}.
	Under the pinhole camera assumption,
	for a given direction and position on a pixel sensor,
	the irradiance is proportional to radiance from a single direction.
	One then integrates over the pixel area  
	to get total irradiance (i.e., flux or power), 
	which is integrated over time to measure radiant energy
	at that pixel (for a given exposure).
}

Clearly, for a given pixel with an eye ray through it,
corresponding to an oriented point $(p,v)$,
we want to obtain the light coming back to the eye:
$ \mathcal{L}(q(p,v), -v) $.
We use a straightforward form of solution, 
detailed in Fig.~\ref{fig:algolight:pt}: 
an eye ray tracer or backward path tracer%
\footnote{
	We mean ``backward'' here in the sense of rays starting from the camera and bouncing  to a light source 
	(i.e., the reverse of what physical light is thought to do) 
	\cite{veach1995bidirectional}, 
	but note that this nomenclature is not uniformly followed 
	(e.g., \cite{arvo1986backward}).
}, which is a form of \acrshort{mc} rendering.
Benefits of such \acrshort{mc} path tracing methods include 
excellent generality 
(in terms of the scene descriptions and lighting effects they can handle), 
relative simplicity of implementation,	
and lack of statistical bias
(i.e., zero error in expectation),
while downsides include the 
slow convergence (depending on the algorithm and scene)
and 
image noise 
\cite{veach1998robust}.

We briefly describe the stochastic path tracing algorithm 
(see Figs.\ \ref{fig:algolight:pt} and \ref{fig:algolight:trace} for details). 
For a given pixel and camera (eye), 
we cast a ray into the scene, finding the scene point $q$ and normal $n$ via the \acrshort{ddf}. 
Based on $\Psi$, we sample new outgoing directions from the hemisphere about $q$, 
and use the DDF to find the next surface in that direction.
These ``bounces'' are the segments of a light path, being built backwards from the eye to the light source.
This continues until the ray misses any local geometry (i.e., $\xi = 0$),
returning a value from $E_L$ in that direction.
The final sampled pixel irradiance corresponds to this value 
(attenuated by the material $f_B$ through the bounces),
plus any emissions (from $M_L$) encountered along the way. 
\SetKwInput{Models}{Models}

\SetKwInput{HyperParameters}{Hyper-Parameters}
\begin{figure} \removelatexerror
\begin{algorithm}[H]
	\DontPrintSemicolon %
	\SetNoFillComment
	\caption{$\mathrm{PathTracePixelDDF}(u,\Pi)$}	
	\KwIn{ Pixel $u$ and Camera $\Pi$ }
	\Models{\acrshort{ddf} $(d,\xi)$, Appearance $(f_B,\Psi)$, and Illumination $(M_L,E_L)$}
	\HyperParameters{Maximum bounces $n$, Number of \acrshort{mc} Samples $m$}
	\KwOut{The pixel colour $c$ at $u$}
	\vspace{0.25\baselineskip}
	$ p,v \gets \mathrm{GetOrientedPoint}(u,\Pi) $ \tcp*{Camera ray for $u$}
	\If(\tcp*[f]{Check for ray intersection with scene} ){$ \xi(p,v) < 0.5 $}{ 
			\Return{$E_L(v)$} \tcp*{Return environment light if no intersection} 
		} 
	$ q \gets p + d(p,v) v $ \tcp*{Scene point for the camera ray} 
	$ n \gets \mathrm{Unit}(\nabla_p d(p,v)) $ \tcp*{Scene point surface normals} 
	$ \omega_o \gets -v $ \tcp*{Direction back to the eye} 
	$ c_E \gets M_L(q, \omega_o) $ \tcp*{Colour emitted from $q$ towards $\omega_o$}
	\For(\tcp*[f]{Iterate over MC samples} ){$k \gets 1$ \textbf{\textnormal{to}} $m$} {
			$ v^{(k)}_n \sim \Psi(q,\omega_o \,|\, n) $ 
			\tcp*{Sample light direction from $q$ about $n$}
			$ \omega^{(k)}_i \gets - v^{(k)}_n $ 
			\tcp*{New \textit{incoming} direction, into $q$}
			$ B\gets f_B(q,\omega^{(k)}_i,\omega_o) $ 
			\tcp*{Query the BRDF field}
			$ c_k \gets B \odot \mathrm{Trace}(q, v^{(k)}_n, \ell = 1) $ 
			\tcp*{Obtain colour by ``bouncing'' the ray}
			$ w_k \gets \Psi(v^{(k)}_n \,|\, q,\omega_o)^{-1} \; 
			| n\cdot \omega_i^{(k)} | $
			\tcp*{Weight in MC averaging}
		}
	$ \displaystyle c = c_E + \frac{1}{m}\sum_{i=1}^m w_i c_i $ 
	\tcp*{Final pixel colour}
	\Return{$c$}
	\label{pddf:algopt}
\end{algorithm}
\caption[DDF-Based Light Transport Algorithm]{
	Algorithm used for DDF rendering via backward path tracing.
	We devise a simple approach based on stochastic backward ray tracing,
	where we start rays from the eye and follow them into the scene
	until a light source is encountered.\protect\footnotemark~ 
	For efficiency, 
	we only use multiple Monte Carlo (\acrshort{mc}) samples 
	for the first bounce,
	since the contribution afterwards often falls off quickly.
	Notice that 
	(i) we can compute the colour of each pixel \textit{independently}, 
	using a series of differentiable field 
	(i.e., \acrshort{ddf}, BRDF field, and lighting field) 
	calls per pixel ray
	(in a parallelizable manner similar to \acrshort{nerf}s),
	(ii) substituting the \acrshort{ddf} 
	as the geometric representation does not preclude utilizing the many advances in path tracing methods, and
	(iii) this framework naturally enables handling reflections, shadows, and glowing objects. %
	See Fig.~\ref{fig:algolight:trace} for the definition of
	the recursive $\mathrm{Trace}$ function and 
	Fig.~\ref{fig:lightbouncing} for a visual depiction.
}
\label{fig:algolight:pt}
\end{figure}
\footnotetext{While $p$ may initially be the camera position (i.e., $C_p$ in $\Pi$), recall from \S\ref{sec:app:rendering} that rendering proceeds by first projecting $p$ along the camera ray to the domain of the \acrshort{pddf}, denoted $\mathcal{B}$, and then utilizing that boundary point as $p$ instead.
}
\begin{figure} \removelatexerror
	\begin{algorithm}[H]
		\DontPrintSemicolon %
		\caption{$\mathrm{Trace}(p,v,\ell)$}
		\KwIn{Initial Scene Point $p$, Direction $v$, Bounce Level $\ell$}
		\Models{\acrshort{ddf} $(d,\xi)$, 
				Appearance $(f_B,\Psi)$, and Illumination $(M_L,E_L)$}
		\HyperParameters{Maximum bounces $n$}
		\KwOut{Colour $c$ at $p$ from direction $v$}
		\vspace{0.25\baselineskip}
		\If(\tcp*[f]{Check for ray intersection with scene}){$ \xi(p,v) < 0.5 $}{ 
			\Return{$E_L(v)$} \tcp*{Return environment light if no intersection} 
		} 
		\If(\tcp*[f]{Check current bounce level}){$\ell \geq n$}{
				\Return{$\myvec[1.8mu]{0}$} \tcp*{Stop bouncing after level $n$} 
			} 
			$ q \gets p + d(p,v) v $  \tcp*{Scene point along ray from $(p,v)$} 
			$ n \gets \mathrm{Unit}(\nabla_p d(p,v)) $   \tcp*{Scene point surface normals} 
			$ \omega_o \gets -v $  \tcp*{Direction back to $p$ from $q$} 
			$ c_E \gets M_L(q, \omega_o) $ \tcp*{Colour emitted from $q$ towards $\omega_o$} 
			$ v_n \sim \Psi(q,\omega_o \,|\, n) $ \tcp*{Sample light direction from $q$ about $n$}
			$ \omega_i \gets - v_n $ \tcp*{New \textit{incoming} direction (i.e., into $q$)}
			$ B\gets f_B(q,\omega_i,\omega_o) $ \tcp*{Query the BRDF field}
			$ c \gets B \odot \mathrm{Trace}(q, v_n, \ell+1) $ 
			\tcp*{Recursive colour from $q$ (via $\omega_i$) to $p$} 
			\Return{$c_E + \Psi(v_n \,|\, q,\omega_o)^{-1} \; 
				| n\cdot \omega_i | \; c$}\tcp*{Single-sample MC irradiance at $p$}
		\end{algorithm}
		\caption{
			DDF-based recursive path tracing process, 
			used by the algorithm in Fig.~\ref{fig:algolight:pt}.
			Each recursive call corresponds to a single bounce of the light path 
			(and thus one call to the \acrshort{ddf}).
			Notice that only a single forward and backward pass is 
			needed from the \acrshort{ddf} per bounce.
			See also Fig.~\ref{fig:lightbouncing} for a visual depiction.
		}
		\label{fig:algolight:trace}
	\end{figure}
	
	\subsection{Basic Intrinsic Appearance Models}
	\label{pddf:light:materials}

	In our simplified setting, we merely need to specify the BRDF $f_B$ and bounce sampler $\Psi$ 
	(see \S\ref{pddf:light:iaddf}).
	We consider three basic cases, 
	but note that any differentiable field 
	(ideally satisfying certain priors, particularly those of physical realism) $f_B$ and 
	sampling procedure (with accompanying density function) $\Psi$ could be used, 
	which would be necessary for modelling arbitrarily complex observed materials.
	Note that we treat $\mathcal{C} = \real^3$; 
	instead of including a continuous dependence on wavelength, which would be more realistic, 
	we instead output three discrete colour values corresponding to the RGB channels in the end.
	
	The first case is Lambertian (or diffuse) material:
	\begin{align}
		\label{pddf:eq:lambert}
		f_{B,\mathrm{Lam}}(q,\omega_i,\omega_o) 
		&= \frac{ \rho_a }{ \pi } \\
		\Psi_\mathrm{Lam}(q,\omega_o\,|\,n) 
		&= \mathbb{U}[ \Omega[n(q)] ],
	\end{align}
	where $\mathbb{U}$ is the uniform distribution and $\rho_a \in [0,1]^3$ is the per-channel albedo.
	We write $\Psi(q,\omega_o|n) $ to make the dependence on $n(q)$ more clear, 
	but note\footnote{Additionally, since the \acrshort{ddf} is not view consistent, $n$ is not necessarily either.} that $n$ depends on $q$, through $d$.
	
	The second case is a mirror material 
	(e.g., see \cite{pharr2023physically,jain1995machine}):
	\begin{align}
		\label{pddf:eq:mirror}
		f_{B,\mathrm{Mir}}(q,\omega_i,\omega_o) 
		&= \frac{ \delta(\omega_r(\omega_i,n) - \omega_o) \rho_m }{ |\omega_r(\omega_i,n) \cdot n| } \\
		\Psi_\mathrm{Mir}(q,\omega_o\,|\,n) 
		&= \mathbb{P}_\delta[\omega_r(-\omega_o,n)],
	\end{align}
	where $\delta$ is the Dirac Delta function (distribution) on $\mathbb{S}^2$,
	$\mathbb{P}_\delta$ is the Dirac probability measure,
	$\rho_m$ is the albedo (Fresnel reflectivity),
	and 
	$\omega_r(u,v)$ is the outgoing reflection direction of an incoming direction $u$ about $n$.
	Recall that, when we sample $v\sim\Psi_\mathrm{Mir}(q,\omega_o\,|\,n)$ (see Fig.~\ref{fig:algolight:pt}),
	we are interested in sampling directions that will contribute to outgoing light back to the eye 
	($\omega_o$); thus, for a mirror, we should of course sample the reflection of $\omega_o$. 
	Notice that the BRDF contains a cosine term that cancels out the geometric attenuation 
	from Lambert's cosine law;
	this is needed to ensure energy conservation in the rendering integral
	(i.e., we do not need to correct for the differential contribution of directions based on foreshortening of the area element, 
	because no scattering occurs about the reflection direction). 
	We remark that modelling reflections is an important and difficult topic in \acrshort{cv};
	modern \abs\ methods based on \acrshort{nerf}s cannot easily model them without significant modification
	(e.g., \cite{guo2022nerfren,zhu2022neural}), 
	despite applications in terms of editing
	(including, e.g., removing reflections from glass or making an object reflective).
	Moving to a path tracing framework, such as the \acrshort{iaddf} or an extension that can model transmission, 
	immediately enables disentangling or generating reflections.
	
	The third case is a simple glossy material, based on the classical Phong model \cite{phong75}.
	This formulation, while non-physical,
	simply mixes a Lambertian and specular reflection term,
	as has been previously applied in computer graphics 
	(e.g., see \cite{montes2012overview,jain1995machine}):
	\begin{align}
		\label{pddf:eq:glossy}
		f_{B,\mathrm{G}}(q,\omega_i,\omega_o) 
		&= \eta_L \frac{\rho_a}{\pi} + 			
		\frac{
			(1 - \eta_L) \rho_m
		}{
			N(\alpha, \omega_i, n)	
		}\,
		\left( \omega_o \cdot \omega_r(\omega_i,n) \right)^\alpha
		\\
		\Psi_\mathrm{G}(q,\omega_o) 
		&= 
		\eta_L \mathbb{U}[ \Omega[n(q)] ]
		+
		(1 - \eta_L)\, P_{\mathrm{Proj}}(q,\omega_o), 
	\end{align}
	where $\eta_L \in (0,1)$ balances diffuseness and specularity, 
	$N(\alpha, \omega_i, n) = \pi^{ \frac{1}{1 + \alpha} } \, |\omega_i \cdot n|^{ \frac{\alpha}{1 + \alpha} }$,
	$P_{\mathrm{Proj}}(q,\omega_o) = \mathcal{N}_{\mathrm{Proj}}(\omega_r(-\omega_o,n(q)), \Sigma_p(\alpha))$,
	$\alpha\in[0,\infty)$ is the specular exponent (controlling ``glossiness''),
	and $ \mathcal{N}_{\mathrm{Proj}} $ is a distribution centered 
	about the reflection direction, based on a spherical projection of a Gaussian.
	In the second line, we use the sum to denote a convex combination (mixture) of the two measures.
	We relegate the details of the Gaussian model to \S\ref{pddf:appendix:light:material}, 
	but remark that its sampling procedure can be back-propagated through\footnote{Notice that such samples depend on the \acrshort{ddf} through $n$ and material parameters, like $\alpha$, can be back-propagated to via pathwise derivatives.}, 
	via the reparameterization trick.
	The second term effectively ``interpolates'' between a mirror ($\alpha\rightarrow\infty$) and a 
	Lambertian material ($\alpha\rightarrow 0$).
	Notice that, in order to obtain visually reasonable specular highlights,
	the \acrshort{ddf} has to provide fairly accurate surface normals 
	(i.e., the derivatives of the mapping need to be correct), just as with the mirror case.

	\subsection{Path Tracing Post-Processing}
	\label{appendix:pt:postproc}
	We remark that our algorithm records the light flowing through a scene at a pixel,
	effectively measuring irradiance at that position.
	It is thus akin to a raw image (in ``linear RGB''),
	which is often not perceptually agreeable.
	In real cameras, of course, the standard
	camera image processing pipeline 
	transforms such images to standard RGB,
	through a complex series of specialized algorithms
	(e.g., \cite{karaimer2016software,yu2021reconfigisp}).
	In our case, we apply a very simplified differentiable post-processing for visualization,
	applying denoising (Gaussian filtering),
	min-max normalization,
	and then a gamma transform via $ \mathrm{rgb} = \mathrm{rgb}_\mathrm{linear}^{1/2.2} $.
	In order to fit to real images, 
	one would likely need a learnable mapping to mimic the camera pipeline
	(to produce the images to which we would be fitting).	
	
	\subsection{Additional Discussion}
	\label{pddf:light:adddisc}
	In conclusion, we have devised a simple approach to integrating \acrshort{ddf}s
	into a path tracing framework, via the \acrshort{iaddf}.
	The interpretation of recursive \acrshort{ddf} applications as covering the space of light paths 
	lends itself well to this combination.
	The \textit{internal structure modelling}
	capability of \acrshort{pddf}s,
	which other directed fields often ignore, is fundamental  
	to our light transport algorithm;
	without it, we could not model the intra-scene paths between surfaces.
	
	Similar to \acrshort{nerf}s, 
	one can differentiably render each pixel and \acrshort{mc} sample independently
	(i.e., in parallel), with a fixed number of queries 
	per ray.
	More exactly, we incur $\mathcal{O}(nm)$ \acrshort{ddf} calls per pixel,
	for $n$ bounces and $m$ \acrshort{mc} samples.
	(see Fig.~\ref{fig:algolight:pt}).
	While this is distinctly more costly than the \textit{single} call used for 
	\textit{geometric} neural rendering of \acrshort{ddf}s
	(and then using a simple \acrshort{inr}, such a texture field \cite{oechsle2019texture}),
	it enables a principled connection between \acrshort{ddf}s and appearance modelling.
	Moreover, rendering via light transport in this manner enables
	(a) decomposition of illumination, reflectance, and shape;
	(b) automatic handling of many natural phenomena, including 
	surface inter-reflections, glowing objects, multiple light sources, 
	complex material reflectances,
	and mirrors;
	and (c) differentiability through the rendering process 
	(i.e., potentially enabling gradient-based learnability).
	When editing \acrshort{iaddf} scene descriptions,
	one can easily relight scenes
	or 
	alter intrinsic material appearance of objects
	(even changing surfaces to mirrors or enabling them to emit light).
	The compositionality of \acrshort{ddf}s
	suggests inserting new objects should be relatively straightforward as well.
	Non-local changes in appearance due to such edits, 
	including shadows or inter-reflections,
	will be naturally accounted for by the path tracing algorithm.
	This is in contrast to, e.g., standard NeRF editing procedures,
	for which such non-local physical effects are not easily modelled or corrected.

	The most important challenge is in deriving 
	\acrshort{iaddf}s from image data.
	The \abs\ multiview image setup, as in NeRFs, is likely to be amenable to inverse rendering with \acrshort{iaddf}s, but powerful priors will be needed to constrain the inherent ambiguities between shape (shading), illumination, and reflectance, 
	as well as neural architectures able to learn higher dimensional (here, 5D) scene descriptions and learning methods for view consistency.
	For instance, priors on smoothness and material-type parsimony have already proven effective
	in decomposing lighting from reflectance 
	(e.g., \cite{barron2014shape,zhang2021nerfactor}); 
	constraining the space of BRDFs will also necessitate learning which materials are likely to appear in the world.
	Similarly, more sophisticated learned models of lighting and intrinsic material appearance will be needed, compared to the simple ones used here.
	In addition, 
	the use of a \acrshort{ddf} does not preclude most techniques 
	for improving \acrshort{mc} path tracing,
	(e.g., bidirectional tracing \cite{veach1995bidirectional}, 
	importance sampling \cite{veach1997metropolis,muller2019neural}, 
	radiosity techniques \cite{hadadan2021neural}).
	Research on such methods 
	has been a major topic within computer graphics for decades
	(and increasingly enmeshed with differentiable and/or learning-based techniques);
	combining such advances with the IADDF %
	should hopefully lead to a disentangled neural scene representation
	that     (i) can be efficiently rendered, 
	(ii) interoperates with other \acrshort{ml} algorithms,
	(iii) models a myriad of physical phenomena,
	(iv) can be learned in a weakly supervised \abs\ manner,
	and (v) is easily edited 
	in terms of shape, material, and lighting.

%% file: supp-theory.tex
\section{Theoretical Details: View Consistency of DDFs}
\label{suppmat:theory}

This section is an extended and more complete version of the theoretical analysis in \S\ref{sec:theory} of the main paper.

\subsection{Preliminaries}

In this section, our intention is to explore the conditions under which a DDF, or one of its constituent fields, represents a shape. 
In particular, we show an equivalence between certain local properties of such fields and the existence of a 3D shape to which the field has been perfectly ``fitted''.
We begin by defining the basic setting, including the domain in which our fields operate, the oriented points that characterize our ray-based geometric representation, 
and the precise definition of a shape.

\begin{mydef}{Domain $\bo$, $\boe$, $\Gamma$, and $\Gamma_\varepsilon$}{pddf:prelim:domain}
	We consider fields with a geometrically simple domain, $\bo \subset \real^3$.
	In particular, we assume that \mbo \ is convex, compact, and has a smooth boundary, $\partial \bo$,
	with outward-facing surface normals.
	Finally, 
	for $\varepsilon > 0$,
	we denote $\boe$ as the $\varepsilon$-domain of $\bo$, defined via 
	\begin{equation}
		\boe = \left\{ p \in \bo \,\middle\vert\, 
		\min_{b\in\partial\bo} || p - b || \geq \varepsilon \right\},
	\end{equation}
	which will be useful for ensuring certain boundary conditions on our fields of interest.
	
	Often, we will be interested in the 5D Cartesian product space, $\Gamma = \bo \times \mathbb{S}^2$.
	Similarly, we can define $\Gamma_\varepsilon = \boe \times \mathbb{S}^2$.
\end{mydef}

The role of $\boe$ is to enable us to define a notion of 
``away from the boundary''
(i.e., $\bo\setminus\boe$ is a thin $\varepsilon$-width shell around the edges of the domain).
Notice that we restrict our attention to a small part of $\real^3$, 
where our shape of interest will reside,
but the domain of our fields will be the 5D space $\Gamma$, 
comprising oriented points or, equivalently, rays.

\begin{mydef}{Oriented Points and Rays}{pddf:prelim:orienrays}
	We generally denote an \textit{oriented point} via
	$\tau = (p,v) \in \Gamma$ (or in $\Gamma_\varepsilon$ when specified).
	Any oriented point induces (or is equivalent to) a 3D \textit{ray}: $r_\tau(t) = p + tv,\, t\geq 0$.
	In a slight abuse of notation, we may refer to the oriented point and ray forms of such 5D elements interchangeably. 
\end{mydef}

Fundamentally, our interest is in representing 3D shapes, 
which are defined simply as follows.

\begin{mydef}{Shapes}{pddf:prelim:shape}
	We define a \textit{shape} to be a compact set $S\subset \bo$.
	Often, we will be interested in shapes 
	$S \subset \boe$,
	for $\varepsilon > 0$.
\end{mydef}

Note that using $\boe$ does not strongly constrain the shape: for example, one can simply use the bounding sphere of the given point set, and then inflate it enough to satisfy the $\varepsilon$ condition.

Next, for the sake of notational simplicity, we define an ``along-ray'' form of any function (or field) that operates on oriented points (i.e., in 5D), by considering its restriction to a given ray.

\begin{mydef}{Along-Ray Functions}{pddf:prelim:alongray}
	We define the ``along-ray'' form of a function
	$ g : \Gamma \rightarrow \mathcal{X} $,
	which maps into a set $\mathcal{X}$, to be:
	\begin{equation}
		f_g(s\mid \tau) := g(p + sv, v) = g(r_\tau(s), v),
	\end{equation}
	where $s \geq 0$, $\tau = (p,v) \in \Gamma$, and
	$r_\tau$ is the ray form of $\tau$.
\end{mydef}

Finally, we introduce notation for the intersections of rays and point sets.
\begin{mydef}{Intersecting Rays and Point Sets}{pddf:prelim:intersections}
	Consider a point set $S\subseteq \bo$ and an oriented point $\tau\in\Gamma$.
	
	\textit{Intersected Points}.
	Let $S_\tau\subseteq S$ denote the set of points in $S$ intersected by $r_\tau$
	(i.e., $q\in S_\tau$ iff $\,\exists\; t\geq 0$ such that $r_\tau(t) = q$).
	For notational clarity, we may write $[S]_\tau$ as well.
	
	\textit{Intersecting Rays}.
	Let $\mathcal{I}_S \subseteq \Gamma$ be the set of rays 
	that intersect at least one member of $S$
	(i.e., $\tau \in \mathcal{I}_{S}$ if $\exists\; q\in S$ such that $r_\tau(s) = q$ for some $s\geq 0$).
	For $q\in\bo$, we denote $ \mathcal{I}_{q} := \mathcal{I}_{\{q\}} $.
\end{mydef}

\subsection{View Consistency for Simple DDFs}
\label{appendix:theory:simpleddfs}

Our first consideration will be a simplified form of DDF, 
which does not utilize a visibility field.

\begin{mydef}{NNBC Field}{pddf:simple:nnbc} \index{NNBC field}
	An \textit{NNBC field} on $\Gamma$ is a \underline{n}on-\underline{n}egative, \underline{b}ounded scalar field that is piece-wise \underline{c}ontinuously differentiable along rays, 
	written $ d : \Gamma \rightarrow \real_{\geq 0} $. 
	The latter condition means the along-ray form is piecewise $C^1$ 
	(i.e., it satisfies $f_d(s\mid p,v) = d(p + sv,v) \in C^1$).
\end{mydef} 

In general, such fields need not have an obvious connection to any shapes (in the sense of the definition above).
In this section, we treat such fields as \textit{putative} representations for shapes -- the goal is to understand the exact conditions under which such fields are ``equivalent to'' (or ``represent'') some shape.

\begin{mycdef}{Positive Boundary Condition for Simple DDFs (BC$_d$)}{pddf:simple:bcd}
	An NNBC field $d$ satisfies the \textit{Positive Boundary Condition} BC$_d$ iff
	\begin{equation}
		\inf_{v\in\mathbb{S}^2} d(p,v) > 0 \;\forall\; p\in \bo\setminus\boe.		
	\end{equation}	
\end{mycdef}

Recall that $\bo\setminus\boe$ is the ``outer shell'' of the bounding domain. In other words, BC$_d$ demands that $d$ cannot have any zeroes close to $\partial\bo$. 

\begin{mycdef}{Directed Eikonal Condition for Simple DDFs (DE$_d$)}{pddf:simple:ded}
	An NNBC field $d$ satisfies the \textit{Directed Eikonal Constraint} DE$_d$ if
	$\partial_s f_d(s\mid\tau) = -1$, except at along-ray zeroes of $d$ 
	(i.e., at any $(q,v)$ such that  
	$\lim_{\epsilon \downarrow 0} \inf_{s\in (0,\epsilon)}
	d(q - sv, v) = 0$).
\end{mycdef}

The second constraint is on the derivative of $d$:
it says that, along any ray, $d$ must decrease linearly, at unit rate, unless $d=0$ (a value at which it may potentially stay).

\begin{mycdef}{Isotropic Opaqueness Condition for Simple DDFs (IO$_d$)}{pddf:simple:iod}
	An NNBC field $d$ satisfies the \textit{Isotropic Opaqueness Constraint} at a point 
	$q\in\bo$ if
	\begin{align}
		& \left(
		\lim_{\epsilon \downarrow 0} 
		\inf_{ \substack{s\in (0,\epsilon)\\ v\in\mathbb{S}^2 } }
		d(q - sv, v) = 0
		\right)
		\,\implies\, \nonumber \\
		& \;\;\;\;\; \left(
		\lim_{\epsilon \downarrow 0} \inf_{s\in (0,\epsilon)}
		d(q - sv, v) = 0
		\,\forall\, v\in\mathbb{S}^2\right).    
		\label{eq:iod}
	\end{align}
	We say $d$ is isotropically opaque, denoted IO$_d$, if it satisfies the Isotropic Opaqueness Constraint everywhere in $\bo$.
\end{mycdef}

Notice that this is stronger than (i.e., it implies) the following constraint,
which refers only to local properties 
in the neighbourhood of some oriented point $(q,v)$:
\begin{align}
	&\left(
	\exists\,v\in\mathbb{S}^2 \;\mathrm{s.t.}\; 
	\lim_{\epsilon \downarrow 0} \inf_{s\in (0,\epsilon)}
	d(q - sv, v) = 0
	\right)
	\,\implies\, \nonumber \\
	& \;\;\;\;\; \left(
	\lim_{\epsilon \downarrow 0} \inf_{s\in (0,\epsilon)}
	d(q - sv, v) = 0
	\,\forall\, v\in\mathbb{S}^2
	\right).
\end{align}

Namely, if the field approaches zero at some position $q$ from \textit{any} direction $v$, then it must approach zero at that point from \textit{all} directions.
Note that, for the left-hand side of Eq.~\ref{eq:iod}, 
a single particular direction $v$ along which $d$ converges to zero need not exist 
(see Supp.~\S\ref{pddf:theory:convergenceexample} for an example).
The weaker, second form of IO$_d$ means that convergence to a zero from one direction must be accompanied by a zero from all directions.

Together, these field constraints enable us to define a \textit{simple DDF}, which we will show is capable of representing shapes with theoretical consistency guarantees.

\begin{mydef}{Simple DDF}{pddf:simple:simpleddf} \index{simple DDF}
	We define a simple DDF to be an NNBC field such 
	that IO$_d$, BC$_d$, and DE$_d$ are satisfied.
\end{mydef}

The remainder of this section will be devoted to showing that
the three field constraints, which define a simple DDF,
are sufficient to ensure a form of ``view consistency''.

Before that, however, we define some notational constructs around Simple DDFs.
First,
we define a simple function based on $d$,
which outputs the \textit{position}
that our depth field takes us to, along a given ray.
\begin{mydef}{Positional Field}{pddf:simple:q}
	Any simple DDF has an associated positional field, or $q$-mapping, defined by:
	$q(p,v) := p + d(p,v) v$.
\end{mydef}

Next, because DDFs can be discontinuous at zeroes,
for convenience of notation, 
whenever a simple DDF $d$ approaches zero along a ray, 
we denote this as a zero of the field (regardless of the nature of any discontinuities at such points).

\begin{mydef}{Left-Continuity Convention}{pddf:simple:leftcont}
	For any $q\in\boe$, we write:
	\begin{equation}
		d(q,v) = 0 
		\,\iff\,
		\lim_{\epsilon \downarrow 0} \inf_{s\in (0,\epsilon)}
		d(q - sv, v) = 0. \label{eq:contconv}
	\end{equation}
\end{mydef}

In words, left continuity for $d$ defines a slightly looser notion of a zero along rays (here, we say ``left'' with respect to the direction $v$).
We also apply this convention to $q$, so that if $d(p,v)=0$, we write $q(p,v) = p$.
Finally, we define a special set of points that is associated to any simple DDF.

\begin{mydef}{Positional Zero Set for Simple DDFs}{pddf:simple:zeroes}
	We denote the set of \textit{positional zeroes} of a Simple DDF as
	\begin{equation}
		Q_d = %
		\left\{ p\in\bo \;\middle\vert\; %
		\lim_{\epsilon \downarrow 0} 
		\inf_{ \substack{s\in (0,\epsilon)\\ v\in\mathbb{S}^2 } }
		d(p - sv, v) = 0 \right\}.
	\end{equation}
\end{mydef}

Note that the condition for being in $Q_d$ is slightly weaker than that of being a directional zero, 
in the sense of Eq.~\ref{eq:contconv}.
I.e., there may be $q \in Q_d$ such that 
there is no $v$ for which $d(q,v) = 0$.
However, any $q$ that satisfies $d(q,v) = 0$ for some $v$ is necessarily in $Q_d$.

Our first lemma concerns these positional zeroes, 
namely that it is a closed set bounded in $\boe$.

\begin{restatlem}{Properties of Simple DDF Positional Zeroes}{pddf:simple:closedinboe}
	Let $d$ be a simple DDF.
	Then $Q_d \subseteq \boe$ and $Q_d$ is a closed set.
\end{restatlem}
\begin{proof}
	See Supp.~\S\ref{pddf:theory:propposzeroes}.
\end{proof}

Notice this shows that $Q_d$ is a shape 
(i.e., compact point set).
We will use these properties later, 
connecting this point set to the shape 
represented by the field (as we will later show).
The first step along this line is to define what it means 
for a simple DDF to ``exactly'' represent a shape.

\begin{mydef}{Shape-Induced Simple DDFs}{pddf:simple:shapeinduction}
	Let $S$ be a shape such that $S\subseteq\boe$. 
	We define a set of ``induced fields'' associated to $S$, which each act as a form of \textit{representation} for the shape.
	Recall that $S_\tau$ denotes the points in $S$ intersected by $r_\tau$.
	Then $ S $ \textit{induces} a field $ d : \Gamma \rightarrow \real_{{\geq}0} $
	iff \\
	(i) for any $\tau=(p,v)\in\Gamma$ such that $S_\tau \neq \varnothing$,
	$d(\tau) = \min_{q \in S_\tau} ||q - p||$
	and \\
	(ii) $\exists\; \varepsilon_E > 0$ such that
	for any $\tau\in\Gamma$ with $S_\tau = \varnothing$, 
	we have that $d(\tau) > \varepsilon_E $, $q(\tau) \notin \bo $, and $f_d$ is $C^1$ along $r_\tau$ with 
	$ \partial_s f_d(s \mid \tau) = -1 $.
	
\end{mydef}

Intuitively, a point set $S$ induces a field $d$ if (i) the $q$-mapping always takes $p$ to the closest point in $S$ along $v$, unless (ii) no intersection exists, in which case $q$ must predict a point \textit{outside} the valid domain. 
Notice, of course, that a large class of $d$ can therefore satisfy this for a given $S$, differing only in where outside \mbo\  they output.\footnote{
	An alternative form of shape-induction (see Supp.~\S\ref{pddf:theory:altfieldind}) avoids defining notions of equivalence by including the boundary $\partial\bo$ with the shape $S$. This enables a simpler form of equality, but the approach above is more analogous to what we will do in the case of the full DDF.
}
We next show that any induced field is a Simple DDF.

\begin{restatthm}{Shape-Induced Fields Are Simple DDFs}{pddf:simple:sifsasdds}
	Consider a field $d$ that is induced by some shape $S$.
	Then $d$ is a simple DDF. 
	Further, the positional zeroes of $d$ in $\bo$ satisfy $S = Q_d$.
\end{restatthm}
\begin{proof}
	See Supp.~\S\ref{pddf:theory:sifasd}.
\end{proof}

Recall that the definition of a shape-induced field need not be unique (as two induced fields may differ on rays that do not intersect $S$) and therefore allows a given $S$ to induce a \textit{set} of $d$ fields, which differ only in terms of their out-of-domain behaviour. 
Hence, we need a measure of ``equivalence'' between Simple DDFs.
A given shape $S$ can then induce an equivalence class of Simple DDFs.
To define this notion of equivalence, we first define the set of empty rays.

\begin{mydef}{Empty Rays}{pddf:simple:emptyrays}
	Let $d$ be a Simple DDF.
	Then we let
	$\mathcal{E}[d] \subseteq \Gamma$ be the set of \textit{empty rays}, defined by 
	$ \mathcal{E}[d] = \{ \tau \in \Gamma \mid q(\tau)\notin \bo \} $, where $q$ is the $q$-mapping of $d$.
	Notice that an empty ray $\tau \in \mathcal{E}[d]$ must have 
	$f_d(s\mid \tau) > 0$ as long as $r_\tau(s) \in \bo$.
	(I.e., we cannot have a zero of $d$ on an empty ray). 
\end{mydef}

For a given $d$, 
the empty rays are those that $d$ maps outside of the domain.
To check equality,
we can ignore the depth \textit{values} of empty rays,
as long as \textit{which} rays are empty is the same.

\begin{mydef}{Equivalence of Simple DDFs}{pddf:simple:equiv}
	Consider two Simple DDFs, $d_1$ and $d_2$. 
	Then $d_1$ and $d_2$ are \textit{equivalent} iff
	(i)
	both fields have the same set of empty rays, written $\mathcal{E}$ 
	(i.e., $ \mathcal{E} := \mathcal{E}[d_1] = \mathcal{E}[d_2] $)
	and
	(ii) 
	they agree on non-empty rays
	(i.e., $d_1(\tau) = d_2(\tau)\;\forall\; \tau \in \Gamma \setminus \mathcal{E}$).
\end{mydef}

Later we will see a slightly different form of equivalence 
between \textit{full} DDFs 
(i.e., including a visibility field), 
similar to that of Simple DDFs.
First, however, we formally define what it means for a simple DDF to be a shape representation.

\begin{mydef}{Shape Representations as View-Consistent Simple DDFs}{pddf:simple:shaperep}
	A Simple DDF $d$ is View Consistent (VC)
	iff
	there exists a shape $S$ such that 
	$S$ induces $d$.
	
	We can therefore call any such $d$ a \textit{shape representation} for the point set $S$.
\end{mydef}

Recall that an NNBC field (see Def.~\ref{df:pddf:simple:nnbc})
maps rays (in 5D) to positive scalars, with very little constraints.
Most such fields will not be view-consistent, 
in the sense that they will not represent a shape (point set), 
in a self-consistent manner
(e.g., violating the basic inequality\footnote{
	Recall that if $q(p_1,v_1)$ is a predicted surface point,
	then $d(p_2,v_2) \leq || p_2 - q(p_1,v_1) ||$, 
	for any ray $r_{(p_2,v_2)}$ that intersects $q(p_1,v_1)$. 
} 
from \S\ref{pddf:appendix:viewconsis}).
The definition above formalizes what it means for a field to 
be view-consistent.

The next theorem is the main result of this section.
It shows that the local\footnote{Local in the sense that each constraint can be defined about a single point, particularly IO$_d$ (a point that is zero along one ray must be zero on all of them) and DE$_d$ (the along-ray derivative at any point is $-1$). } field constraints (Constraints \ref{cdf:pddf:simple:bcd}, \ref{cdf:pddf:simple:ded}, and \ref{cdf:pddf:simple:iod}) are sufficient to ensure a Simple DDF represents a shape (i.e., in a view-consistent manner).

\begin{restatthm}{Every Simple DDF is a Shape Representation}{pddf:simple:esdiasr}
	Any simple DDF $d$ 
	is view consistent 
	(i.e., there exists a shape $S \subseteq \boe$ such that $d$ is equivalent to any field induced by $S$).
	Further, 
	$d$ is a shape representation for the set formed by its positional zeroes, $Q_d$
	(i.e., $Q_d$ induces $d$).
\end{restatthm}
\begin{proof}
	See Supp.~\S\ref{pddf:theory:esdiasr}.
\end{proof}

The two theorems above
(\ref{th:pddf:simple:sifsasdds} and \ref{th:pddf:simple:esdiasr})
show that being a simple DDF 
(i.e., non-negativity, no zeroes close to the boundary, isotropic zeroes, and unit along-ray derivatives) 
is equivalent to view consistency (i.e., to being induced by some shape).
In other words, 
the constraints on the implicit field of being a simple DDF is synonymous with being a shape representation.
Later, these results will be useful when showing that \textit{full} DDFs 
can be constrained in a similar manner, 
in order to guarantee view consistency.

\subsection{View Consistency for Visibility Fields}
\label{appendix:theory:visfields}

We next look at the second constituent field in a DDF: the visibility field, which demarcates whether or not a shape is present along a given ray. Similar to the previous section, our interest is in determining sufficient constraints on a given visibility field such that one can be assured it is ``view-consistent'', meaning compatible with some 3D shape. 

As for simple DDFs, we begin by defining a relatively unconstrained field,
for which we shall find constraints relating to view consistency.

\begin{mydef}{BOZ Field}{pddf:vis:bozf} \index{BOZ field}
	Let $\xi : \Gamma \rightarrow \{ 0, 1 \}$ be a binary-valued field on $\Gamma$. 
	We restrict $\xi$ such that its zero set is open; i.e.,
	$\forall \; \tau \in \Gamma$, if $\xi(\tau) = 0$, then $\exists\; \varepsilon > 0$ such that $\xi(\widetilde{\tau}) = 0 \;\forall\; \widetilde{\tau} \in B_{\varepsilon}(\tau) \cap \Gamma $, where $B_{\varepsilon}(\tau)$ is the open ball centered at $\tau$ of radius $\varepsilon$.
	
	We call any such \underline{b}inary field, with an \underline{o}pen \underline{z}ero set, a BOZ field.
\end{mydef}

Similar to NNBC fields (Def.~\ref{df:pddf:simple:nnbc}), a BOZ field does not have an obvious connection to 3D shapes or their multiview silhouettes -- one can  imagine BOZ fields that do not represent a consistent 3D point set from all viewpoints.
Our goal is to understand when (or under what conditions) a BOZ field acts as a continuous representation of the silhouettes of some coherent shape
(i.e., when it assigns every ray a binary indication 
as to whether a shape point exists along that ray or not).
It turns out a very similar situation to that of Simple DDFs arises.
Thus, similarly, we begin by defining field constraints,
which will eventually connect BOZ fields to coherent silhouettes of shapes. 

\begin{mycdef}{Non-Visible Boundary Condition for Visibility Fields (BC$_\xi$)}{pddf:vis:bcxi}
	A BOZ field $\xi$ satisfies the 
	\textit{Non-Visible Boundary Condition},
	denoted BC$_\xi$, if
	$\xi(p,v) = 0 \;\forall\; (p,v)\in 
	\mathcal{O}[\bo\setminus\boe, \boe]$,
	where 
	$\mathcal{O}[V,U] = \{ \tau=(p,v) \in\Gamma \mid p\in V, r_\tau(s) \notin U \;\forall\; s\geq 0 \}$ is the set of ``outward rays'' from a given set.
\end{mycdef}
Recall that $\bo\setminus\boe$ is the ``outer shell'' of the bounding domain and, essentially, $\mathcal{O}[V,U]$ contains oriented points (or, equivalently, rays) that \textit{start} in $V$ and do \textit{not} intersect $U$.
In other words, BC$_\xi$ demands that any rays that (i) start at a point close to the boundary and (ii) do not intersect the inner part of the domain (i.e., $\boe$) must have a value of zero. 
Intuitively, if we imagine $\xi$ to be representing an object in $\boe$, 
this condition prevents the object from being visible on rays 
that start outside of $\boe$ and ``look away'' from $\boe$.

\begin{mycdef}{Directed Eikonal Constraint for Visibility Fields (DE$_\xi$)}{pddf:vis:dexi}
	A BOZ field $\xi$ satisfies the \textit{Directed Eikonal Constraint} DE$_\xi$ 
	if
	$\xi$ is always non-increasing along a ray
	(i.e., $ f_\xi(s_1|\tau) \leq f_\xi(s_2|\tau) \;\forall \tau \in \Gamma, s_1 > s_2$ ).
\end{mycdef}
Intuitively, along a ray, this constraint means that the visibility can only ever go from ``seeing'' to ``not seeing'' an object 
(i.e., along a ray, one should not suddenly see a new point in the shape become visible).

\begin{mycdef}{Isotropic Opaqueness for Visibility Fields (IO$_\xi$)}{pddf:vis:ioxi}
	A BOZ field $d$ satisfies the \textit{Isotropic Opaqueness Constraint} at a point $q\in\bo$ if
	the following implication holds:
	\begin{align}		
		\exists\; v\;\mathrm{s.t.}\,
		&\left(
		\lim_{\epsilon\downarrow 0} \sup_{s\in(0,\epsilon)}
		\xi(q - sv,v) = 1\right)
		\;\land\; \nonumber \\ 
		&%
		\left(
		\lim_{\epsilon\downarrow 0} 
		\inf_{s\in(0,\epsilon)}
		\xi(q + sv,v) = 0 \vphantom{\sup_{s\in(0,\epsilon)}}
		\right) \nonumber \\
		\;\implies\; &
		\big(
		\xi(\tau) = 1
		\;\forall\;\tau \in \mathcal{I}_q 
		\big), 
	\end{align}
	where $\mathcal{I}_q $ is the set of rays in $\bo$
	(i.e., $r_\tau$, for oriented points $\tau\in\bo$)
	that intersect $q$.
	An \textit{isotropically opaque field} (IO$_\xi$) satisfies the Isotropic Opaqueness Constraint everywhere.	
\end{mycdef}

In words, if there is \textit{any} direction 
along which the field flips from one to zero,
then the point at which that occurs must be ``isotropically opaque''
(i.e., from any direction, any ray that hits that point must produce a visibility value of one).
Intuitively, IO$_\xi$ says that 
such ``one-to-zero`` discontinuities in $\xi$ are special. 
As we shall see, they act similar to surface points 
(or zeroes of $d$, in the case of simple DDFs).
In particular, for any point $q\in\bo$, 
if even a single direction $v$ exists such that 
a one-to-zero discontinuity occurs along $v$ at $q$, then any ray that contains $q$ must also be visible.    

As for simple DDFs, with these three constraints,
we can now define an appropriately limited set of fields, 
called visibility fields,
which we will show is intimately tied to shapes.

\begin{mydef}{Visibility Field}{pddf:vis:visibilityfields}
	We define a Visibility Field to be a BOZ field such that
	IO$_\xi$, BC$_\xi$, and DE$_\xi$ are satisfied.	
\end{mydef}

Analogously to simple DDFs with left continuity (Def.~\ref{df:pddf:simple:leftcont}), 
we utilize a notational convention for visibility, 
which slightly broadens when an oriented point is considered visible.

\begin{mydef}{Upper-Continuity Convention for Visibility Fields}{pddf:vis:uppercont}
	Whenever a visibility field $\xi$ approaches one, 
	we denote the field value as one, regardless of the nature of any discontinuities at such points.
	In particular, for any $q\in\bo$, 
	we write:
	\begin{equation}
		\xi(q,v) = 1 
		\,\iff\,
		\lim_{\epsilon \downarrow 0} 
		\sup_{ \substack{ || \delta || < \epsilon \\ q + \delta \in \bo } } 
		\xi(q + \delta, v) = 1.
		\label{eq:contconv:vis}
	\end{equation}
\end{mydef}
In other words, we assume ``upper continuity'' for $\xi$, along all directions.
Notice that this is consistent with the requirement on BOZ fields that the zero set $\xi^{-1}(0)$ is open.
Further, its complement, $\xi^{-1}(1) \subseteq \Gamma = \bo\times\mathbb{S}^2$, is closed (using the fact that $\bo$ is compact).

\begin{mydef}{Positional Discontinuities of the Visibility Field}{pddf:vis:posdiscont}
	Consider a visibility field $\xi$. 
	Then we define its positional discontinuity set as follows:
	\begin{align}
		Q_\xi :=
		\closure & %
		\left(
		\left\{ 
		q \in \bo
		\;\middle\vert\;
		\exists\; v\;\mathrm{s.t.}\; 
		\lim_{\epsilon\downarrow 0} \sup_{s\in(0,\epsilon)}
		\xi(q - sv,v) = 1
		\right.\right. \nonumber
		 \\
		\land\;
		&
		\left.\left.
		\lim_{\epsilon\downarrow 0} \inf_{s\in(0,\epsilon)}
		\xi(q + sv,v) = 0
		\right\}
		\right), %
	\end{align}  
	where $\closure$ is the closure operator on sets.
\end{mydef}
Intuitively, given a visibility field $\xi$, the set $Q_\xi$
simply contains all of the one-to-zero discontinuities in $\xi$.
More specifically, for any point $q\in\bo$, 
if there exists some direction $v$ such that the field $\xi$
flips from one to zero 
(i.e., from ``seeing something'' to ``seeing nothing'') 
at $q$,
then $q\in Q_\xi$.

\begin{restatlem}{Properties of Positional Discontinuities of Visibility Fields}{pddf:vis:posdiscont}
	A visibility field $\xi$ satisfies 
	(i) $Q_\xi \subseteq \boe$,
	(ii) $Q_\xi$ is a shape, and 
	(iii) there cannot be discontinuities in $\xi$ from zero-to-one along a ray.
\end{restatlem}
\begin{proof}
	See Supp.~\S\ref{pddf:theory:posdiscont}.
\end{proof}

Given these properties, the discontinuities $Q_\xi$
are a good candidate for a potential shape that
$\xi$ is implicitly representing.
First, like for simple DDFs, 
we need to define what it means for a shape 
to \textit{induce} a field 
(i.e., ``exactly'' represent it).

\begin{mydef}{Shape-Induced Binary Fields}{pddf:vis:sibfs}
	Let $S\subseteq\boe$ be a shape. 
	Recall that $\mathcal{I}_{S} \subseteq\Gamma$ is the set of rays that intersect $S$.
	Then we say a BOZ field $\xi$ %
	is \textit{induced} by $S$ if the following holds:
	\begin{equation}
		\xi(\tau) = 1 \;\iff\; \tau \in \mathcal{I}_{S}.
	\end{equation}
	Further, given a fixed shape $S$, its associated induced field $\xi$ is unique.
\end{mydef}

Notice that the induction relation is an ``iff'', 
meaning $\xi(\tau) = 0$ implies that \textit{no} $q\in S$
can be present along the ray $r_\tau$.
In addition, importantly, while the $S$-induced field $\xi$ is unique,
the inducing shape for a given $\xi$ is \textit{not} necessarily unique.
For instance, imagine a sphere with another shape (say, another sphere) inside it -- such a shape will induce the same visibility field as using the outermost sphere alone for the inducement. 
Given this definition,
our first result in this section shows that any shape-induced field 
satisfies the constraints of a visibility field.
\begin{restatthm}{Shape-Induced Binary Fields Are Visibility Fields}{pddf:vis:sibfavf}
	Let $S\subseteq \boe$ be a shape, with an induced field $\xi$.
	Then $\xi$ is a visibility field 
	(i.e., IO$_\xi$, BC$_\xi$, and DE$_\xi$ hold).
	Also, the positional discontinuities of $\xi$ satisfy $Q_\xi \subseteq S$.
\end{restatthm}
\begin{proof}
	See Supp.~\S\ref{pddf:theory:sibfavf}.
\end{proof}

So far we have shown that any shape can be used to generate or induce 
a visibility field (in the sense of Def.~\ref{df:pddf:vis:visibilityfields}).
We can use this to define a notion of shape representation for visibility fields.

\begin{mydef}{Shape {\Xirepname}s as View-Consistent Visibility Fields}{pddf:vis:shaperep}
	A visibility field $\xi$ is View Consistent (VC) iff there exists a shape $S$ 
	such that $S$ induces $\xi$.
	We can therefore call  
	any such $\xi$ a \textit{shape} \textit{{\xirepname}} for the point set $S$.
\end{mydef} \index{shape indicator}
Note we call it a shape indicator instead of a representation because we cannot (necessarily) reconstruct $S$ from $\xi$ (i.e., it would be an incomplete representation). Instead, we call it an {\xirepname} because it ``indicates'' whether or not the shape (or some shape) exists along the given ray
(i.e., acts as an indicator function for each ray). 
We next show that the local constraints of a visibility field
are sufficient for it to be a representation of some shape.
\begin{restatthm}{Every Visibility Field is a Shape \Xirepname}{pddf:vis:evfias}
	Let $\xi$ be a visibility field.
	Then there exists a point set $S$ such that $\xi$ is induced by $S$.
	Further, we can take the positional discontinuities as an inducer; 
	i.e., $S := Q_\xi$ induces $\xi$.
\end{restatthm}
\begin{proof}
	See Supp.~\S\ref{pddf:theory:evfias}.
\end{proof}

So far, we have shown a close duality between visibility fields
and shape indicators (i.e., between binary fields with specific local constraints on the field, namely IO$_\xi$, DE$_\xi$, and BC$_\xi$, and binary fields constructed from a given point set).
However, it is clear that many inducing point sets can induce the same field.
Unlike the case with simple DDFs, 
this is not merely a question of behaviour outside the domain
(e.g., recall the example of a sphere within a sphere inducing the same field as the outer sphere alone).
We can therefore ask for a ``minimal example'' over an equivalence class of inducing point sets
(i.e., given many shapes $S$ that all induce the same $\xi$, which is the simplest?).

This leads us to the following corollary, which shows that
the simplest (i.e., smallest) shape (i.e., closed point set)
that induces a given
visibility field is the set of positional discontinuities.

\begin{restatcor}{Minimal Characterization of Visibility Fields}{pddf:vis:minimalcharvis}
	Consider the set of shapes 
	$ \mathfrak{S}[\xi] = \{ S \mid S\subseteq\bo \;\text{induces}\; \xi \}$ 
	that all induce the same field $\xi$.
	I.e., $\mathfrak{S}[\xi]$ is an equivalence class of inducing shapes.
	Then, $Q_\xi \in \mathfrak{S}[\xi]$ is the smallest closed point set among all such inducers.
\end{restatcor}
\begin{proof}
	See Supp.~\S\ref{pddf:theory:minimalcharvis}.
\end{proof}

This shows that $Q_\xi$ is special among shapes that induce $\xi$,
in that it forms a subset of any other shape that also induces $\xi$.
Next, we provide some additional intuition concerning $Q_\xi$, 
which is derived from the visibility field itself.
Instead, given any shape $S$, we can describe which subset of points
actually forms $Q_\xi \subseteq S$, 
without referring to any $\xi$ directly.

\begin{mydef}{Directly Lit Points}{pddf:vis:dlp} \index{Directly Lit Points}
	Consider a shape $S$ contained in $\boe$.
	Imagine a set of uniform lights placed at each $p\in\partial\bo$
	(or, equivalently, imagine each such $p$ casting rays in all possible directions.
	We denote the set of points in $S$ that are ``directly lit'' by such lights
	(or intersected \textit{first} by such cast rays) as
	$\mathfrak{D}_\ell(S)$.
	Mathematically, the directly lit points (DLPs) can be written
	\begin{align}
		\mathfrak{D}_\ell(S) = 
		\closure &
		\left(
		\left\{
		\vphantom{\arg\min_{ \widetilde{q} \in S_\tau }}
		q \in S \;\middle\vert \;
		\exists\;\tau = (p,v)\in\Gamma_\partial
		\,\;\mathrm{s.t.} \right.\right. \nonumber \\ & \left. \left. 
		q = \arg\min_{ \widetilde{q} \in S_\tau } || p - \widetilde{q} ||
		\right\}\right),
	\end{align}
	where $\Gamma_\partial = \partial\bo\times \mathbb{S}^2$ and 
	$S_\tau$ is the set of points in $S$ that are intersected by the ray $r_\tau$.
\end{mydef}

Intuitively, any point in $S$ that can be hit by a light ray from the boundary, 
directly and without occlusion,
is a DLP.
One could, equivalently, call $q\in \mathfrak{D}_\ell(S)$ 
an \textit{observable} point from the boundary.
Similar to $Q_\xi$, 
we include a closure operator, 
to ensure that $ \mathfrak{D}_\ell(S) $ 
contains its limit points and boundary.
Without this, the directly lit set is not necessarily closed.

We next show that directly lit points are an alternative characterization of the positional discontinuities.
Intuitively, it relates the local field property of a one-to-zero discontinuity
to the ``observable'' behaviour of the shape from the boundary $\partial \bo$.
In other words, DLPs are defined via the point set $S$, 
\textit{not} its induced field $\xi$. 
However, we next show that $\mathfrak{D}_\ell(S)$ is a special subset of $S$,
in that it forms a minimal inducer for the equivalence class $\mathfrak{S}[\xi]$,
and it is therefore equal to $Q_\xi$.

\begin{restatcor}{Directly Lit Points Minimally Characterize Visibility Fields}{pddf:vis:dlpsareqxi}
	Given any shape $S$ in the equivalence class $\mathfrak{S}[\xi]$ of inducers for the field $\xi$,
	the set of DLPs $\mathfrak{D}_\ell(S)$ is a \textit{minimal} inducer of $\xi$
	and furthermore $\mathfrak{D}_\ell(S) = Q_\xi$.
\end{restatcor}
\begin{proof}
	See Supp.~\S\ref{pddf:theory:dlpsareqxi}.
\end{proof}

In words, to get the minimal inducing point set from a shape $S$, one need not construct the induced field and then find its discontinuities. 
Instead, one can compute the DLPs directly from $S$, by ``looking inwards'' from the boundary and finding the observable (i.e., directly lit) points.

\textbf{Summary.}
We have shown that 
(i) we can devise a notion of ``inducement'', under which any 3D shape gives rise to an associated visibility field, $\xi$, which acts as its binary indicator function along rays;
(ii) the local constraints that define a visibility field $\xi$ 
(i.e., IO$_\xi$, DE$_\xi$, BC$_\xi$) are sufficient to guarantee that 
\textit{some} point set must exist that induces $\xi$
(meaning, every visibility field is the indicator of some shape);
(iii) the positional discontinuities $Q_\xi$ play a special role, as the \textit{minimal} closed subset of $S$ that can induce a given field $\xi$
(in some sense, $Q_\xi$ represents the best reconstruction of the initial shape that one can obtain from $\xi$); and
(iv) an alternative but equivalent characterization of the minimal closed inducing subset of $S$ can be given by the set of directly lit points (DLPs).

\subsubsection{Shape-from-Silhouette, Visual Hulls, and Visibility Fields}
\label{appendix:theory:sfsvhvf}

The connection to DLPs (or boundary-observable points) allows us to interpret $\xi$ more intuitively as the continuous field of silhouette images.
In this setting, each pixel is associated to a ray that starts, say, from a camera outside the domain,
and assigns a binary pixel value that decides whether or not the given shape is intersected by that ray.
As already noted, given $\xi$, one can compute $Q_\xi$, which recovers as much information about the shape (or, rather, the equivalence class of shapes) that could have induced it as is possible.
This is therefore closely related to 
prior work on shape-from-silhouette and visual hulls, 
which has seen significant theoretical investigation in prior works. 

In particular, estimating 3D shape from visibility images alone (i.e., silhouettes) has been studied in computer vision for decades \cite{baumgart1974geometric} 
(see also \cite{cheung2005shape}), 
complemented by research into how humans overcome 
the inherent ambiguity of the problem
(e.g., \cite{marr1980visual,willatsl1992seeing}).
Given that an infinite set of shapes can give rise to an identical silhouette,
additional constraints are needed for shape inference, 
either in the form of purely geometric considerations \cite{richards1987inferring}, 
priors \cite{li2018single}, or simply more viewpoints. 
In general,
the basic idea is to use back-projection from the silhouettes into 3D space,
with each such constraint forming a 3D \textit{visual cone},
in which the shape must lie.
The volumetric intersection of such cones generates the \textit{visual hull} of the object, a point set that contains the shape of interest with increasing precision as the number of views increases.
Significant theoretical work has characterized the geometry of visual hulls,
particularly in the shape-from-silhouette context
(e.g., \cite{laurentini1994visual,laurentini1995far,laurentini1997many,laurentini1999computing,petitjean1998computational,trager2016consistency,kutulakos1997shape}).
Numerous techniques have been applied to visual hull computation
(e.g., \cite{lazebnik2001computing,lazebnik2007projective,landabaso2008shape,yi2007efficiently}),
as well as applications thereof
(e.g.,  %
\cite{matusik2000image,furukawa2009carved,franco2006visual}).
Generalizations of the shape-from-silhouette problem consider 
errors in the camera poses, inconsistencies or noise in the silhouettes themselves,
or issues related to non-convexity of shapes
(e.g., \cite{bottino2003introducing,haro2012shape,cheung2000real,landabaso2008shape,zheng19923d,zheng1994acquiring}).
Space carving \cite{kutulakos2000theory} 
generalizes shape-from-silhouette geometric constraints 
to utilize a notion of photo-consistency, 
producing a ``photo-hull'' that is more constrained than the visual hull;
this relates more closely to recent AbS approaches to multiview 3D scene reconstruction,
such as NeRFs. %

In contrast, our work focuses on potentially under-constrained 5D visibility fields,
which can be rendered into silhouettes from any continuously varying viewpoint, 
and the conditions under which such fields properly correspond to some underlying shape
(i.e., when $\xi$ is a shape representation, guaranteeing view consistency).
The most common shape-from-silhouette situation involves so-called \textit{external} visual hulls, which involve only viewpoints (i.e., cameras) outside the convex hull of the shape \cite{laurentini1994visual}.
In the context of this thesis, a visibility field $\xi$ considers the presence of a shape along any possible ray.
This enables $\xi$ 
to represent aspects of shape that are missed by external visual hulls,
such as concavities,
and is thus more similar to \textit{internal} visual hulls 
(e.g., see \cite{petitjean1998computational,laurentini1994visual}),
which assigns the ``viewing region'' to be anywhere outside the shape 
(not just its convex hull).

Nevertheless, questions still arise regarding visibility fields and their representational capacity.
One aspect of shapes ``missed'' by visibility fields is \textit{internal structure}:
shape points that are completely surrounded by other shape points cannot be preserved by the field.
In other words, given a shape $S$ with internal structure, 
any $S$-induced field will not be able to recover such structure.\footnote{
	This is why there is an equivalence class of shapes that induce a given visibility field 
	(i.e., $\mathfrak{S}[\xi]$).}
One can see this by noting that 
the directly lit points of $\xi$ 
(equivalently, the set of one-to-zero discontinuities)
will not include such internal structure.
These points, however, fully encode the field (as they induce it).
Combining a visibility field with a distance field, as we will do in the next section, enables a complete shape representation (i.e., including internal structure).

\subsection{View Consistency for Directed Distance Fields}

In this section, we investigate the \textit{combination} of a visibility field and a slightly modified form of simple DDF,
which define the full DDF discussed in the other sections of this chapter.
Similar to the prior two sections, 
we define view consistency through a notion of ``shape representation''; 
i.e., whether or not some shape (point set) exists that induces the DDF.
In this case, the visibility field will be used to control where 
(in $\Gamma$) 
the distance field needs to be strongly constrained. 
This is different from the simple DDF case,     
in that it alleviates constraints on $d$.
Given a simple DDF, 
one can imagine constructing a visibility field based on the presence or absence of a zero of the $d$-field along the ray (i.e., whether $q$ maps a ray outside of $\bo$).
However, this section considers the opposite: rather than deriving $\xi$ from $d$, we specify $\xi$ first, and use it to derive where and how $d$ must be constrained (reducing the demands on $d$).
As a result of this, 
we need additional conditions to ensure that the visibility field and depth field are compatible with one each other.

\begin{mydef}{Visible and Non-Visible Ray Sets}{pddf:full:vnvrs}
	Let $\xi$ be a visibility field.
	We define the sets of visible and non-visible rays, respectively, as
	\begin{align}
		\visrays[\xi]    &= \{ \tau \in \Gamma \mid \xi(\tau) = 1 \} \\
		\nonvisrays[\xi] &= \{ \tau \in \Gamma \mid \xi(\tau) = 0 \}, 
	\end{align}
	written via upper-continuity (Def.~\ref{df:pddf:vis:uppercont}).
	Further, $\visrays[\xi]$ is closed, 
	while $\nonvisrays[\xi]$ is open.
\end{mydef}

The closed or open nature of these sets
is due to the BOZ field definition, which requires %
$\xi^{-1}(0)$ to be open,
and so its complement 
$\xi^{-1}(1)\subseteq \Gamma = \bo\times \mathbb{S}^2$ is closed,
since $\Gamma$ is compact.

Define $\mathcal{A}_{d,\xi}(q)$
as the \textit{visible} infimum of $d$ at $q$:
$$
\mathcal{A}_{d,\xi}(q) = 
\lim_{\epsilon \downarrow 0} 
\inf_{ \substack{s\in (0,\epsilon)\\ 
		v\in\mathbb{S}^2 \,\mid\, \xi(q,v) = 1 } }
d(q - sv, v).
$$

\begin{mydef}{$\xi$-Coherent Simple DDFs}{pddf:full:xicoherent}	
	Given a visibility field $\xi$,
	an NNBC field $d$ is a $\xi$-coherent simple DDF
	iff it satisfies %
	\begin{enumerate}
		\item (IO$_{d,\xi}$)
		\textit{Isotropic opaqueness on visible rays.}
		\begin{align}
			\forall\; &
			(q,v)\in\visrays[\xi]: \nonumber \\ & 
			\left(\mathcal{A}_{d,\xi}(q) = 0\right)
			\,\implies\,
			\left(d(q,v) = 0\right). %
		\end{align}
		\item (BC$_{d,\xi}$)
		\textit{Positive boundary condition on visible rays.}
		\begin{equation}
			\big(
			(p,v)\in \visrays[\xi] \,\land\, 
			p\in \bo\setminus\boe
			\big)
			\,\implies\,
			d(p,v) > 0 .
		\end{equation}
		\item (DE$_{d,\xi}$)
		\textit{Directed Eikonal constraint on visible rays.}
		
		For any $\tau\in\visrays[\xi]$ such that
		$d(\tau) > 0$, we have $ \partial_s f_d(s\mid\tau)|_{s=0} = -1 $.
	\end{enumerate}	
\end{mydef}

In words, a $\xi$-coherent depth field $d$ satisfies the
isotropic opaqueness, boundary, and directed Eikonal constraints
whenever $\xi = 1$ (i.e., it is a simple DDF on $\visrays[\xi]$).
Outside of the visible rays $\visrays[\xi]$, $d$ is essentially unconstrained.

\begin{mydef}{Locally Visible Depth Zeroes}{pddf:full:lvdz}	
	Let $\xi$ be a visibility field and $d$ be $\xi$-coherent.
	Then the \textit{locally visible zeroes} are given by
	\begin{equation}
		\lvz = 
		\closure
		\left(\left\{
		\widetilde{q} \in Q_d  
		\; \middle\vert \;
		\exists\, v \in \mathbb{S}^2 
		\;\mathrm{s.t.}\;
		\xi(\widetilde{q}, v) = 1
		\right\}\right),
	\end{equation}
	where 
	$\closure$ is closure and
	we use upper-continuity %
	for $\xi$ (Def.~\ref{df:pddf:vis:uppercont}). %
\end{mydef}
This set includes any zero of $d$ 
(i.e., $q\in Q_d$; see Def.~\ref{df:pddf:simple:zeroes})
that is ``locally visible'' along some direction 
(meaning, 
$\lim_{\epsilon\downarrow 0}\sup_{||\delta|| < \epsilon} \xi(q + \delta,v) = 1$).
The locally visible zeroes ($\lvz$) will play an analogous role
to the zeroes $Q_d$ for simple DDFs
and the discontinuities $Q_\xi$ for visibility fields.

\begin{mydef}{Directed Distance Field (DDF)}{pddf:full:fullddf}	
	A BOZ field $\xi$ and an NNBC field $d$ form a 
	\textit{Directed Distance Field} $(\xi,d)$ iff:
	\begin{enumerate}
		\item 
		$\xi$ is a visibility field.
		\item 
		$d$ is a $\xi$-coherent simple DDF.
		\item 
		Locally visible depth zeroes are isotropically opaque with respect to $\xi$:\\
		$q \in \lvz$ $\implies$ 
		$ \tau \in\visrays[\xi] \;\forall\; \tau\in\mathcal{I}_q $,
		where 
		$\mathcal{I}_q\subset \Gamma$ is the set of rays that intersect $q$.
		\item 
		Every visible ray must hit a locally visible depth zero:\\
		$ \tau \in\visrays[\xi] \implies \bo_\tau \cap \lvz \ne \varnothing$,
		where $\bo_\tau$ is the set of points in $\bo$ 
		intersected by $r_\tau$.
	\end{enumerate}
\end{mydef}

The definition of a DDF can be intuited as combining a visibility field with a simple DDF that is constrained only along visible rays, 
along with two additional restrictions that force the two to be consistent.
The consistency constraints hinge on special zeroes of $d$ 
that are ``locally visible'' 
(i.e., at least one ray exists, 
along which $q\in Q_d$ is visible at an infinitesimal distance).
Such zeroes are (i) always visible and (ii) every visible ray must intersect at least one.
The next lemma clarifies the relation between the fundamental point-sets associated with these various fields: the zeroes of $d$, the visibility discontinuities, and the locally visible depth zeroes.

\begin{restatlem}{Fundamental Point Sets}{pddf:full:fundps}
	Let $(\xi,d)$ be a DDF. Then $Q_{\xi} \subseteq \lvz \subseteq Q_d$.
\end{restatlem}
\begin{proof}
	See Supp.~\S\ref{pddf:theory:fundps}.
\end{proof}

Note that locally visible zeroes (i.e., points in $\lvz$) require all intersecting rays to be visible rays, but they do \textit{not} require such points to be one-to-zero visibility discontinuities (i.e., $Q_\xi$). 
However, such discontinuities \textit{do} have to be 
zeroes of the depth field
(i.e., in $Q_d$).

Next, we connect the locally visible depth zeroes to a closely related set of points, the directly visible zeroes, where the relation is more non-local.
Later, this relation will be useful in determining the relationship between shape-induced field tuples (soon to be defined, in Def.~\ref{df:pddf:full:inducedfieldpair}) and DDFs.

\begin{mydef}{Directly Visible Depth Zeroes}{pddf:full:dvdz}	
	Let $(\xi,d)$ form a DDF.
	Then the set of \textit{directly visible depth zeroes} is given by
	\begin{equation}
		\dvdz = 
		\closure
		\left(\big\{
		\widetilde{q} \in\bo  
		\; \big\vert \;
		\exists\,  (p,v) \in \visrays[\xi] 
		\;\mathrm{s.t.}\;
		q(p,v) = \widetilde{q}
		\big\}\right),
	\end{equation}
	where $q(p,v)$ is the $q$-mapping of $d$.
	
\end{mydef}
In words, $\dvdz\subseteq \bo$ records positions where
at least one ray $\tau$ exists such that
(i) $\xi(\tau) = 1$ and
(ii) $\widetilde{q}$ is the \textit{first} zero (i.e., member of $Q_d$) along $r_\tau$.
Thus, equivalently, using the properties of $\xi$-coherence,
we may write
\begin{align}
	\dvdz = 
	\mathrm{Closure}\big(\big\{
	\widetilde{q} \in\bo  
	\; \big\vert \; &
	\exists\, \tau \in \visrays[\xi] 
	\;\mathrm{s.t.} 
	\nonumber \\ 
	&\widetilde{q}
	=
	p + \widetilde{s}(\tau\mid Q_d) v
	\big\}\big),
\end{align}
where $\widetilde{s}(\tau\mid S) = \min_{q \in S_\tau} || p - q ||$, $\tau = (p,v)$,
and $S_\tau$ is the set of points in $S\subseteq\bo$ that are intersected by $r_\tau$.
This latter form helps illuminate the connection to shape induction 
(e.g., recall Def.~\ref{df:pddf:simple:shapeinduction} 
for simple DDFs).
First, however, we show that not only are
the locally and directly visible depth zeroes 
equivalent, but 
that the outputs of $d$ and $q$ (the positional field of $d$)
are effectively controlled by
them.

\begin{restatlem}{DDFs Map to Locally Visible Depth Zeroes}{pddf:full:dmdvdz}
	Let $(\xi,d)$ be a DDF. Then $\lvz = \dvdz$.
	Further:
	\begin{align}
		\forall\;\tau = (p,v) \in\visrays[\xi]: &\;
		d(\tau) = \min_{ \widetilde{q} \in [\dvdz]_\tau } || \widetilde{q} - p ||
		\,\;\text{and}\;\, \nonumber \\
		q(\tau) \in \lvz,
	\end{align}
	where $[\dvdz]_\tau$ is the set of points in $\dvdz$ that intersect $r_\tau$ and $q$ is the $q$-mapping of $d$.
\end{restatlem}
\begin{proof}
	See Supp.~\S\ref{pddf:theory:dmdvdz}.
\end{proof}
This lemma will be used later, to link DDFs to shape representations.
Following the previous two sections, on simple DDFs and visibility fields, 
we next define an approach to ``field induction'': 
given a shape $S$, how can we construct a DDF, 
consisting of a ``field pair'' $(\xi,d)$,
that appropriately represents $S$?
We then show that inducing a field in this manner, versus having a DDF with constraints on local field behaviour 
(i.e., Def.~\ref{df:pddf:full:fullddf})
are equivalent.

\begin{mydef}{Induced Field Pair}{pddf:full:inducedfieldpair}	
	Consider a shape $S \subseteq \boe$.
	By Def.~\ref{df:pddf:vis:sibfs}, $S$ induces a unique visibility field, $\xi$.
	We also define an induced distance field $d$ to be any NNBC field that satisfies 
	\begin{equation}
		d(\tau) = \min_{q\in S_\tau} ||q - p||
		\;\forall\; 
		\tau\in\visrays[\xi],
	\end{equation}
	where $S_\tau$ is the set of points in $S$ that are intersected by $r_\tau$.
	
	Then, given a shape $S$, 
	we define any such pair $(\xi,d)$, 
	to be an \textit{induced field pair}.
\end{mydef}

Notice that this construct for $d$ is similar to inducement of simple DDFs; 
the main difference is that $d(\tau)$ is unconstrained when $\xi(\tau) = 0$ 
(i.e., $d$ no longer needs to predict distances outside the domain).
Further, the induced visibility $\xi(\tau)$ is one iff $\tau$ intersects $S$. 
Hence, for a visible ray $\tau\in\visrays[\xi]$, 
$d$ is constrained to always predict the distance to 
the closest $q\in S_\tau$.

Thus, similar to Theorems \ref{th:pddf:simple:sifsasdds} and \ref{th:pddf:vis:sibfavf}, 
when a shape $S$ induces a field pair $F$, 
we want to show that $F$ follows the field requirements of DDFs 
(Def.~\ref{df:pddf:full:fullddf}).

\begin{restatthm}{Induced Field Pairs are Directed Distance Fields}{pddf:full:ifpaddfs}
	Let $S$ be a shape and $F=(\xi, d)$ be an $S$-induced field pair.
	Then $F$ is a Directed Distance Field, which satisfies $\lvz = S$.	
\end{restatthm}
\begin{proof}
	See Supp.~\S\ref{pddf:theory:ifpaddfs}.
\end{proof}

Now that we know that any field pair generated from a shape is a DDF,
we next want to work towards showing the converse, 
that any DDF is a shape representation of some shape.
First, similar to simple DDFs, the presence of the visibility field
means we need to define a notion of equivalence between fields
(where values on non-visible rays can be ignored).

\begin{mydef}{Equivalence of Directed Distance Fields}{pddf:full:equivddf}	
	Consider two DDFs, $F_1 = (\xi_1,d_1)$ and $F_2 = (\xi_2,d_2)$.
	Then $F_1$ and $F_2$ are equivalent iff
	(i) $\xi_1(\tau) = \xi_2(\tau)\;\forall\; \tau\in\Gamma$
	and 
	(ii) $d_1(\tau) = d_2(\tau) \;\forall\; \tau\in\visrays[\xi_1]$.
\end{mydef}

This definition of field equality enables us to establish when a DDF is a shape representation.

\begin{mydef}{View Consistent Directed Distance Fields as Shape Representations}{pddf:full:shapereps}	
	A DDF $F$ is view consistent (VC) iff it is equivalent to a DDF that has been induced by a shape $S$.
	I.e., $F$ is VC iff there exists a shape $S$ such that 
	$F$ is an $S$-induced field pair.
	In such a case, 
	we say that \textit{$F$ is a shape representation for $S$}.
\end{mydef}

Finally, 
analogously to Theorems~\ref{th:pddf:simple:esdiasr} and \ref{th:pddf:vis:evfias},
we can assert that the field constraints of Def.~\ref{df:pddf:full:fullddf}
are sufficient to guarantee that a DDF represents some shape.

\begin{restatthm}{Every Directed Distance Field is a Shape Representation}{pddf:full:eddfiasr}
	Let $F=(\xi,d)$ be a DDF. %
	Then $F$ must be view consistent 
	(i.e., there exists a shape $S$ such that $F$ is an $S$-induced field pair).
	Further, $\lvz$ is a shape that induces $F$.
\end{restatthm}
\begin{proof}
	See Supp.~\S\ref{pddf:theory:eddfiasr}.
\end{proof}

This last theorem finally links \textit{full} DDFs to shape representations, by combining view-consistent visibility fields with depth fields that are constrained (in the same manner as simple DDFs) along visible rays (i.e., in $\visrays[\xi]$) and \textit{un}constrained along non-visible rays (i.e., in $\nonvisrays[\xi]$).
In other words, full DDFs are simple DDFs, 
but with looser requirements on $d$ when $\xi = 0$ (controlled instead by a visibility field, $\xi$).
The concept of shape representation can therefore be equivalently derived in two ways: (i) by starting from a point set and inducing a field pair from it, or (ii) by starting from a visibility field and a distance field that are ``compatible'' with one another, and satisfying the basic requirements on the fields (BC, IO, and DE).

The latter case, in particular, is important, as it suggests that
one need only ensure certain local constraints,
in order to obtain a view consistent 5D field.
Specifically, for the boundary conditions (BC), 
one should demand that, 
near the boundary of the domain,
$d$ does not predict surface points (i.e., zeroes) and
the visibility field returns zero whenever it ``looks outward'' from the domain (i.e., the shape cannot be visible outside the domain).
For the directed Eikonal (DE) constraints, one requires $\partial_s f_d = -1$ and $f_\xi$ non-increasing, along visible and all rays, respectively.
Isotropic opaqueness (IO)
asks that (visible) zeroes of $d$ are zero from all directions and, similarly, that one-to-zero ``flips'' in $\xi$ are visible from all directions.
Finally,
two demands are placed to ensure consistency between the visibility and depth fields, relying on the set of locally visible zeroes 
(i.e., depth zeroes that are visible at infinitesimally close points along some ray):
(a) any ray that intersects such a zero must be visible, 
and 
(b) any visible ray must hit such a zero.
Together, the previous theorems ensure that a field satisfying these conditions must be a shape representation (and vice-versa).

This is reminiscent of solving the Eikonal equation to obtain a distance field,
which is loosely followed by some modern distance-based isosurface \acrshort{isf} training methods.
Notice, for instance, that BC and IO act like boundary conditions, 
while DE forms the constraint that needs to be solved across space.
We expect that improvements in techniques 
for fitting higher-dimensional \acrshort{inr}s with differential constraints
(in this case, equivalent to solving a 5D partial differential equation)
will make neural implementation of (P)DDFs more viable.

In summary, these theorems give requirements for generic learned fields 
(i.e., implemented as neural networks $(\xi_\theta,d_\theta)$ with weights $\theta$)
that ensure view-consistency if satisfied.
By careful design of the field architectures and losses, 
to ensure fitting to the data and satisfaction of these conditions, 
we should converge to the desired shape,
given enough network capacity and sufficient supervision.

\section{Proofs for View Consistency of DDFs}
\label{suppmat:proofs}

This section provides the proofs for the Lemmas, Theorems, and Corollaries of 
\S\ref{sec:theory} in the main paper and Supp.~\S\ref{suppmat:theory}.
It also includes an alternative definition of field induction for simple DDFs
(\S\ref{pddf:theory:altfieldind})
and an example of why the positional zeroes need to be defined as converging over all directions, rather than just a single one (\S\ref{pddf:theory:convergenceexample}).
Note that we denote a shape $S$ to be a compact point set 
in the domain $\boe$.

\subsection{Proofs for Simple DDFs}

\subsubsection{Alternative Formulation of Field Induction}
\label{pddf:theory:altfieldind}

\begin{mydef}{Field Induction via Boundary-Subsumed Shapes}{pddf:simple:altboundsubsume}
	Let $S$ be a shape (closed point set) such that $S\subseteq\boe$.
	Define $ S^\partial = S \cup \partial\bo$.
	Let $S_\tau^\partial$ be the set of points in $ S^\partial $ that are intersected by $r_\tau$ (i.e., for any $q\in S_\tau^\partial$, there exists $t\geq 0$ such that $r_\tau(t) = q$).	
	Then $ S^\partial $ \textit{induces} a field  $d$ in the boundary-subsumed sense 
	iff
	$d(\tau) = \min_{q \in S_\tau^\partial} ||q - p||$,
	for any $\tau\in\Gamma$.
\end{mydef}

Note that $S$ is closed by definition, while $S_\tau^\partial$ must be closed and non-empty (because we include the boundary in the definition of $S_\tau^\partial$, there must be at least one intersection); hence, the minimum must exist.   
In this case, the induced field is unique and equality may be used instead of equivalence, for simple DDFs.
Finally, notice that saying $S$ induces $d$ in the boundary-subsumed sense is equivalent to saying $S^\partial = S\cap \partial\bo$ induces $d$ (in the non-boundary-subsumed sense).
However, in the boundary-subsumed case, $S$ has a \textit{unique} induced field, for a fixed $\bo$.
On one hand, this simplifies the notion of equivalence between simple \acrshort{ddf}s.
However, in the main thesis, we utilize the more complex notion of equivalence, as it 
(i) keeps the shape separate from the domain boundary and
(ii) is more analogous to the scenario we encounter for full \acrshort{ddf}s.

\subsubsection{Properties of Positional Zeroes}\hfill
\label{pddf:theory:propposzeroes}
\myrecall{pddf:simple:closedinboe}
\begin{proof}
	By BC$_d$, $d(p,v) > 0$ for all $p \in \bo\setminus\boe,\,v\in\mathbb{S}^2$,
	so there cannot be any zeros of the field outside of $\boe$.
	
	Next we show that the set is closed.
	Since $d$ is a simple DDF, it satisfies IO$_d$, BC$_d$, and DE$_d$.
	
	Suppose that $Q_d$ is not closed.
	Then there must exist a $p$ s.t.\ there is a sequence $(p_k)_{k=1}^\infty$ 
	that converges to $p$, where every 
	$p_k \in Q_d$, but $p\notin Q_d$ 
	(i.e., $p$ is the limit of points $p_k$, all of which are in $Q_d$).
	Since any $p_k\in Q_d$, by IO$_d$, we have that $d(p_k,v) = 0 \;\forall\; v\in\mathbb{S}^2$.
	
	Let $\varepsilon > 0$.
	Choose $p_k$ s.t.\ $|| p_k - p || < \varepsilon $
	(such a $p_k$ must exist as the sequence is converging to $p$)
	and set $v_k := \mathrm{Unit}(p_k - p) \in\mathbb{S}^2$ to be the direction vector pointing from $p$ back to $p_k$.
	By IO$_d$, $d(p_k,v_k) = 0$ and so, by DE$_d$, we must have that
	$ d(p, v_k) \leq || p_k - p || < \varepsilon $.
	But since $\varepsilon$ is arbitrary, we must have that 
	\begin{equation}
		\lim_{\epsilon \downarrow 0} 
		\inf_{ \substack{s\in (0,\epsilon)\\ v\in\mathbb{S}^2 } }
		d(p - sv, v) = 0,
	\end{equation}
	which implies $p\in Q_d$.
	
	This is a contradiction; hence, $Q_d$ must be closed.
	
\end{proof}

\subsubsection{Shape-Induced Fields Are Simple DDFs}
\label{pddf:theory:sifasd}
\myrecall{pddf:simple:sifsasdds}
\begin{proof}
	We need to show that $d$, induced by $S$, 
	satisfies the three constraints of a simple \acrshort{ddf}.
	\begin{itemize}
		\item 
		Let $d$ be induced by $S$. 
		Let $\tau = (p,v)\in \Gamma$ be such that $S_\tau \ne \varnothing$, and 
		$\tau_e = (p_e, v_e) \in\Gamma$ be such that $ S_{\tau_e} = \varnothing $. 
		\item 
		Since $ S \subseteq \boe $ is compact, $d(\tau) = \min_{q\in S_\tau} ||q - p|| \geq 0$ is both non-negative and bounded. 
		By definition, $d(\tau_e) > 0$ is also non-negative and bounded.
		Finally, for any $\tau$ (with $S_\tau \ne \varnothing$), 
		$ d(\tau) = \min_{q\in S_\tau} ||q - p|| $ is clearly $C^1$ along the ray, with potential discontinuities at zeroes of $d$; for any $\tau_e$, $d$ is defined to be $C^1$ along the ray as well. %
		
		\item 
		(IO$_d$) %
		First, note that there can be no zeros of $d$ except on rays that intersect $S$, since
		a ray with no intersections (e.g., $\tau_e$) is defined to have positive $d$;
		further, those zeroes must be \emph{at} the intersection with $S$, since, by definition, at all other points, $d$ measures the distance to the closest of those points along the ray. 
		So we need only check points in $S$.
		
		So, consider any $q_0 \in S$, and choose an arbitrary $v\in\mathbb{S}^2$.
		Consider the oriented point $ \tau(s) := (q_0 - sv, v) $,
		so that $S_{\tau(s)}$ is the set of points in $S$
		intersected by the ray $r_{\tau(s)}$,
		which is a function of $s$.
		Looking at $d$, as we move along $v$ towards $q_0$, shows us that
		\begin{align}
			&\lim_{\epsilon \downarrow 0} \inf_{s\in (0,\epsilon)} d(q_0-sv,v) 
			\\ &= \lim_{\epsilon \downarrow 0} \inf_{s\in (0,\epsilon)} \min_{q\in S_{\tau(s)}} 
			|| q - (q_0 - sv) || 
			\\ &\leq \lim_{\epsilon \downarrow 0} \inf_{s\in (0,\epsilon)} ||q_0 - q_0 + sv|| 
			= 0.
		\end{align}
		Since $d \geq 0$ as well, we must have that $ \lim_{\epsilon \downarrow 0} \inf_{s\in (0,\epsilon)} d(q_0-sv,v) = 0 $. 
		Since this holds for arbitrary $v$, IO$_d$ also holds.

		\item 
		(BC$_d$) %
		By definition, $S$ is a subset of $\boe$.
		Hence, the induced field $d$ cannot have any zeroes in $\bo\setminus\boe$.
		Since $d$ is non-negative, we must have $d(p,v) > 0$ for any $v\in\mathbb{S}^2$, $ p\in\bo\setminus\boe $.
		Thus, BC$_d$ holds.
		
		\item 
		(DE$_d$) %
		Consider any intersecting oriented point, 
		$\tau = (p,v) \in \Gamma$, 
		s.t.\ $d(p,v) = \min_{q\in S_\tau} ||q - p|| = ||q_0 - p|| > 0$
		(i.e., $ d(\tau)v + p = q_0 \in S$).
		First, note that 
		$ f_d(s\mid \tau) = ||q_0 - (p+sv)|| = || d(\tau)v + p - (p+sv) || = || [d(\tau) - s]v || = d(\tau) - s $
		for $s\in [0,d(\tau)]$.
		Then, clearly, $\partial_s f_d(s\mid\tau) = \partial_s (d(\tau) - s) = -1 $, as required.
		For a non-intersecting ray, by construction, 
		we have $d(p,v) > 0$ and $\partial_s f_d(s\mid p,v) = -1$.
		Hence, $d$ satisfies DE as well.
		
		\item 
		Finally, as noted above (see the section on IO$_d$), we know that the zeros of an induced field inside $\bo$ can only occur at intersections with $S$. 
		Hence, any $q\in \bo$ s.t.\  $q\in Q_d$ must also satisfy $q \in S$.
		Since $S\subseteq\boe$, this also restricts $Q_d$ to points in $\boe$ (recall, by the lemma above, that $Q_d\subseteq\boe$ for any simple DDF). 
		There may be additional zeros of $d$ outside the domain $\bo$, 
		which we can ignore.
		Hence, we have $ S = Q_d $.
		
		\item 
		Altogether, we have shown that any closed point set in $\boe$ 
		can induce or generate a simple DDF.
		Further, the zeros of the induced field (in the domain) coincide with the $S$ itself.
		
	\end{itemize}	
\end{proof}

\subsubsection{Every Simple DDF is a Shape Representation}
\label{pddf:theory:esdiasr}
\myrecall{pddf:simple:esdiasr}
\begin{proof}
	Let $d$ be a simple DDF on $\bo$.
	We show that its positional zeroes, $Q_d$, in $\boe$ form the inducing shape. 
	Recall that any inducing shape must be in $\boe$ and compact
	(and note that $\boe$ itself is compact).
	\begin{itemize}
		\item \textbf{$Q_d$ is a Shape}
		
		By Lemma~\ref{lma:pddf:simple:closedinboe}, 
		$Q_d\subseteq \boe$ and $Q_d$ is closed; 
		hence, $Q_d$ is a compact point set (i.e., shape).
		Thus, we can simply set $S = Q_d$.
		
		Now, we also already know that any $S$-induced field, say $d_s$, must be a simple DDF. 
		We simply need to show that $d_s$ (which was induced by $Q_d$) is equivalent to $d$.
		
		\item
		\textbf{Empty rays}
		
		Suppose $\mathcal{E}[d_s] \ne \mathcal{E}[d]$.
		Then either (a) $\exists\; \tau_a\in \mathcal{E}[d_s]\setminus\mathcal{E}[d]$
		or (b) $\exists\; \tau_b\in \mathcal{E}[d]\setminus\mathcal{E}[d_s]$.
		
		Case (a): $ \tau_a\notin\mathcal{E}[d] \;\implies\; \exists\; q = p + d(\tau_a)v \in \boe $.
		(Note that $q\notin\bo\setminus\boe$ due to BC$_d$.) 
		But by DE$_d$, $d(q,v) = 0\;\implies\; q\in Q_d$ and thus $q\in S$, 
		contradicting $\tau_a \in \mathcal{E}[d_s]$.
		(I.e., $\tau_a$ cannot be in $\mathcal{E}[d_s]$ if it is not in $\mathcal{E}[d]$.)
		
		Case (b): $ \tau_b\notin\mathcal{E}[d_s] \;\implies\; \exists\; q = p + d(\tau_b)v \in \boe $.
		In particular, $\tau_b$ must intersect some $q\in S$ along $r_{\tau_b}$ 
		(where $S\subseteq \boe$).
		But this implies $q = q(\tau_b) \in \boe$, 
		which contradicts $\tau_b\in \mathcal{E}[d]$.
		(I.e., $\tau_b$ cannot be in $\mathcal{E}[d]$ if it is not in $\mathcal{E}[d_s]$.)

		Thus, both cases lead to a contradiction.  
		Hence, the empty sets coincide: $ \mathcal{E}[d_s] = \mathcal{E}[d] $.
		
		\item
		\textbf{Intersecting rays}
		
		Consider an intersecting ray $\tau=(p,v)\in\Gamma$ (i.e., non-empty ray).
		For the shape-induced field, we have 
		$ d_s(p,v) = \min_{q\in S_\tau} || q - p || = || q_s - p || $,
		where $q_s$ is the closest intersection point along $r_\tau$.
		We therefore know $q_s\in S = Q_d$ and there cannot be any other $q\in Q_d$
		(i.e., zeros of $d$) between $p$ and $q_s$ along $\tau$.
		Thus, by DE$_d$, along $\tau$, $d$ must decrease at a unit rate, 
		until it reaches the zero at $q_s$
		(or, equivalently, $f_d$ must increase at a unit rate, going from $q_s$ to $p$).
		Hence, $d(p,v) = ||q_s - p || = d_s(p,v)$, as required.
		
		This shows that $d$ and $d_s$ are equivalent.
		In other words, given a Simple DDF $d$, 
		we are guaranteed that $d$ 
		is a view-consistent shape representation of its zero-set, $Q_d$.
	\end{itemize}
\end{proof}

\subsection{Proofs for Visibility Fields}

\subsubsection{Properties of Positional Discontinuities of Visibility Fields}\hfill
\label{pddf:theory:posdiscont}
\myrecall{pddf:vis:posdiscont}
\begin{proof}
	Note that the closure operator in Def.~\ref{df:pddf:vis:posdiscont} ensures the set $Q_\xi$ is closed by definition.
	Also, by DE$_\xi$, there cannot ever be discontinuities from zero to one along a ray.
	Further, since $\xi$ satisfies BC$_\xi$, note that $Q_\xi \cap (\bo\setminus\boe) = \varnothing$,
	because (by IO$_\xi$) any ray $\tau$ that intersects a $q\in Q_\xi$ 
	(i.e., satisfies $\tau\in \mathcal{I}_{q}$)
	must have $\xi(\tau) = 1$ \textit{including outward-facing rays},
	which would violate BC$_\xi$ if they were present.
	Hence, $Q_\xi \subseteq \boe$.
	This shows that $Q_\xi$ is closed and bounded, and hence is a shape.
\end{proof}

\subsubsection{Shape-Induced Binary Fields Are Visibility Fields}
\label{pddf:theory:sibfavf}
\myrecall{pddf:vis:sibfavf}
\begin{proof}
	We need to show that a shape-induced binary field satisfies
	the constraints of a visibility field.
	\begin{itemize}
		\item 
		(BC$_\xi$) 
		Since $S\subseteq \boe$, any outward-pointing ray $\tau\in\bo\setminus\boe$ 
		in the $\epsilon$-boundary
		cannot intersect a $q\in S$.
		Hence, such rays satisfy $\xi(\tau)=0$ as required.
		
		\item 
		(DE$_\xi$)
		Consider a ray $ \tau_E \notin \mathcal{I}_{S} $ that does not intersect $S$.
		Then $f_\xi(s\mid\tau_E) = 0\;\forall\;s\geq 0$, 
		meaning $\xi$ is non-increasing along $\tau_E$
		(since no point in $S$ can appear along $r_{\tau_E}$).
		
		Next, consider a ray $\tau=(p,v)\in \mathcal{I}_{S} $ that does intersect $q\in S$.
		We must have $f_\xi(s\mid\tau) = 1$ until at least $s_q = ||p - q||$.
		After $s_q$, we either must have $f_\xi(s\mid\tau) = 0$ iff there are no more points in $S$ along $r_\tau$ or $f_\xi(s\mid\tau) = 1$ iff additional intersection points exist.
		In the former case, by the same reason as for $\tau_E$ above,
		$f_\xi(s'\mid\tau) = 0$ for all $s\geq s_q$, 
		because of the assumption that no further points exist.
		
		\item 
		($Q_\xi \subseteq S$)
		Consider a point $q$ with a one-to-zero discontinuity
		(i.e., at $q$, there exists a $v$ such that 
		$\lim_{\epsilon\downarrow 0} \sup_{s\in(0,\epsilon)}
		\xi(q - sv,v) = 1
		\;\land\;
		\lim_{\epsilon\downarrow 0} \inf_{s\in(0,\epsilon)}
		\xi(q + sv,v) = 0$).
		This means that there exists a ray that, when approaching $q$,
		has a visibility of one,
		but has a visibility of zero upon exiting it.
		By the upper continuity convention,
		we have %
		$\xi(q,v) = 1$ but $\xi(q+sv,v) = 0$ for any $s > 0$
		(recalling also that $\xi$ cannot increase along a ray, 
		and hence it must stay at zero thereafter).
		Recall that $\xi(\tau) = 1$ iff $\tau$ intersects $q_s \in S$,
		but also $\xi(\tau_E) = 0$ iff $\tau_E$ does \textit{not} intersect $S$.
		This implies we must have $q \in S$.
		Thus, any positional discontinuity of an shape-induced field must be an element of the shape itself.
		
		\item 
		(IO$_\xi$)
		By the definition of $Q_\xi$, IO$_\xi$ can be written as
		$q\in Q_\xi \implies \xi(\tau) = 1 \;\forall\; \tau\in \mathcal{I}_q $,
		where $\mathcal{I}_q $ is the set of rays that intersect $q$.
		Now, suppose that $q\in Q_\xi$. But, since $Q_\xi \subseteq S$, then $q\in S$.
		Hence, by the definition of inducement, any ray $\tau$ containing $q$ must satisfy $\xi(\tau) = 1$.
		
		\item 
		Together, these imply that any shape-induced field is a visibility field.
		
	\end{itemize}
\end{proof}

\subsubsection{Every Visibility Field is a Shape \Xirepname}
\label{pddf:theory:evfias}
\myrecall{pddf:vis:evfias}
\begin{proof}
	Our strategy is to show that $Q_\xi$ induces $\xi$ directly.
	\begin{itemize}
		\item 
		
		We have already shown that $Q_\xi$ is a shape in Lemma~\ref{lma:pddf:vis:posdiscont}.
		
		Next, we simply need to show that $Q_\xi$ induces $\xi$.
		By definition, $\xi$ is a BOZ field that satisfies IO$_\xi$, BC$_\xi$, and DE$_\xi$.
		For any $\tau\in\Gamma$, we need to show that $\xi(\tau) = 1 \iff \tau\in \mathcal{I}_{Q_\xi} $.
		
		\begin{enumerate}
			\item 
			Suppose $\xi(\tau) = 1$, where $(p,v)=\tau$. 
			Consider $f_\xi(s' \mid \tau)$, which is the visibility along the ray $r_\tau$.
			By BC$_\xi$, there must exist an $s$ such that $f(s\mid\tau) = 0$ 
			(since any outward ray in $\bo\setminus\boe$ must have a visibility of zero).
			This means there must exist $s_q \in [0,s]$ such that $q = p + s_q v$ is a positional discontinuity 
			(i.e., $q\in Q_\xi$). By DE$_\xi$, there can only be one such one-to-zero discontinuity.
			But this means that $r_\tau$ intersects $q\in Q_\xi$, 
			so $\tau \in \mathcal{I}_{Q_\xi} $.
			
			Notice also that, if $\xi(\tau) = 0$, along the ray $r_\tau$, 
			we can never have a discontinuity from zero to one, by DE$_\xi$. 
			Hence, $\xi(\tau) = 0$ implies $r_\tau$ cannot intersect any point in $Q_\tau$ as well.
			
			\item 
			Suppose $ \tau = (p,v)\in\mathcal{I}_{Q_\xi} $.
			This means that there exists $q\in Q_\xi$ along the ray $r_\tau$
			(i.e., there exists $s$ such that $q = p + s v \in Q_\xi$).
			By the definition of $Q_\xi$,
			at $q$, there must exist at least one $v$ such that that a one-to-zero 
			discontinuity in $\xi$ occurs along $v$ at $q$.
			Then, the implication of IO$_\xi$ is satisfied, so 
			we must have that $\xi(\tau) = 1$ 
			(since $ \tau \in\mathcal{I}_{q} $).
			
		\end{enumerate}
		Together, these show that $\xi(\tau) = 1 \iff \tau\in\mathcal{I}_{Q_\xi} $,
		meaning $Q_\xi$ induces $\xi$, as required.
		
	\end{itemize}
	
\end{proof}

\subsubsection{Minimal Characterization of Visibility Fields}
\label{pddf:theory:minimalcharvis}
\myrecall{pddf:vis:minimalcharvis}
\begin{proof}
	We have already shown that $Q_\xi$ induces $\xi$ (i.e., $Q_\xi \in \mathfrak{S}[\xi]$) in Theorem~\ref{th:pddf:vis:evfias}.
	However, we have also shown that \textit{any} $S$-induced field $\xi$ has $Q_\xi \subseteq S$ in Theorem~\ref{th:pddf:vis:sibfavf}.
	This means $Q_\xi$ is at \textit{at most} as large as the smallest member of $\mathfrak{S}[\xi]$.
	Hence, it is the smallest such inducing point set.

\end{proof}

\subsubsection{Directly Lit Points Minimally Characterize Visibility Fields}\hfill
\label{pddf:theory:dlpsareqxi}
\myrecall{pddf:vis:dlpsareqxi}
\begin{proof}
	Given Cor.~\ref{mc:pddf:vis:minimalcharvis}, 
	we simply need to show that $Q_\xi = \mathfrak{D}_\ell(S)$.
	\begin{itemize}
		\item 
		Let $S$ be a shape and $\xi$ its induced field.
		By definition, both $Q_\xi$ and $ \mathfrak{D}_\ell(S) $ are closed sets.
		
		(i) 
		Consider $q\in Q_\xi$ (and since $Q_\xi \subseteq S$, we also have $q\in S$). 
		This means there exists a $v$ such that $\xi(p,v)=1$ (by upper-continuity) but
		$\lim_{\epsilon\downarrow 0} \inf_{s\in(0,\epsilon)} \xi(q+sv,v) = 0$.
		Since by this latter zero value, 
		by DE$_\xi$, there can be no $\widetilde{q}\in Q_\xi$ between $q$ and the boundary along $v$.
		There also cannot be any 
		$s > 0$ such that $q + sv \in S$
		(else $\xi$ would be one).
		Let $b\in\partial\bo$ be the intersection point of $r_{(q,v)}$ with the boundary.
		This means that the first intersection with $S$, from the ray $r_{(b,-v)}$, is $q$.
		Hence $q$ is directly lit; i.e., $q\in Q_\xi\implies q\in \mathfrak{D}_\ell(S)$.
		
		(ii) 
		Consider $q\in \mathfrak{D}_\ell(S)$, with $b\in \partial\bo$ being a boundary point from which $q$ can be observed. Let $v := \mathrm{Unit}(q-b)$ point from $b$ to $q$. 
		We make the same argument as in (i), but effectively in reverse.
		By the definition of a DLP, $q\in S$ 
		and there can be no other $q'\in S$ between $q$ and $b$.
		Hence, $ \xi(q,-v) = 1 $ (notated via upper-continuity), 
		but $\lim_{\epsilon\downarrow 0} \inf_{s\in(0,\epsilon)} \xi(q-sv,-v) = 0$.
		Thus, $q\in Q_\xi$ is a positional discontinuity;
		i.e., $ q\in \mathfrak{D}_\ell(S) \implies q\in Q_\xi $.
		
		Together, these show that $\mathfrak{D}_\ell(S) = Q_\xi$.
		Since $Q_\xi$ is a minimal inducer, so too is $\mathfrak{D}_\ell(S)$.
		
	\end{itemize}
\end{proof}

\subsection{Proofs for Directed Distance Fields (DDFs)}

\subsubsection{Fundamental Point Sets}\hfill
\label{pddf:theory:fundps}
\myrecall{pddf:full:fundps}
\begin{proof}
	The second relation is true by definition, so we focus on the first one.
	Consider a location with a visibility discontinuity,
	$q \in Q_\xi$.
	By the definition of $Q_\xi$ (and the upper continuity convention), 
	we may write $\xi(q,v) = 1$ for some $v\in\mathbb{S}^2$.
	Consider the visible ray $(q,v) = \tau \in\visrays[\xi]$, 
	which starts at $q$ and goes along $v$ (so that $\xi(\tau) = 1$).
	By property 4 of the DDF (Def.~\ref{df:pddf:full:fullddf}), 
	there must be a $\widetilde{q}\in \lvz$ 
	along $r_\tau$ (i.e., $\exists s \geq 0$ s.t.\ 
	$r_\tau(s) = \widetilde{q}$).
	But, since $q\in Q_\xi$, we must have that
	$ \xi(q + sv,v) = 0 $ for any $s > 0$.
	That is, there is \textit{no} locally visible zero along $r_\tau(s)$,
	for any $s > 0$.
	Hence, we must have a zero at $s=0$, meaning $q \in \lvz$;
	i.e., any $q\in Q_\xi$ also satisfies $q\in \lvz$,
	as required.
\end{proof}

\subsubsection{DDFs Map to Locally Visible Depth Zeroes}
\label{pddf:theory:dmdvdz}
\myrecall{pddf:full:dmdvdz}
\begin{proof}
	First, we show that locally visible and directly visible depth zeroes are equivalent; i.e., 
	$\lvz = \dvdz$.
	The rest of the lemma follows straightforwardly.
	\begin{itemize}
		\item	
		Consider $q\in \lvz$.
		Then, $d(q,v) = 0$ and $\xi(q,v) = 1$.
		This means  $\tau = (q,v) \in\visrays[\xi]$ and
		$q(q,v) = q + 0 v= q$, which implies $q\in \dvdz$.
		
		Consider $q\in \dvdz$.
		Then there must exist $\tau=(p,v)\in\visrays[\xi]$ such that
		$q = p + d(p,v)v$ 
		(note that this relation ensures $d(\tau) = ||q - p||$).
		Along $\tau$, by DE$_{d,\xi}$, we have that $f_d(s\mid\tau)$ must be linearly decreasing until $s = ||q - p||$, at which point  
		$d(p+sv,v) = d(q,v) = 0$.
		We therefore only need to show that $\xi(q,v) = 1$.
		We know that $\xi(p,v) = 1$.
		However, we have already shown that $Q_\xi\subseteq \lvz$ in 
		Lemma~\ref{lma:pddf:full:fundps}.
		Thus, we can only have a one-to-zero discontinuity \textit{at a zero of $d$}.
		(We cannot change from zero to one because $\xi$ satisfies DE$_\xi$).
		Thus, there are only two options:
		(i) there is a discontinuity in $\xi$ at $q$, but this exactly means $q\in \lvz$ (and $\xi(q,v) = 1$ by upper continuity), or
		(ii) there is no discontinuity between $p$ and $q$, in which case $\xi = 1$ throughout the ray (i.e., $f_\xi(s\mid(p,v)) = 1\;\forall\;s\in[0,d(p,v)]$.
		In either case, $\xi(q,v) = 1$ as required.
		
		Hence, $\lvz = \dvdz$.
		
		\item 
		Consider a visible ray $\tau=(p,v)\in\visrays[\xi]$.
		We want to show that $d$ computes the distance to the closest member of $\lvz$ (equivalently, $\dvdz$).
		Since $(\xi,d)$ is a DDF and $\tau\in\visrays[\xi]$, 
		there must be a closest zero, $q \in \lvz$, along $\tau$ 
		(by requirement four of Def.~\ref{df:pddf:full:fullddf}).
		By the same logic as the previous part of the proof, 
		we must have $\xi = 1$ along the ray until at least $q$
		(i.e., a one-to-zero discontinuity cannot occur until $q$, 
		at the closest).
		By definition, $d(q,v) = 0$.
		Thus, along the ray (where $\xi=1$), we must have that DE$_{d,\xi}$ holds, 
		which ensures that $d(p,v) = ||q - p||$ at $p$.
		Thus, $d$ computes the distance to the closest member of $\lvz = \dvdz$, as required.
		
		\item 
		
		Since $d(p,v) = \min_{q\in[\lvz]_\tau} ||p-q||$, where
		$\tau=(p,v)\in\visrays[\xi]$, we have that
		$q(\tau) = p + d(p,v)v = p + (\min_{q\in[\lvz]_\tau} ||p - q||) v = p + ||q^* - p|| v = q^* \in \lvz $, where $q^* = \arg\min_{q\in[\lvz]_\tau} ||p - q|| $ and using the fact that $q^*$ lies on $r_\tau$.
		
	\end{itemize}
\end{proof}

\subsubsection{Induced Field Pairs are Directed Distance Fields}
\label{pddf:theory:ifpaddfs}
\myrecall{pddf:full:ifpaddfs}
\begin{proof}
	We need to show that $F$ satisfies the properties of a DDF.
	\begin{enumerate}
		\item 
		$\xi$ is necessarily a visibility field, because, as shown in Theorem~\ref{th:pddf:vis:sibfavf}, any shape-induced binary field is a visibility field.
		\item 
		The induced $d$ is defined in the same way as a simple DDF, except that it is unconstrained on non-visible rays.
		On visible rays, the same properties therefore apply (recalling that we have previously shown that shape-induced depth fields are simple DDFs; see Theorem \ref{th:pddf:simple:sifsasdds}).
		However, restricting IO$_d$, BC$_d$, and DE$_d$ to $\visrays[\xi]$ is exactly the definition of $\xi$-coherence, as required.
		\item 
		Next, we show that $\lvz = S$.
		
		(i) %
		Let $q\in \lvz$ be a locally visible depth zero.
		We know $\xi(q,v)=1$ and $d(q,v) = 0$, for some $v\in\mathbb{S}^2$.
		But for a visible ray $(q,v) = \tau\in\visrays[\xi]$, 
		the only way $f_d(s\mid\tau)$ can approach zero
		is if $q\in S$, since $d$ is induced and thus constrained along $\tau$.
		Hence, $q\in \lvz$ implies $q\in S$.
		
		(ii) %
		Consider $q\in S$.
		By the definition of induced field pair, 
		$\xi(q,v) = 1$ and $d(q,v) = 0$, 
		for any $v\in\mathbb{S}^2$.
		This immediately implies $q\in \lvz$.
		
		Together, these show that $\lvz = S$ for induced field pairs.
		
		\item 
		Next, let $q\in \lvz$ and 
		consider $\mathcal{I}_q$, the set of rays that intersect $q$.
		We need to show that every $\tau \in \mathcal{I}_q$ is visible 
		(i.e., $\tau\in\visrays[\xi]$).
		But $q\in S$ as well, so (since $\xi$ is induced by $S$), 
		we must have that $\xi(p,v) = 1$ 
		for any $(p,v)\in \mathcal{I}_q$.
		
		\item 
		Finally, consider $\tau \in\visrays[\xi]$.
		But $\xi$ is induced by $S$, so there must exist some 
		$q\in S$ such that $\tau \in \mathcal{I}_q$.
		But since $S = \lvz$, $q\in \lvz$ as well.
		This implies
		$\bo_\tau \cap \lvz \ne \varnothing$, 
		as required.
	\end{enumerate}
\end{proof}

\subsubsection{Every Directed Distance Field is a Shape Representation}\hfill
\label{pddf:theory:eddfiasr}
\myrecall{pddf:full:eddfiasr}
\begin{proof}
	Our strategy is to directly show that $\lvz$ is a shape that induces $F$.
	\begin{itemize}
		\item 
		We first note that $\lvz$ is a shape.
		Since $\lvz \subseteq Q_d$ and $Q_d\subseteq \boe$ is bounded, so too is $\lvz$.
		The closure operator ensures that $\lvz$ is a closed set by definition.
		Hence, $\lvz$ is a compact point set and therefore a shape.
		
		\item 
		We first show that $\xi$ is induced by $\lvz$.
		We need to show $\xi(\tau) = 1 \iff \tau\in\mathcal{I}_{\lvz}$,
		where $\mathcal{I}_{\lvz} \subseteq \Gamma$ is the set of rays starting within $\bo$ that intersect $\lvz$.
		
		Let $\tau\in\visrays[\xi]$.
		Since $(\xi,d)$ is a DDF, we have that 
		$r_\tau$ must intersect $\lvz$, by
		requirement four of Def.~\ref{df:pddf:full:fullddf}.
		
		In the other direction, suppose 
		$\tau\in\mathcal{I}_{\lvz} $.
		But by requirement three (of Def.~\ref{df:pddf:full:fullddf}), every locally visible depth zero is isotropically opaque with respect to $\xi$. Hence, $\xi(\tau) = 1$, as required.
		
		Hence $\xi$ is induced by $\lvz$, and uniquely so. 
		This is the first element of the field pair.
		
		\item 
		Next, we need to show that $d$ is induced by $\lvz$ on visible rays; 
		i.e., $d(\tau) = \min_{q\in [\lvz]_\tau} || q - p ||$ for any $\tau\in\visrays[\xi]$.
		However, 
		using the directly visible depth zeroes,
		we have already shown this in Lemma~\ref{lma:pddf:full:dmdvdz}. 
		Since $\mathcal{D}_{d,\xi} = \lvz$,
		we have that
		$d(\tau) = \min_{q\in [\mathcal{D}_{d,\xi}]_\tau} || q - p ||
		= \min_{q\in [\lvz]_\tau} || q - p || $.
		This means that any $\widetilde{d}$ induced by $\lvz$ matches $d$ on visible rays;
		hence, any induced pair $(\xi, \widetilde{d})$ is equivalent to $(\xi,d)$.
		
		\item 
		Together, these show that any DDF actually forms a shape-induced field pair, via the locally visible zeros $\lvz$.
		In other words, any DDF is a representation of the shape $\lvz$.
		
	\end{itemize}
	
\end{proof}

\subsection{Convergence Example}
\label{pddf:theory:convergenceexample}

We demonstrate the need for defining positional zeroes ($Q_d$) with
\begin{equation}
	\lim_{\epsilon \downarrow 0} 
	\inf_{ \substack{s\in (0,\epsilon)\\ v\in\mathbb{S}^2 } }
	d(q - sv, v) = 0,
	\label{app:pzeros1}
\end{equation} 
rather than the simpler form
\begin{equation}
	\exists\; v \;\,\mathrm{ s.t. }\;
	\lim_{\epsilon \downarrow 0} \inf_{s\in (0,\epsilon)}
	d(q - sv, v) = 0.
	\label{app:pzeros2}
\end{equation} 
The latter scenario requires specifying a single fixed direction 
along which convergence occurs.  
In the former case, however, such a direction need not necessarily exist
(e.g., via a sequence where $d$ converges to zero, 
but there is no fixed direction along which the converge occurs).
This is useful to ensure that the positional zeros $Q_d$ always form a closed set.

Consider the following 2D point set $S$:
\begin{equation}
	p_k = \rho^k R(\theta_k) [1,\, 0]^T 
\end{equation}
where $k\in\{1,2,\ldots\}$, $\rho < 1$, $R(\theta)$ is a 2D rotation matrix with angle $\theta$, 
and $\theta_k = \pi / (2k)$.
Note that $p_k$ converges to $q = [0,\, 0]^T$ 
(but $q\notin \{p_k\}_k$), 
while $\theta_k$ converges to 0.
Let $v_k = R(\theta_k) [1,\, 0]$. As $k$ increases, $v_k$ converges to $u = [1,\,0]^T$.

Consider the simple DDF $d$, which has been induced by $S = \{ p_k \}_k$.\footnote{Note that we define shapes to be compact sets but here $S$ is not closed. However, that is not an issue for this example because we merely want to show that \textit{some} simple DDF can have an open set as its positional zeroes. I.e., we do not need to obtain it by inducement.
}
Is $q\in Q_d$? 
Note that $p_k \in Q_d\;\forall\; k$, so if not, then $Q_d$ will not be a compact set (shape).
Further, note that $d$ necessarily satisfies IO$_d$, in addition to BC$_d$ and DE$_d$.
The former means that if a point is not a zero from one direction, 
then it cannot be a zero from any direction.

First, consider the definition using Eq.~\ref{app:pzeros1}.
In this case, $q \in Q_d$, 
because 
(similar to the proof in \S\ref{pddf:theory:propposzeroes} above) %
for any $\epsilon > 0$, we can always choose a $p_k$ s.t.\ 
$|| p_k - q || < \epsilon$ and 
let $\widetilde{v} := \mathrm{Unit}(p_k - q)$.
This converges to zero, as $\epsilon$ shrinks, in the limit.
Hence the equation is satisfied. 
However, note that $\widetilde{v}$ changes continuously; 
there is not one single direction that is used.

Next, consider the definition using Eq.~\ref{app:pzeros2}.
But notice that $\tau = (q,u)$ is an empty ray, 
since $q\notin \{p_k\}_k$.
Thus, $d(q,u) > 0$ (by IO$_d$), meaning $q\notin Q_d$.
This situation is what we wish to avoid, since it permits the set of positional zeroes to not be closed.